 \documentclass[fleqn,10pt]{wlpeerj} % for preprint submissions

\usepackage{ucs}
\usepackage{amsmath,tabu}
\usepackage{amssymb}
\usepackage{amsfonts}
\usepackage{subfigure}
\usepackage{graphicx,multicol}
\usepackage{color}
\usepackage{algpseudocode,algorithm,algorithmicx}
\usepackage[toc,page]{appendix}
\usepackage{url}
\usepackage{rotating}
\usepackage{soul}

\newtheorem{lemma}{Lemma}
\newtheorem{theorem}{Theorem}

\newcommand{\Var}{\operatorname{Var}}
\newcommand{\AlphaLim}{\underset{\alpha \rightarrow 0}{\lim}}
\newcommand{\E}{\mathbb{E}}
\newcommand{\R}{\mathbb R}

\newcommand\independent{\protect\mathpalette{\protect\independenT}{\perp}}
\def\independenT#1#2{\mathrel{\rlap{$#1#2$}\mkern2mu{#1#2}}}

\title{Analysis of cause-effect inference by comparing regression errors}

\author[1]{Patrick Bl\"obaum}
\author[2]{Dominik Janzing}
\author[1]{Takashi Washio}
\author[3]{Shohei Shimizu}
\author[2]{Bernhard Sch\"olkopf}

\affil[1]{Osaka University, Japan}
\affil[2]{MPI for Intelligent Systems, T\"ubingen, Germany}
\affil[3]{Shiga University, Japan}

\begin{abstract}
We address the problem of inferring the causal direction between two variables by comparing the least-squares errors of the predictions in both possible directions. Under the assumption of an independence between the function relating cause and effect, the conditional noise distribution, and the distribution of the cause, we show that the errors are smaller in causal direction if both variables are equally scaled and the causal relation is close to deterministic. Based on this, we provide an easily applicable algorithm that only requires a regression in both possible causal directions and a comparison of the errors. The performance of the algorithm is compared with various related causal inference methods in different artificial and real-world data sets.
\end{abstract}

\begin{document}

\flushbottom
\maketitle
\thispagestyle{empty}

\section{Introduction}
\label{sec:introduction}
Causal inference \citep{spirtes2000causation,Pearl:2009:CMR:1642718} is becoming an increasingly popular topic in machine learning. The results are often not only of interest in predicting the result of potential interventions, but also in general statistical and machine learning applications \citep{causality_book}. While the causal relationship between variables can generally be discovered by performing specific randomized experiments, such experiments can be very costly, infeasible or unethical.\footnote{Further discussions about ethics in randomized experiments, especially in the context of clinical trials, can be found in \cite{rosner1987ethics}.} In particular, the identification of the causal direction between two variables without performing any interventions is a challenging task. However, recent research developments in causal discovery allow, under certain assumptions, inference of the causal direction between two variables purely based on observational data \citep{Kano2003,Comley,shimizu2006linear,SunLauderdale,zhang2009identifiability,hoyer2009nonlinear,SecondOrder,6620,discreteAN,janzing2012information,sgouritsa2015inference,Mooijetal16,8215503}. As regards the present work, we further contribute to the causal discovery in an unconfounded bivariate setting based on observational data, where one variable is the cause and the other variable is the effect. That is, given observed data $X, Y$ that are drawn from a joint distribution $p_{X,Y}$, we are interested in inferring whether $X$ caused $Y$ or $Y$ caused $X$. In this sense, we define $X$ as the cause and $Y$ as the effect if intervening on $X$ changes the distribution of $Y$. In the following, we use the term 'causal inference' to refer to the identification of the true causal direction.
 
\begin{figure}[t]
\subfigure[]{
  \centering
  \includegraphics[width=0.49\columnwidth]{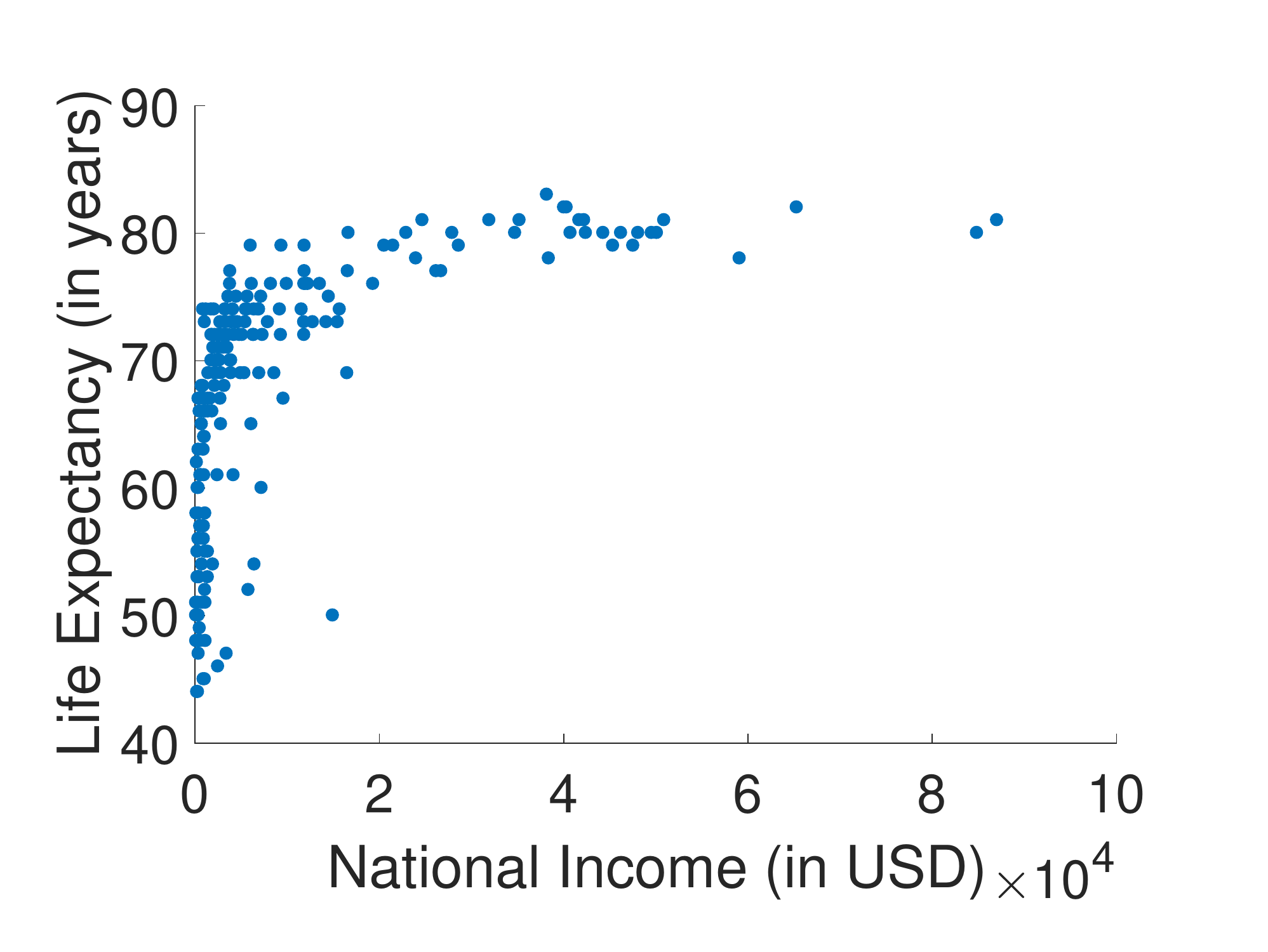}
  \label{fig:income1}
}
\subfigure[]{
  \centering
  \includegraphics[width=0.49\columnwidth]{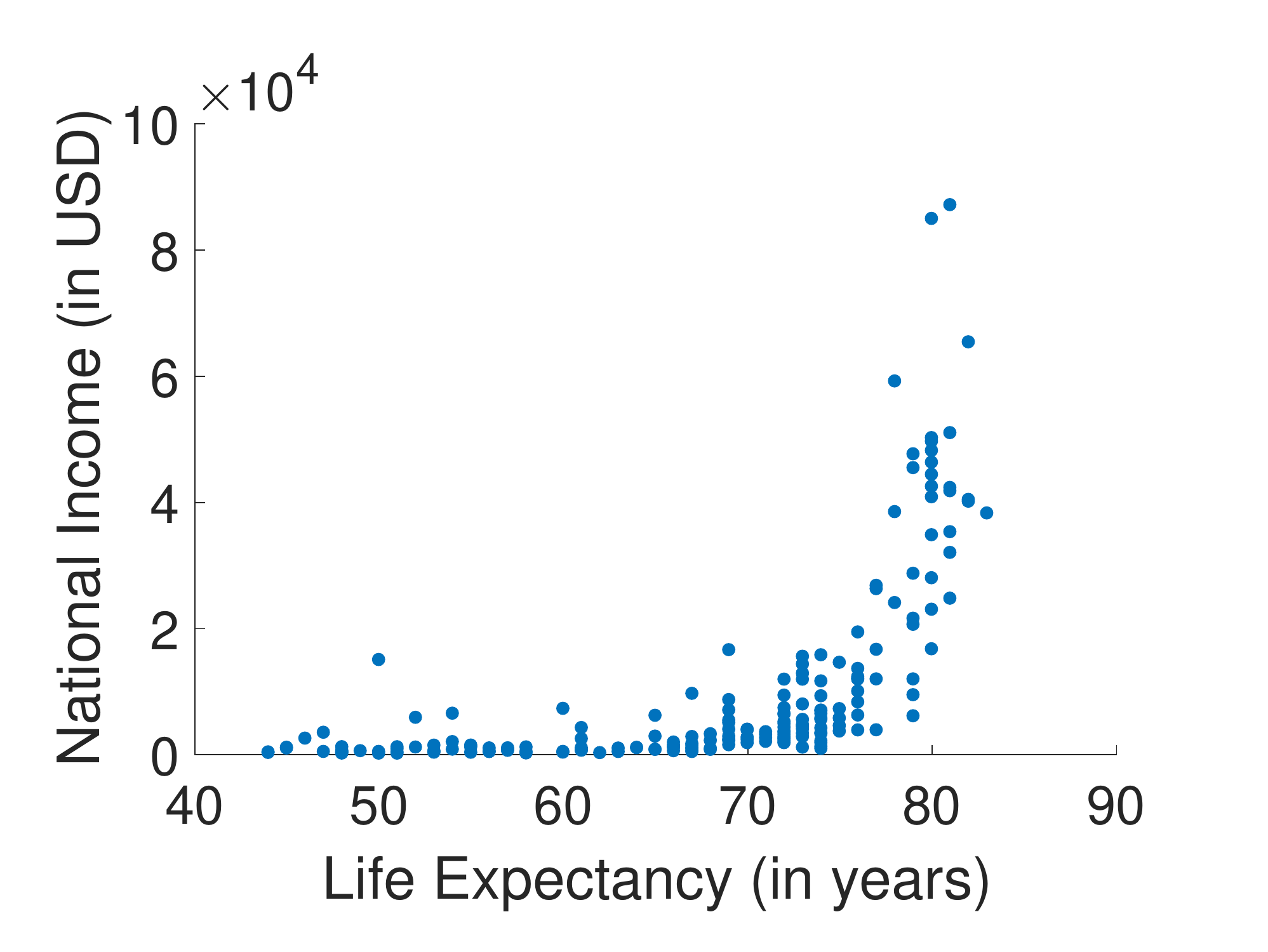}
  \label{fig:income2}
}
\caption{A comparison of the national income of 194 countries and the life expectancy at birth. \textbf{(a)} The national income on the x-axis and the life expectancy on the y-axis. \textbf{(b)} The life expectancy on the x-axis and the national income on the y-axis.}
\label{fig:income}
\end{figure}

A possible application is the discovery of molecular pathways, which relies on the identification of causal molecular interactions in genomics data \citep{Statnikov2012}. Other examples in biomedicine where observational data can be used for causal discovery are discussed in the work by \cite{Ma2017}. An example for a bivariate relationship is provided in Figure \ref{fig:income}, where the national income of countries are compared with the life expectancy at birth.\footnote{The data is taken from https://webdav.tuebingen.mpg.de/cause-effect/ and further discussed in \cite{Mooijetal16}.} Here, a clear statement about the causal relationship is not obvious. It has been argued that richer countries have a better health care system than poorer countries. Hence, a higher national income leads to a higher life expectancy \citep{Mooijetal16}. Based on the plots, this causal relationship is not clear at all. Nevertheless, we provide a way to correctly determine the causal direction by only using these data points.

Conventional approaches to causal inference rely on conditional independences and therefore require at least three observed variables. Given the observed pattern of conditional dependences and independences, one infers a class of directed acyclic graphs (DAGs) that is compatible with the respective pattern (subject to Markov condition and faithfulness assumption \citep{spirtes2000causation,Pearl:2009:CMR:1642718}). Whenever there are causal arrows that are common to all DAGs in the class, conditional (in)dependences yield definite statements about causal directions. In a bivariate setting, however, we rely on asymmetries between cause and effect that are already apparent in the bivariate distribution alone.

One kind of asymmetry is given by restricting the structural equations relating cause and effect to a certain function class: For linear relations with non-Gaussian independent noise, the linear non-Gaussian acyclic model (LiNGAM) \citep{shimizu2006linear} provides a method to identify the correct causal direction. For nonlinear relations, the additive noise model (ANM) \citep{hoyer2009nonlinear} and its generalization to post-nonlinear models (PNL) \citep{zhang2009identifiability} identify the causal direction by assuming an independence between cause and noise, where, apart from some exceptions such as bivariate Gaussian, a model can only be fit in the correct causal direction such that the input is independent of the residual.

Further recent approaches for the bivariate setting are based on an {\it informal} independence assumption stating that the distribution of the cause (denoted by $p_C$) contains no information about the conditional distribution of the effect given the cause (denoted by $p_{E|C}$). Here, the formalization of `no information' is a challenging task. For the purpose of foundational insights (rather than for practical purposes), \cite{Algorithmic} and \cite{LemeireJ2012} formalize the idea via {\it algorithmic information} and postulate that knowing $p_C$ does not enable a shorter description of $p_{E|C}$ and vice versa. Using algorithmic information theory, one can, for instance, show that the algorithmic independence of $p_C$ and $p_{E|C}$ implies
\begin{equation}
\label{eq:kolmo}
	K(p_C)+K(p_{E|C}) \leq K(p_E) + K(p_{C|E}),
\end{equation}
if $K$ denotes the description length of a distribution in terms of its Kolmogorov complexity (for details see Section 4.1.9 in \cite{causality_book}). In this sense, appropriate independence assumptions between $p_C$ and $p_{E|C}$ imply that $p_{E,C}$ has a simpler description in causal direction than in anticausal direction. An approximation of \eqref{eq:kolmo} is given by the SLOPE algorithm in the work by \cite{8215503}, where regression is utilized to estimate and compare the approximated Kolmogorov complexities. For this, a logarithmic error is used, which is motivated by a minimum description length perspective. Another work that is inspired by the independence assumption is the information-geometric approach for causal inference (IGCI) \citep{janzing2012information}. IGCI provides a method to infer the causal direction in deterministic nonlinear relationships subject to a certain independence condition between the slope of the function and the distribution of the cause. A related but different independence assumption is also used by a technique called unsupervised inverse regression (CURE) \citep{sgouritsa2015inference}, where the idea is to estimate a prediction model of both possible causal directions in an unsupervised manner, i.e. only the input data is used for the training of the prediction models. With respect to the above independence assumption, the effect data may contain information about the relation between cause and effect that can be employed for predicting the cause from the effect, but the cause data alone does not contain any information that helps the prediction of the effect from the cause (as hypothesized in \cite{ScholkopfJPSZMV2013}). Accordingly, the unsupervised regression model in the true causal direction should be less accurate than the prediction model in the wrong causal direction.

For our approach, we address the causal inference problem by exploiting an asymmetry in the mean-squared error (MSE) of predicting the cause from the effect and the effect from the cause, respectively, and show, that under appropriate assumptions and in the regime of almost deterministic relations, the prediction error is smaller in causal direction. A preliminary version of this idea can be found in \cite{blobaum2017error,blobaum2017causal} but in these works the analysis is based on a simple heuristic assuming that the regression of $Y$ on $X$ and the regression of $X$ on $Y$ yield functions that are inverse
to each other, which holds approximately in the limit of small noise. Moreover, the analysis is also based on the assumption of an additive noise model in causal direction and on having prior knowledge about the functional relation between $X$ and $Y$, which makes it impractical for generic causal inference problems.
 
In this work, we aim to generalize and extend the two aforementioned works in several ways: 1) We explicitly allow a dependency between cause and noise. 2) We give a proper mathematical proof of the theory that justifies the method subject to clear formal assumptions. 3) We perform extensive evaluations for the application in causal inference and compare it with various related approaches. The theorem stated in this work might also be of interest for general statistical purposes. A briefer version of this work with less extensive experiments, lesser details and without detailed proofs can be found in \cite{confver}.

This paper is structured as follows: In Section \ref{sec:prelim}, we define the problem setting and introduce the used notations and assumptions, which are necessary for the main theorem of this work stated in Section \ref{sec:theory}. An algorithm that utilizes this theorem is proposed in Section \ref{sec:algorithm} and evaluated in various artificial and real-world data sets in Section \ref{sec:experiments}. 

\section{Preliminaries}
\label{sec:prelim}
In the following, we introduce the preliminary problem setting, notations and assumptions.

\subsection{Problem setting and notation}
\begin{figure}[t!]
\centering
\includegraphics[width=0.8\columnwidth]{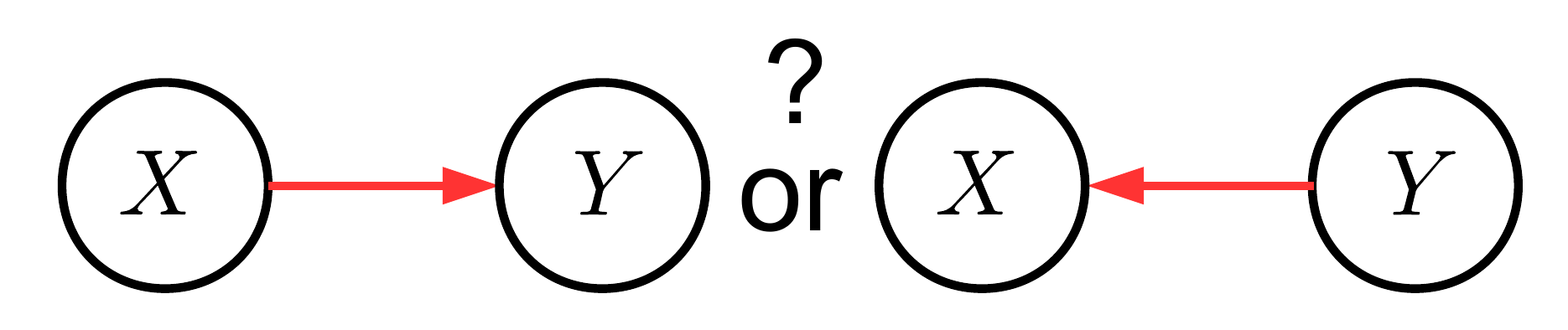}
\caption{An illustration of the goal of our proposed method. It aims to identify the causal DAG of two variables, where either $X$ causes $Y$ or $Y$ causes $X$.}
\label{fig:identifydag}
\end{figure}
In this work, we use the framework of structural causal models \citep{Pearl:2009:CMR:1642718} with the goal of correctly identifying cause and effect variables of given observations from $X$ and $Y$. As illustrated in Figure \ref{fig:identifydag}, this can be described by the problem of identifying whether the causal DAG of $X \rightarrow Y$ or $X \leftarrow Y$ is true. Throughout this paper, a capital letter denotes a random variable and a lowercase letter denotes values attained by the random variable. Variables $X$ and $Y$ are assumed to be real-valued and to have a joint probability density (with respect to the Lebesgue measure), denoted by $p_{X,Y}$. By slightly abusing terminology, we will not further distinguish between a distribution and its density since the Lebesgue measure as a reference is implicitly understood. The notations $p_{X}$, $p_{Y}$, and  $p_{Y|X}$ are used for the corresponding marginal and conditional densities, respectively. The derivative of a function $f$ is denoted by $f'$.

\subsection{General idea}
\label{sec:generalIdea}
As mentioned before, the general idea of our approach is to simply compare the MSE of regressing $Y$ on $X$ and the MSE of regressing $X$ on $Y$. If we denote  cause and effect by $C,E\in \{X,Y\}$, respectively, our approach explicitly reads as follows. Let $\phi$ denote the function that minimizes the expected least squares error when predicting $E$ from $C$, which implies that $\phi$ is given by the conditional expectation $\phi(c)=\E[E|c]$. Likewise, let $\psi$ be the minimizer of the least squares error for predicting $C$ from $E$, that is, $\psi(e)=\E[C|e]$. Then we will postulate assumptions that imply
\begin{equation}\label{eq:compareerrors}
	\E[(E-\phi(C))^2] \leq \E[(C-\psi(E))^2]
\end{equation}
in the regime of almost deterministic relations. This conclusion certainly relies on some kind of scaling convention. For our theoretical results we will assume that both $X$ and $Y$ attain values between $0$ and $1$. However, in some applications, we will also scale $X$ and $Y$ to unit variance to deal with unbounded variables. Eq. \eqref{eq:compareerrors} can be rewritten in terms of conditional variance as
\begin{equation*}
	\E[\Var[E|C]] \leq \E[\Var[C|E]].
\end{equation*}

\subsection{Assumptions}
\label{sec:assumptions}
First, recall that we assume throughout the paper that either $X$ is the cause of $Y$ or vice versa in an unconfounded sense, i.e. there is no common cause. Therefore, the general structural equation is defined as
\begin{equation}
\label{eq:structuralEq}
E = \zeta(C, \tilde N),
\end{equation}
where $C \independent \tilde N$. For our analysis, we first define a function $\phi$ to be the conditional expectation of the effect given the cause, i.e.
\begin{equation*}
\phi(c) := \E[E|c]
\end{equation*}
and, accordingly, we define a noise variable $N$ as the residual
\begin{equation}
\label{eq:residualDef}
N := E - \phi(C).
\end{equation}
Note that \eqref{eq:residualDef} implies that $\E[N|c] = 0$. The function $\phi$ is further specified below. Then, to study the limit of an almost deterministic relation in a mathematically precise way, we consider a family of effect variables $E_\alpha$ by
\begin{equation}\label{eq:model}
	E_\alpha := \phi(C) + \alpha N,
\end{equation}
where $\alpha\in \R^+$ is a parameter controlling the noise level and $N$ is a noise variable that has some (upper bounded) joint density $p_{N,C}$ with $C$. Note that $N$ here does not need to be statistically independent of $C$ (in contrast to ANMs), which allows the noise to be non-additive. Therefore, \eqref{eq:model} does not, a priori, restrict the set of possible causal relations, because for any pair $(C,E)$ one can always define the noise $N$ as \eqref{eq:residualDef} and thus obtain $E_{\alpha = 1} = E$ for any arbitrary function $\phi$.\footnote{Note that although the form of \eqref{eq:model} is similar to that of ANM, the core assumption of ANM is an independence between cause and noise, which we do not need in our approach. Therefore, we assume the general structural equation defined in \eqref{eq:structuralEq}, whereas ANM assumes a more restrictive structural equation of the form $E = \zeta(C) + \tilde N$ with $C \independent \tilde N$.}

For this work, we make use of the following assumptions:
\begin{enumerate}
	\item \textbf{Invertible function:} $\phi$ is a strictly monotonically increasing two times differentiable function $\phi: [0,1] \rightarrow [0,1]$. For simplicity, we assume that $\phi$ is monotonically increasing with $\phi(0)=0$ and $\phi(1)=1$ (similar results for monotonically decreasing functions follow by reflection $E \to 1-E$). We also assume that $\phi^{-1'}$ is bounded.
	 \item \textbf{Compact supports:} The distribution of $C$ has compact support. Without loss of generality, we assume that $0$ and $1$ are, respectively, the smallest and the largest values attained by $C$. We further assume that the distribution of $N$ has compact support and that there exist values $n_+>0>n_-$ such that for any $c$, $[n_-,n_+]$ is the smallest interval containing the support of $p_{N|c}$. This ensures that we know $[\alpha n_-,1+\alpha n_+]$ is the smallest interval containing the support of $p_{E_\alpha}$. Then the shifted and rescaled variable 
\begin{equation}\label{eq:tilde} 
	\tilde{E}_\alpha :=  \frac{1}{1 + \alpha n_+ - \alpha n_-}(E_\alpha - \alpha n_-)
\end{equation}
attains $0$ and $1$ as minimum and maximum values and thus is equally scaled as $C$.
	\item \textbf{Unit noise variance:} The expected conditional noise variance is ${\E[\Var[N|C]] = 1}$ without loss of generality, seeing that we can scale the noise arbitrary by the parameter $\alpha$ and we are only interested in the limit $\alpha\to 0$.
	\item \textbf{Independence postulate:} While the above assumptions are just technical, we now state the essential assumption that generates the asymmetry between cause and effect. To this end, we consider the unit interval $[0,1]$ as probability space	with uniform distribution as probability measure. The functions $c\mapsto \phi'(c)$ and $c \mapsto \Var[N|c] p_C(c)$ define random variables on this space, which we postulate to be uncorrelated, formally stated as
\begin{equation}\label{eq:shortpost}
	\operatorname{Cov}[\phi', \Var[N|c] p_C] = 0.
\end{equation}
More explicitly, \eqref{eq:shortpost} reads:
\begin{equation}
	\int_0^1 \phi'(c) \Var[N|c] p_C(c) dc - \int_0^1 \phi'(c) dc \int_0^1 \Var[N|c] p_C(c) dc=0.
 	\label{eq:indep}	
\end{equation}
\end{enumerate}
The justification of \eqref{eq:shortpost} is not obvious at all. For the special case where the conditional variance $\Var[N|c]$ is a constant in $c$ (e.g. for ANMs), \eqref{eq:shortpost} reduces to
\begin{equation}\label{eq:igcilike}
	\operatorname{Cov}[\phi',  p_C] = 0,
\end{equation}
which is an independence condition for deterministic relations stated in \cite{ScholkopfJPSZMV2013}. Conditions of similar type as \eqref{eq:igcilike} have been discussed and justified in \cite{janzing2012information}.
They are based on the idea that $\phi$ contains no information about $p_C$. This, in turn, relies on the idea that the conditional $p_{E|C}$ contains no information about $p_C$. 

To discuss the justification of \eqref{eq:indep}, observe first that it {\it cannot} be justified as stating some kind of  `independence' between $p_C$ and $p_{E|C}$. To see this, note that \eqref{eq:indep} states an uncorrelatedness of the two functions $c \mapsto \phi'(c)$ and $c \mapsto \Var[N|c]p_C(c)$. While $\phi'$ depends only on the conditional $p_{E|C}$ and not on $p_C$, the second function depends on both $p_{C|E}$ and $p_E$, since $\Var[N|c]$ is a property of $p_{E|C}$. Nevertheless, to justify \eqref{eq:indep} we assume that the function $\phi$ represents a law of nature that persists when $p_C$ and $N$ change due to changing background conditions. From this perspective, it becomes unlikely that they are related to the background condition at hand. This idea follows the general spirit of `modularity and autonomy' in structural equation modeling, that some structural equations may remain unchanged when other parts of a system change (see Chapter~2 in \cite{causality_book} for a literature review).\footnote{Note, however, that the assignment \eqref{eq:model} is not a structural equation in a strict sense, because then $C$ and $N$ would need to be statistically independent.} To further justify \eqref{eq:shortpost}, one could think of a scenario where someone changes $\phi$ independently of $p_{N, C}$, which then results in vanishing correlations. Typically, this assumption would be violated if $\phi$ is adjusted to $p_{N, C}$ or vice versa. This could happen due to an intelligent design by, for instance, first observing $p_{N, C}$ and then defining $\phi$ or due to a long adaption process in nature (see \cite{janzing2012information} for further discussions of possible violations in a deterministic setting).

A simple implication of \eqref{eq:indep} reads
\begin{equation}\label{eq:just1}
	\int_0^1 \phi'(c) \Var[N|c] p_C(c) dc = 1,
\end{equation}
due to $\int_0^1 \phi'(c) dc = 1$ and $\int_0^1 \Var[N|c] p_C(c) dc = \E[\Var[N|C]] =1$. 

In the following, the term \textit{independence postulate} is used to refer to the aforementioned postulate and the term \textit{independence} to a statistical independence, which should generally become clear from the context.

\section{Theory}
\label{sec:theory}
As introduced in Section \ref{sec:generalIdea}, we aim to exploit an inequality of the expected prediction errors in terms of $\E[\Var[E|C]] \leq \E[\Var[C|E]]$ to infer the causal direction. In order to conclude this inequality and, thus, to justify an application to causal inference, we must restrict our analysis to the case where the noise variance is sufficiently small, since a more general statement is not possible under the aforementioned assumptions. The analysis can be formalized by the ratio of the expectations of the conditional variances in the limit $\alpha \rightarrow 0$.

We will then show 
\begin{equation*}
	\AlphaLim \frac{\E[\Var[C|\tilde{E}_\alpha]] }{\E[\Var[\tilde{E}_\alpha |C]]} \geq 1.
\end{equation*}

\subsection{Error asymmetry theorem}
For our main theorem, we first need an important lemma:
\begin{lemma}[Limit of variance ratio]\label{lem:lim}
	Let the assumptions 1-3 in Section \ref{sec:assumptions} hold. Then the following limit holds:
	\begin{equation}\label{eq:lem}
	\AlphaLim \frac{\E[\Var[C | \tilde{E}_\alpha]]}{\E[\Var[\tilde{E}_\alpha | C]]} = \int_0^1 \frac{1}{\phi'(c)^2} \Var[N | c] p_C(c) dc
\end{equation}
\end{lemma}
\textbf{Proof:} We first give some reminders of the definition of the conditional variance
and some properties.
For two random variables $Z$ and $Q$ the conditional variance of $Z$, given $q$ is defined by
\begin{align*}
	\Var[Z|q] & := \E[(Z - \E[Z|q])^2 | q], 
\end{align*}	
while $\Var[Z|Q]$ is the random variable attaining the value $\Var [Z|q]$ when $Q$ attains the value $q$. Its expectation reads
\begin{align*}	
	\E[\Var[Z|Q]] & := \int \Var[Z | q] p_Q(q) dq. 
\end{align*}
For any $a\in \R$, we have
\begin{align*}
	\Var\left[\frac{Z}{a} \middle| q \right] & = \frac{\Var[Z|q]}{a^2}.
\end{align*}
For any function $h$, we have
\begin{align*}
	\Var[h(Q) + Z | q] & = \Var[Z | q], 
\end{align*}
which implies $\Var[h(Q)|q]=0$.
Moreover, we have 
\[
	\Var[Z|h(q)] = \Var[Z | q],
\]
if $h$ is invertible. 

To begin the main part of the proof, we first observe
\begin{equation}\label{eq:alpha2}
\E[\Var[E_\alpha|C]] = \E[\Var[\phi(C) + \alpha N|C]] = \alpha^2 \underbrace{\E[\Var[N|C]]}_{= \ 1 \text{ (Assumpt. 3))}} = \alpha^2. 
\end{equation}
Moreover, one easily verifies that
\begin{equation}\label{eq:removetilde}
\AlphaLim \frac{\E[\Var[C | \tilde{E}_\alpha ]]}{\E[\Var[\tilde{E}_\alpha | C]]} = \AlphaLim \frac{\E[\Var[C | E_\alpha]]}{\E[\Var[E_\alpha | C]]},
\end{equation}
due to \eqref{eq:tilde}
provided that these limits exist. Combining \eqref{eq:alpha2} and \eqref{eq:removetilde}
yields
\begin{equation}
\label{eq:reducedLimit}
	\AlphaLim \frac{\E[\Var[C | \tilde{E}_\alpha]]}{\E[\Var[\tilde{E}_\alpha | C]]} = \AlphaLim \frac{\E[\Var[C | E_\alpha]]}{\alpha^2} = \AlphaLim \E\left[\Var\left[\frac{C}{\alpha} \middle| E_\alpha \right] \right].
\end{equation}

Now, we can rewrite \eqref{eq:reducedLimit} as
\begin{align}
	\AlphaLim \E\left[\Var\left[\frac{C}{\alpha} \middle| E_\alpha \right] \right] & = \AlphaLim \E\left[\Var\left[\frac{\phi^{-1}(E_\alpha - \alpha N)}{\alpha} \middle| E_\alpha \right] \right] \nonumber \\
	& = \AlphaLim \int_{\phi(0) + \alpha n_-}^{\phi(1) + \alpha n_+} \Var\left[\frac{\phi^{-1}(e - \alpha N)}{\alpha} \middle| e \right] p_{E_\alpha}(e) de \nonumber \\
	& = \AlphaLim \int_{\phi(0)}^{\phi(1)} \Var\left[\frac{\phi^{-1}(e - \alpha N)}{\alpha} \middle| e \right] p_{E_\alpha }(e) de.
\label{eq:int1}
\end{align}
In the latter step, $\alpha n_+$ and $-\alpha n_-$ vanishes in the limit seeing that the function
\[
e \mapsto \Var\left[\phi^{-1}(e - \alpha N) / \alpha \middle| e \right] p_{E_\alpha}(e)
\]
is uniformly bounded in $\alpha$. This is firstly, because $\phi^{-1}$ attains only values in $[0,1]$, and hence the variance is bounded by $1$. Secondly, $p_{E_\alpha}(e)$ is uniformly bounded due to  
\begin{eqnarray*}
p_{E_\alpha} (e) &=& \int_{n_-}^{n_+} p_{\phi(C),N} (e-\alpha n,n) dn = \int_{n_-}^{n_+} p_{C,N}(\phi^{-1}(e-\alpha n),n)
\phi^{-1'}(e-\alpha n) dn \\ &\leq & 
\|\phi^{-1'}\|_\infty \|p_{C,N}\|_\infty (n_+-n_-).
\end{eqnarray*}
Accordingly, the bounded convergence theorem states 
\[
\AlphaLim \int_{\phi(0)}^{\phi(1)} \Var\left[\frac{\phi^{-1}(e - \alpha N)}{\alpha} \middle| e \right] p_{E_\alpha }(e) de =
 \int_{\phi(0)}^{\phi(1)} \AlphaLim \left( \Var\left[\frac{\phi^{-1}(e - \alpha N)}{\alpha} \middle| e \right] p_{E_\alpha }(e) \right) de.
\]

 To compute the limit of
 \[
 \Var\left[\frac{\phi^{-1}(e - \alpha N)}{\alpha} \middle| e \right],
\] 
we use Taylor's theorem to obtain
\begin{equation}\label{eq:taylor}
	\phi^{-1}(e - \alpha n) = \phi^{-1}(e) - \alpha n \phi^{-1}{'}(e) - \frac{\alpha^2 n^2 \phi^{-1}{''}(E_2(n, e))}{2},
\end{equation}
where $E_2(n, e)$ is a real number in the interval $(e-\alpha n, e)$. 
Since \eqref{eq:taylor} holds for every $n\in [-\frac{e}{\alpha},\frac{1-e}{\alpha}]$ 
(note that $\phi$ and $\phi^{-1}$ are bijections of $[0,1]$, thus $e-\alpha n$ lies in $[0,1]$)
 it also holds for the random variable $N$ if $E_2(n,e)$ is replaced with the random variable
$E_2(N,e)$ (here, we have implicitly assumed that the map $n \mapsto e_2(n,e)$ is measurable).
Therefore, we see that
\begin{eqnarray}
\AlphaLim \Var\left[\frac{\phi^{-1}(e - \alpha N)}{\alpha} \middle| e \right] &= &\AlphaLim \Var\left[- N \phi^{-1}{'}(e) - \frac{\alpha N^2 \phi^{-1}{''}(E_2(N, e))}{2} \middle| e \right] \nonumber\\
  &=& \phi^{-1'}(e)^2 \Var [N|e] \label{eq:mainred}.
\end{eqnarray}
Moreover, we have
\begin{equation}\label{eq:e0}
\lim_{\alpha \to 0} p_{E_\alpha}(e) = p_{E_0}(e). 
\end{equation}
Inserting \eqref{eq:e0} and \eqref{eq:mainred} into \eqref{eq:int1} yields
\begin{eqnarray*}
\AlphaLim \E\left[\Var\left[\frac{C}{\alpha} \middle| E_0 \right] \right]&=&
\int_{\phi(0)}^{\phi(1)} \phi^{-1'}(e)^2 \Var [N|e]  p_{E_0}(e) de \\
&=& \int_0^1 \phi^{-1'}(\phi(c))^2 \Var [N|\phi(c)]  
  p_C(c) dc\\
&=& \int_0^1 \frac{1}{\phi'(c)^2} \Var [N|c]  
  p_C(c) dc,
\end{eqnarray*}
where the second equality is a variable 
substitution using the deterministic relation $E_0=\phi(C)$ (which implies 
$p_{E_0}(\phi(c)) =p_C(c)/\phi'(c)$ or, equivalently, the simple symbolic equation
$p_{E_0}(e) de = p_C(c) dc$).  
This completes the proof due to \eqref{eq:reducedLimit}. 
\begin{flushright}
$\square$
\end{flushright}

While the formal proof is a bit technical, the intuition behind this idea is quite simple: just think of the scatter plot of an almost deterministic relation as a thick line. Then $\Var[E_\alpha | c]$ and $\Var[C | E_\alpha = \phi(c)]$ are roughly the squared widths of the line at some point $(c,\phi(c))$ measured in vertical and horizontal direction, respectively. The quotient of the widths in vertical and horizontal direction is then given by the slope. This intuition yields the following approximate identity for small $\alpha$:
\begin{equation}\label{eq:approx}
\Var[C|\tilde{E}_\alpha=\phi(c)] \approx \frac{1}{(\phi'(c))^2} \Var[\tilde{E}_\alpha|C=c]    = \alpha^2  \frac{1}{(\phi'(c))^2} \Var[N|c] .
\end{equation}
Taking the expectation of \eqref{eq:approx} over $C$ and recalling that Assumption~3 implies $\E[\Var[\tilde{E}_\alpha| C]] = \alpha^2 \E [\Var [N|C]] = \alpha^2$
already yields \eqref{eq:lem}.

With the help of Lemma~\ref{lem:lim}, we can now formulate the core theorem of this paper:

\begin{theorem}[Error Asymmetry]\label{thm:main}
Let the assumptions 1-4 in Section \ref{sec:assumptions} hold. Then the following limit always holds
\begin{equation*}
	\AlphaLim \frac{\E[\Var[C | \tilde{E}_\alpha]]}{\E[\Var[\tilde{E}_\alpha | C]]} \geq 1,
\end{equation*}
with equality only if the function stated in Assumption 1 is linear.
\end{theorem}
\textbf{Proof:} We first recall that Lemma~\ref{lem:lim} states
\begin{align*}
	\AlphaLim \frac{\E[\Var[C | \tilde{E}_\alpha]]}{\E[\Var[\tilde{E}_\alpha | C]]} = \int_0^1 \frac{1}{\phi'(c)^2} \Var[N | c ] p_C(c) dc.
\end{align*}
We then have
\begin{align}
	& \int_0^1 \frac{1}{\phi'(c)^2} \Var[N | c ] p_C(c) dc \nonumber \\
	= & \int_0^1 \frac{1}{\phi'(c)^2} \Var[N | c ] p_C(c) dc \cdot \underbrace{\int_0^1 \Var[N | c ] p_C(c) dc}_{= \ 1 \text{ (Assumpt. 3)}} \nonumber \\
	= & \int_0^1 \sqrt{\left(\frac{1}{\phi'(c)}\right)^2 \Var[N | c ]}^2 p_C(c) dc \cdot \int_0^1 \sqrt{\Var[N | c]}^2 p_C(c) dc \nonumber \\
	\geq & \left(\int_0^1 \sqrt{\left(\frac{1}{\phi'(c)}\right)^2 \Var[N | c ]} \sqrt{\Var[N | c ]} p_C(c) dc\right)^2 \nonumber \\
	= & \left(\int_0^1 \frac{1}{\phi'(c)} \Var[N | c ] p_C(c) dc\right)^2,
	\label{eq:ineq1}
\end{align}
where the inequality is just the Cauchy Schwarz inequality applied to the bilinear form $f,g \mapsto \int f(c) g(c) p_C(c) dc$ for the space of functions $f$ for which $\int f^2(c) p_c(c) dc$ exists. Note that if $\phi$ is linear, \eqref{eq:ineq1} becomes $1$, since $\phi' = 1$ according to Assumpt. 1. We can make a statement about \eqref{eq:ineq1} in a similar way by using \eqref{eq:just1} implied by the independence postulate and  using Cauchy Schwarz:
\begin{align}
	& \int_0^1 \frac{1}{\phi'(c)} \Var[N | c ] p_C(c) dc \nonumber \\
	= & \int_0^1 \frac{1}{\phi'(c)} \Var[N | c ] p_C(c) dc \cdot \underbrace{\int_0^1 \phi'(c) \Var[N | c ] p_C(c) dc}_{= \ 1 \text{ \eqref{eq:just1}}} \nonumber \\
	= & \int_0^1 \sqrt{\frac{1}{\phi'(c)} \Var[N | c ]}^2 p_C(c) dc \cdot \int_0^1 \sqrt{\phi'(c) \Var[N | c ]}^2 p_C(c) dc \nonumber  \\
	\geq & \left(\int_0^1 \sqrt{\frac{1}{\phi'(c)} \Var[N | c ]} \sqrt{\phi'(c) \Var[N | c ]} p_C(c) dc\right)^2 \nonumber \\
	= & \left(\underbrace{\int_0^1 \Var[N | c] p_C(c) dc}_{= \ 1 \text{ (Assumpt. 3)}} \right)^2 = 1.
	\label{eq:ineq2}
\end{align}
Combining \eqref{eq:ineq1} and \eqref{eq:ineq2} with Lemma~\ref{lem:lim}
completes the proof.
\begin{flushright}
$\square$
\end{flushright}

\subsection{Remark}
Theorem $1$ states that the inequality holds for all values of $\alpha$ smaller than a certain finite threshold. Whether this threshold is small or whether the asymmetry with respect to regression errors already occurs for large noise cannot be concluded from the theoretical insights. Presumably, this depends on the features of $\phi$, $p_C$, $p_{N|C}$ in a complicated way. However, the experiments in Section \ref{sec:experiments} suggest that the asymmetry often appears even for realistic noise levels. 

If the function $\phi$ is non-invertible, there is an information loss in anticausal direction, since multiple possible values can be assigned to the same input. Therefore, we can expect that the error difference becomes even higher in these cases, which is supported by the experiments in Section \ref{sec:cepNoise}.

\section{Algorithm}
\label{sec:algorithm}
A causal inference algorithm that exploits Theorem~\ref{thm:main} can be formulated in a straightforward manner. Given observations $X, Y$ sampled from a joint distribution $p_{X,Y}$, the key idea is to fit regression models in both possible directions and compare the MSE. We call this approach Regression Error based Causal Inference (RECI) and summarize the algorithm in Algorithm~1.

Although estimating the conditional expectations $\E[Y|X]$ and $\E[X|Y]$ by regression is a standard task in machine learning, we should emphasize that the usual issues of over- and underfitting are critical for our purpose (like for methods based on ANMs or PNLs), because they under- or overestimate the noise levels. It may, however, happen that the method even benefits from underfitting: if there is a simple regression model in causal direction that fits the data quite well, but in anticausal relation the conditional expectation becomes more complex, a regression model with underfitting increases the error even more for the anticausal direction than for the causal direction.
\begin{algorithm}[t]
\label{alg:algo}
\caption{The proposed causal inference algorithm.}
\begin{algorithmic}
\Function{RECI}{$X$, $Y$} \Comment{$X$ and $Y$ are the observed data.}
\State $(X, Y) \gets \text{RescaleData}(X, Y)$
\State $f \gets \text{FitModel}(X, Y)$ \Comment{Fit regression model $f\colon X \rightarrow Y$}
\State $g \gets \text{FitModel}(Y, X)$ \Comment{Fit regression model $g\colon Y \rightarrow X$ }
\State $\text{MSE}_{Y|X} \gets \text{MeanSquaredError}(f, X, Y)$
\State $\text{MSE}_{X|Y} \gets \text{MeanSquaredError}(g, Y, X)$
\If{$\text{MSE}_{Y|X} < \text{MSE}_{X|Y}$}
	\State \Return \text{$X$ causes $Y$}
\ElsIf{$\text{MSE}_{X|Y} < \text{MSE}_{Y|X}$}
	\State \Return \text{$Y$ causes $X$}
\Else
	\State \Return \text{No decision}
\EndIf
\EndFunction
\end{algorithmic}
\end{algorithm}
This speculative remark is related to \eqref{eq:kolmo} and somehow supported by our experiments, where we observed that simple models performed better than complex models, even though they probably did not represent the true conditional expectation.

Also, an accurate estimation of the MSE with respect to the regression model and appropriate preprocessing of the data, such as removing isolated points in low-density regions, might improve the performance. While Algorithm 1 only rejects a decision if the error is equal, one could think about utilizing the error difference as a rejection criteria of a decision. For instance, if the error difference is smaller than a certain threshold, the algorithm returns 'no decision'. This idea is further evaluated in Section \ref{sec:errorRatio}.

\section{Experiments}
\label{sec:experiments}
In this section, we compare our algorithm with five different related methods for inferring the causal direction in various artificially generated and observed real-world data sets. In each evaluation, observations of two variables were given and the goal was to correctly identify cause and effect variable.

\subsection{Causal inference methods for comparison}
In the following, we briefly discuss and compare the causal inference methods which we used for the evaluations.

\paragraph{LiNGAM} The model assumptions of LiNGAM \citep{shimizu2006linear} are
\begin{equation*}
 E = \beta C + N,
\end{equation*}
where $\beta \in \mathbb{R}$, $C \independent N$ and $N$ is non-Gaussian. While LiNGAM is especially suitable for linear functional relationships with non-Gaussian noise, it performs poorly if these assumptions are violated. The computational cost is, however, relatively low.

For the experiments, we used a state-of-the-art implementation of LiNGAM that utilizes an entropy based method for calculating the likelihood ratios of the possible causal directions, instead of an independent component analysis based algorithm as in the original version \citep{hyvarinen2013pairwise}. For this, Eq. (3) in \cite{hyvarinen2013pairwise} is used in order to estimate the likelihood ratio Eq. (2) in \cite{hyvarinen2013pairwise}.

\paragraph{ANM} The ANM \citep{hoyer2009nonlinear} approach assumes that
\begin{equation*}
 E = f(C) + N,
\end{equation*}
where $f$ is nonlinear and $C \independent N$. An asymmetry between cause and effect is achieved by the assumption of an independence between cause and residual. Therefore, this method requires fitting a regression function and performing an additional evaluation of the relation between inputs and residual, which lead to a high computational cost. Note that the choice of the evaluation method is crucial for the performance.

We used an implementation provided by \cite{Mooijetal16}, which uses a Gaussian process regression for the prediction and provides different methods for the evaluation of the causal direction. For the experiments, we chose different evaluation methods; \textit{HSIC} for statistical independence tests, an entropy estimator for the estimation of the mutual information between input and residuals (denoted as \textit{ENT}) and a Bayesian model comparison that assumes Gaussianity (denoted as \textit{FN}). Implementation details and parameters can be found in Table $2$ of \cite{Mooijetal16}.

\paragraph{PNL} Post non-linear models \citep{zhang2009identifiability} are a generalization of ANMs. Here, it is assumed that
\begin{equation*}
	E = g(f(C) + N),
\end{equation*}
where $g$ is nonlinear and $C \independent N$. Due to the additional nonlinearity coming from $g$, this allows a non-additive influence of the noise as in contrast to an ANM. For inferring the causal direction, the idea remains roughly the same; fit a PNL in both possible directions and check for independence between input and disturbance. However, the disturbance here is different from the regression residual and fitting a PNL model is a significantly harder problem than fitting an ANM.
 
In the experiments, we used an implementation provided by the authors \cite{zhang2009identifiability}, where a constrained nonlinear independent component analysis is utilized for estimating the disturbances and HSIC for statistical independence tests.

\paragraph{IGCI} The IGCI \citep{janzing2012information} approach is able to determine the causal relationship in a deterministic setting
\begin{equation*}
	E = f(C),
\end{equation*}
under the `independence assumption' $\operatorname{Cov}[\log f', p_C] = 0$, i.e. the (logarithmic) slope of the function and the cause distribution are uncorrelated. The causal direction can then be inferred if the Kullback Leibler divergence between a reference measure and $p_X$ is bigger or smaller than the Kullback Leibler divergence between the same reference measure and $p_Y$, respectively. The corresponding algorithm has been applied to noisy causal relations with partial success (and some heuristic justifications \citep{janzing2012information}), but generalizations of IGCI for non-deterministic relations are actually not known and we consider Assumption 4 in Section~\ref{sec:assumptions} as first step towards a possibly more general formulation. The computational cost depends on the utilized method for estimating the information criterion, but is generally low. Therefore, IGCI is the fastest of the methods.

For the experiments, we also used an implementation provided by \cite{Mooijetal16}, where we always tested all possible combinations of reference measures and information estimators. These combinations are denoted as IGCI-ij, where $i$ and $j$ indicate:
\begin{itemize}
	\item $i = \text{U}$: Uniform reference measure (normalizing $X$ and $Y$)
	\item $i = \text{G}$: Gaussian reference measure (standardizing $X$ and $Y$)
	\item $j = 1$: Entropy estimator using Eq. (12) in \cite{6620}
	\item $j = 2$: Integral approximation of Eq. (13) in \cite{6620}
	\item $j = 3$: Integral approximation of Eq. (22) in \cite{Mooijetal16}
\end{itemize}

\paragraph{CURE} CURE \citep{sgouritsa2015inference} is based on the idea that an unsupervised regression of $E$ on $C$ by only using information from $p_C$ performs worse than an unsupervised regression of $C$ on $E$ by only using information from $p_E$. CURE implements this idea in a Bayesian way via a modified Gaussian process regression. However, since CURE requires the generation of Markov-Chain-Monte-Carlo (MCMC) samples, the biggest drawback is a very high computational cost. 

An implementation of CURE by the authors has been provided for our experiments. Here, we used similar settings as described in Section 6.2 of \cite{sgouritsa2015inference}, where $200$ data samples were used and $10000$ MCMC samples were generated. The number of internal repetitions depends on the experimental setting.

\paragraph{SLOPE} The SLOPE approach by \cite{8215503} is essentially motivated by \eqref{eq:kolmo} and compares an estimation of $(K(p_X) + K(p_{Y|X}) / (K(p_X) + K(p_Y))$ with an estimation of $(K(p_Y) + K(p_{X|Y}) / (K(p_X) + K(p_Y))$ based on the minimum description length principle \citep{RISSANEN1978465}. This approach uses a global and multiple local regression models to fit the data, where the description length of the fitted regression models and the description length of the error with respect to the data can be used to approximate $K(p_{Y|X})$ and $K(p_{X|Y})$, respectively. Seeing that multiple regression models need to be fit depending on the structure of the data, the computational costs can vary between data sets.

For our experiments, we used the implementation provided by the authors with the same parameters as used in their experiments with real-world data.

\paragraph{RECI} Our approach addresses non-deterministic nonlinear relations and, in particular, allows a dependency between cause and noise. Since we only require the fitting of a least-squares solution in both possible causal directions, RECI can be easily implemented. It does not rely on any independence tests and has, depending on the regression model and implementation details, a low computational cost.

In the experiments, we have always used the same class of regression function for the causal and anticausal direction to compare the errors, but performed multiple experiments with different function classes. For each evaluation, we randomly split the data into training and test data, where we tried different ratios and selected the best performing model on the test data. The used percentage of training data were 70\%, 50\% or 30\%, where the remaining data served as test data. In each run, we only randomly split the data once. The utilized regression models were:
\begin{itemize}
	\item a logistic function (LOG) of the form $a + (b - a) / (1 + \operatorname{exp}(c \cdot (d - x)))$
	\item shifted monomial functions (MON) of the form $a x^n + b$ with $n \in [2, 9]$
	\item polynomial functions (POLY) of the form $\sum_{i = 0}^k a_i x^i$ with $k \in [1, 9]$
	\item support vector regression (SVR) with a linear kernel
	\item neural networks (NN) with different numbers of hidden neurons: 2, 5, 10, 20, 2-4, 4-8, where '-' indicates two hidden layers
\end{itemize}
The logistic and monomial functions cover rather simple regression models, which are probably not able to capture the true function $\phi$ in most cases. On the other hand, support vector regression and neural networks should be complex enough to capture $\phi$. The polynomial functions are rather simple too, but more flexible than the logistic and monomial functions.

We used the standard Matlab implementation of these methods and have always chosen the default parameters, where the parameters of LOG, MON and POLY were fitted by minimizing the least-squares error.

During the experiments, we observed that the MSE varied a lot in many data sets due to relatively small sample sizes and the random selection of training and test data. Therefore, we averaged the MSE over all performed runs within the same data set first before comparing them, seeing that this should give more accurate estimations of $\E[\Var[Y|X]]$ and $\E[\Var[X|Y]]$ with respect to the class of the regression function. Although the choice of the function class for each data set is presumably a typical model selection problem, we did not optimize the choice for each data set individually. Therefore, we only summarize the results of the best performing classes with respect to the experimental setup in the following. The estimated MSE in each data set were averaged over all performed runs. A detailed overview of all results, including the performances and standard deviations of all function classes when estimating the MSE in single and multiple runs, can be found in the appendix. For the normalization of the data we used
\begin{align*}
	\hat C & := \frac{C - \operatorname{min}(C)}{\operatorname{max}(C) - \operatorname{min}(C)} \\
	\hat E & := \frac{E - \operatorname{min}(E)}{\operatorname{max}(E) - \operatorname{min}(E)}
\end{align*}
and for the standardization we used
\begin{align*}
	\hat C & := \frac{C - \E[C]}{\sqrt{\Var[C]}} \\
	\hat E & := \frac{E - \E[E]}{\sqrt{\Var[E]}}.
\end{align*}

\paragraph{General Remark} Each evaluation was performed in the original data sets and in preprocessed versions where isolated points (low-density points) were removed. For the latter, we used the implementation and parameters from \cite{sgouritsa2015inference}, where a kernel density estimator with a Gaussian kernel is utilized to sort the data points according to their estimated densities. Then, data points with a density below a certain threshold (0.1 in our experiments) are removed from the data set. In this way, outliers should have a smaller impact on the performance. It also shows how sensitive each approach is to outliers. However, removing outliers might lead to an underestimation of the noise in a heavy tail noise distribution and is, therefore, not always the best choice as a preprocessing step. Note that CURE per default uses this preprocessing step, also in the original data. In all evaluations, we forced a decision by the algorithms, where in case of ANM the direction with the highest score of the independence test was taken.

Except for CURE, we averaged the performances of each method over 100 runs, where we uniformly sampled 500 data points for ANM and SVR if the data set contains more than 500 data points. For CURE, we only performed
four internal repetitions in the artificial and eight internal repetitions in the real-world data sets due to the high computational cost. As performance measure, we consider the accuracy of correctly identifying the true causal direction. Therefore, the accuracy is calculated according to
\begin{equation} \label{eq:accuracy}
	\text{accuracy} = \frac{\sum_{m = 1}^M w_m \delta_{\hat d_m, d_m}}{\sum_{m = 1}^M w_m },
\end{equation}
where $M$ is the number of data sets, $w_m$ the weight of data set $m$, $d_m$ the correct causal direction and $\hat d_m$ the inferred causal direction of the corresponding method. Note that we consider $w_m = 1$ for all artificial data sets, while we use different weights for the real-world data sets. Since we use all data points for SLOPE, IGCI and LiNGAM, these methods have a consistent performance over all runs. An overview of all utilized data sets with their corresponding number of cause-effect pairs and data samples can be found in Table \ref{tab:datasets} in the appendix.

\subsection{Artificial data}
For experiments with artificial data, we performed evaluations with simulated cause-effect pairs generated for a benchmark comparison in \cite{Mooijetal16}. Further, we generated additional pairs with linear, nonlinear invertible and nonlinear non-invertible functions where input and noise are strongly dependent.

\subsubsection{Simulated benchmark cause-effect pairs}
\label{sec:sim}
The work of \cite{Mooijetal16} provides simulated cause-effect pairs with randomly generated distributions and functional relationships under different conditions. As pointed out by \cite{Mooijetal16}, the scatter plots of these simulated data look similar to those of real-world data. We took the same data sets as used in \cite{Mooijetal16} and extend the reported results with an evaluation with SLOPE, CURE, LiNGAM, RECI and further provide results in the preprocessed data.

\begin{figure}[t!]
\subfigure[Original {\ttfamily SIM}]{
  \label{fig:SIMo}
  \centering
  \includegraphics[width=0.5\columnwidth]{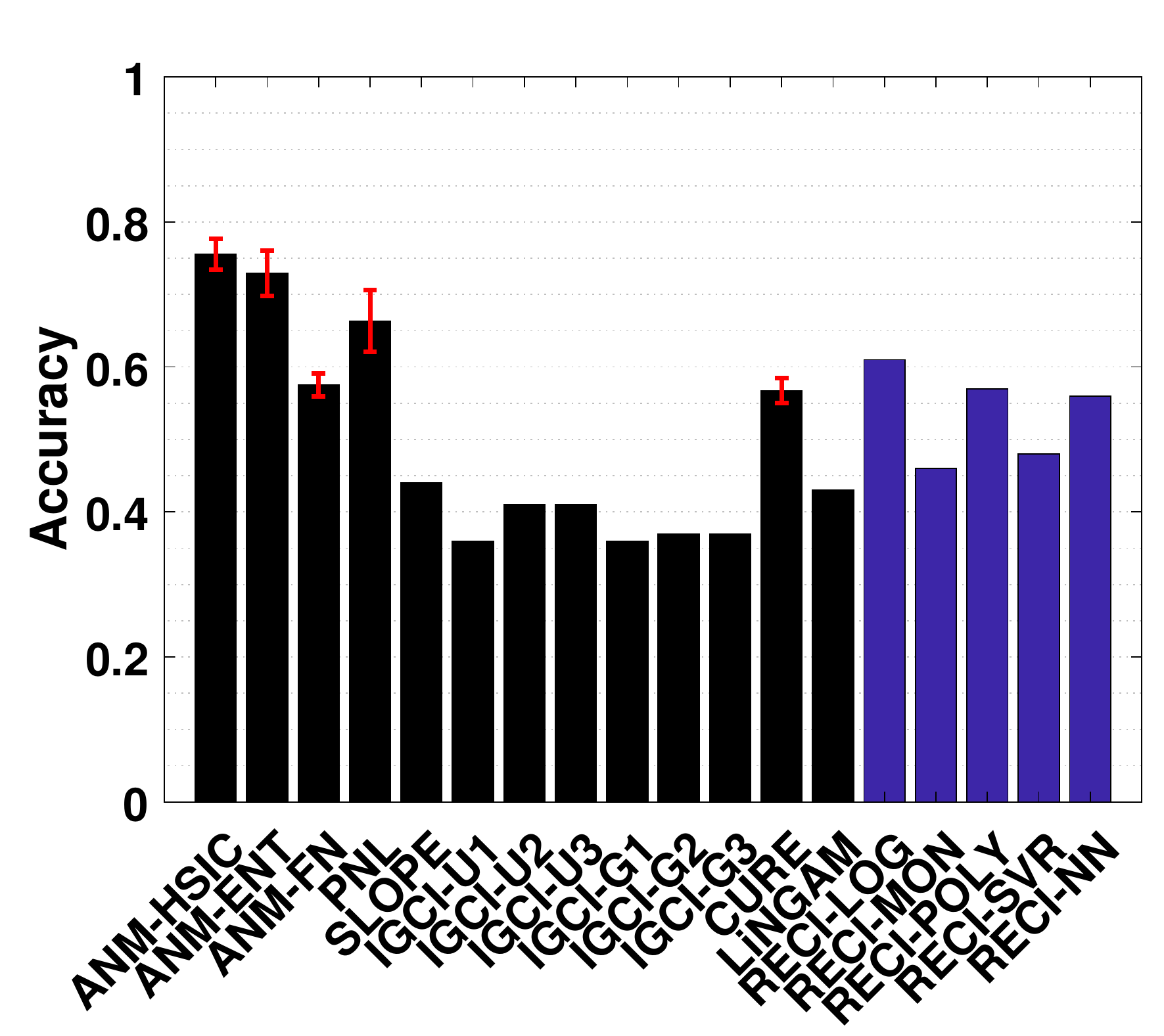}
}
\subfigure[Preprocessed {\ttfamily SIM}]{
  \label{fig:SIMp}
  \centering
  \includegraphics[width=0.5\columnwidth]{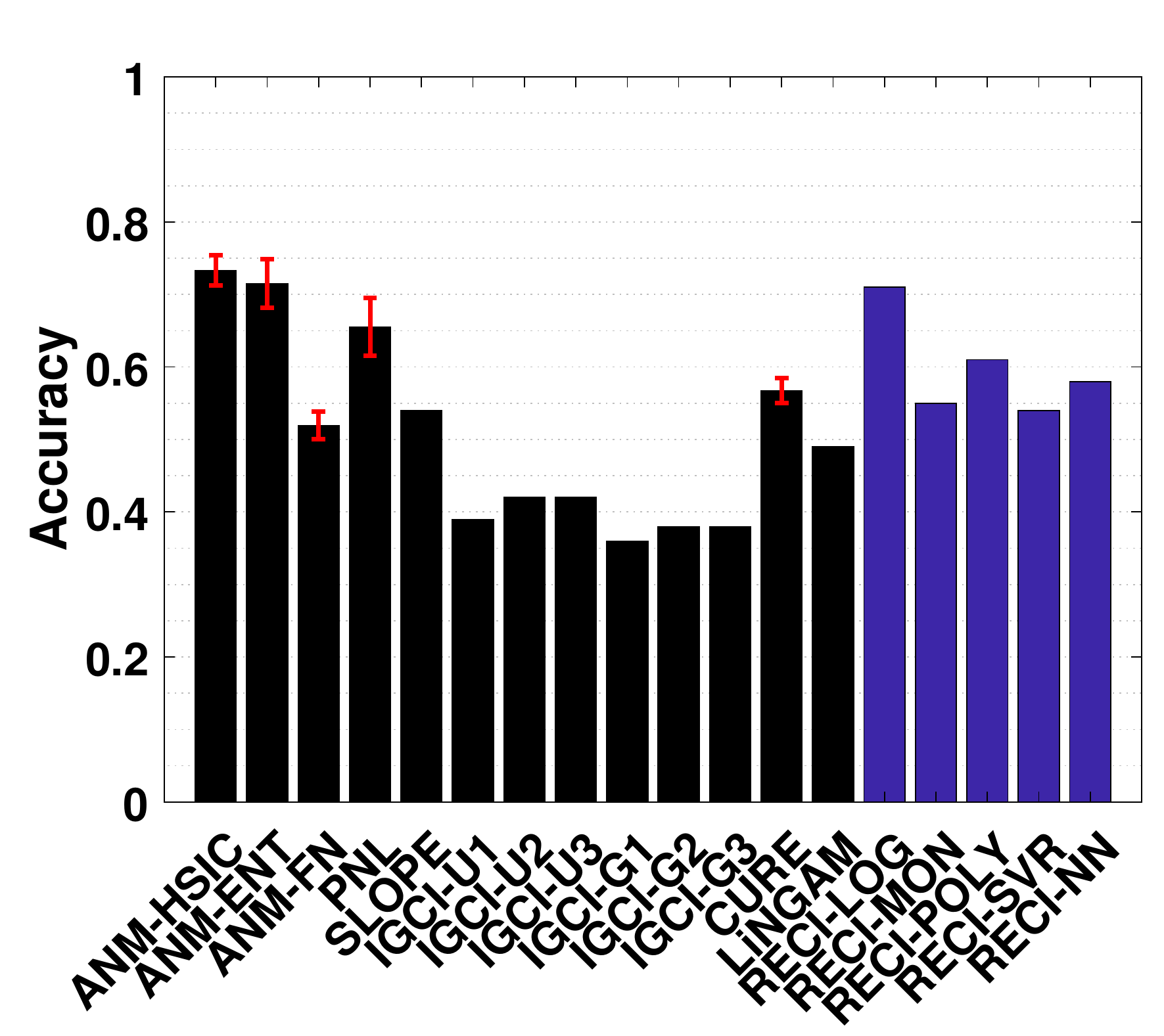}
}
\subfigure[Original {\ttfamily SIM-c}]{
  \label{fig:SIMCo}
  \centering
  \includegraphics[width=0.5\columnwidth]{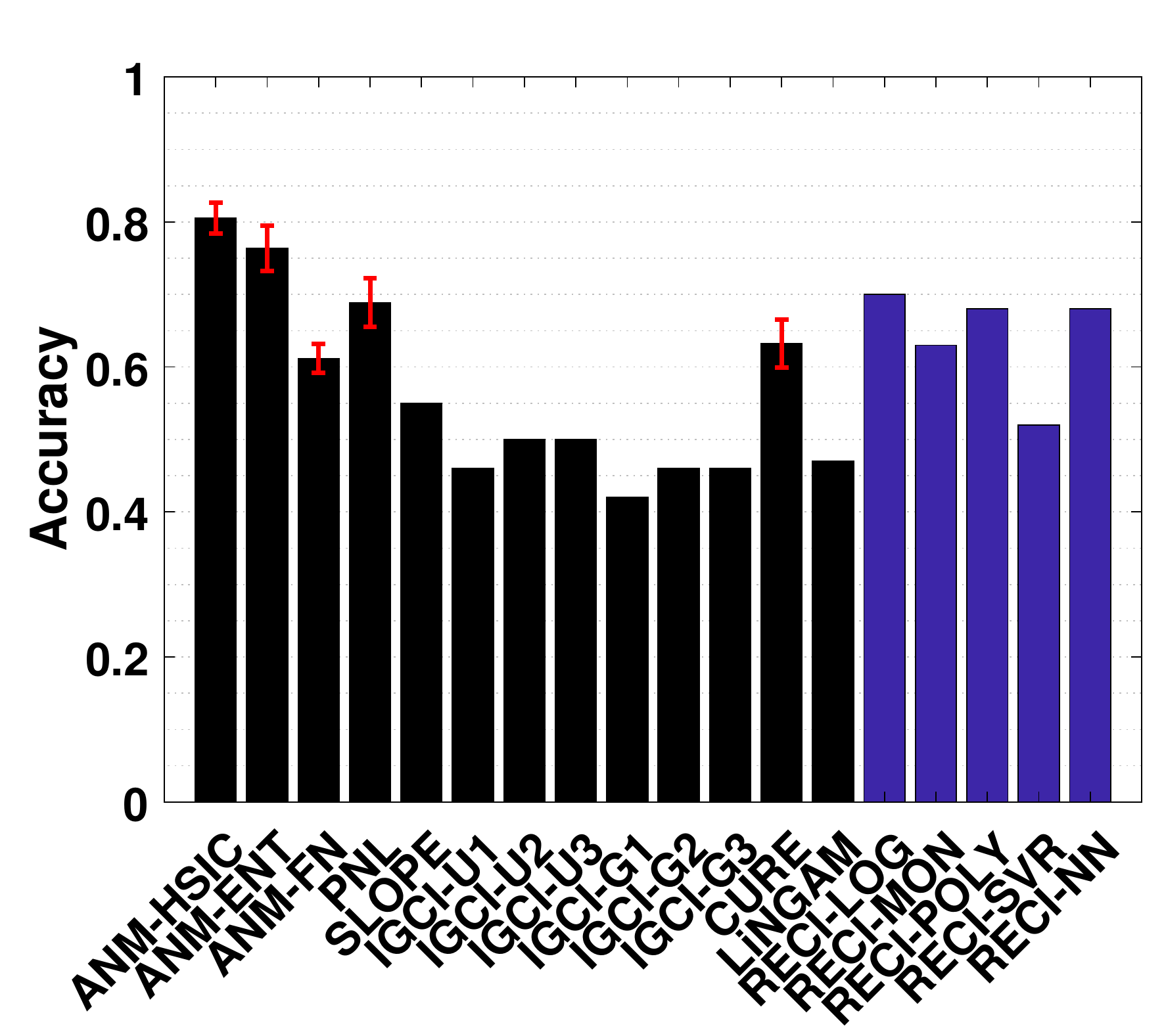}
}
\subfigure[Preprocessed {\ttfamily SIM-c}]{
  \label{fig:SIMCp}
  \centering
  \includegraphics[width=0.5\columnwidth]{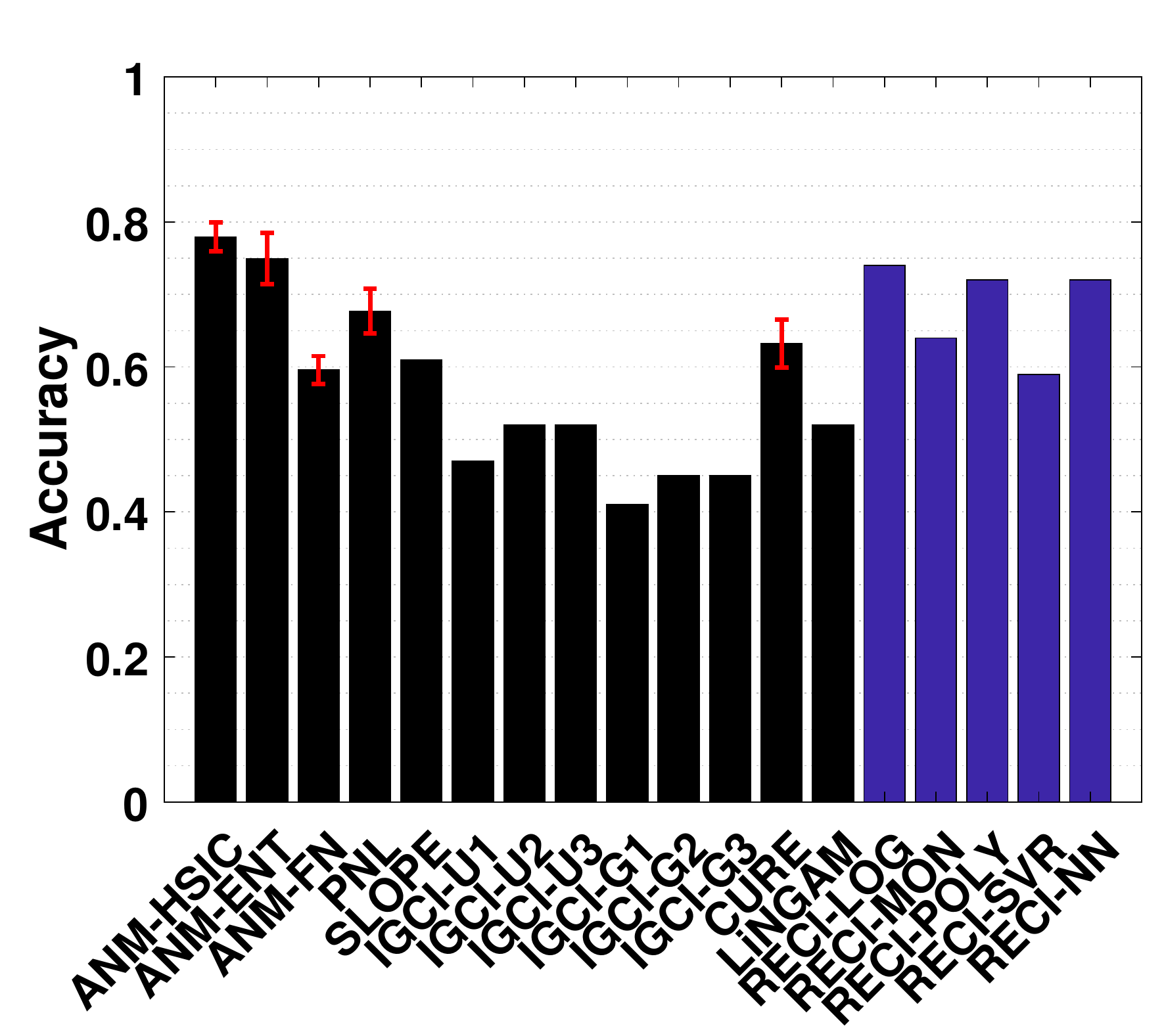}
}
\caption{Evaluation results of all methods in the {\ttfamily SIM} and {\ttfamily SIM-c} data sets. The figures on the left side show the results of the evaluations in the original data and on the right side the results in the preprocessed versions where low-density points were removed.}
\end{figure}

\begin{figure}[t!]
\subfigure[Original {\ttfamily SIM-ln}]{
  \label{fig:SIMLNo}
  \centering
  \includegraphics[width=0.5\columnwidth]{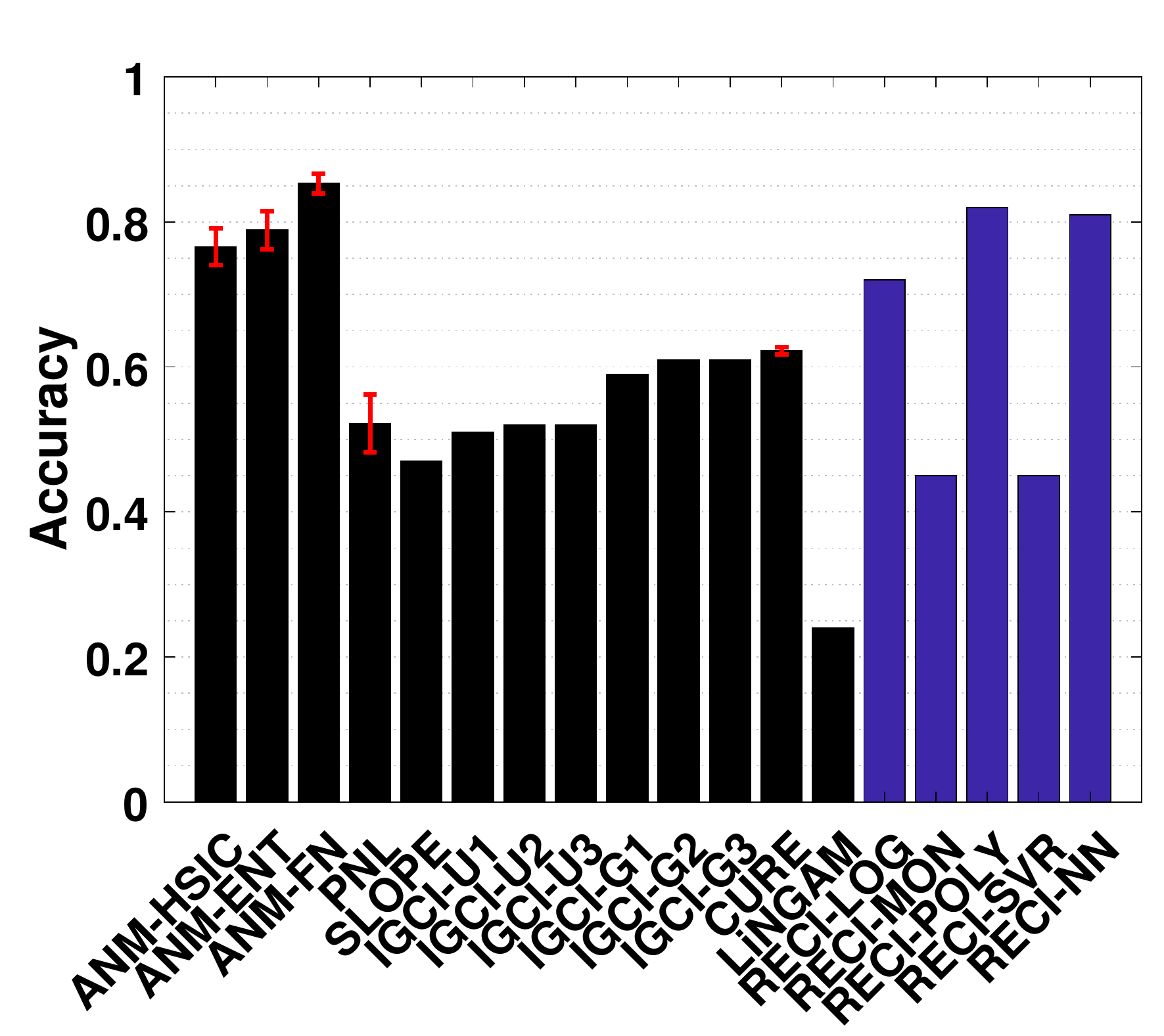}
}
\subfigure[Preprocessed {\ttfamily SIM-ln}]{
  \label{fig:SIMLNp}
  \centering
  \includegraphics[width=0.5\columnwidth]{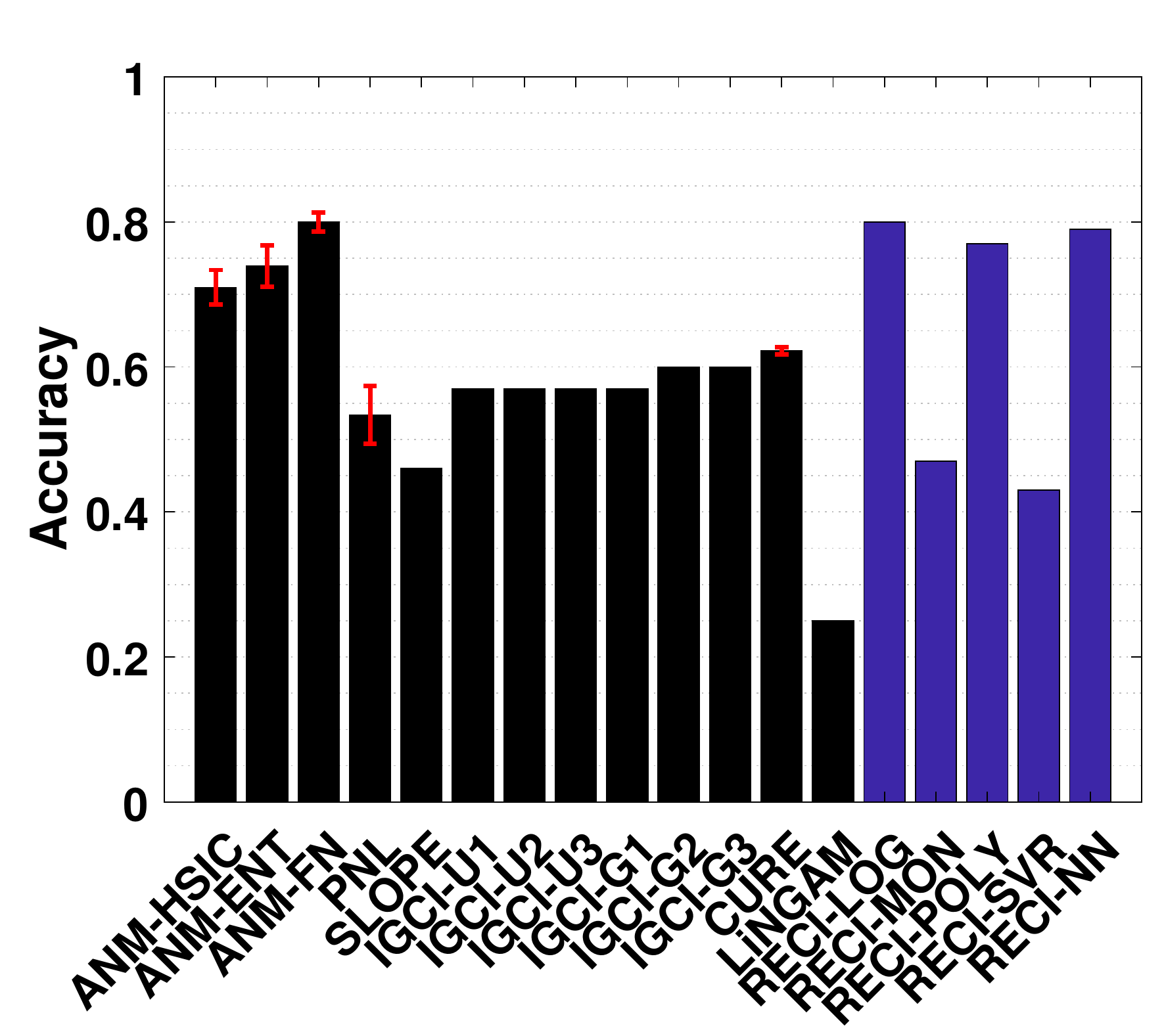}
}
\subfigure[Original {\ttfamily SIM-G}]{
  \label{fig:SIMGo}
  \centering
  \includegraphics[width=0.5\columnwidth]{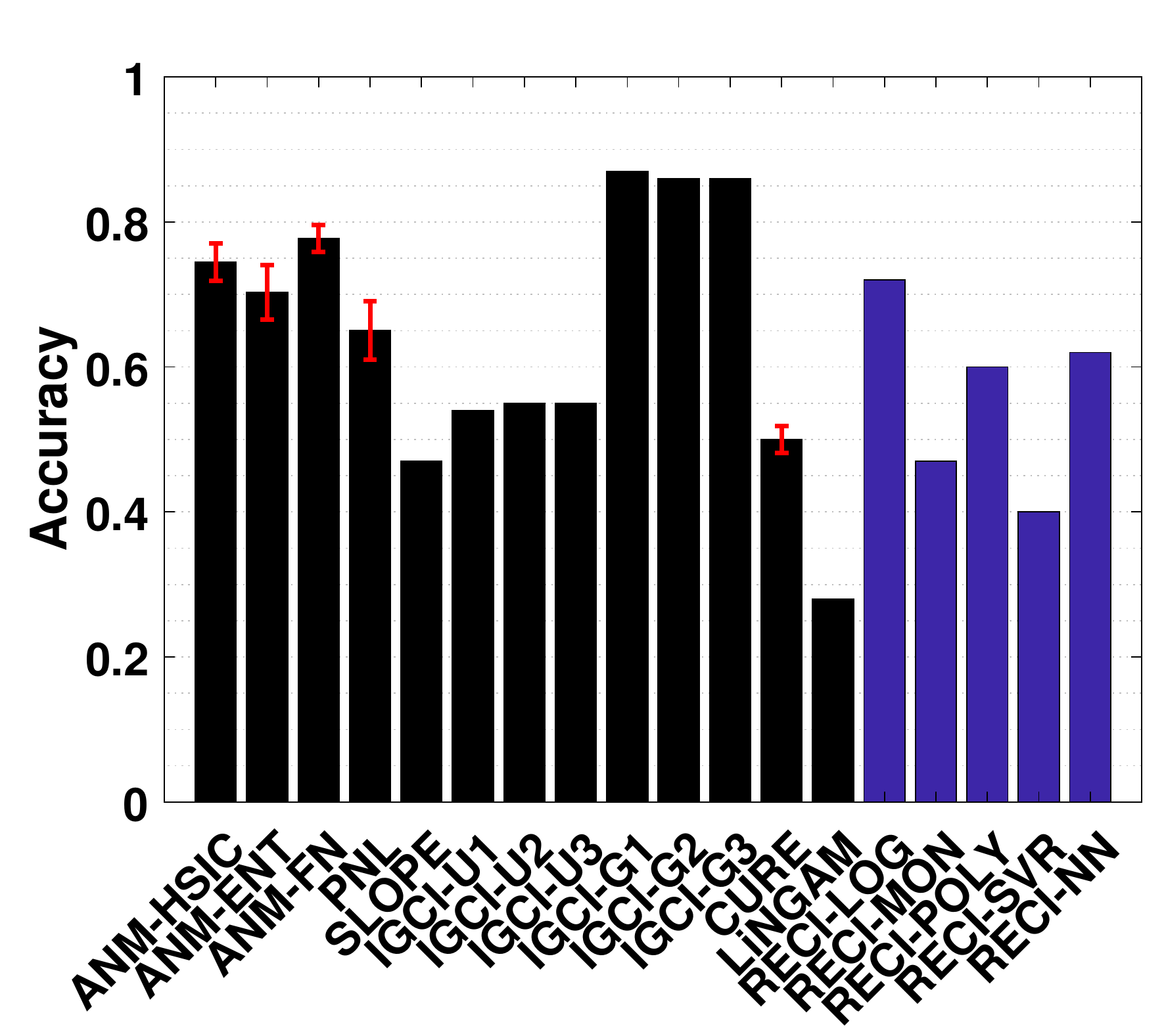}
}
\subfigure[Preprocessed {\ttfamily SIM-G}]{
  \label{fig:SIMGp}
  \centering
  \includegraphics[width=0.5\columnwidth]{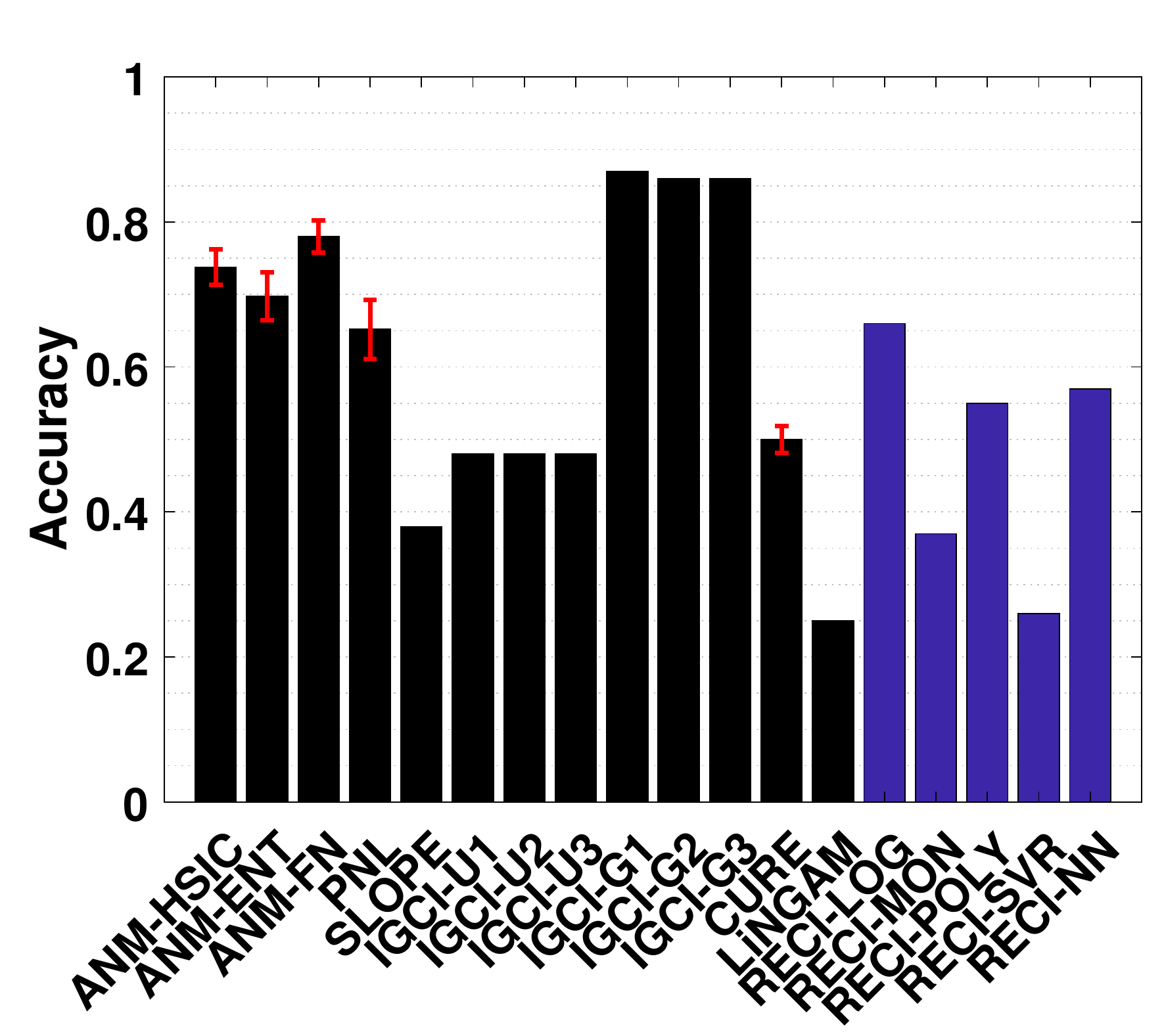}
}
\caption{Evaluation results of all methods in the {\ttfamily SIM-ln} and {\ttfamily SIM-G} data sets. The figures on the left side show the results of the evaluations in the original data and on the right side the results in the preprocessed versions where low-density points were removed.}
\end{figure}

The data sets are categorized into four different categories:
\begin{itemize}
	\item {\ttfamily SIM}: Pairs without confounders. The results are shown in Figure \ref{fig:SIMo}-\ref{fig:SIMp}
	\item {\ttfamily SIM-c}: A similar scenario as {\ttfamily SIM}, but with one additional confounder. The results are shown in Figure \ref{fig:SIMCo}-\ref{fig:SIMCp}
	\item {\ttfamily SIM-ln}: Pairs with low noise level without confounder. The results are shown in Figure \ref{fig:SIMLNo}-\ref{fig:SIMLNp}
	\item {\ttfamily SIM-G}: Pairs where the distributions of $C$ and $N$ are almost Gaussian without confounder. The results are shown in Figure \ref{fig:SIMGo}-\ref{fig:SIMGp}
\end{itemize}
The general form of the data generation process without confounder but with measurement noise\footnote{Note, however, that adding noise to the cause (as it is done here) can also be considered as a kind of confounding. Actually, $C'$ is the cause of $E$ in the below generating model, while the noisy version $C$ is not the cause of $E$. Accordingly, $C'$ is the hidden common cause of $C$ and $E$. Here we refer to the scenario as an unconfounded case with measurement noise as in \cite{Mooijetal16}.} is
\begin{align*}
	C' & \sim p_C, N \sim p_N \\
	N_C & \sim \mathcal{N}(0, \sigma_C), N_E \sim \mathcal{N}(0, \sigma_E) \\
	C & = C' + N_C \\
	E & = f_E(C', N) + N_E
\end{align*}
and with confounder
\begin{align*}
	C' & \sim p_C, N \sim p_N, Z \sim p_Z \\
	C'' & = f_C(C', Z) \\
	N_C & \sim \mathcal{N}(0, \sigma_C), N_E \sim \mathcal{N}(0, \sigma_E) \\
	C & = C'' + N_C \\
	E & = f_E(C'', Z, N) + N_E,
\end{align*}
where $N_C, N_E$ represent independent observational Gaussian noise and the variances $\sigma_C$ and $\sigma_E$ are chosen randomly with respect to the setting. Note that only $N_E$ is Gaussian, while the regression residual is non-Gaussian due to the nonlinearity of $f_E$ and non-Gaussianity of $N, Z$. Thus, the noise in {\ttfamily SIM}, {\ttfamily SIM-c} and {\ttfamily SIM-G} is non-Gaussian. More details can be found in Appendix C of \cite{Mooijetal16}.

Generally, ANM performs the best in all data sets. However, the difference between ANM and RECI, depending on the regression model, becomes smaller in the preprocessed data where isolated points were removed. According to the observed performances, removing these points seems to often improve the accuracy of RECI, but decrease the accuracy of ANM. In case of the preprocessed {\ttfamily SIM-G} data sets, the accuracy of RECI is decreased seeing that in these nearly Gaussian data the removal of low-density points leads to an underestimation of the noise distribution.

In all data sets, except for {\ttfamily SIM-G}, RECI always outperforms SLOPE, IGCI, CURE and LiNGAM if a simple logistic or polynomial function is utilized for the regression. However, in the {\ttfamily SIM-G} data set, our approach performs comparably poor, which could be explained by the violation of the assumption of a compact support. In this nearly Gaussian setting, IGCI performs the best with a Gaussian reference measure. However, we also evaluated RECI with standardized data in the {\ttfamily SIM-G} data sets, which is equivalent to a Gaussian reference measure for IGCI. A summary of the results can be found in Figures \ref{fig:addResults}(c)-\ref{fig:addResults}(d) in the appendix and more detailed results in Table \ref{tab:meanStandard} and Table \ref{tab:nonMeanStand} in the appendix. These results are significantly better than normalizing the data in this case. However, although our theorem only justifies a normalization, a different scaling, such as standardization, might be a reasonable alternative.

Even though Theorem 1 does not exclude cases of a high noise level, it makes a clear statement about low noise level. Therefore, as expected, RECI performs the best in {\ttfamily SIM-ln}, where the noise level is low. In all cases, LiNGAM performs very poorly due to the violations of its core assumptions. Surprisingly, although PNL is a generalization of ANM, we found that PNL performs generally worse than ANM, but better than SLOPE, CURE and LiNGAM.

ANM and RECI require a least-squares regression, but ANM additionally depends on an independence test, which can have a high computational cost and a big influence on the performance. Therefore, even though RECI does not outperform ANM, it represents a competitive alternative with a lower computational cost, depending on the regression model and MSE estimation. Also, seeing that RECI explicitly allows both cases, a dependency and an independency between $C$ and $N$ and ANM only the latter, it can be expected that RECI performs significantly better than ANM in cases where the dependency between $C$ and $N$ is strong. This is evaluated in Section \ref{sec:cepNoise}. In comparison with PNL, SLOPE, IGCI, LiNGAM and CURE, RECI outperforms in almost all data sets. Note that \cite{Mooijetal16} performed more extensive experiments and showed more comparisons with ANM and IGCI in these data sets, where additional parameter configurations were tested. However, they reported no results for the preprocessed data.

\subsubsection{Simulated cause-effect pairs with strong dependent noise}
\label{sec:cepNoise}
\begin{figure}[t!]
	\centering
 \includegraphics[width=0.6\columnwidth]{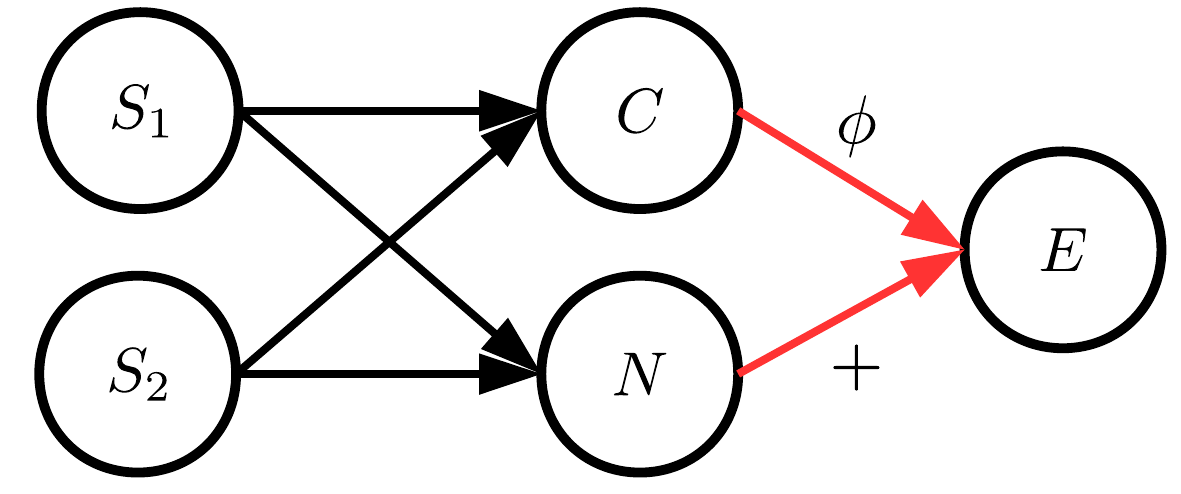}
\caption{The general structure of the data generation process where $C$ and $N$ are dependent. In order to achieve this, cause and noise consist of a mixture of two sources $S_1$ and $S_2$.}
  \label{fig:experiemts}
\end{figure}

Since the data sets of the evaluations in Section \ref{sec:sim} are generated by structural equations with independent noise variables, we additionally performed evaluations with artificial data sets where the input distribution and the noise distribution are strongly dependent. For this, we considered a similar data generation process as described in the work by \cite{6620}. We generated data with various cause and noise distributions, different functions and varying values for $\alpha \in [0, 1]$. In order to ensure a dependency between $C$ and $N$, we additionally introduced two unobserved source variables $S_1$ and $S_2$ that are randomly sampled from different distributions. Variables $C$ and $N$ then consist of a randomly weighted linear combination of $S_1$ and $S_2$. The general causal structure of these data sets is illustrated in Figure \ref{fig:experiemts}. Note that $S_1$ and $S_2$ can be seen as hidden confounders affecting both $C$ and $E$.

\begin{table}
\caption{All distributions $p_S$, functions $\phi$ and functions $f$ that were used for the generation of the {\ttfamily Linear}, {\ttfamily Non-invertible} and {\ttfamily Invertible} data sets. In case of the functions for {\ttfamily Non-invertible}, rescale$(X, -n, n)$ denotes a rescaling of the input data $X$ on $[-n, n]$. $GM_{\boldsymbol \mu, \boldsymbol \sigma}$ denotes a Gaussian mixture distribution with density $p_{GM_{\boldsymbol \mu, \boldsymbol \sigma}}(c) = \frac{1}{2}(\varphi(c|\mu_1, \sigma_1) + \varphi(c|\mu_2, \sigma_2))$ and Gaussian pdf $\varphi(c|\mu, \sigma)$.}
\begin{center}
\begin{tabular}{c} 
$p_S$ \\ 
\hline  
$U(0, 1)$ \\  
$\mathcal{N}(0, \sigma^2)$ \\  
$\mathcal{N}(0.5, \sigma^2)$ \\  
$\mathcal{N}(1, \sigma^2)$ \\  
$GM_{[0.3, 0.7]^\text{T}, [0.1, 0.1]^\text{T}}$ \\  
\end{tabular} 
\quad
\begin{tabular}{c|c}
Data set & $\phi(C)$ \\ 
 \hline
 {\ttfamily Linear} & $C$ \\
 {\ttfamily Invertible} & $s_5(C)$\\ 
 {\ttfamily Non-invertible} & $\text{rescale}(C, -2, 2)^2$ \\ 
 & $\text{rescale}(C, -2, 2)^4$\\ 
 & $\text{sin}(\text{rescale}(C, -2 \cdot \pi, 2 \cdot \pi))$\\ 
\end{tabular}
\quad
\begin{tabular}{c}
\\
$f(X)$ \\ 
 \hline
 $X$ \\
 $\text{exp}(X)$ \\ 
 $s_5(X)$\\ 
\end{tabular}
\end{center}
\label{table:distfuncs}
\end{table}

Apart from rather simple functions for $\phi$, \cite{6620} proposed to generate more general functions in the form of convex combinations of mixtures of cumulative Gaussian distribution functions $\psi(C|\mu_i, \sigma_i)$:
\begin{equation*}
	s_n(C) = \sum_{i = 1}^n \beta_i \psi(C|\mu_i, \sigma_i),
\end{equation*}
where $\beta_i, \mu_i \in [0, 1]$ and $\sigma_i \in [0, 0.1]$. For the experiments, we set $n = 5$ and chose the parameters of $s_5(C)$ randomly according to the uniform distribution. Note that $\psi(C|\mu_i, \sigma_i)$ is always monotonically increasing and thus $s_5(C)$ can have an arbitrary random shape while being monotonically increasing.

Cause-effect pairs were then generated in the following way
\begin{align*}
	w_1, w_2 & \sim U(0, 1) \\
	S_1, S_2 & \sim p_S \\
	S_1 & = S_1 - \E[S_1] \\	
	S_2 & = S_2 - \E[S_2] \\
	C' & = w_1 \cdot f_1(S_1) + (1 - w_1) \cdot f_2(S_2) \\
	N' & = w_2 \cdot f_3(S_1) + (1 - w_2) \cdot f_4(S_2) \\
	C & = \operatorname{normalize}(C') \\
	N & = \alpha \cdot \operatorname{standardize}(N') \\
	E & = \phi(C) + N,
\end{align*}
where the distributions of $S_1$ and $S_2$ and the functions $f_1, f_2, f_3, f_4$ were chosen randomly from $p_S$ and $f$ in Table \ref{table:distfuncs}, respectively. Note that $S_1$ and $S_2$ can follow different distributions. The choice of $\phi$ depends on the data set, where we differentiated between three data sets:
\begin{itemize}
	\item {\ttfamily Linear}: Only the identity function $\phi(C) = C$
	\item {\ttfamily Invertible}: Arbitrary invertible functions $\phi(C) = s_5(C)$
	\item {\ttfamily Non-invertible}: Functions that are not invertible on the respective domain
\end{itemize}

In total, we generated $100$ data sets for each value of parameter $\alpha$, which controls the amount of noise in the data. In each generated data set, we randomly chose different distributions and functions. For {\ttfamily Linear} and {\ttfamily Non-invertible} the step size of $\alpha$ is $0.1$ and for {\ttfamily Invertible} $0.025$. Here, we only performed one repetition on each data set for all algorithms. Figures \ref{fig:linear}-\ref{fig:noninv} summarize all results and Table \ref{table:noise} in the appendix shows the best performing functions and parameters of the different causal inference methods. Note that we omitted experiments with CURE in these data sets due to the high computational cost.

{\ttfamily Linear}: As expected, ANM, PNL, SLOPE, IGCI and RECI perform very poorly, since they require nonlinear data. In case of RECI, Theorem 1 states an equality of the MSE if the functional relation is linear and, thus, the causal direction can not be inferred. While LiNGAM performs well for $\alpha = 0.1$, and probably for smaller values of $\alpha$ too, the performance drops if $\alpha$ increases. The poor performances of LiNGAM and ANM can also be explained by the violation of its core assumption of an independence between cause and input.

{\ttfamily Invertible}: In this data set, IGCI performs quite well for small $\alpha$, since all assumptions approximately hold, but the performance decreases when the noise becomes stronger and violates the assumption of a deterministic relation. In case of RECI, we made a similar observation, but it performs much better than IGCI if $\alpha < 0.5$. Aside from the assumption of linear data for LiNGAM, the expected poor performance of LiNGAM and ANM can be explained by the violation of the independence assumption between $C$ and $N$. In contrast to the previous results, PNL performs significantly better in this setting than ANM, although, likewise LiNGAM and ANM, the independence assumption is violated.

{\ttfamily Non-invertible}: These results seem very interesting, since it supports the argument that the error asymmetry becomes even clearer if the function is not invertible due to an information loss of regressing in anticausal direction. Here, IGCI and SLOPE perform reasonably well, while ANM and LiNGAM perform even worse than a baseline of just guessing. Comparing ANM and PNL, PNL has a clear advantage, although the overall performance is only slightly around 60\% in average. The constant results of each method can be explained by the rather simple and similar choice of data generating functions.

While the cause and noise also have a dependency in the {\ttfamily SIM-c} data sets, the performance gap between ANM and RECI is vastly greater in {\ttfamily Invertible} and {\ttfamily Non-invertible} than in {\ttfamily SIM-c} due to a strong violation of the independent noise assumption. Therefore, RECI might perform better than ANM in cases with a strong dependency between cause and noise.

\begin{figure}[h!]
\subfigure[{\ttfamily Linear}]{
  \label{fig:linear}
  \centering
  \includegraphics[width=0.5\columnwidth]{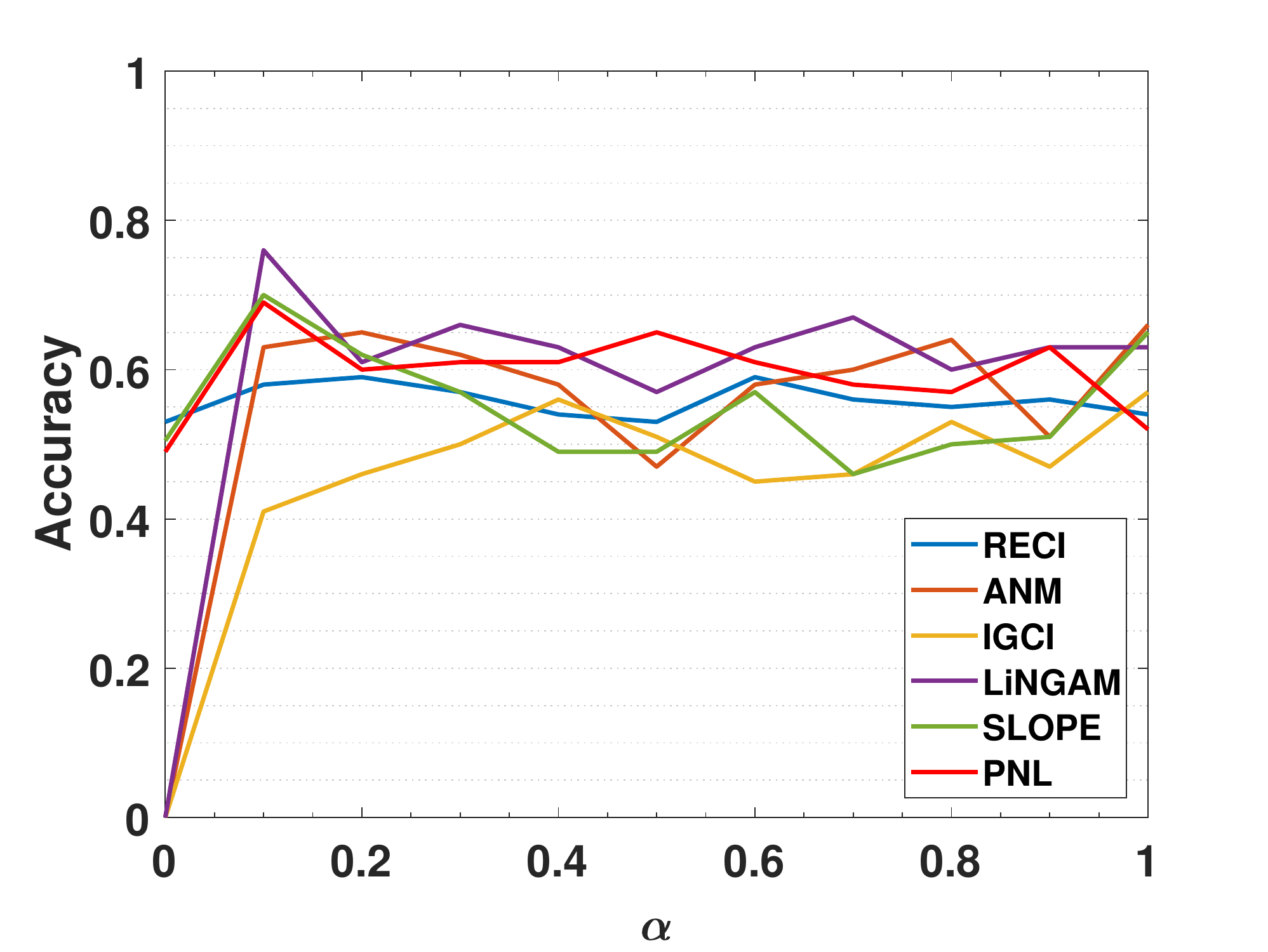}
}
\subfigure[{\ttfamily Invertible}]{
  \label{fig:inv}
  \centering
  \includegraphics[width=0.5\columnwidth]{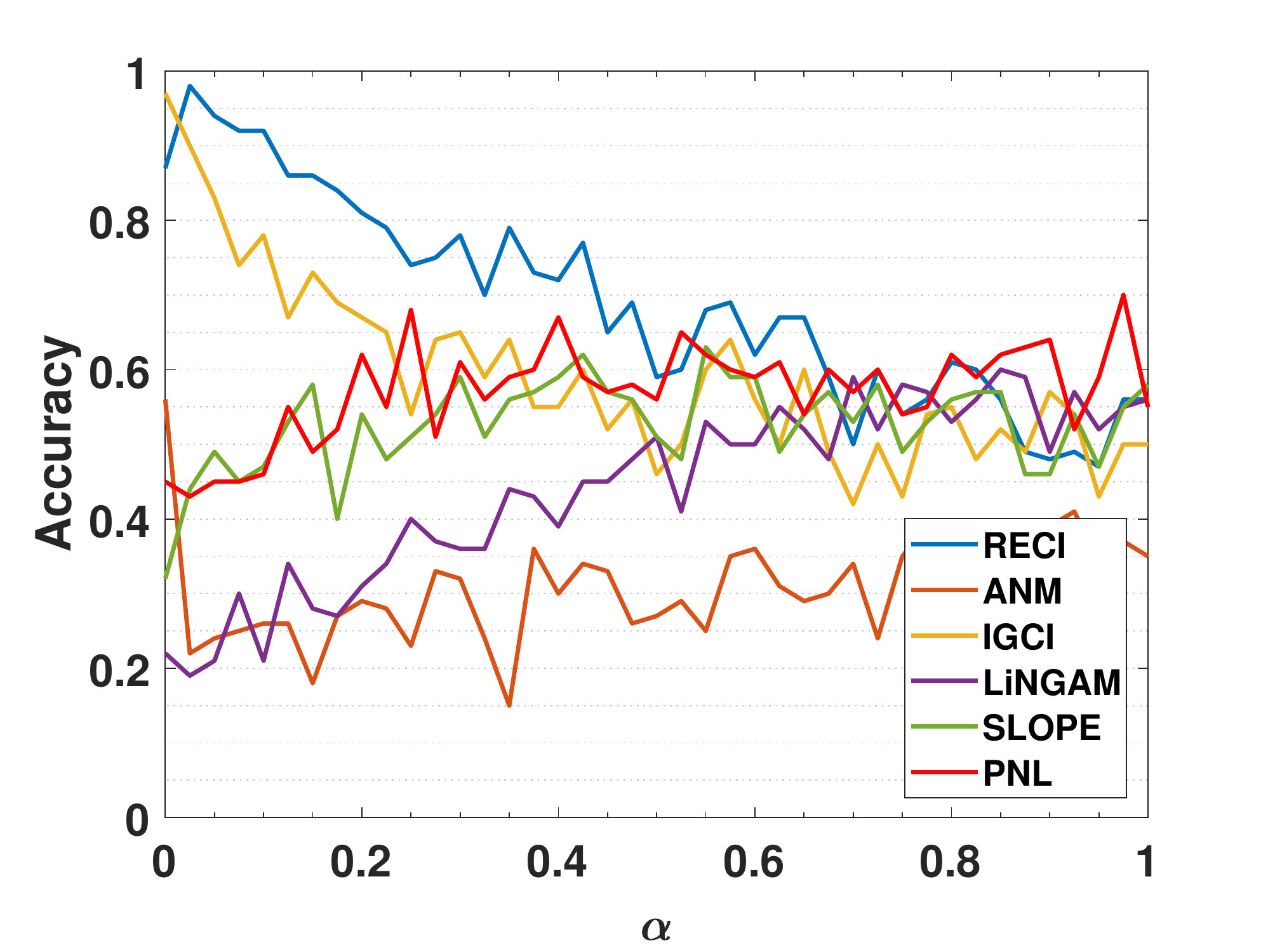}
}
\begin{center}
\subfigure[{\ttfamily Non-Invertible}]{
  \label{fig:noninv}
  \centering
  \includegraphics[width=0.5\columnwidth]{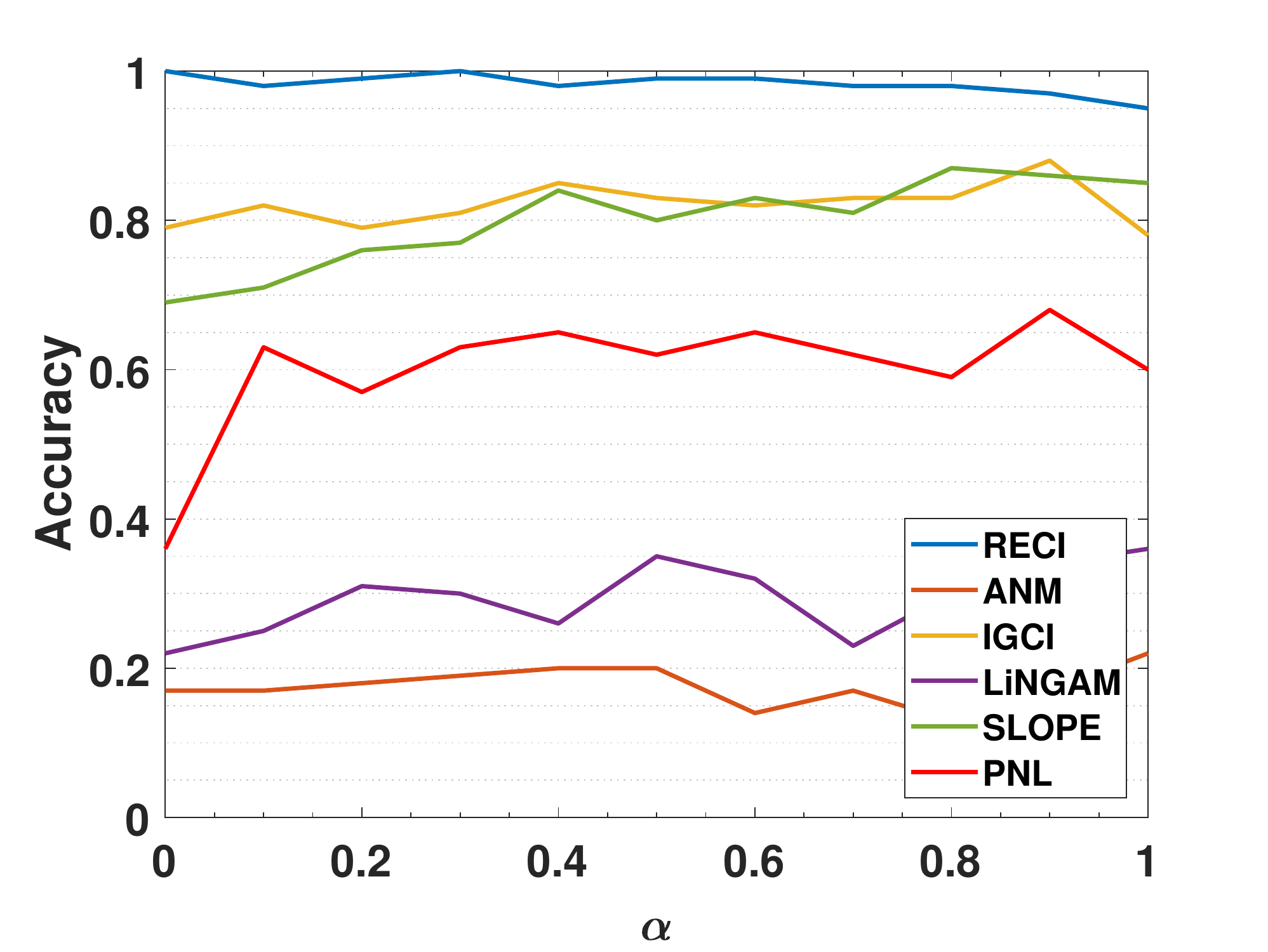}
}
\end{center}
\caption{Evaluation results of all methods in the {\ttfamily Linear}, {\ttfamily Invertible} and {\ttfamily Non-Invertible} data sets. The parameter $\alpha$ controls the amount of noise in the data.}
\end{figure}

\subsection{Real-world data}
\begin{figure}[h!]
\subfigure[Original {\ttfamily CEP}]{
  \label{fig:cepo}
  \centering
  \includegraphics[width=0.5\columnwidth]{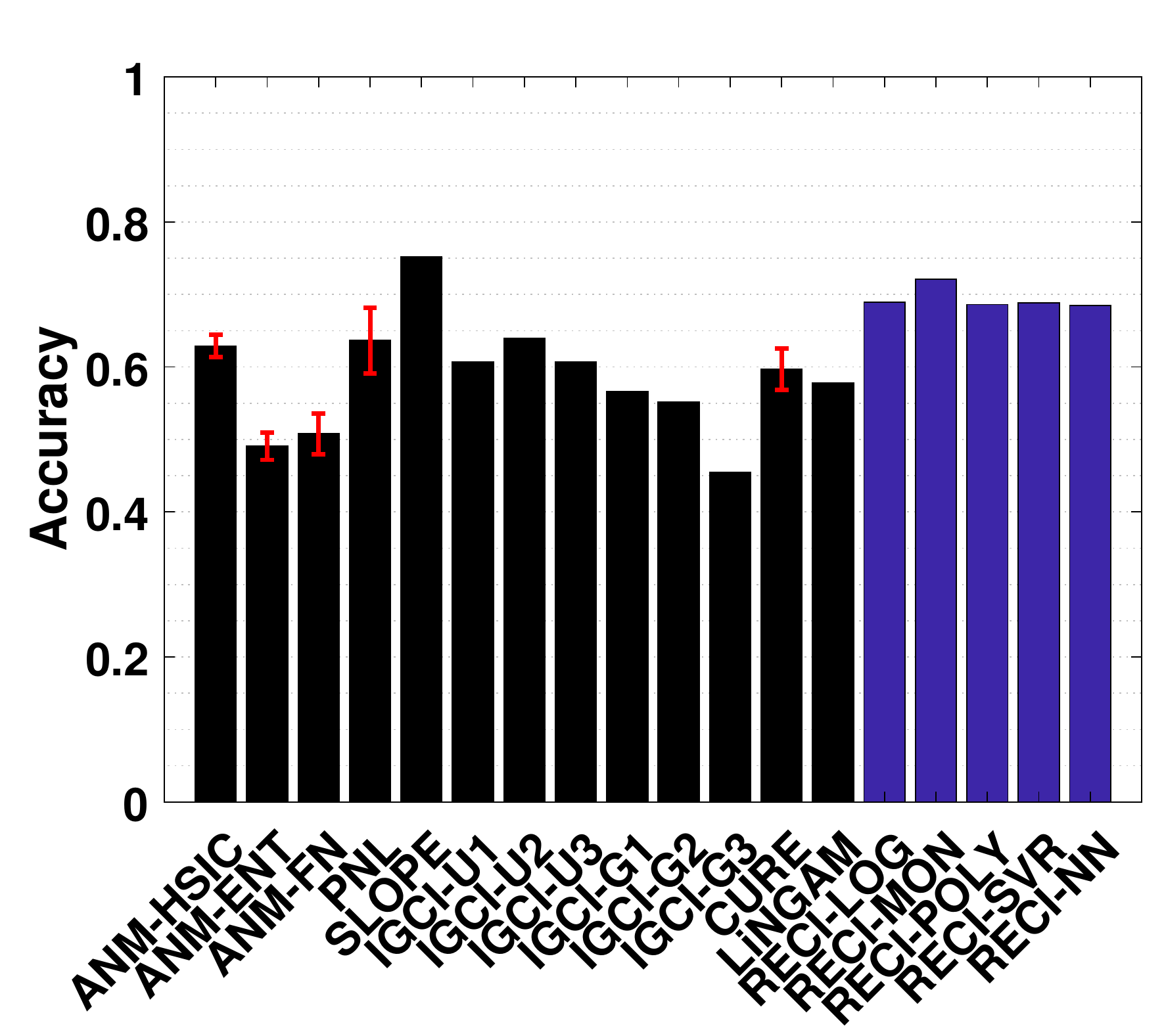}
}
\subfigure[Preprocessed {\ttfamily CEP}]{
  \label{fig:cepp}
  \centering
  \includegraphics[width=0.5\columnwidth]{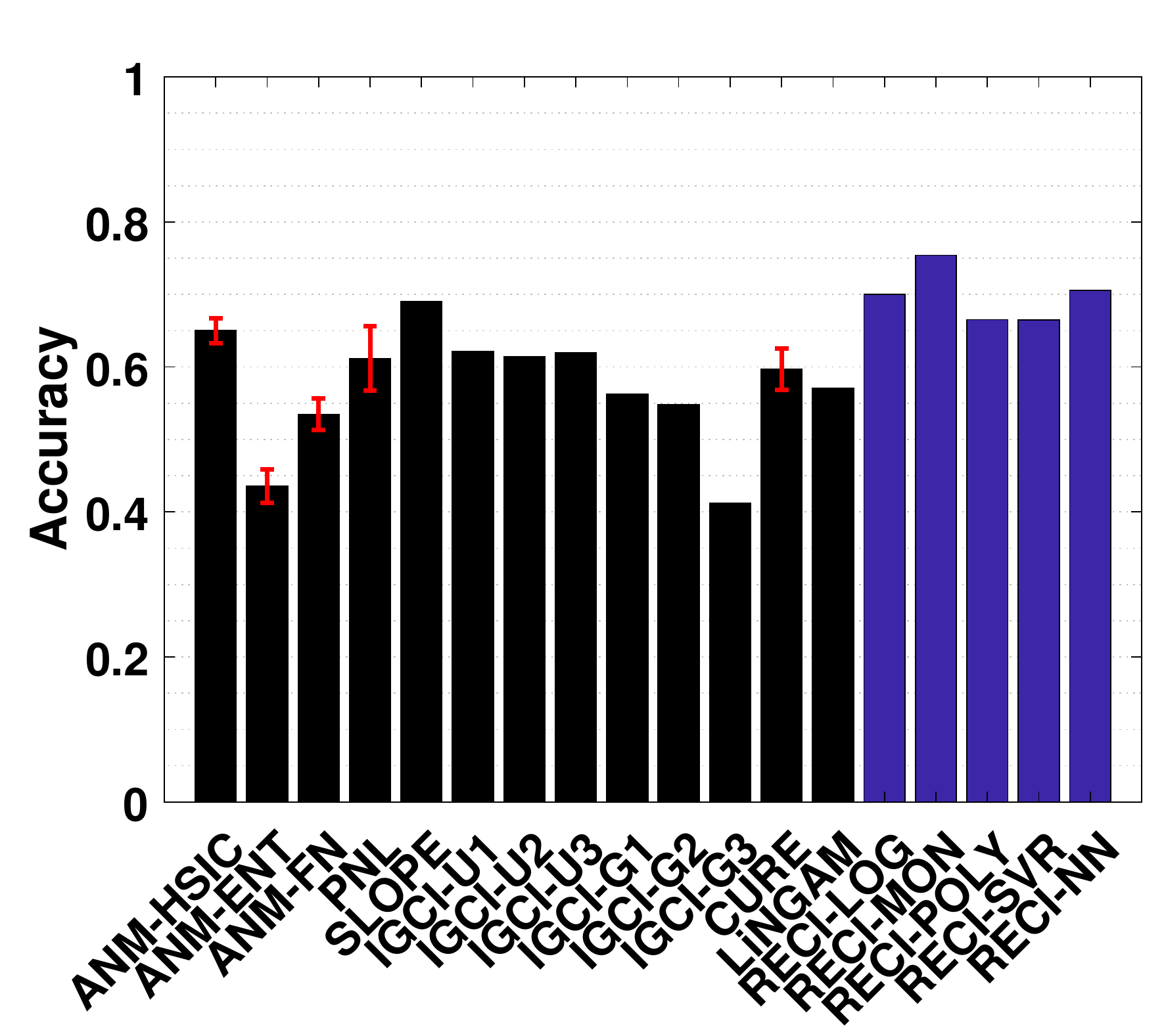}
}
\caption{Evaluation results of all methods in the real-world {\ttfamily CEP} data sets. The figure on the left side shows the result of the evaluations in the original data and on the right side the results in the preprocessed versions where low-density points were removed.}
\end{figure}

In real-world data, the true causal relationship generally requires expert knowledge and can still remain unclear in cases where randomized controlled experiments are not possible. For our evaluations, we considered the commonly used cause-effect pairs ({\ttfamily CEP}) benchmark data sets for inferring the causal direction in a bivariate setting. These benchmark data sets provided, at the time of these evaluations, 106 data sets with given cause and effect variables and can be found on https://webdav.tuebingen.mpg.de/cause-effect/. However, since we only consider a two variable problem, we omit six multivariate data sets, which leaves $100$ data sets for the evaluations. These data sets consist of a wide range of different scenarios, such as data from time dependent and independent physical processes, sociological and demographic studies or biological and medical observations. An extensive analysis and discussion about the causal relationship of the first 100 data sets can be found in the work by \cite{Mooijetal16}. Each data set comes with a corresponding weight determined by expert knowledge. This is because several data sets are too similar to consider them as independent examples, hence they get lower weights. Therefore, the weight $w_m$ in \eqref{eq:accuracy} depends on the corresponding data set. The evaluation setup is the same as for the artificial data sets, but we doubled the number of internal repetition of CURE to eight times in order to provide the same conditions as in \cite{sgouritsa2015inference}.

Figure \ref{fig:cepo} and \ref{fig:cepp} shows the results of the evaluations in the original and preprocessed data, respectively. In all cases, SLOPE and RECI perform significantly better than ANM, PNL, IGCI, CURE and LiNGAM. While SLOPE performs slightly better in the original data sets than RECI, RECI performs better overall in the preprocessed data.\footnote{The work of \cite{Mooijetal16} provides further evaluations of ANM and IGCI in the original {\ttfamily CEP} data set with parameter configurations that reached slightly higher accuracies than the presented results in this work. Regarding CURE, we had to use a simplified implementation due to the high computational cost, which did not perform as well as the results reported in \cite{sgouritsa2015inference}.} Further, the performance gap even increases in the preprocessed data with removed low-density points. Surprisingly, as Table \ref{tab:meanAll} in the appendix indicates, the simple shifted monomial function $a x^2 + c$ performs the best, even though it is very unlikely that this function is able to capture the true function $\phi$. We obtained similar observations in the artificial data sets, where the simple logistic function oftentimes performs the best.

In order to show that RECI still performs reasonably well under a different scaling, we also evaluated RECI in the real-world data set with standardized data. These results can be found summarized in Figures \ref{fig:addResults}(a)-\ref{fig:addResults}(b) in the appendix and more detailed in Table \ref{tab:meanStandard} and Table \ref{tab:nonMeanStand} in the appendix. While standardizing the data improves the performance in the {\ttfamily SIM-G} data, it slightly decreases the performance in the real-world data as compared to a normalization of the data, but still performs reasonably well. This shows some robustness with respect to a different scaling.

\subsection{Error ratio as rejection criterion}
\label{sec:errorRatio}
\begin{algorithm}[t]
\caption{Causal inference algorithm that uses \eqref{eq:errorRatio} as rejection criterion.}
\begin{algorithmic}
\Function{RECI}{$X$, $Y$, $t$} \Comment{$X$ and $Y$ are the observed data and $t \in [0, 1]$ is the confidence threshold for rejecting a decision.}
\State $(X, Y) \gets \text{RescaleData}(X, Y)$
\State $f \gets \text{FitModel}(X, Y)$ \Comment{Fit regression model $f\colon X \rightarrow Y$}
\State $g \gets \text{FitModel}(Y, X)$ \Comment{Fit regression model $g\colon Y \rightarrow X$ }
\State $\text{MSE}_{Y|X} \gets \text{MeanSquaredError}(f, X, Y)$
\State $\text{MSE}_{X|Y} \gets \text{MeanSquaredError}(g, Y, X)$
\State $\xi \gets 1 - \frac{\operatorname{min}(\text{MSE}_{X|Y}, \text{MSE}_{Y|X})}{\operatorname{max}(\text{MSE}_{X|Y}, \text{MSE}_{Y|X})}$
\If{$\xi \geq t$}
	\If{$\text{MSE}_{Y|X} < \text{MSE}_{X|Y}$}
		\State \Return \text{$X$ causes $Y$}
	\Else
		\State \Return \text{$Y$ causes $X$}
	\EndIf
\Else
	\State \Return \text{No decision}
\EndIf
\EndFunction
\end{algorithmic}
\end{algorithm}

\begin{figure}
\begin{center}
\subfigure[{\ttfamily CEP}-LOG]{
  \label{fig:decCEP}
  \centering
  \includegraphics[width=0.22\columnwidth]{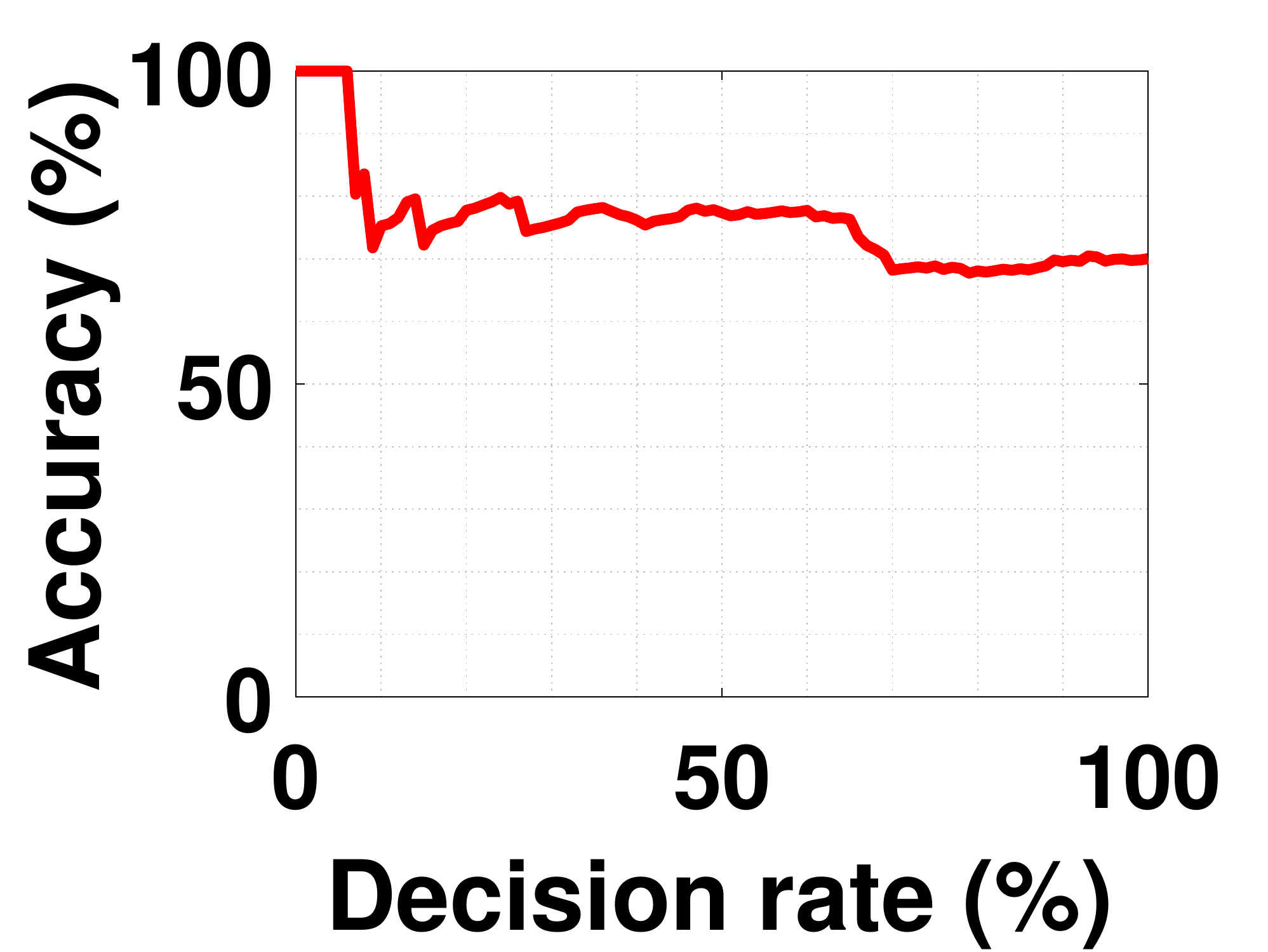}
}
\subfigure[{\ttfamily SIM}-LOG]{
  \label{fig:decSIM}
  \centering
  \includegraphics[width=0.22\columnwidth]{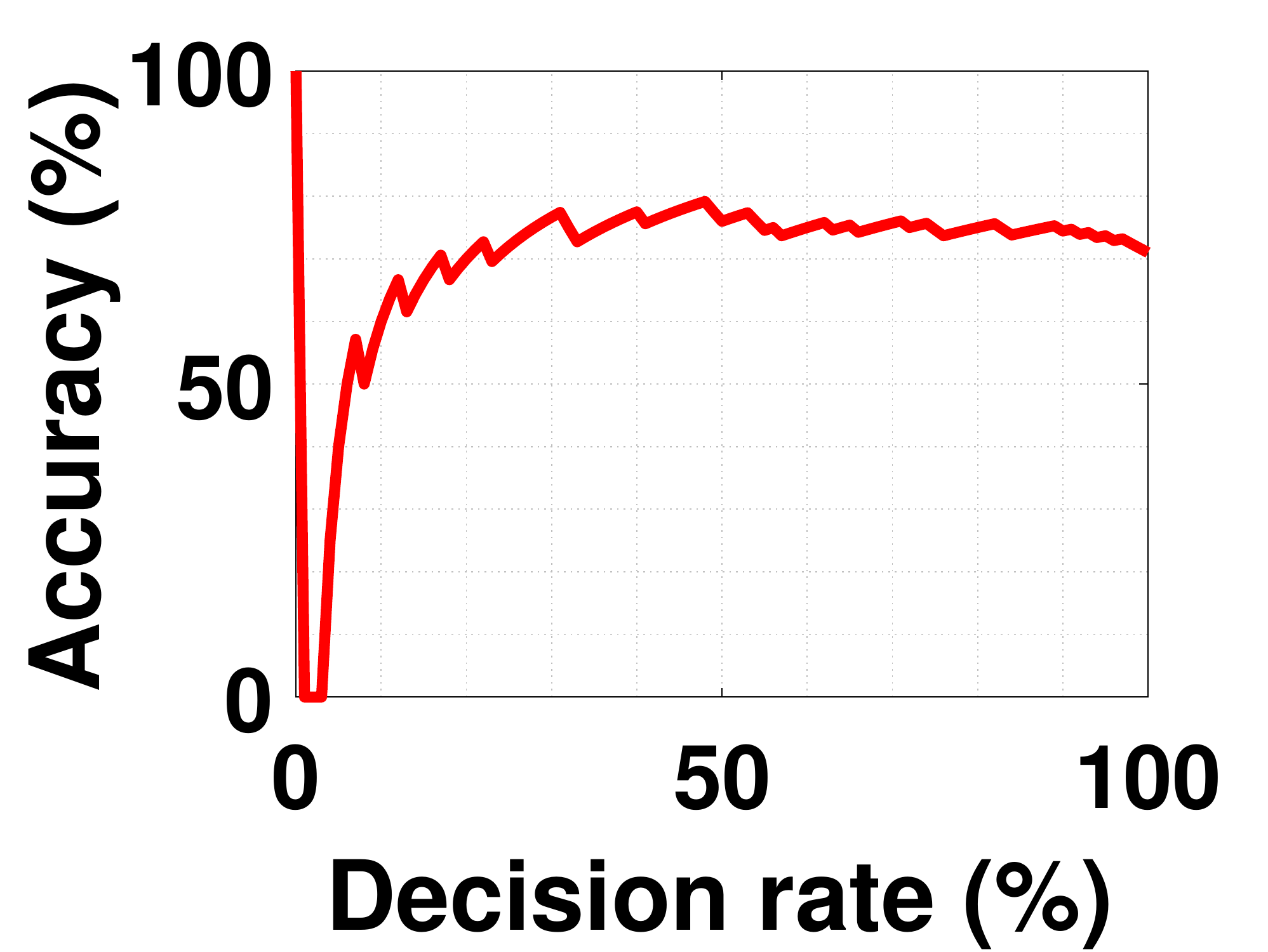}
}
\subfigure[{\ttfamily SIM-c}-NN]{
  \label{fig:decSIMC}
  \centering
  \includegraphics[width=0.22\columnwidth]{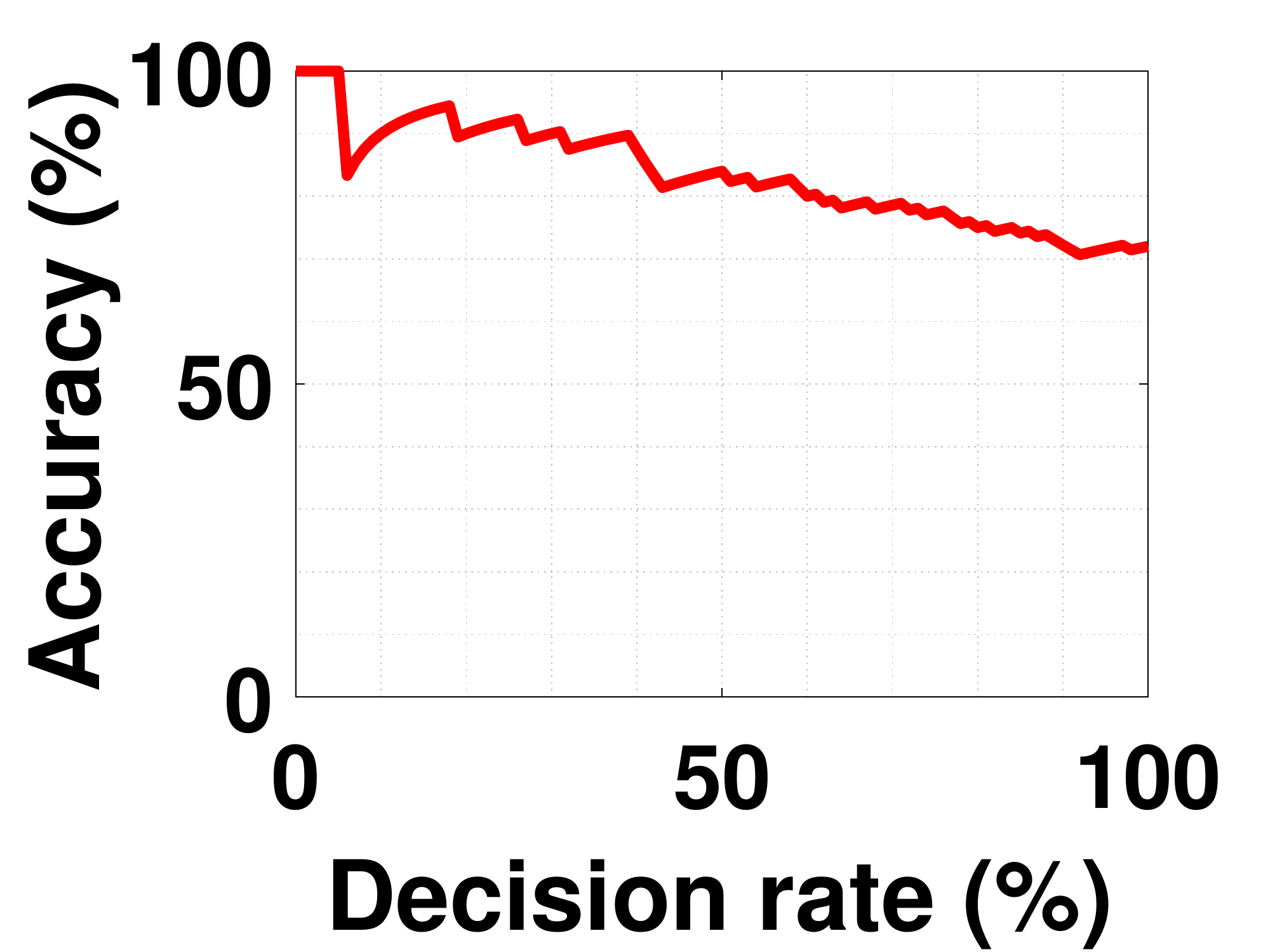}
}
\subfigure[{\ttfamily SIM-ln}-POLY]{
  \label{fig:decSIMLN}
  \centering
  \includegraphics[width=0.22\columnwidth]{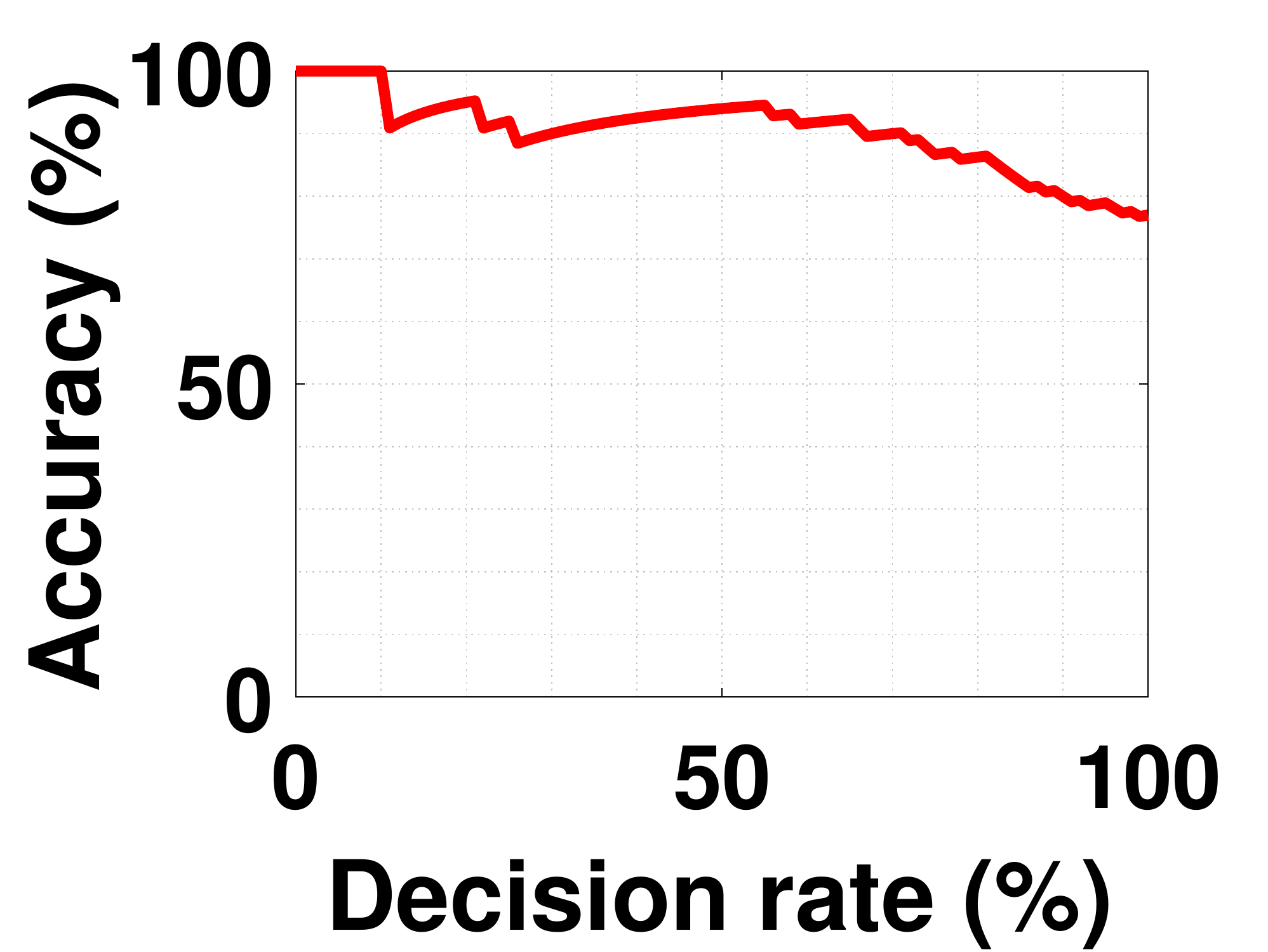}
}
\end{center}
\caption{The exemplary performance of RECI if a certain decisions rate is forced. Here, the decisions are ranked according to the confidence measure defined in \eqref{eq:errorRatio}.}
\end{figure}

It is not clear how a confidence measure for the decision of RECI can be defined. However, since Theorem \ref{thm:main} states that the correct causal direction has a smaller error, we evaluated the idea of utilizing the error ratio for a confidence measure in terms of:
\begin{equation}
\label{eq:errorRatio}
\text{confidence} = 1-\frac{\operatorname{min}(\E[\Var[X|Y]], \E[\Var[Y|X]])}{\operatorname{max}(\E[\Var[X|Y]], \E[\Var[Y|X]])},
\end{equation}
The idea is that, the smaller the error ratio, the higher the confidence of the decision, due to the large error difference. Note that we formulated the ratio inverse to Theorem \ref{thm:main} in order to get a value on $[0, 1]$. Algorithm 1 can be modified in a straight forward manner to utilize this confidence measure. The modification is summarized in Algorithm 2.\footnote{Algorithm 1 and Algorithm 2 are equivalent if $t = 0$.}

We re-evaluated the obtained results by considering only data sets where Algorithm 2 returns a decision with respect to a certain confidence threshold. Figures \ref{fig:decCEP}-\ref{fig:decSIMLN} show some examples of the performance of RECI if we use \eqref{eq:errorRatio} to rank the confidence of the decisions. A decision rate of $20\%$, for instance, indicates the performance when we only force a decision on $20\%$ of the data sets with the highest confidence. In this sense, we can get an idea of how useful the error ratio is as rejection criterion. While Figures \ref{fig:decCEP}, \ref{fig:decSIMC} and \ref{fig:decSIMLN} support the intuition that the smaller the error ratio, the higher the chance of a correct decision, Figure \ref{fig:decSIM} has a contradictive behavior. In Figure \ref{fig:decSIM}, it seems that a small error ratio (big error difference) is rather an indicator for an uncertain decision (probably caused by over- or underfitting issues). Therefore, we can not generally conclude that \eqref{eq:errorRatio} is a reliable confidence measure in all cases, but it seems to be a good heuristic approach in the majority of the cases. More plots can be found in the appendix, where Figure \ref{fig:fig2} shows the plots in the original data, Figure \ref{fig:fig3} in the preprocessed data and Figure \ref{fig:fig4} in the standardized data. Note that a deeper analysis of how to define an appropriate confidence measure for all settings is beyond the scope of this paper and we rather aim to provide some insights of utilizing the error ratio for this purpose.

\subsection{Run-time comparison}
\begin{table}
\caption{The average run-time in seconds of each method using all {\ttfamily CEP} data sets.}
\begin{center}
\renewcommand{\arraystretch}{1.2}
\begin{tabular}{c|c}
\textbf{Method} & \textbf{Time (s)} \\
\hline
ANM-HSIC & $2551.8 \pm 161.5$\\  
ANM-ENT &  $2463.1 \pm 39$ \\  
ANM-FN & $2427.6 \pm 40.2$\\  
PNL & $6019.7 \pm 118.49$\\ 
SLOPE & $1654.9 \pm 10.01$ \\  
IGCI-U1 & $0.0385 \pm 0.0028$ \\  
IGCI-U2 & $0.0384 \pm 0.0024$ \\  
IGCI-U3 & $0.5843 \pm 0.0327$\\  
IGCI-G1 & $0.0414 \pm 0.0025$\\   
\end{tabular} 
\hspace*{1.5 cm}
\renewcommand{\arraystretch}{1.2}
\begin{tabular}{c|c}
\textbf{Method} & \textbf{Time (s)} \\
\hline
IGCI-G2 & $0.0429 \pm 0.0028$\\
IGCI-G3 & $0.5866 \pm 0.0329$ \\  
CURE & $> 9999$ \\  
LiNGAM & $0.1459 \pm 0.0053$ \\  
RECI-LOG & $63.16 \pm 3.35$ \\  
RECI-MON & $4.65 \pm 0.28$ \\  
RECI-POLY & $2.78 \pm 0.1$ \\  
RECI-SVR & $87.33 \pm 33.43$ \\  
RECI-NN  & $46.62 \pm 0.28$\\  
\end{tabular}
\end{center}
\label{tab:runtime}
\end{table}
In order to have a brief overview and comparison of the run-times, we measured the execution durations of each method in the evaluation of the original {\ttfamily CEP} data sets. All measures were performed with a Intel Xeon E5-2695 v2 processor. Table \ref{tab:runtime} summarizes the measured run-times, where we stopped the time measurement of CURE after 9999 seconds. As the table indicates, IGCI is the fastest method, followed by LiNGAM and RECI. The ranking of ANM, PNL, SLOPE and RECI is not surprising; ANM and PNL need to evaluate the independence between input and residual on top of fitting a model. In case of SLOPE, multiple regression models need to be fitted depending on a certain criterion that requires to be evaluated. Therefore, by construction, RECI can be expected to be faster than ANM, PNL and SLOPE.

\subsection{Discussion}
Due to the greatly varying behavior and the choice of various optimization parameters, a clear rule of which regression function is the best choice for RECI remains an unclear and difficult problem. Overall, it seems that simple functions are better in capturing the error asymmetries than complex models. However, a clear explanation for this is still lacking. A possible reason for this might be that simple functions in causal direction already achieve a small error, while in anticausal direction, more complex models are required to achieve a small error. To justify speculative remarks of this kind raises deep questions about the foundations of causal inference. According to \eqref{eq:kolmo}, it is possible to conclude that the joint distribution has a simpler description in causal direction than in anticausal direction. Seeing this, a model selection based on the regression performance and model complexity considered in a dependent manner might further improve RECI's practical applicability. Regarding the removal of low-density points, the performance of methods that are based on the Gaussiantiy assumption, such as FN and IGCI with Gaussian reference measure, seems not to be influenced by the removal. On the other hand, the performance of HSIC, ENT, and IGCI with uniform measure is negatively affected, while the performance of LiNGAM and RECI increases. In case of RECI, this can be explained by a better estimation of the true MSE with respect to the regression function class.

Regarding the computational cost, we want to emphasize again that RECI, depending on the implementation details, can have a significantly lower computational cost than ANM, SLOPE and CURE, while providing comparable or even better results. Further, it can be easily implemented and applied.

\section{Conclusion}
\label{sec:conclusion}
We presented an approach for causal inference based on an asymmetry in the prediction error. Under the assumption of an independence among the data generating function, the noise, and the distribution of the cause, we proved (in the limit of small noise) that the conditional variance of predicting the cause by the effect is greater than the conditional variance of predicting the effect by the cause. For instance, in the example shown in Figure \ref{fig:income}, the regression error in the true causal direction is smaller than the error in the anticausal direction. In our work, the additive noise is not assumed to be independent of the cause (in contrast to so-called additive noise models). The stated theorem might also be interesting for other statistical applications.

We proposed an easily implementable and applicable algorithm, which we call RECI, that exploits this asymmetry for causal inference. The evaluations in artificial and real-world data sets show supporting results and leave room for further improvements. By construction, the performance of RECI depends on the regression method. According to our limited experience so far, regression with simple model classes (that tend to underfit the data) performs reasonably well. To clarify whether this happens because the conditional distributions tend to be simpler -- in a certain sense -- in causal direction than in anticausal direction has to be left for the future.

\section*{Acknowledgement}
This work was supported by JST CREST Grant Number JPMJCR1666 and JSPS KAKENHI Grant Number JP17K00305, Japan. The final publication is available at \url{https://doi.org/10.7717/peerj-cs.169}.

\bibliography{Literature}

\section*{Appendix}
\subsection*{Results of standardized data}
\begin{figure}[H]
	\subfigure[Org. {\ttfamily CEP} with standardized variables]{
		\label{fig:cepStand}
		\centering
		\includegraphics[width=0.5\columnwidth]{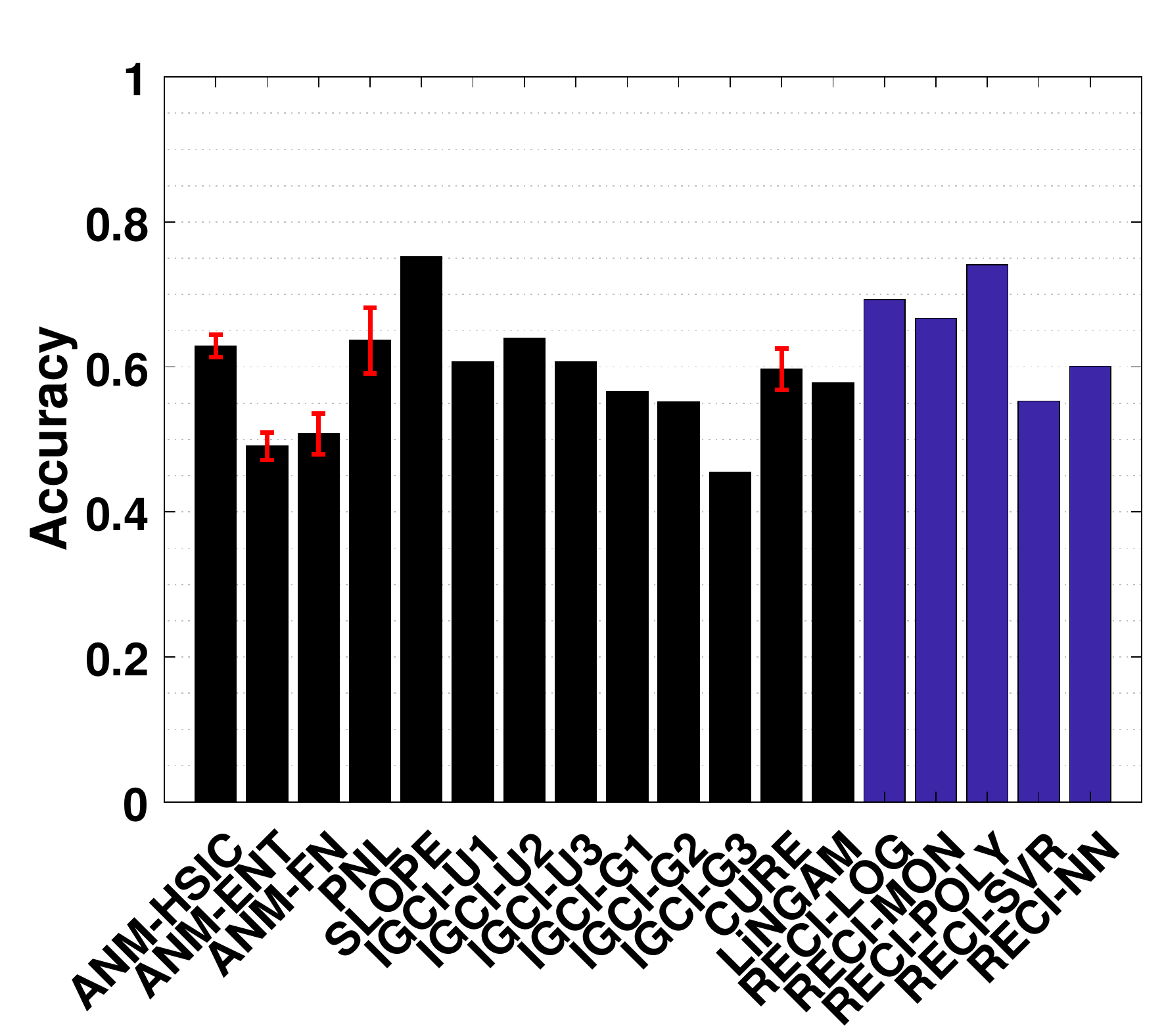}
	}
	\subfigure[Prep. {\ttfamily CEP} with standardized variables]{
		\label{fig:ceppStand}
		\centering
		\includegraphics[width=0.5\columnwidth]{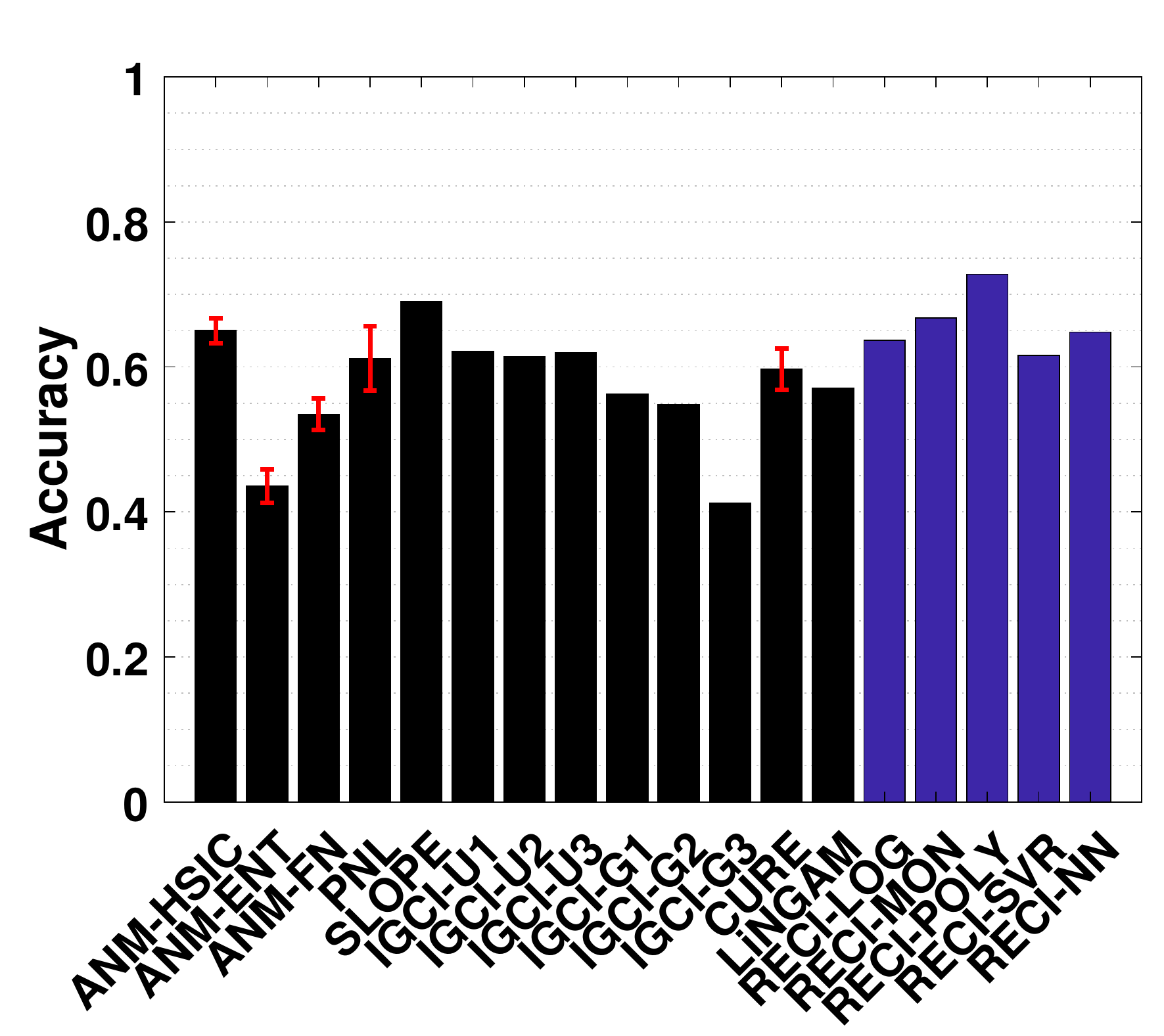}
	}
	\subfigure[Org. {\ttfamily SIM-G} with standardized variables]{
		\label{fig:simGStand}
		\centering
		\includegraphics[width=0.5\columnwidth]{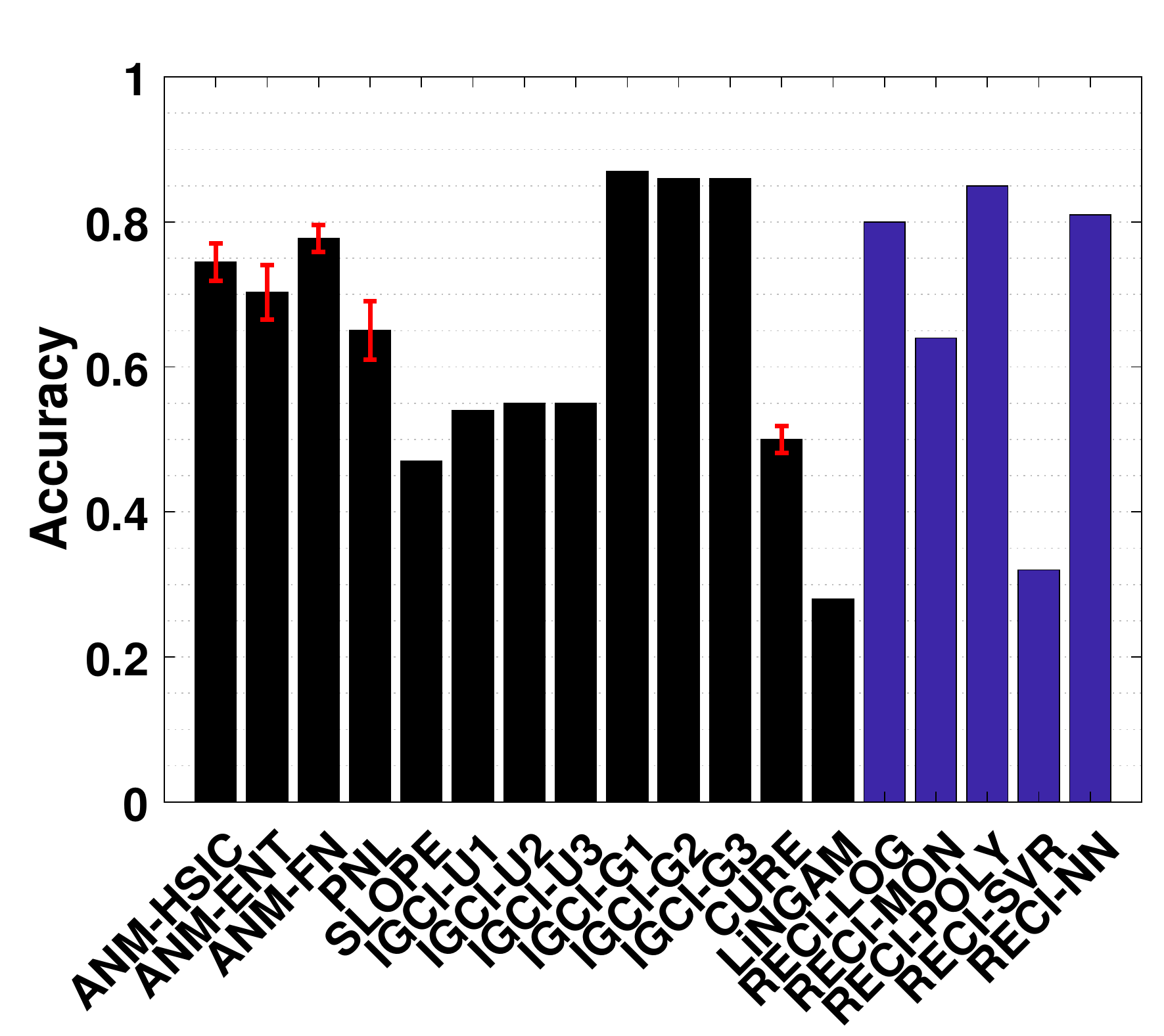}
	}
	\subfigure[Prep. {\ttfamily SIM-G} with standardized variables]{
		\label{fig:simGpStand}
		\centering
		\includegraphics[width=0.5\columnwidth]{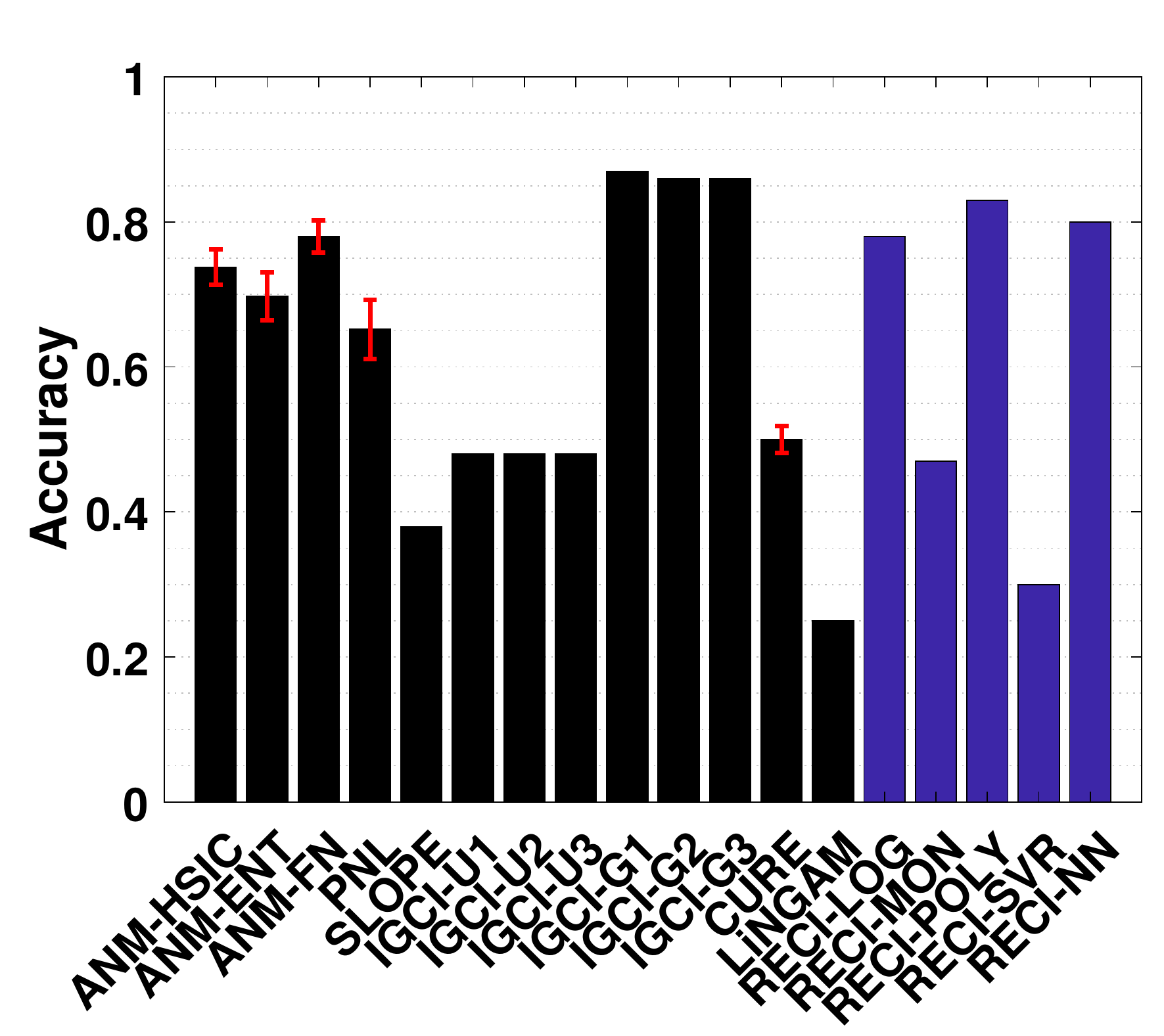}
	}
	\caption{Evaluation results of all methods in the real-world {\ttfamily CEP} and {\ttfamily SIM-G} data sets. Here, a standardization instead of a normalization was used for the scaling in RECI. The figure on the left side shows the results of the evaluations in the original data and on the right side the results in the preprocessed versions where low-density points were removed.\label{fig:addResults}}
\end{figure}

\subsection*{Tables}
\label{sec:tables}
\begin{sidewaystable}[h!]
	\caption{The performances of all used classes of regression functions for RECI in the artificial and real-world benchmark data sets when averaging the MSE over all runs.}
	\label{tab:meanAll}
	\scriptsize
	\centering
	\renewcommand{\arraystretch}{1.2}
	\begin{tabular}{c|c|c|c|c|c|c|c|c|c|c}
		& {\ttfamily CEP} & Prep. {\ttfamily CEP} & {\ttfamily SIM} & Prep. {\ttfamily SIM} & {\ttfamily SIM-c} & Prep. {\ttfamily SIM-c} & {\ttfamily SIM-ln} & Prep. {\ttfamily SIM-ln} & {\ttfamily SIM-G} & Prep. {\ttfamily SIM-G} \\ 
		\hline 
		LOG & $68.93\%$ & $70.02\%$ & $\mathbf{61\%}$ & $\mathbf{71\%}$ & $\mathbf{70\%}$ & $\mathbf{74\%}$ & $72\%$ & $\mathbf{80\%}$ & $\mathbf{72\%}$ & $\mathbf{66\%}$ \\ 
		\hline 
		$ax^2 + c$ & $\mathbf{72.13\%}$ & $\mathbf{75.41\%}$ & $43\%$ & $52\%$ & $57\%$ & $63\%$ & $44\%$ & $47\%$ & $44\%$ & $33\%$ \\ 
		\hline 
		$ax^3 + c$ & $69.07\%$ & $72.67\%$ & $45\%$ & $55\%$ & $54\%$ & $60\%$ & $45\%$ & $47\%$ & $43\%$ & $34\%$ \\ 
		\hline 
		$ax^4 + c$ & $68.84\%$ & $72.67\%$ & $42\%$ & $53\%$ & $56\%$ & $58\%$ & $45\%$ & $47\%$ & $43\%$ & $37\%$ \\ 
		\hline 
		$ax^5 + c$ & $69.92\%$ & $71.99\%$ & $43\%$ & $54\%$ & $60\%$ & $58\%$ & $44\%$ & $44\%$ & $45\%$ & $35\%$ \\ 
		\hline 
		$ax^6 + c$ & $68.84\%$ & $70.16\%$ & $45\%$ & $55\%$ & $61\%$ & $63\%$ & $40\%$ & $41\%$ & $45\%$ & $36\%$ \\ 
		\hline 
		$ax^7 + c$ & $67.75\%$ & $68.93\%$ & $46\%$ & $54\%$ & $62\%$ & $64\%$ & $42\%$ & $40\%$ & $47\%$ & $35\%$ \\ 
		\hline 
		$ax^8 + c$ & $67.61\%$ & $67.42\%$ & $46\%$ & $55\%$ & $62\%$ & $60\%$ & $38\%$ & $39\%$ & $44\%$ & $34\%$ \\ 
		\hline 
		$ax^9 + c$ & $66.56\%$ & $69.93\%$ & $46\%$ & $53\%$ & $63\%$ & $62\%$ & $37\%$ & $39\%$ & $45\%$ & $36\%$ \\ 
		\hline 
		$\sum_{i=0}^1 a_i x^i$ & $67.47\%$ & $66.51\%$ & $46\%$ & $54\%$ & $51\%$ & $58\%$ & $45\%$ & $43\%$ & $41\%$ & $26\%$ \\ 
		\hline 
		$\sum_{i=0}^2 a_i x^i$ & $67.7\%$ & $64.26\%$ & $53\%$ & $58\%$ & $65\%$ & $68\%$ & $69\%$ & $70\%$ & $54\%$ & $49\%$ \\ 
		\hline 
		$\sum_{i=0}^3 a_i x^i$ & $68.61\%$ & $64.49\%$ & $57\%$ & $59\%$ & $67\%$ & $71\%$ & $78\%$ & $76\%$ & $57\%$ & $52\%$ \\ 
		\hline 
		$\sum_{i=0}^4 a_i x^i$ & $66.63\%$ & $64.26\%$ & $57\%$ & $61\%$ & $68\%$ & $72\%$ & $80\%$ & $77\%$ & $60\%$ & $55\%$ \\ 
		\hline 
		$\sum_{i=0}^5 a_i x^i$ & $66.10\%$ & $66.54\%$ & $55\%$ & $59\%$ & $68\%$ & $71\%$ & $80\%$ & $75\%$ & $60\%$ & $53\%$ \\ 
		\hline 
		$\sum_{i=0}^6 a_i x^i$ & $66.08\%$ & $67\%$ & $52\%$ & $57\%$ & $68\%$ & $70\%$ & $\mathbf{82\%}$ & $76\%$ & $59\%$ & $54\%$ \\ 
		\hline 
		$\sum_{i=0}^7 a_i x^i$ & $66.36\%$ & $65.59\%$ & $57\%$ & $57\%$ & $68\%$ & $72\%$ & $81\%$ & $74\%$ & $60\%$ & $54\%$ \\ 
		\hline 
		$\sum_{i=0}^8 a_i x^i$ & $65.55\%$ & $64.26\%$ & $56\%$ & $59\%$ & $68\%$ & $69\%$ & $\mathbf{82\%}$ & $75\%$ & $60\%$ & $54\%$ \\ 
		\hline 
		$\sum_{i=0}^9 a_i x^i$ & $65.28\%$ & $63.34\%$ & $55\%$ & $61\%$ & $68\%$ & $69\%$ & $78\%$ & $71\%$ & $59\%$ & $55\%$ \\ 
		\hline 
		SVR & $68.84\%$ & $66.51\%$ & $48\%$ & $54\%$ & $52\%$ & $59\%$ & $45\%$ & $43\%$ & $40\%$ & $26\%$ \\ 
		\hline 
		NN 2 & $63.82\%$ & $63.34\%$ & $54\%$ & $57\%$ & $68\%$ & $72\%$ & $81\%$ & $79\%$ & $60\%$ & $56\%$ \\ 
		\hline 
		NN 5 & $68.48\%$ & $65.85\%$ & $56\%$ & $55\%$ & $67\%$ & $70\%$ & $79\%$ & $76\%$ & $61\%$ & $56\%$ \\ 
		\hline 
		NN 10 & $64.82\%$ & $66.63\%$ & $54\%$ & $56\%$ & $68\%$ & $69\%$ & $80\%$ & $70\%$ & $62\%$ & $56\%$ \\ 
		\hline 
		NN 20 & $67.81\%$ & $70.53\%$ & $54\%$ & $58\%$ & $66\%$ & $70\%$ & $81\%$ & $72\%$ & $60\%$ & $57\%$ \\ 
		\hline 
		NN 2-4 & $64.28\%$ & $64.12\%$ & $55\%$ & $58\%$ & $68\%$ & $68\%$ & $78\%$ & $76\%$ & $60\%$ & $54\%$ \\ 
		\hline 
		NN 4-8 & $67.47\%$ & $65.85\%$ & $52\%$ & $56\%$ & $68\%$ & $68\%$ & $77\%$ & $71\%$ & $61\%$ & $55\%$ \\ 
	\end{tabular}
\end{sidewaystable}

\begin{sidewaystable}
	\caption{The performances of all used classes of regression functions for RECI in the artificial and real-world benchmark data sets when not averaging the MSE.}
	\scriptsize
	\centering
	\renewcommand{\arraystretch}{1.2}
	\begin{tabular}{c|c|c|c|c|c|c|c|c|c|c}
		& {\ttfamily CEP} & Prep. {\ttfamily CEP} & {\ttfamily SIM} & Prep. {\ttfamily SIM} & {\ttfamily SIM-c} & Prep. {\ttfamily SIM-c} & {\ttfamily SIM-ln} & Prep. {\ttfamily SIM-ln} & {\ttfamily SIM-G} & Prep. {\ttfamily SIM-G} \\ 
		\hline 
		LOG & $63.66\% \pm 3.48$ & $64.34\% \pm 3.57$ & $57.62\% \pm 3.57$ & $60.83\% \pm 3.83$ & $65.02\% \pm 3.82$ & $67.52\% \pm 3.3$ & $71.74\% \pm 4.02$ & $72.71\% \pm 3.49$ & $\mathbf{67.37\%} \pm 4.15$ & $\mathbf{61.95\%} \pm 4.35$ \\ 
		\hline 
		$ax^2 + c$ & $\mathbf{70.73\%} \pm 1.55$ & $\mathbf{72.90\%} \pm 0.00$ & $43.84\% \pm 1.17$ & $52\% \pm 0.00$ & $52.74\% \pm 1.07$ & $60\% \pm 0.00$ & $44.45\% \pm 1.29$ & $47\% \pm 0.00$ & $43.74\% \pm 0.91$ & $33.56\% \pm 1.78$ \\ 
		\hline 
		$ax^3 + c$ & $70.12\% \pm 1.71$ & $72.67\% \pm 0.00$ & $45.47\% \pm 1.78$ & $54\% \pm 0.00$ & $53.37\% \pm 0.94$ & $58\% \pm 0.00$ & $45.08\% \pm 1.28$ & $47.63\% \pm 1.13$ & $42.77\% \pm 1.72$ & $34.80\% \pm 1.69$ \\ 
		\hline 
		$ax^4 + c$ & $69.39\% \pm 1.39$ & $72.67\% \pm 0.00$ & $43.92\% \pm 0.01$ & $53\% \pm 0.00$ & $56.42\% \pm 1.44$ & $58.52\% \pm 1.87$ & $45\% \pm 0.00$ & $47\% \pm 0.00$ & $44.11\% \pm 1.62$ & $34.54\% \pm 1.78$ \\ 
		\hline 
		$ax^5 + c$ & $69.20\% \pm 1.82$ & $70.5\% \pm 2.18$ & $43.66\% \pm 0.02$ & $54\% \pm 0.00$ & $58.11\% \pm 1.78$ & $58.54\% \pm 1.17$ & $43.76\% \pm 1.66$ & $43.22\% \pm 0.02$ & $44.59\% \pm 1.99$ & $35.47\% \pm 2.39$ \\ 
		\hline 
		$ax^6 + c$ & $68.84\% \pm 0.00$ & $69.63\% \pm 2.47$ & $43.68\% \pm 0.02$ & $55\% \pm 0.00$ & $59\% \pm 0.00$ & $62.99\% \pm 0.1$ & $41.42\% \pm 1.83$ & $43.51\% \pm 0.41$ & $44\% \pm 0.00$ & $36.78\% \pm 2.11$ \\ 
		\hline 
		$ax^7 + c$ & $67.35\% \pm 2.54$ & $68.01\% \pm 2.61$ & $44.07\% \pm 0.12$ & $53.92\% \pm 1.73$ & $60\% \pm 0.00$ & $64\% \pm 0.00$ & $41.76\% \pm 2.31$ & $42.17\% \pm 0.11$ & $44\% \pm 0.00$ & $35.25\% \pm 2.44$ \\ 
		\hline 
		$ax^8 + c$ & $67.27\% \pm 2.31$& $67.27\% \pm 2.44$ & $44.12\% \pm 1.24$ & $54.89\% \pm 1.8$ & $60.27\% \pm 1.58$ & $60.18\% \pm 1.01$ & $39.17\% \pm 0.57$ & $41.85\% \pm 0.14$ & $43.27\% \pm 2.26$ & $34\% \pm 1.94$ \\ 
		\hline 
		$ax^9 + c$ & $66.78\% \pm 2.25$ & $66.67\% \pm 2.77$ & $44.47\% \pm 2.12$ & $54.27\% \pm 1.84$ & $61\% \pm 0.00$ & $62\% \pm 0.00$ & $37.69\% \pm 0.00$ & $41.19\% \pm 0.08$ & $43\% \pm 0.14$ & $34.3\% \pm 2.83$ \\ 
		\hline 
		$\sum_{i=0}^1 a_i x^i$ & $67.54\% \pm 0.00$ & $66.28\% \pm 0.00$ & $45.77\% \pm 1.54$ & $54.37\% \pm 1.47$ & $51\% \pm 0.00$ & $58.24\% \pm 1.62$ & $45\% \pm 0.00$ & $43\% \pm 0.00$ & $41\% \pm 0.00$ & $24.13\% \pm 1.05$ \\ 
		\hline 
		$\sum_{i=0}^2 a_i x^i$ & $66.99\% \pm 0.00$ & $66\% \pm 2.64$ & $53.53\% \pm 1.86$ & $59.54\% \pm 2.11$ & $65\% \pm 0.00$ & $68\% \pm 0.00$ & $69\% \pm 0.00$ & $71.47\% \pm 1.04$ & $54.08\% \pm 1.68$ & $49\% \pm 0.00$ \\ 
		\hline 
		$\sum_{i=0}^3 a_i x^i$ & $67.24\% \pm 0.00$ & $65.4\% \pm 1.97$ & $57.67\% \pm 2.4$ & $59.91\% \pm 2.54$ & $\mathbf{67\%} \pm 0.00$ & $\mathbf{70\%} \pm 0.00$ & $75\% \pm 0.00$ & $75.54\% \pm 1.22$ & $57.26\% \pm 1.95$ & $52.01\% \pm 2.31$ \\ 
		\hline 
		$\sum_{i=0}^4 a_i x^i$ & $63.87\% \pm 2.78$ & $66.81\% \pm 0.00$ & $57.47\% \pm 2.37$ & $60.72\% \pm 2.59$ & $68\% \pm 0.00$ & $\mathbf{70\%} \pm 0.00$ & $77\% \pm 0.00$ & $75.63\% \pm 1.45$ & $59.35\% \pm 2.11$ & $54\% \pm 0.00$ \\ 
		\hline 
		$\sum_{i=0}^5 a_i x^i$ & $65.19\% \pm 0.00$ & $66.54\% \pm 0.00$ & $56.59\% \pm 2.48$ & $58.92\% \pm 2.47$ & $\mathbf{67}\% \pm 0.00$ & $\mathbf{70\%} \pm 0.00$ & $78\% \pm 0.00$ & $74.43\% \pm 1.46$ & $59\% \pm 0.00$ & $53.06\% \pm 2.37$ \\ 
		\hline 
		$\sum_{i=0}^6 a_i x^i$ & $64.76\% \pm 2.91$ & $67\% \pm 0.00$ & $54.17\% \pm 1.98$ & $\mathbf{61.81\%} \pm 2.72$ & $\mathbf{67\%} \pm 0.00$ & $69\% \pm 0.00$ & $80\% \pm 0.00$ & $73\% \pm 0.00$ & $58.51\% \pm 2.19$ & $53.82\% \pm 2.41$ \\ 
		\hline 
		$\sum_{i=0}^7 a_i x^i$ & $65.08\% \pm 0.00$ & $66.36\% \pm 0.00$ & $55.39\% \pm 2.05$ & $58.99\% \pm 2.54$ & $\mathbf{67\%} \pm 0.00$ & $69\% \pm 0.00$ & $79\% \pm 0.00$ & $72\% \pm 0.00$ & $58.94\% \pm 2.87$ & $53.01\% \pm 2.89$ \\ 
		\hline 
		$\sum_{i=0}^8 a_i x^i$ & $65.07\% \pm 0.00$ & $66.24\% \pm 0.00$ & $55.19\% \pm 2.23$ & $58.18\% \pm 2.94$ & $62.06\% \pm 2.86$ & $68\% \pm 0.00$ & $79\% \pm 0.00$ & $72\% \pm 0.00$ & $58.47\% \pm 2.14$ & $53.47\% \pm 2.94$ \\ 
		\hline 
		$\sum_{i=0}^9 a_i x^i$ & $64.94\% \pm 0.00$ & $67.21\% \pm 0.00$ & $54.72\% \pm 2.59$ & $61.15\% \pm 2.41$ & $\mathbf{67\%} \pm 0.00$ & $69\% \pm 0.00$ & $77\% \pm 0.00$ & $72\% \pm 0.00$ & $57.57\% \pm 2.61$ & $53\% \pm 0.00$ \\ 
		\hline 
		SVR & $66\% \pm 0.38$ & $64.67\% \pm 2.73$ & $45.47\% \pm 1.49$ & $54.33\% \pm 2.06$ & $52\% \pm 0.00$ & $58\% \pm 0.00$ & $41.37\% \pm 2.1$ & $43\% \pm 0.00$ & $40\% \pm 0.00$ & $25.69\% \pm 1.88$ \\ 
		\hline 
		NN 2 & $63.5\% \pm 1.94$ & $64.03\% \pm 1.57$ & $54.41\% \pm 2.54$ & $58.17\% \pm 2.21$ & $66.54\% \pm 1.78$ & $68.36\% \pm 2$ & $\mathbf{80.97\%} \pm 1.48$ & $\mathbf{76.97\%} \pm 2.31$ & $59.65\% \pm 2.43$ & $54.93\% \pm 2.89$ \\ 
		\hline 
		NN 5 & $63.92\% \pm 2.06$ & $65.69\% \pm 1.52$ & $56.69\% \pm 2.56$ & $57.08\% \pm 2.54$ & $66.49\% \pm 0.97$ & $68.26\% \pm 2.28$ & $79.72\% \pm 1.74$ & $74.58\% \pm 3.17$ & $60.26\% \pm 2.53$ & $54.67\% \pm 3.11$ \\ 
		\hline 
		NN 10 & $64.83\% \pm 2.44$ & $66.07\% \pm 1.46$ & $55.16\% \pm 2.36$ & $58.07\% \pm 2.95$ & $66.82\% \pm 1.1$ & $68.71\% \pm 2.26$ & $76.99\% \pm 2.41$ & $73.47\% \pm 3.29$ & $59.48\% \pm 2.38$ & $54.49\% \pm 2.85$ \\ 
		\hline 
		NN 20 & $64.9\% \pm 3.02$ & $66.87\% \pm 2.19$ & $\mathbf{57.65\%} \pm 2.26$ & $59.72\% \pm 3.23$ & $66.36\% \pm 2.78$ & $69.46\% \pm 2.42$ & $76.36\% \pm 2.84$ & $73.10\% \pm 2.83$ & $58.98\% \pm 2.9$ & $54.83\% \pm 3.21$ \\ 
		\hline 
		NN 2-4 & $63.26\% \pm 1.87$ & $65.33\% \pm 1.61$ & $56.72\% \pm 2.5$ & $57.93\% \pm 2.56$ & $66.52\% \pm 1.02$ & $66.89\% \pm 2.44$ & $80.25\% \pm 1.81$ & $74.04\% \pm 2.16$ & $59.98\% \pm 2.46$ & $54.26\% \pm 2.83$ \\ 
		\hline 
		NN 4-8 & $64.33\% \pm 2.23$ & $66.05\% \pm 1.44$ & $54.96\% \pm 2.73$ & $57.82\% \pm 2.91$ & $66.48\% \pm 1.77$ & $68.5\% \pm 2.59$ & $78.71\% \pm 2.01$ & $73.12\% \pm 2.41$ & $59.76\% \pm 2.51$ & $55.44\% \pm 2.61$ \\ 
	\end{tabular}
\end{sidewaystable}

\begin{sidewaystable}
	\caption{The performances of all used classes of regression functions for RECI in the standardized {\ttfamily CEP} and {\ttfamily SIM-G} data sets when averaging the MSE over all runs.}
	\label{tab:meanStandard}
	\scriptsize
	\centering
	\renewcommand{\arraystretch}{1.2}
	\begin{tabular}{c|c|c|c|c}
		& Standardized {\ttfamily CEP} & Standardized Prep. {\ttfamily CEP} & Standardized {\ttfamily SIM-G} & Standardized Prep. {\ttfamily SIM-G} \\ 
		\hline 
		LOG & $69.29\%$ & $66.87\%$& $80\%$ & $78\%$ \\ 
		\hline 
		$ax^2 + c$ & $63.57\%$ & $66.53\%$& $24\%$ & $21\%$ \\ 
		\hline 
		$ax^3 + c$ & $61.99\%$ & $57.83\%$& $60\%$ & $47\%$ \\ 
		\hline 
		$ax^4 + c$ & $65.44\%$ & $66.79\%$& $29\%$ & $18\%$ \\ 
		\hline 
		$ax^5 + c$ & $64.61\%$ & $62.71\%$& $58\%$ & $43\%$ \\ 
		\hline 
		$ax^6 + c$ & $65.65\%$ & $65.72\%$& $31\%$ & $19\%$ \\ 
		\hline 
		$ax^7 + c$ & $64.49\%$ & $64.26\%$& $58\%$ & $47\%$ \\ 
		\hline 
		$ax^8 + c$ & $66.72\%$ & $65.76\%$& $36\%$ & $26\%$ \\ 
		\hline 
		$ax^9 + c$ & $65.86\%$ & $65.84\%$& $64\%$ & $44\%$ \\ 
		\hline 
		$\sum_{i=0}^1 a_i x^i$ & $60.83\%$ & $61.25\%$& $31\%$ & $30\%$ \\ 
		\hline 
		$\sum_{i=0}^2 a_i x^i$ & $\mathbf{74.07\%}$ & $\mathbf{72.79\%}$& $83\%$ & $79\%$ \\ 
		\hline 
		$\sum_{i=0}^3 a_i x^i$ & $67.47\%$ & $66.01\%$& $83\%$ & $80\%$ \\ 
		\hline 
		$\sum_{i=0}^4 a_i x^i$ & $67.28\%$ & $66.07\%$& $\mathbf{85\%}$ & $\mathbf{83\%}$ \\ 
		\hline 
		$\sum_{i=0}^5 a_i x^i$ & $62.4\%$ & $66.95\%$& $80\%$ & $82\%$ \\ 
		\hline 
		$\sum_{i=0}^6 a_i x^i$ & $67.97\%$ & $64\%$ & $80\%$ & $81\%$ \\ 
		\hline 
		$\sum_{i=0}^7 a_i x^i$ & $64.43\%$ & $63.06\%$& $78\%$ & $77\%$ \\ 
		\hline 
		$\sum_{i=0}^8 a_i x^i$ & $67.87\%$ & $65.70\%$& $80\%$ & $82\%$ \\ 
		\hline 
		$\sum_{i=0}^9 a_i x^i$ & $68.89\%$ & $68.31\%$& $81\%$ & $81\%$ \\ 
		\hline 
		SVR & $55.30\%$ & $61.58\%$& $32\%$ & $30\%$ \\ 
		\hline 
		NN 2 & $55.19\%$ & $64.82\%$& $80\%$ & $78\%$ \\ 
		\hline 
		NN 5 & $56.85\%$ & $61.35\%$& $81\%$ & $76\%$ \\ 
		\hline 
		NN 10 & $60.57\%$ & $58.90\%$& $80\%$ & $80\%$ \\ 
		\hline 
		NN 20 & $56.75\%$ & $60.89\%$& $79\%$ & $79\%$ \\ 
		\hline 
		NN 2-4 & $60.09\%$ & $57.78\%$& $72\%$ & $77\%$ \\ 
		\hline 
		NN 4-8 & $59.13\%$ & $59.21\%$& $80\%$ & $80\%$ \\ 
	\end{tabular}
\end{sidewaystable}

\begin{sidewaystable}
	\caption{The performances of all used classes of regression functions for RECI in the standardized {\ttfamily CEP} and {\ttfamily SIM-G} data sets when not averaging the MSE.}
	\label{tab:nonMeanStand}
	\scriptsize
	\centering
	\renewcommand{\arraystretch}{1.2}
	\begin{tabular}{c|c|c|c|c}
		& Standardized {\ttfamily CEP} & Standardized Prep. {\ttfamily CEP} & Standardized {\ttfamily SIM-G} & Standardized Prep. {\ttfamily SIM-G} \\ 
		\hline 
		LOG & $53.86\% \pm 5.33$ & $54.18\% \pm 6.12$& $76.67\% \pm 3.26$ & $76.21\% \pm 3.13$ \\ 
		\hline 
		$ax^2 + c$ & $62.41\% \pm 0.00$ & $58.88\% \pm 0.00$& $34.94\% \pm 4.51$ & $26.23\% \pm 3.86$ \\ 
		\hline 
		$ax^3 + c$ & $61.99\% \pm 0.00$ & $56.67\% \pm 0.81$& $54.42\% \pm 3.53$ & $47.91\% \pm 4.12$ \\ 
		\hline 
		$ax^4 + c$ & $65.44\% \pm 0.00$ & $59.68\% \pm 0.00$& $32.44\% \pm 3.18$ & $26.23\% \pm 3.45$ \\ 
		\hline 
		$ax^5 + c$ & $56.80\% \pm 0.00$ & $58.79\% \pm 0.02$& $54.65\% \pm 3.87$ & $45.74\% \pm 4.03$ \\ 
		\hline 
		$ax^6 + c$ & $65.65\% \pm 0.00$ & $60.43\% \pm 0.00$& $36.24\% \pm 2.97$ & $35.60\% \pm 3.80$ \\ 
		\hline 
		$ax^7 + c$ & $60.53\% \pm 0.00$ & $57.39\% \pm 0.00$& $51.44\% \pm 3.63$ & $44.65\% \pm 4.10$ \\ 
		\hline 
		$ax^8 + c$ & $\mathbf{65.73\%} \pm 0.63$ & $62.91\% \pm 0.00$& $38.8\% \pm 3.4$ & $31.20\% \pm 4.89$ \\ 
		\hline 
		$ax^9 + c$ & $63.72\% \pm 0.92$ & $58.79\% \pm 0.00$& $53\% \pm 0.00$ & $50\% \pm 0.00$ \\ 
		\hline 
		$\sum_{i=0}^1 a_i x^i$ & $56.05\% \pm 5.73$ & $57.78\% \pm 6.10$& $43.45\% \pm 0.00$ & $43.87\% \pm 4.04$ \\ 
		\hline 
		$\sum_{i=0}^2 a_i x^i$ & $60.03\% \pm 5.66$ & $61.73\% \pm 0.00$& $\mathbf{83\%} \pm 0.00$ & $77\% \pm 0.00$ \\ 
		\hline 
		$\sum_{i=0}^3 a_i x^i$ & $60.52\% \pm 0.00$ & $59.14\% \pm 4.47$& $\mathbf{83\%} \pm 0.00$ & $79\% \pm 0.00$ \\ 
		\hline 
		$\sum_{i=0}^4 a_i x^i$ & $62.71\% \pm 4.45$ & $61.14\% \pm 4.65$& $\mathbf{83\%} \pm 0.00$ & $\mathbf{83\%} \pm 0.00$ \\ 
		\hline 
		$\sum_{i=0}^5 a_i x^i$ & $61.08\% \pm 4.14$ & $62.07\% \pm 4.54$& $79.27\% \pm 1.34$ & $81\% \pm 0.00$ \\ 
		\hline 
		$\sum_{i=0}^6 a_i x^i$ & $61.93\% \pm 4.87$ & $61.73\% \pm 4.47$& $78.39\% \pm 1.44$ & $80\% \pm 0.00$ \\ 
		\hline 
		$\sum_{i=0}^7 a_i x^i$ & $61.76\% \pm 4.07$ & $61.23\% \pm 4.79$& $77.85\% \pm 1.98$ & $76.59\% \pm 3.17$ \\ 
		\hline 
		$\sum_{i=0}^8 a_i x^i$ & $62.37\% \pm 5.10$ & $\mathbf{63.35\%} \pm 4.20$& $77.8\% \pm 1.78$ & $82\% \pm 0.00$ \\ 
		\hline 
		$\sum_{i=0}^9 a_i x^i$ & $63.70\% \pm 4.53$ & $63.27\% \pm 4.16$& $76.83\% \pm 1.71$ & $81\% \pm 0.00$ \\ 
		\hline 
		SVR & $55.12\% \pm 5.50$ & $57.15\% \pm 5.49$& $44.53\% \pm 4.23$ & $42.69\% \pm 2.01$ \\ 
		\hline 
		NN 2 & $55.89\% \pm 5.30$ & $57.12\% \pm 5.70$& $80.07\% \pm 2.38$ & $80.04\% \pm 2.74$ \\ 
		\hline 
		NN 5 & $55.61\% \pm 5.87$ & $57.31\% \pm 6.08$& $78.10\% \pm 2.23$ & $77.29\% \pm 2.35$ \\ 
		\hline 
		NN 10 & $56.63\% \pm 5.37$ & $57.08\% \pm 3.52$& $76.68\% \pm 2.41$ & $77.10\% \pm 2.50$ \\ 
		\hline 
		NN 20 & $55.93\% \pm 6.55$ & $58.07\% \pm 5.76$& $75.45\% \pm 2.76$ & $75.78\% \pm 2.94$ \\ 
		\hline 
		NN 2-4 & $56.26\% \pm 6.47$ & $56.27\% \pm 6.03$& $78.71\% \pm 2.69$ & $78.26\% \pm 2.62$ \\ 
		\hline 
		NN 4-8 & $56.57\% \pm 3.70$ & $57.65\% \pm 3.47$& $77.04\% \pm 2.19$ & $77.60\% \pm 2.40$ \\ 
	\end{tabular}
\end{sidewaystable}

\begin{sidewaystable}
	\caption{All performances of ANM, PNL, SLOPE, IGCI, CURE and LiNGAM.}
	\label{tab:otherMethods}
	\scriptsize
	\renewcommand{\arraystretch}{1.2}
	\begin{tabular}{c|c|c|c|c|c|c|c|c|c|c}
		& {\ttfamily CEP} & Prep. {\ttfamily CEP} & {\ttfamily SIM} & Prep. {\ttfamily SIM} & {\ttfamily SIM-c} & Prep. {\ttfamily SIM-c} & {\ttfamily SIM-ln} & Prep. {\ttfamily SIM-ln} & {\ttfamily SIM-G} & Prep. {\ttfamily SIM-G} \\ 
		\hline 
		ANM-HSIC & $62.89\% \pm 1.56$ & $65.01\% \pm 1.74$ & $\mathbf{75.54\%} \pm 2.14$ & $\mathbf{73.33\%} \pm 2.08$ & $\mathbf{80.54\%} \pm 2.14$ & $\mathbf{77.93\%} \pm 2.02$ & $76.60\% \pm 2.53$ & $70.98\% \pm 2.40$ & $74.49\% \pm 2.57$ & $73.78\% \pm 2.44$ \\ 
		\hline 
		ANM-ENT & $49.07\% \pm 1.88$ & $43.57\% \pm 2.31$ & $72.91\% \pm 3.09$ & $71.51\% \pm 3.35$ & $76.36\% \pm 3.14$ & $74.95\% \pm 3.57$ & $78.88\% \pm 2.63$ & $73.93\% \pm 2.88$ & $70.32\% \pm 3.77$ & $69.78\% \pm 3.32$ \\ 
		\hline 
		ANM-FN & $50.80\% \pm 2.80$ & $53.50\% \pm 2.18$ & $57.50\% \pm 1.60$ & $51.94\% \pm 1.92$ & $61.15\% \pm 1.99$ & $59.57\% \pm 1.93$ & $\mathbf{85.31\%} \pm 1.36$ & $\mathbf{80.03\%} \pm 1.31$ & $77.72\% \pm 1.90$ & $78\% \pm 2.23$ \\ 
		\hline 
		PNL & $63.66\% \pm 1.65$ & $61.18\% \pm 1.62$ & $66.33\% \pm 4.25$ & $65.53\% \pm 3.99$ & $68.85\% \pm 3.35$ & $67.71\% \pm 3.16$ & $52.21\% \pm 4.02$ & $53.39\% \pm 4.00$ & $65.06\% \pm 4.02$ & $65.20\% \pm 4.08$ \\ 
		\hline 
		SLOPE & $\mathbf{75.19\%} \pm 0.00$ & $\mathbf{69.02\%} \pm 0.00$ & $44\% \pm 0.00$ & $54\% \pm 0.00$ & $55\% \pm 0.00$ & $61\% \pm 0.00$ & $47\% \pm 0.00$ & $46\% \pm 0.00$ & $47\% \pm 0.00$ & $38\% \pm 0.00$ \\ 
		\hline 
		IGCI$_{\text{U},1}$ & $60.66\% \pm 0.00$ & $62.13\% \pm 0.00$ & $36\% \pm 0.00$ & $39\% \pm 0.00$ & $46\% \pm 0.00$ & $47\% \pm 0.00$ & $51\% \pm 0.00$ & $57\% \pm 0.00$ & $54\% \pm 0.00$ & $48\% \pm 0.00$ \\ 
		\hline 
		IGCI$_{\text{U},2}$ & $63.97\% \pm 0.00$ & $61.40\% \pm 0.00$ & $41\% \pm 0.00$ & $42\% \pm 0.00$ & $50\% \pm 0.00$ & $52\% \pm 0.00$ & $52\% \pm 0.00$ & $57\% \pm 0.00$ & $55\% \pm 0.00$ & $48\% \pm 0.00$ \\ 
		\hline 
		IGCI$_{\text{U},3}$ & $60.67\% \pm 0.00$ & $61.99\% \pm 0.00$ & $41\% \pm 0.00$ & $42\% \pm 0.00$ & $50\% \pm 0.00$ & $52\% \pm 0.00$ & $52\% \pm 0.00$ & $57\% \pm 0.00$ & $55\% \pm 0.00$ & $48\% \pm 0.00$ \\ 
		\hline 
		IGCI$_{\text{G},1}$ & $56.62\% \pm 0.00$ & $56.26\% \pm 0.00$ & $36\% \pm 0.00$ & $36\% \pm 0.00$ & $42\% \pm 0.00$ & $41\% \pm 0.00$ & $59\% \pm 0.00$ & $57\% \pm 0.00$ & $\mathbf{87\%} \pm 0.00$ & $\mathbf{87\%} \pm 0.00$ \\ 
		\hline 
		IGCI$_{\text{G},2}$ & $55.15\% \pm 0.00$ & $54.79\% \pm 0.00$ & $37\% \pm 0.00$ & $38\% \pm 0.00$ & $46\% \pm 0.00$ & $45\% \pm 0.00$ & $61\% \pm 0.00$ & $60\% \pm 0.00$ & $86\% \pm 0.00$ & $86\% \pm 0.00$ \\ 
		\hline 
		IGCI$_{\text{G},3}$ & $45.45\% \pm 0.00$ & $41.26\% \pm 0.00$ & $37\% \pm 0.00$ & $38\% \pm 0.00$ & $46\% \pm 0.00$ & $45\% \pm 0.00$ & $61\% \pm 0.00$ & $60\% \pm 0.00$ & $86\% \pm 0.00$ & $86\% \pm 0.00$ \\
		\hline
		CURE & $59.67\% \pm 2.86$ & $59.67\% \pm 2.86$ & $59.62\% \pm 3.30$ & $59.62\% \pm 3.30$ & $68.60\% \pm 3.68$ & $68.60\% \pm 3.68$ & $55.58\% \pm 1.82$ & $55.58\% \pm 1.82$ & $47.46\% \pm 0.96$ & $47.46\% \pm 0.96$ \\
		\hline 
		LiNGAM & $57.76\% \pm 0.00$ & $57.06\% \pm 0.00$ & $43\% \pm 0.00$ & $49\% \pm 0.00$ & $47\% \pm 0.00$ & $52\% \pm 0.00$ & $24\% \pm 0.00$ & $25\% \pm 0.00$ & $28\% \pm 0.00$ & $25\% \pm 0.00$ \\ 
	\end{tabular}
\end{sidewaystable}

\begin{table}
	\caption{The best performing parameters for each causal inference method in the artificial data sets {\ttfamily Linear}, {\ttfamily Invertible} and {\ttfamily Non-invertible}.}
	\label{table:noise}
	\begin{center}
		\renewcommand{\arraystretch}{1.2}
		\begin{tabular}{c|c|c|c}
			Data set & ANM estimator & IGCI configuration & RECI regression model \\ 
			\hline 
			Linear & FN & IGCI-U1 & MON: $a x^9 + c$ \\ 
			\hline 
			Invertible & HSIC & IGCI-U2 & NN 4-8 \\ 
			\hline 
			Non-invertible & HSIC & IGCI-U1 & POLY: $\sum_{i = 0}^4 a_i x^i$ \\ 
		\end{tabular} 
	\end{center}
\end{table}

\begin{table}
	\caption{An overview of all data sets with the corresponding number of cause-effect pairs and data samples.	\label{tab:datasets}}
	\begin{center}
		\renewcommand{\arraystretch}{1.2}
		\begin{tabular}{c|c|c}
			\textbf{Dataset} & \textbf{Number of Cause-Effect Pairs} & \textbf{Number of Samples per Pair}\\ 
			\hline 
			{\ttfamily SIM} & 100 & 1000 \\ 
			\hline 
			{\ttfamily SIM-c} & 100 & 1000 \\ 
			\hline 
			{\ttfamily SIM-ln} & 100 & 1000 \\ 
			\hline 
			{\ttfamily SIM-G} & 100 & 1000 \\  
			\hline 
			{\ttfamily Linear} & 1100 & 500 \\
			\hline 
			{\ttfamily Invertible} & 4100 & 500 \\
			\hline 
			{\ttfamily Non-invertible} & 1100 & 500 \\
		\end{tabular} 
		\renewcommand{\arraystretch}{1.2}
		\begin{flushleft}
			\begin{tabular}{c|c|c|c|c|c|c|c|c|c|c} 
				\textbf{{\ttfamily CEP} Pair} & 1 & 2 & 3 & 4 & 5 & 6 & 7 & 8 & 9 & 10 \\ 
				\hline
				\textbf{Number of Samples}  & 349 & 349 & 349 & 349 & 4177 & 4177 & 4177 & 4177 & 4177 & 4177 \\  
			\end{tabular}
		\end{flushleft}
		\begin{flushleft}
			\begin{tabular}{c|c|c|c|c|c|c|c|c|c|c|c|c|c} 
				11 & 12 & 13 & 14 & 15 & 16 & 17 & 18 & 19 & 20 & 21 & 22 & 23 & 24 \\ 
				\hline 
				4177 & 5000 & 392 & 392 & 392 & 392 & 5000 & 314 & 194 & 349 & 349 & 450 & 450 & 451 \\  
			\end{tabular}
		\end{flushleft}
		\begin{flushleft}
			\begin{tabular}{c|c|c|c|c|c|c|c|c|c|c|c|c|c} 
				25 & 26 & 27 & 28 & 29 & 30 & 31 & 32 & 33 & 34 & 35 & 36 & 37 & 38 \\ 
				\hline 
				1030 & 1030 & 1030 & 1030 & 1030 & 1030 & 1030 & 1030 & 345 & 345 & 345 & 345 & 345 & 757 \\  
			\end{tabular}
		\end{flushleft}
		\begin{flushleft}
			\begin{tabular}{c|c|c|c|c|c|c|c|c|c|c|c|c|c} 
				39 & 40 & 41 & 42 & 43 & 44 & 45 & 46 & 47 & 48 & 49 & 50 & 51 & 52 \\ 
				\hline 
				394 & 733 & 763 & 9162 & 10369 & 10369 & 10369 & 10369 & 254 & 168 & 365 & 365 & 365 & 192\\  
			\end{tabular}
		\end{flushleft}
		\begin{flushleft}
			\begin{tabular}{c|c|c|c|c|c|c|c|c|c|c|c|c|c} 
				53 & 54 & 55 & 56 & 57 & 58 & 59 & 60 & 61 & 62 & 63 & 64 & 65 & 66 \\ 
				\hline 
				192 & 192 & 192 & 192 & 192 & 192 & 192 & 162 & 1331 & 1331 & 1331 & 498 & 16382 & 4499\\  
			\end{tabular}
		\end{flushleft}
		\begin{flushleft}
			\begin{tabular}{c|c|c|c|c|c|c|c|c|c|c|c|c|c} 
				67 & 68 & 69 & 70 & 71 & 72 & 73 & 74 & 75 & 76 & 77 & 78 & 79 & 80 \\ 
				\hline 
				1632 & 5084 & 194 & 205 & 347 & 8401 & 721 & 721 & 721 & 365 & 365 & 365 & 3102 & 994\\  
			\end{tabular}
		\end{flushleft}
		\begin{flushleft}
			\begin{tabular}{c|c|c|c|c|c|c|c|c|c|c|c|c|c} 
				81 & 82 & 83 & 84 & 85 & 86 & 87 & 88 & 89 & 90 & 91 & 92 & 93 & 94 \\ 
				\hline 
				666 & 7753 & 261 & 131 & 1261 & 149 & 150 & 432 & 9504 & 9504 & 9504 & 202 & 94 & 2287\\  
			\end{tabular}
		\end{flushleft}
		\begin{flushleft}
			\begin{tabular}{c|c|c|c|c|c} 
				95 & 96 & 97 & 98 & 99 & 100 \\ 
				\hline 
				2287 & 300 & 109 & 109 & 109 & 114 \\  
			\end{tabular}
		\end{flushleft}
	\end{center}
\end{table}

\clearpage

\subsection*{Plots}
\label{sec:plots}

\begin{figure}[H]
	\begin{multicols}{5}
		\centering
		\includegraphics[width=1\linewidth]{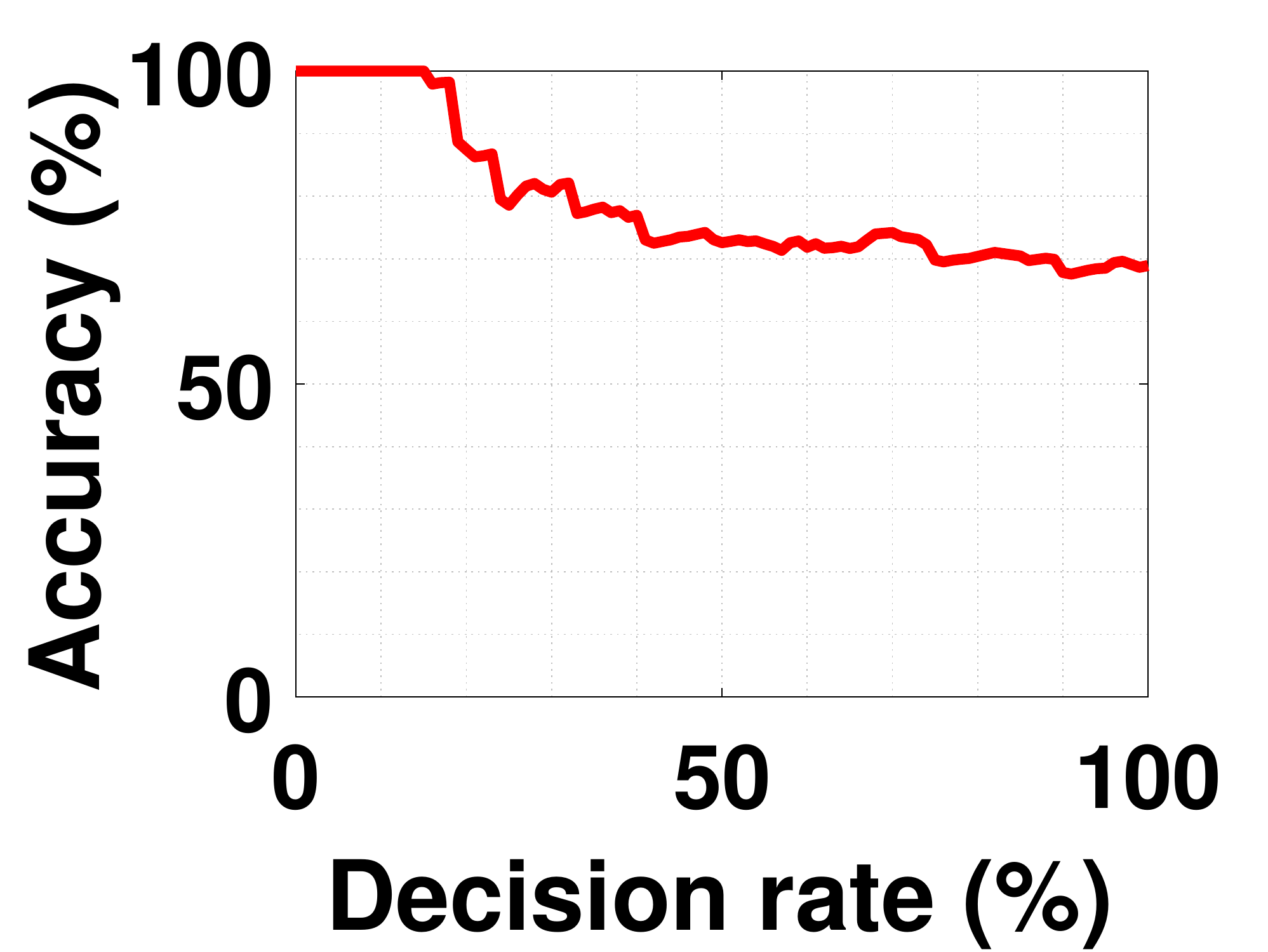}
		{\ttfamily CEP}-LOG
		\includegraphics[width=1\linewidth]{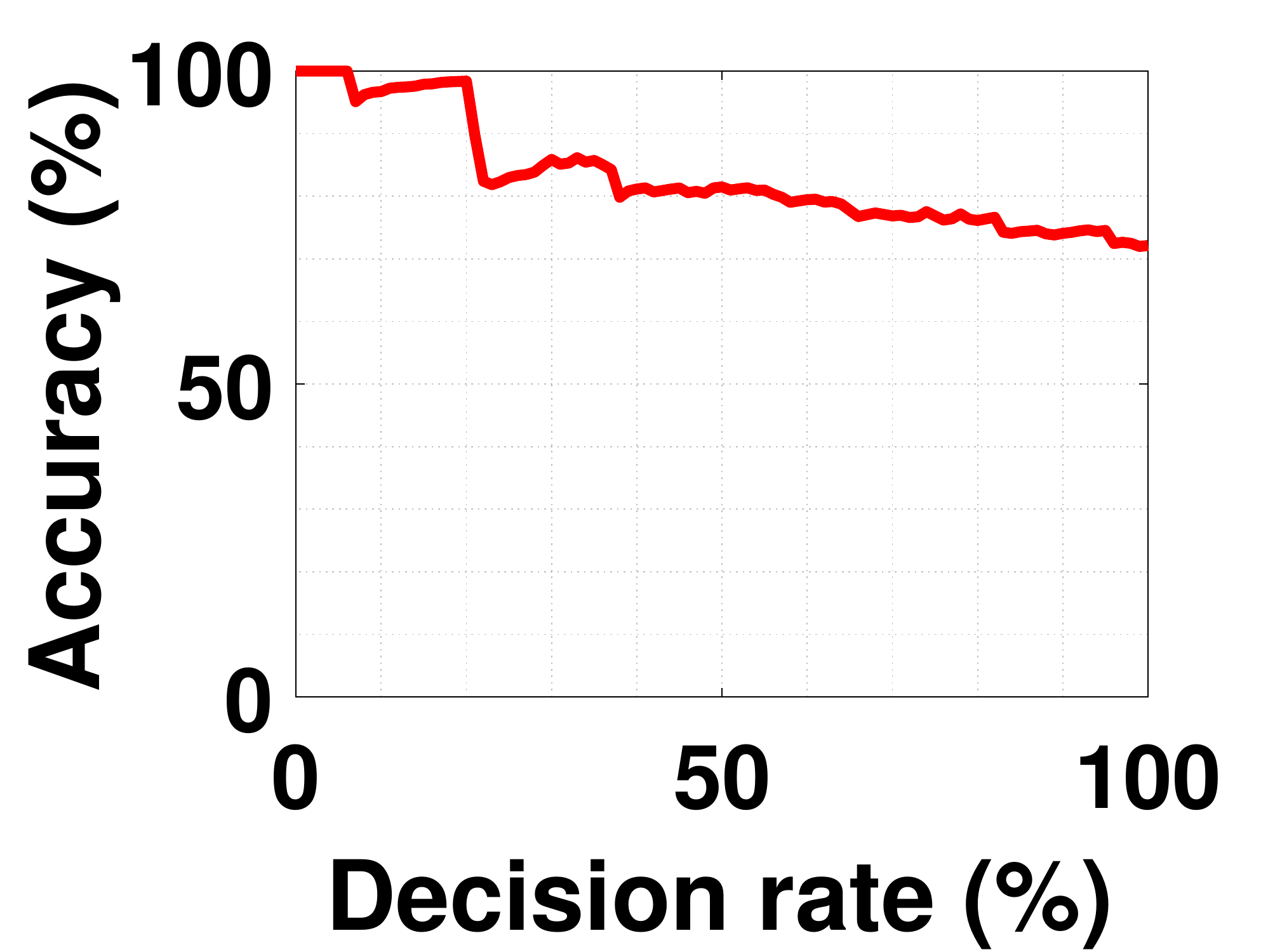}
		{\ttfamily CEP}-MON
		\includegraphics[width=1\linewidth]{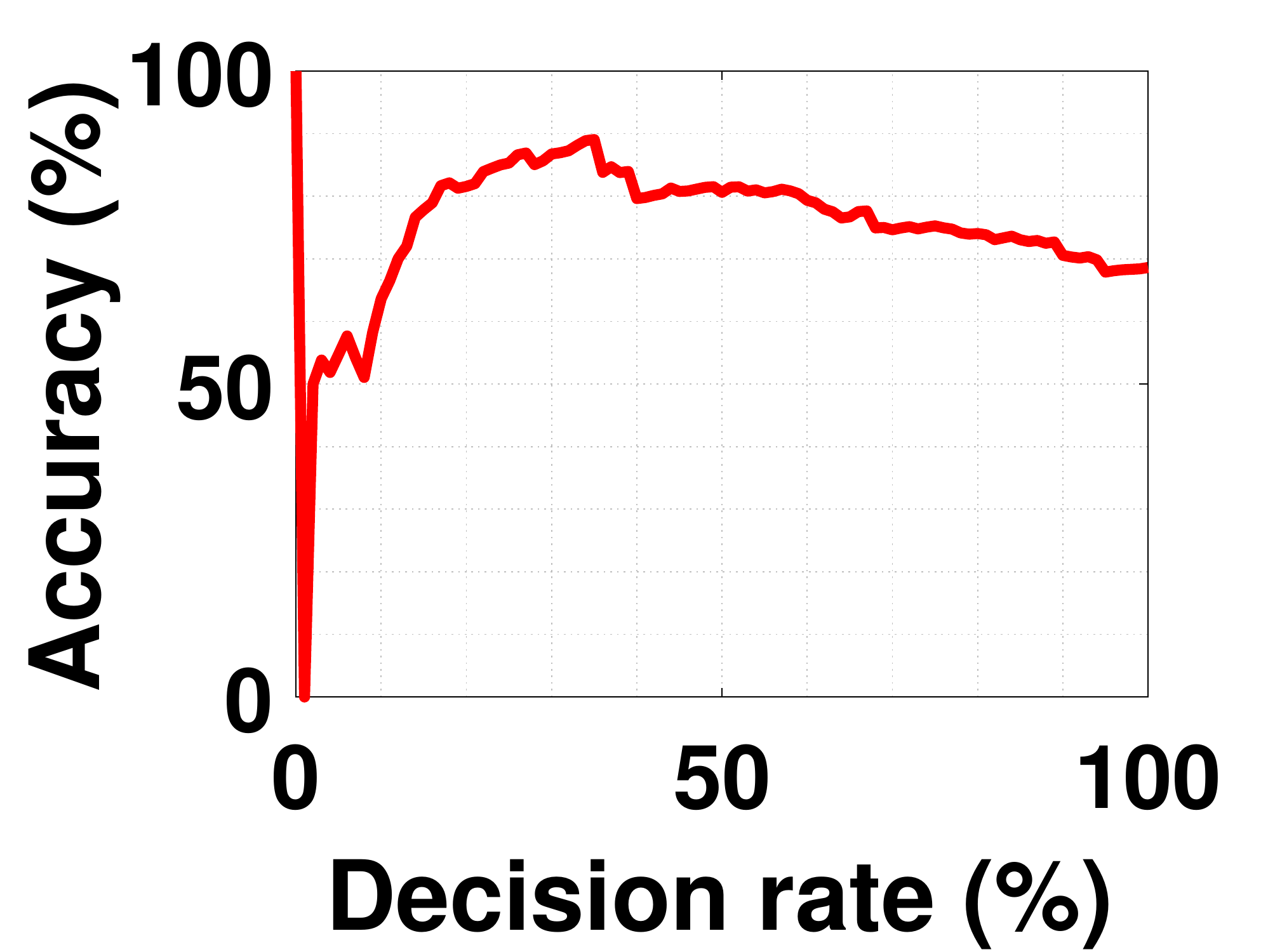}
		{\ttfamily CEP}-POLY
		\includegraphics[width=1\linewidth]{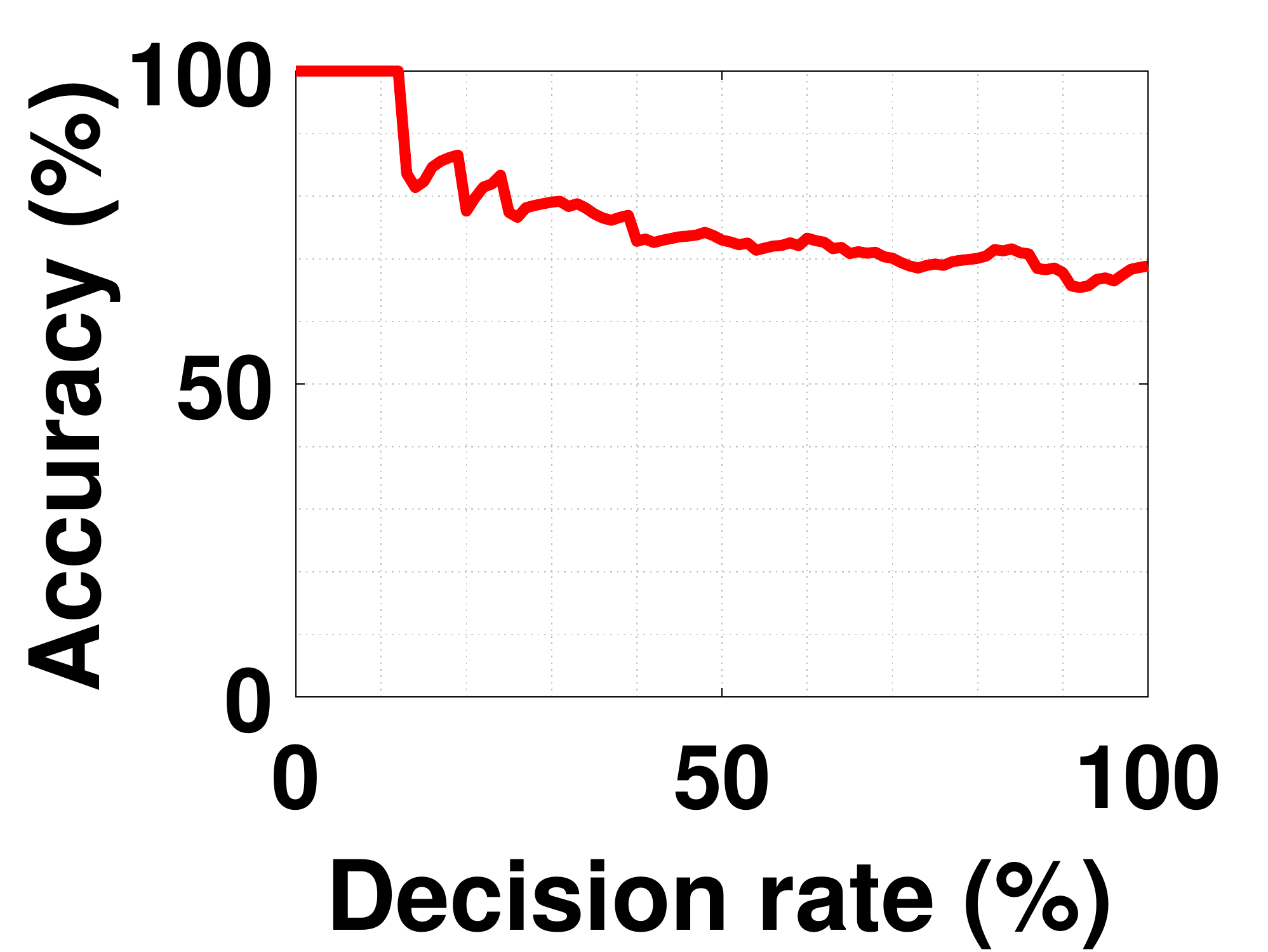}
		{\ttfamily CEP}-SVR
		\includegraphics[width=1\linewidth]{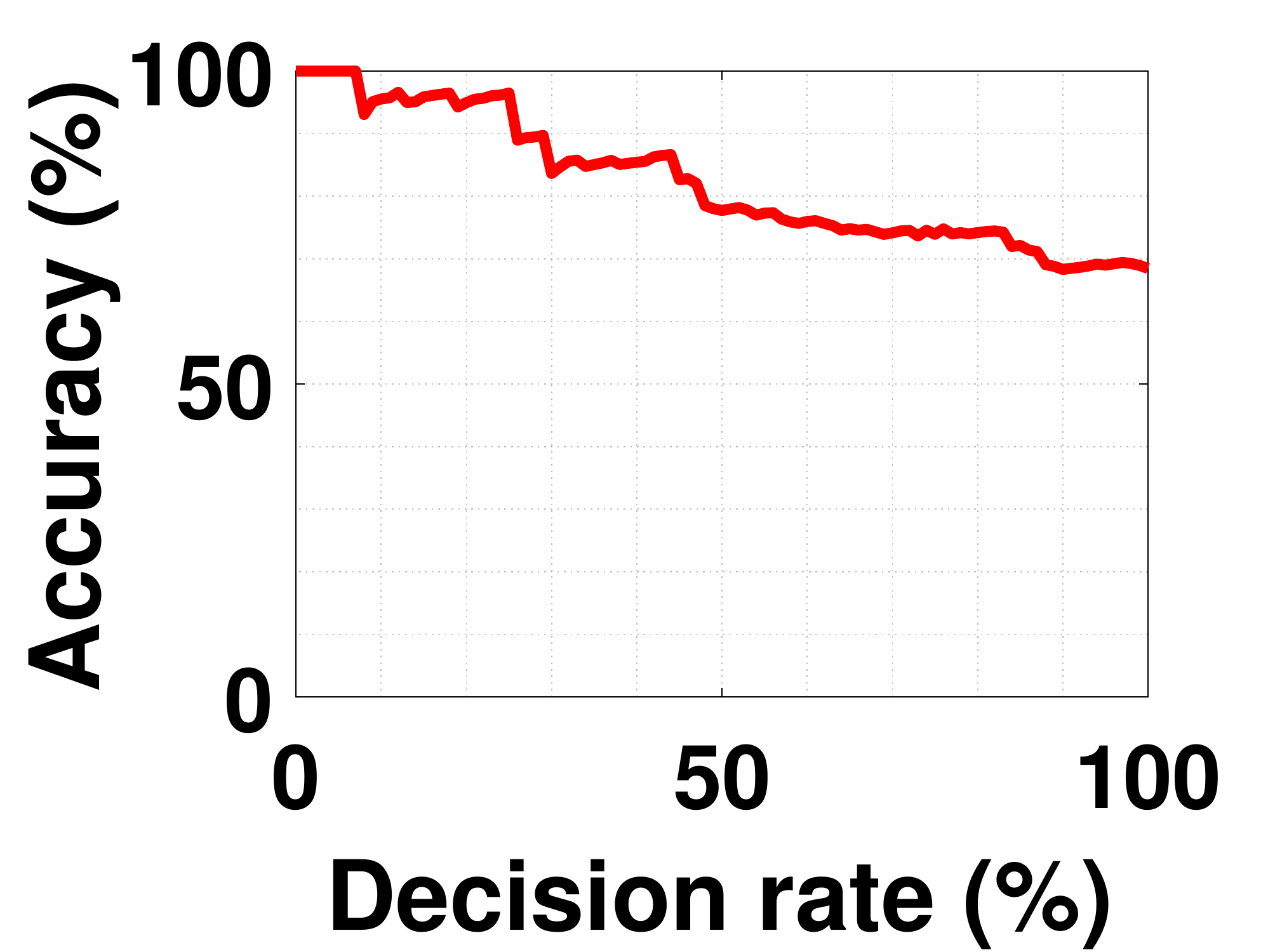}
		{\ttfamily CEP}-NN
	\end{multicols}
	
	\begin{multicols}{5}
		\centering
		\includegraphics[width=1\linewidth]{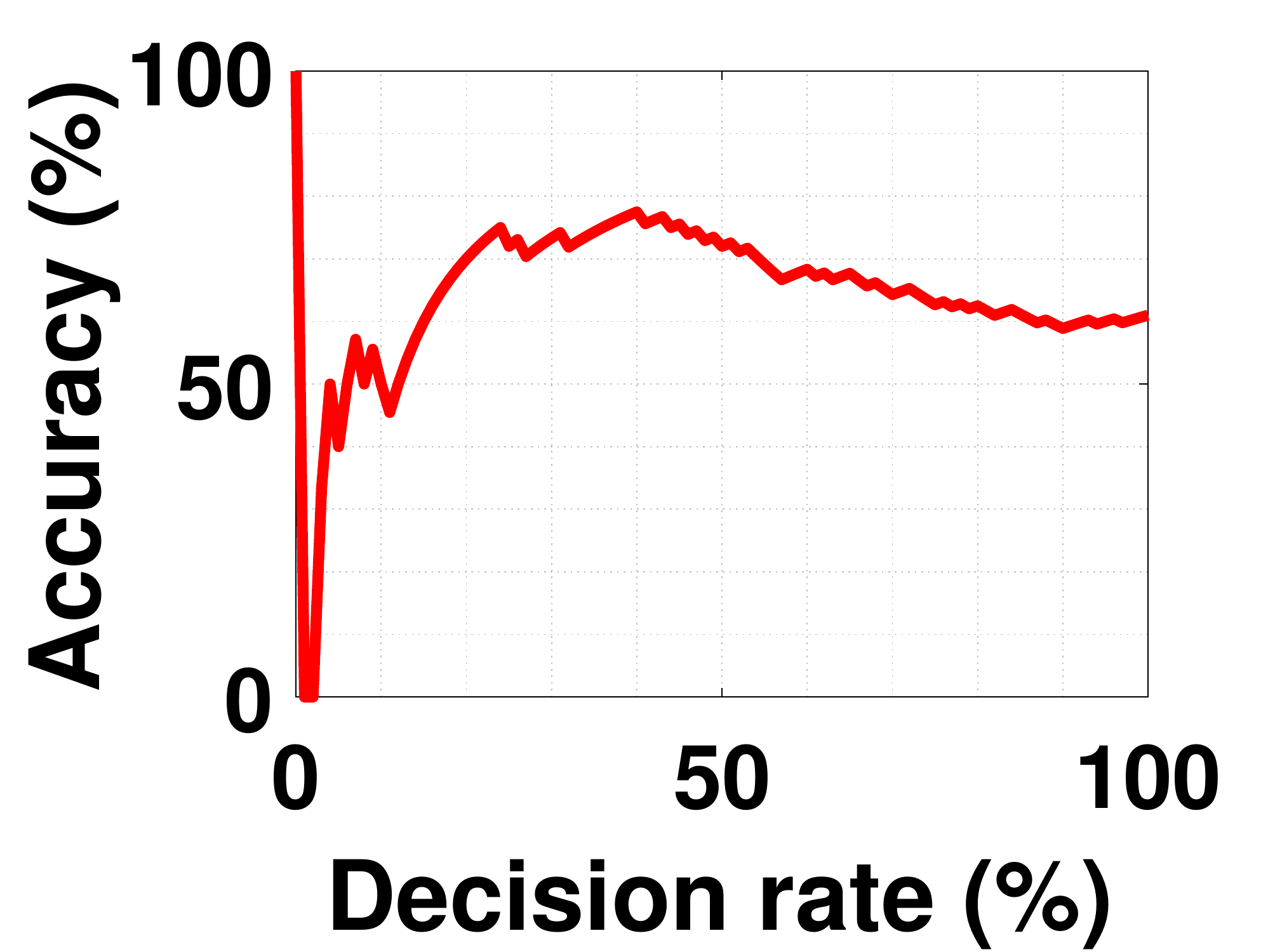}
		{\ttfamily SIM}-LOG
		\includegraphics[width=1\linewidth]{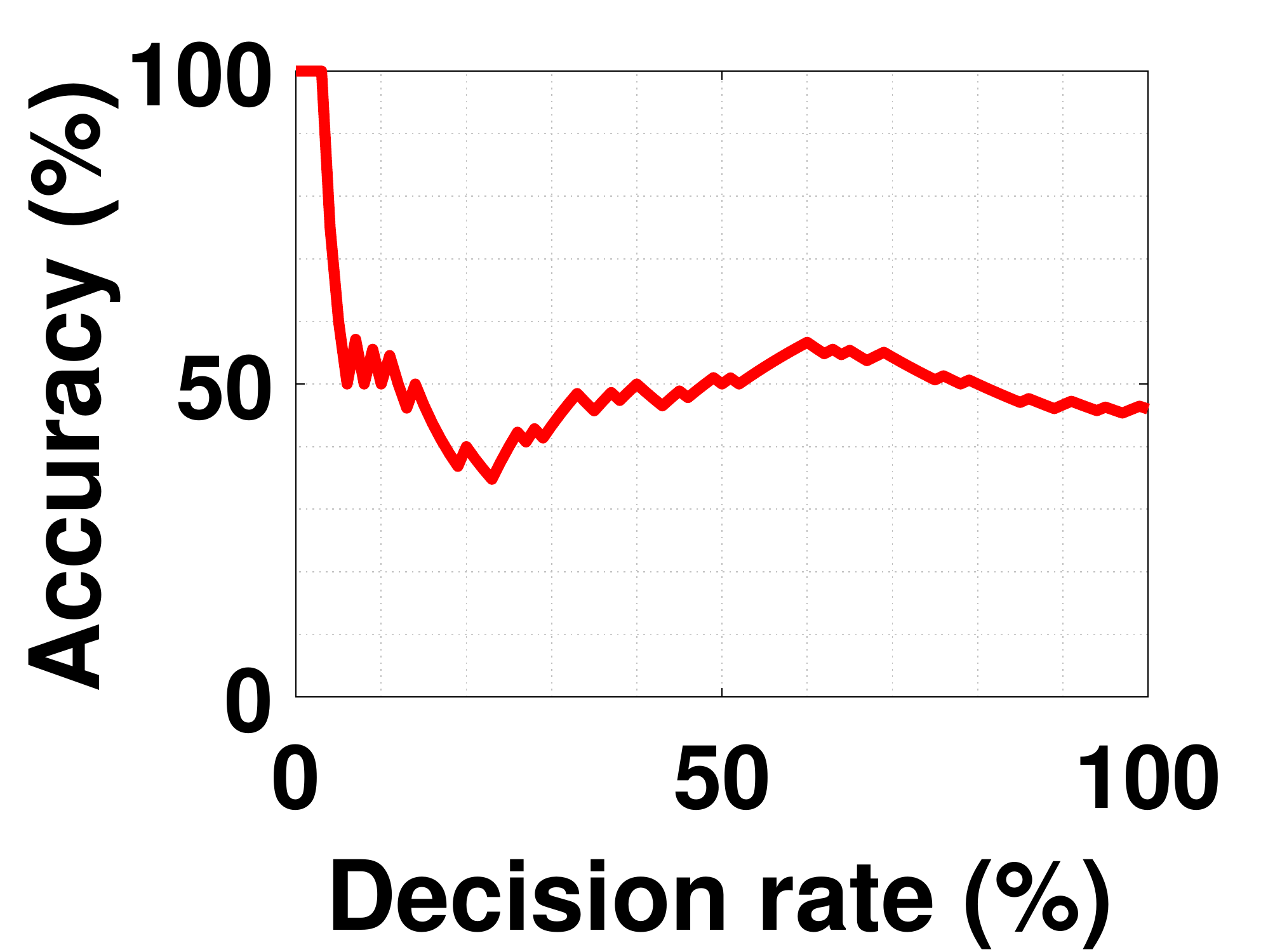}
		{\ttfamily SIM}-MON
		\includegraphics[width=1\linewidth]{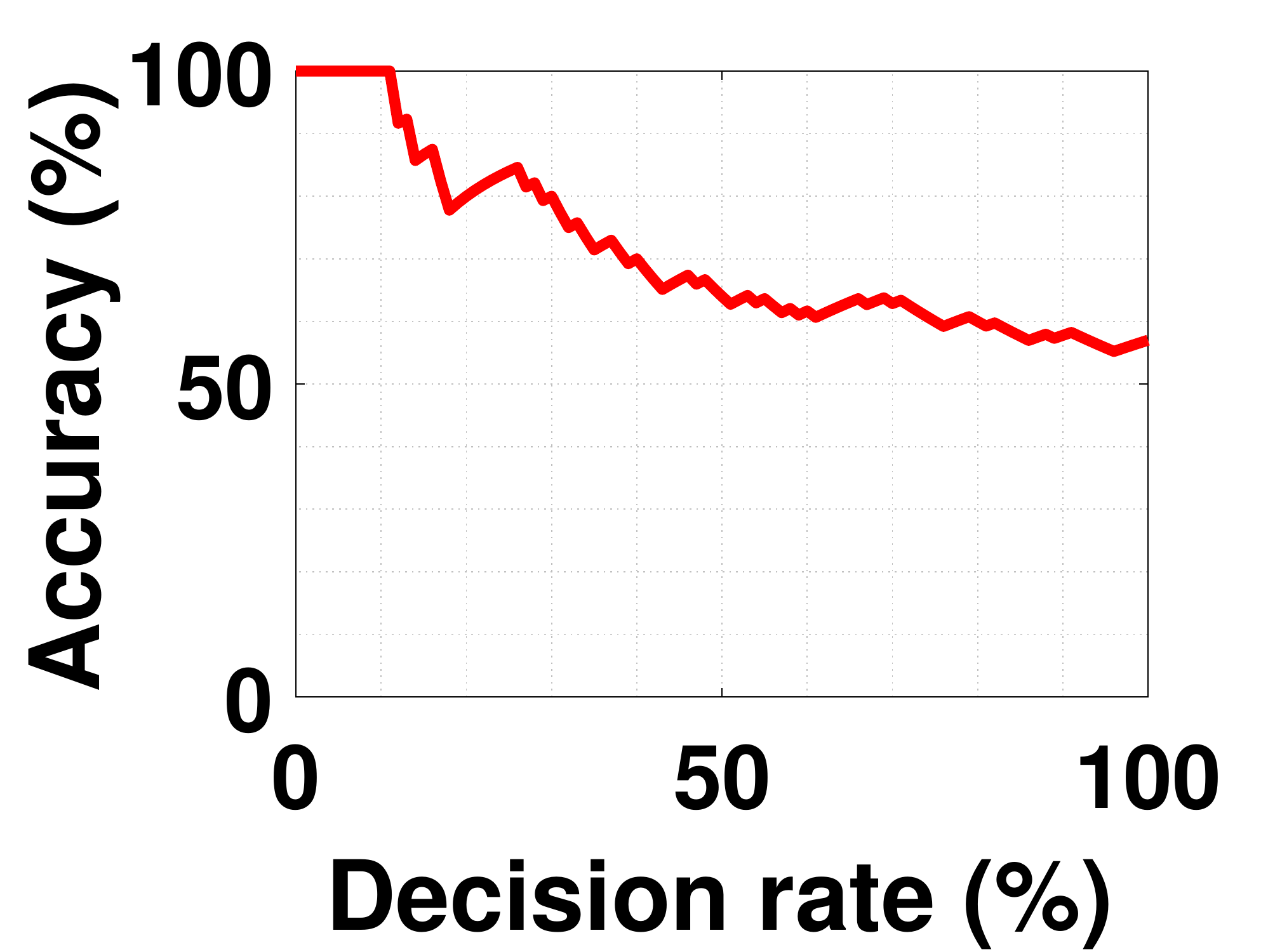}
		{\ttfamily SIM}-POLY
		\includegraphics[width=1\linewidth]{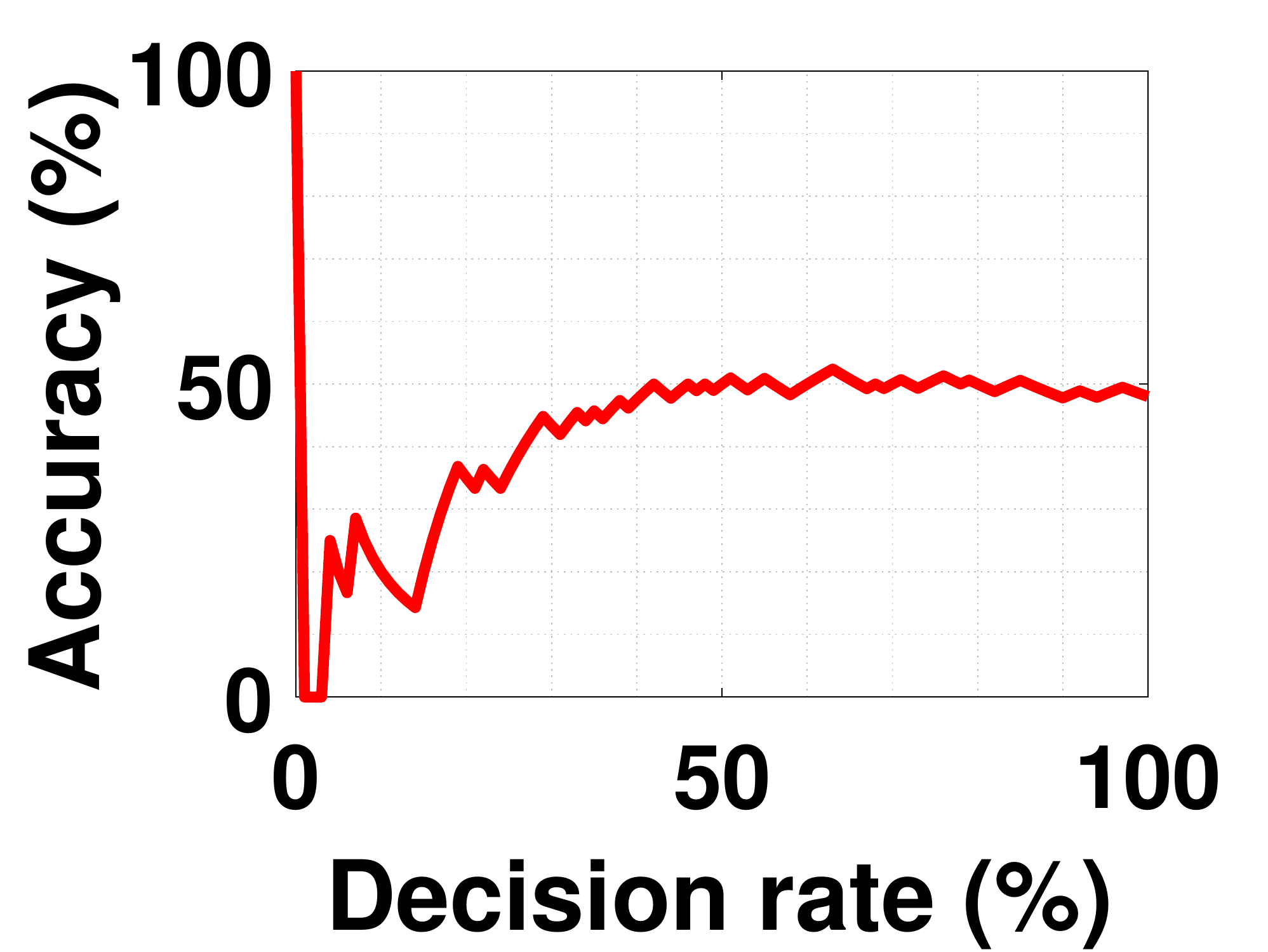}
		{\ttfamily SIM}-SVR
		\includegraphics[width=1\linewidth]{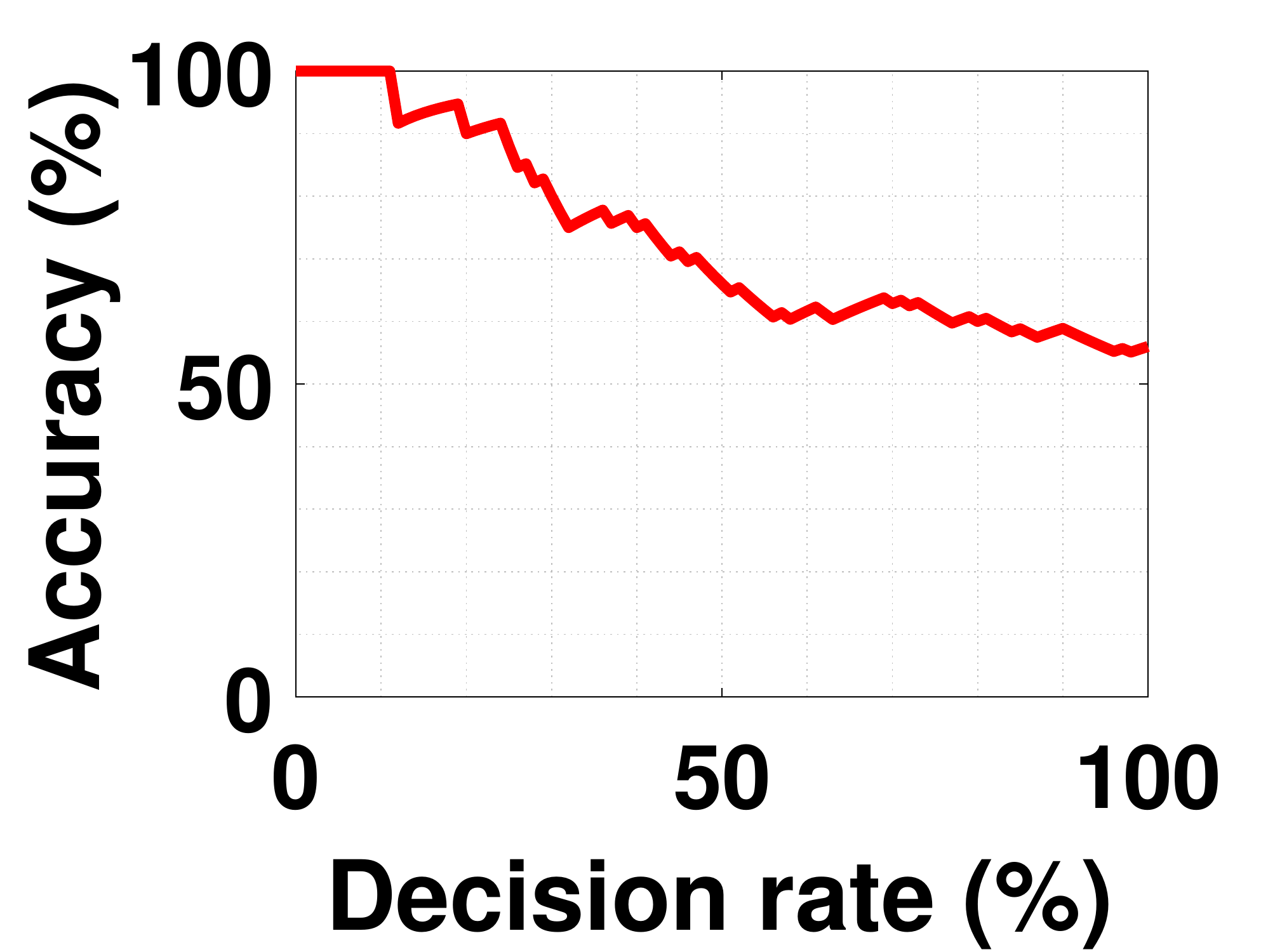}
		{\ttfamily SIM}-NN
	\end{multicols}
	
	\begin{multicols}{5}
		\centering
		\includegraphics[width=1\linewidth]{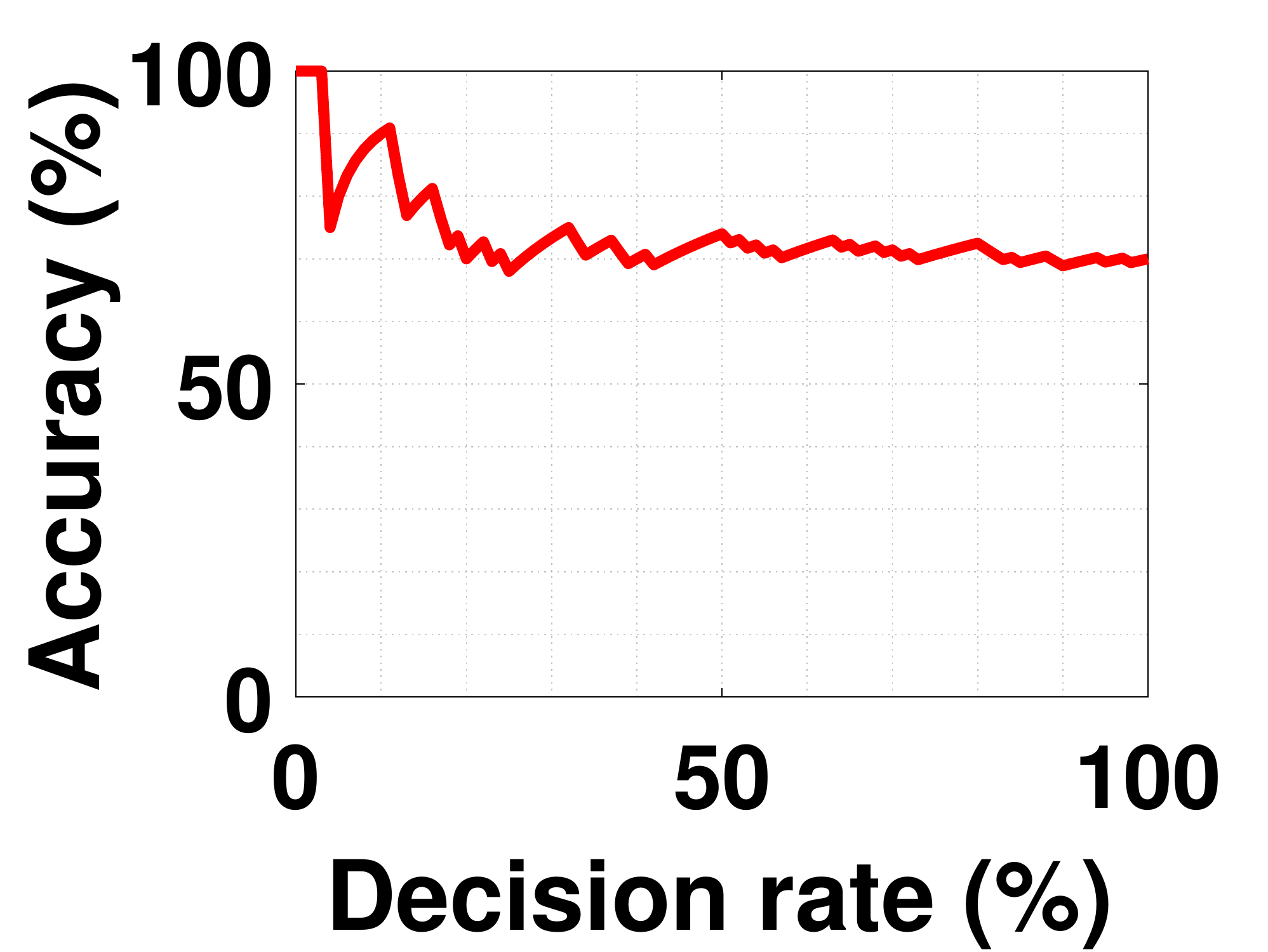}
		{\ttfamily SIM-c}-LOG
		\includegraphics[width=1\linewidth]{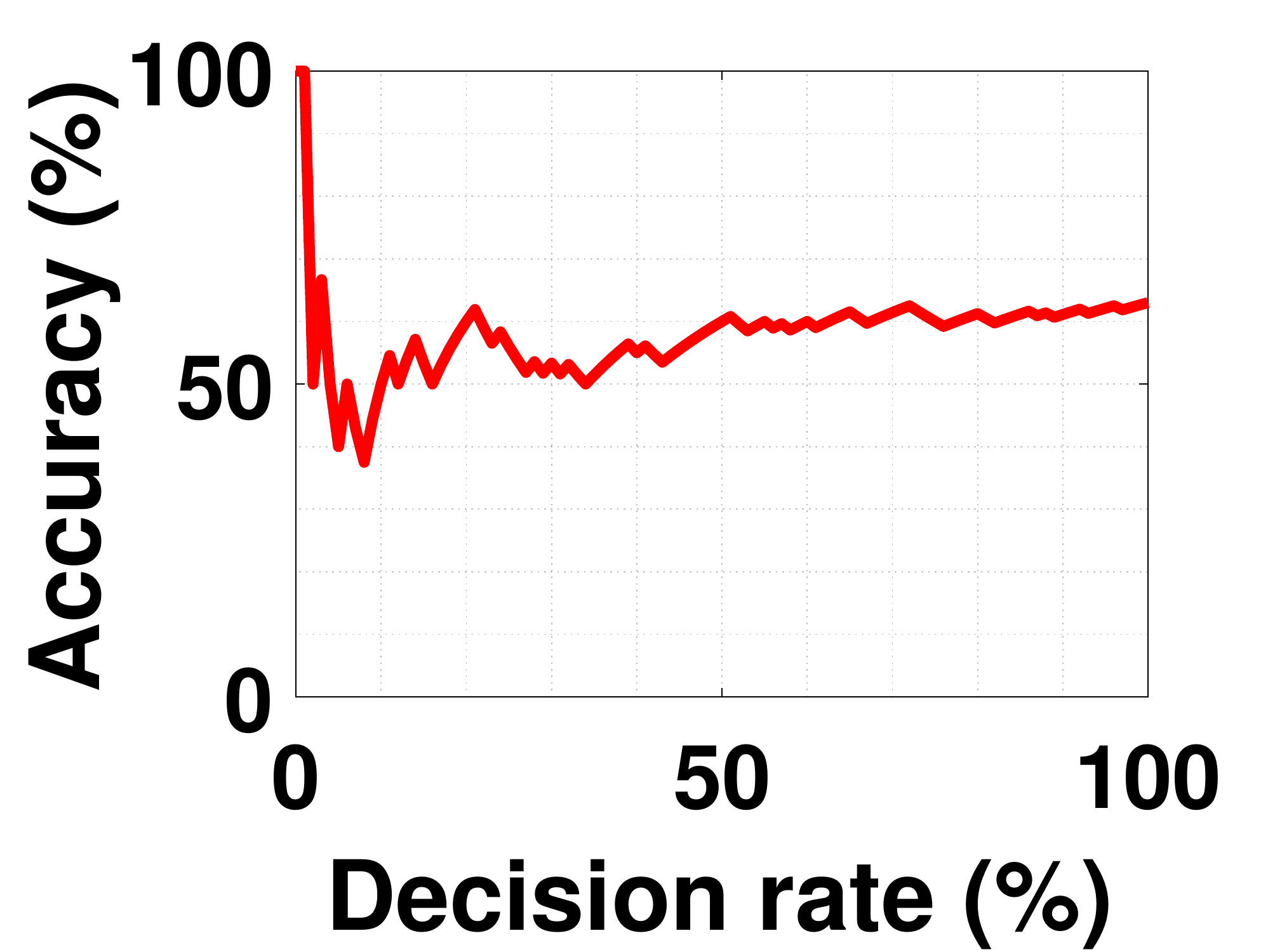}
		{\ttfamily SIM-c}-MON
		\includegraphics[width=1\linewidth]{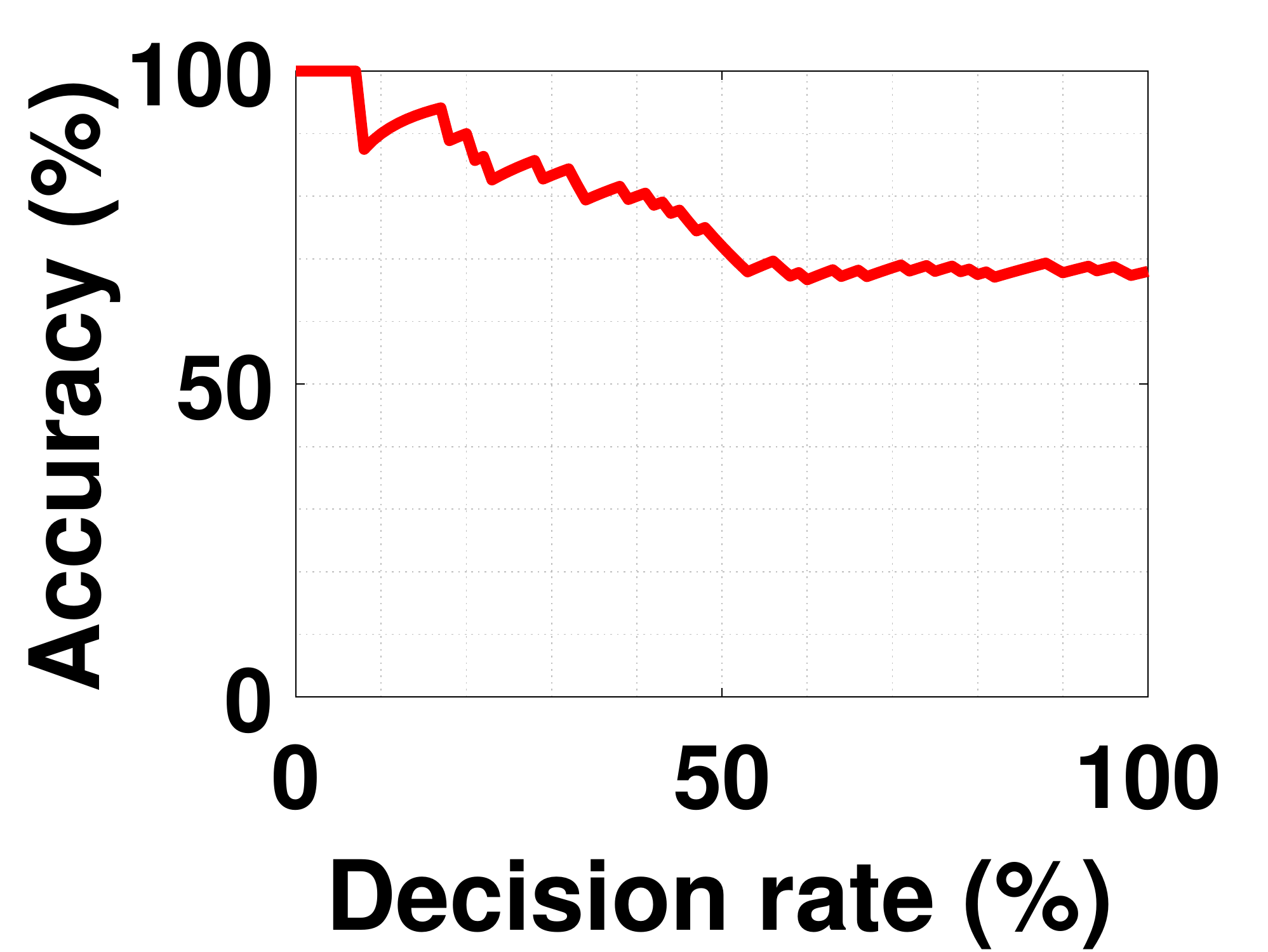}
		{\ttfamily SIM-c}-POLY
		\includegraphics[width=1\linewidth]{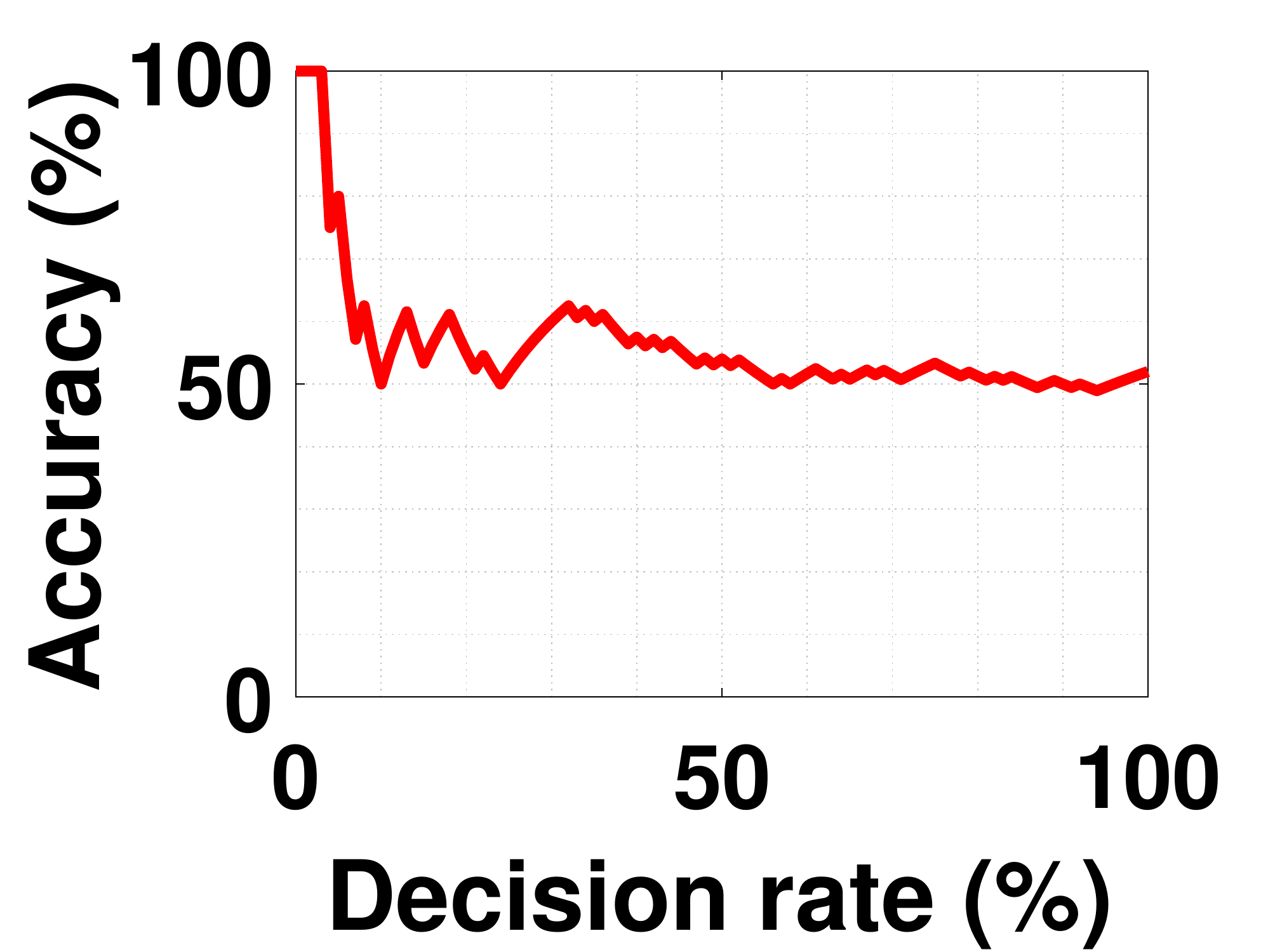}
		{\ttfamily SIM-c}-SVR
		\includegraphics[width=1\linewidth]{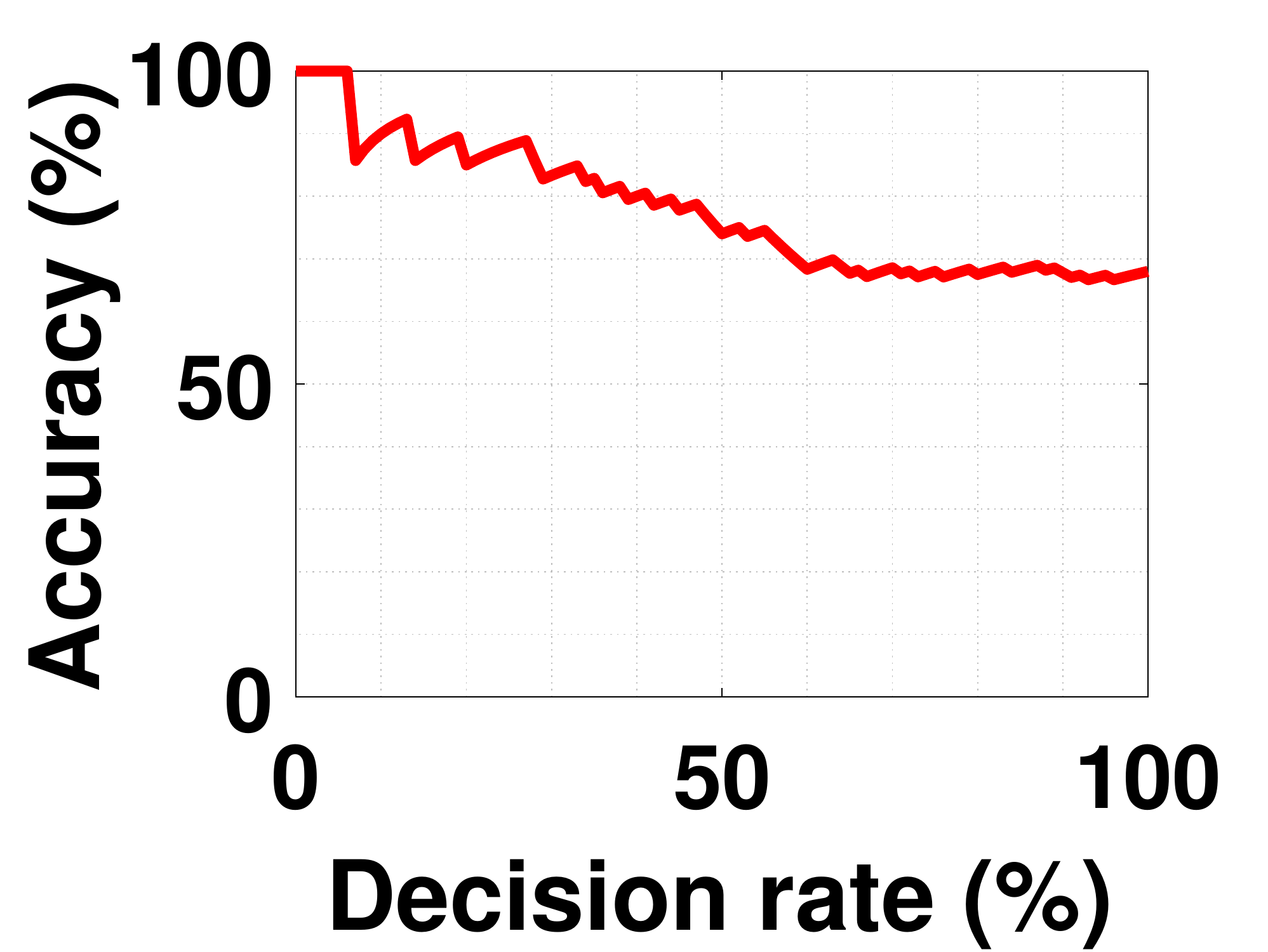}
		{\ttfamily SIM-c}-NN
	\end{multicols}
	
	\begin{multicols}{5}
		\centering
		\includegraphics[width=1\linewidth]{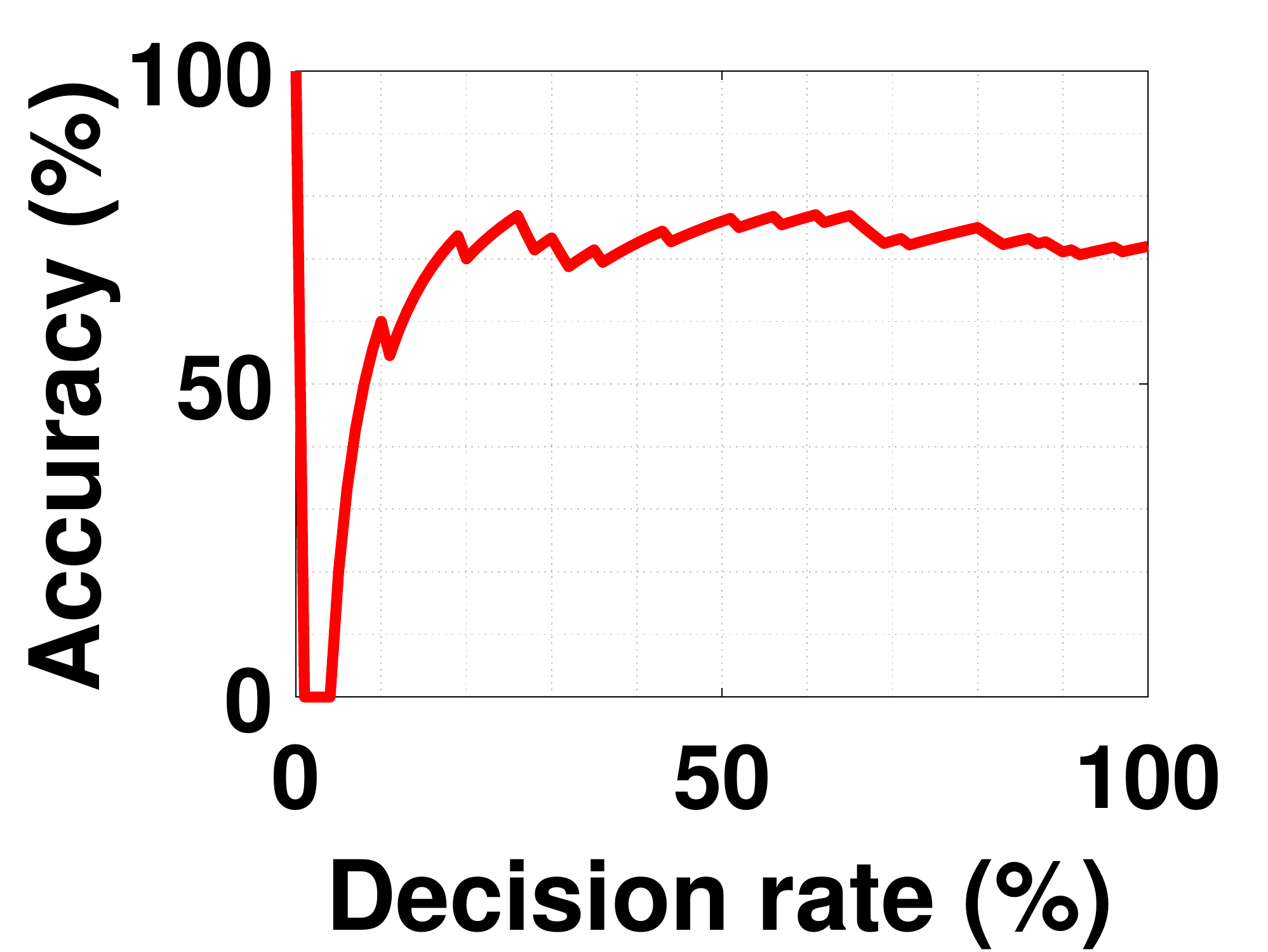}
		{\ttfamily SIM-ln}-LOG
		\includegraphics[width=1\linewidth]{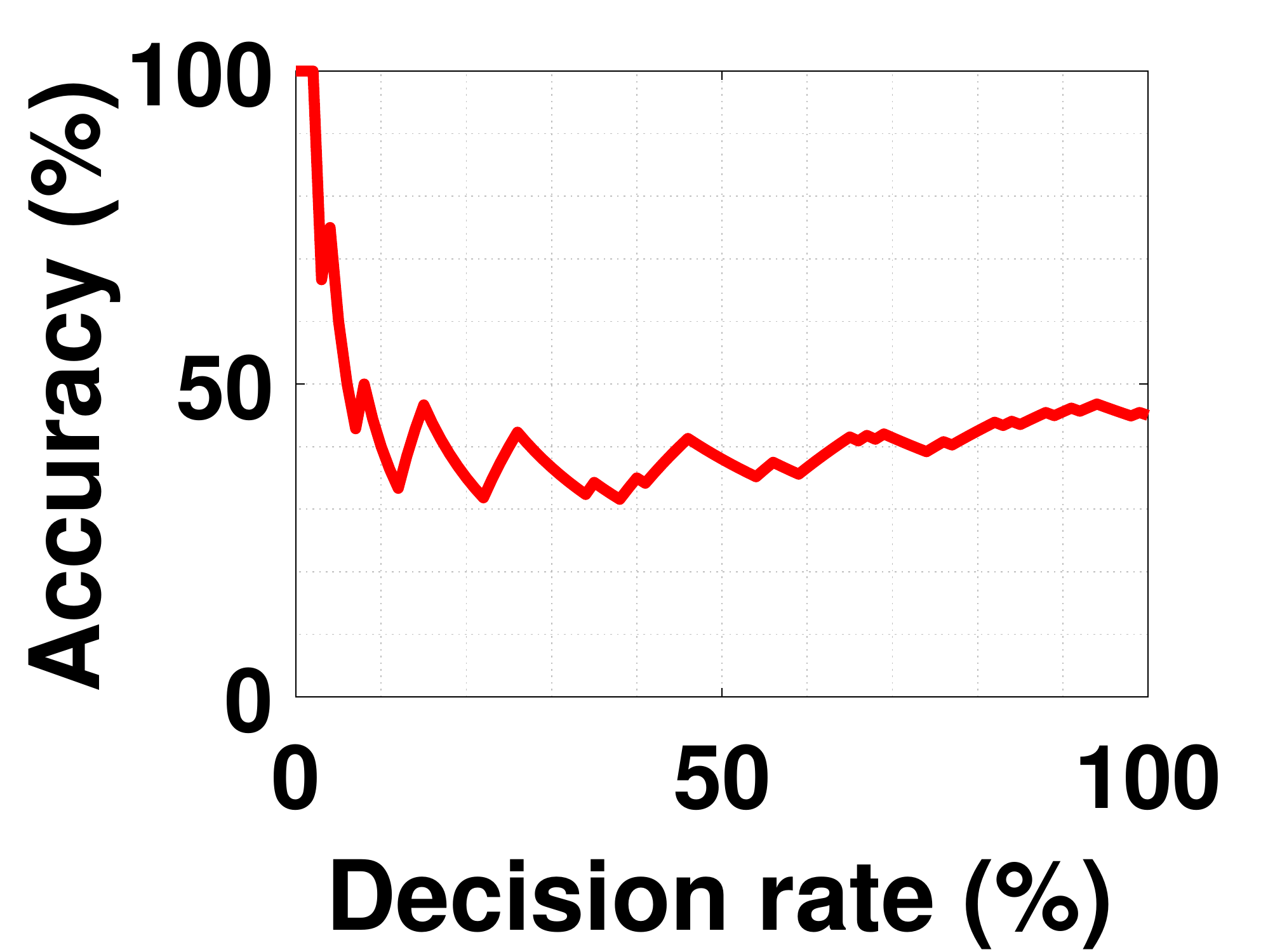}
		{\ttfamily SIM-ln}-MON
		\includegraphics[width=1\linewidth]{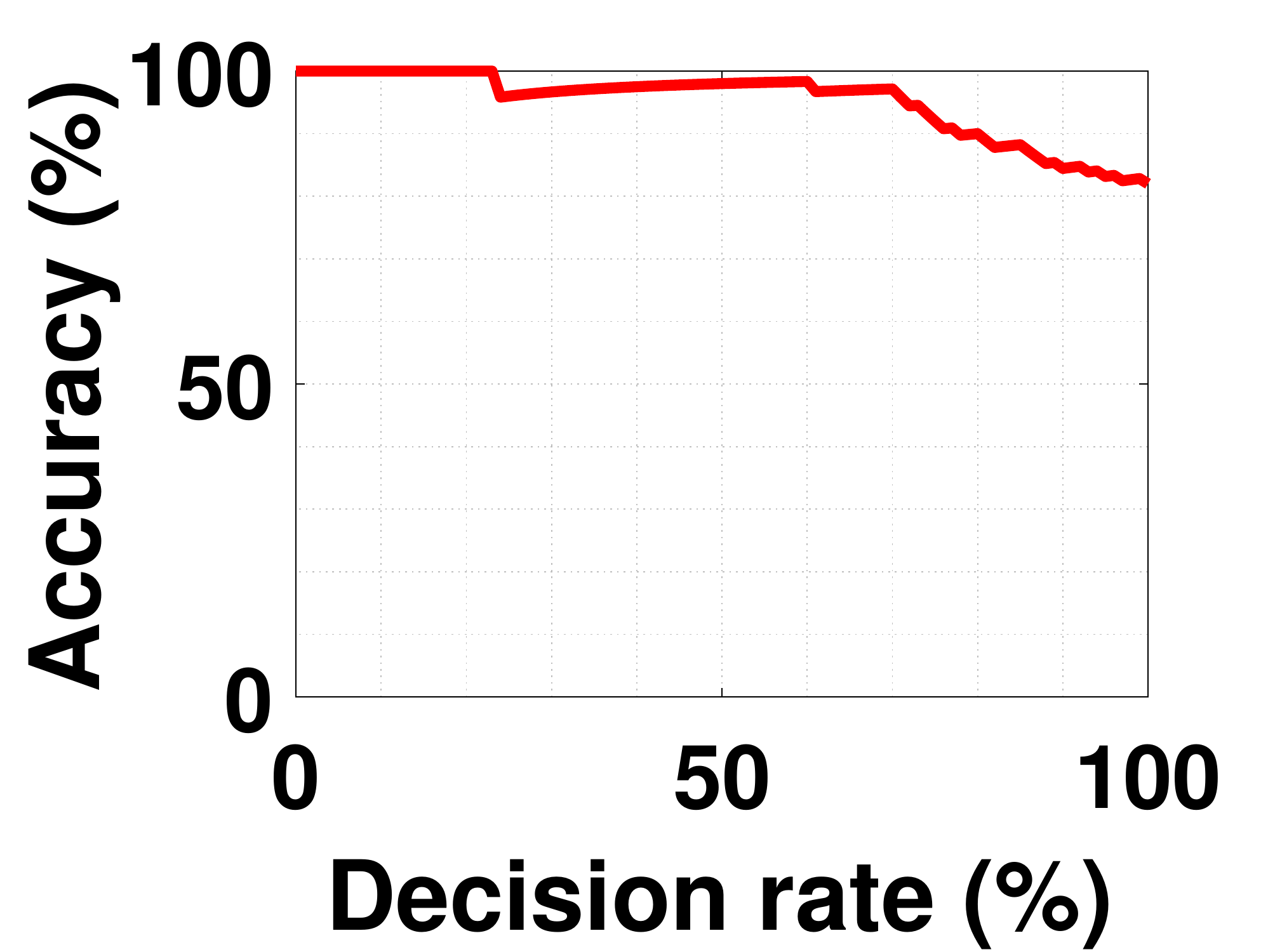}
		{\ttfamily SIM-ln}-POLY
		\includegraphics[width=1\linewidth]{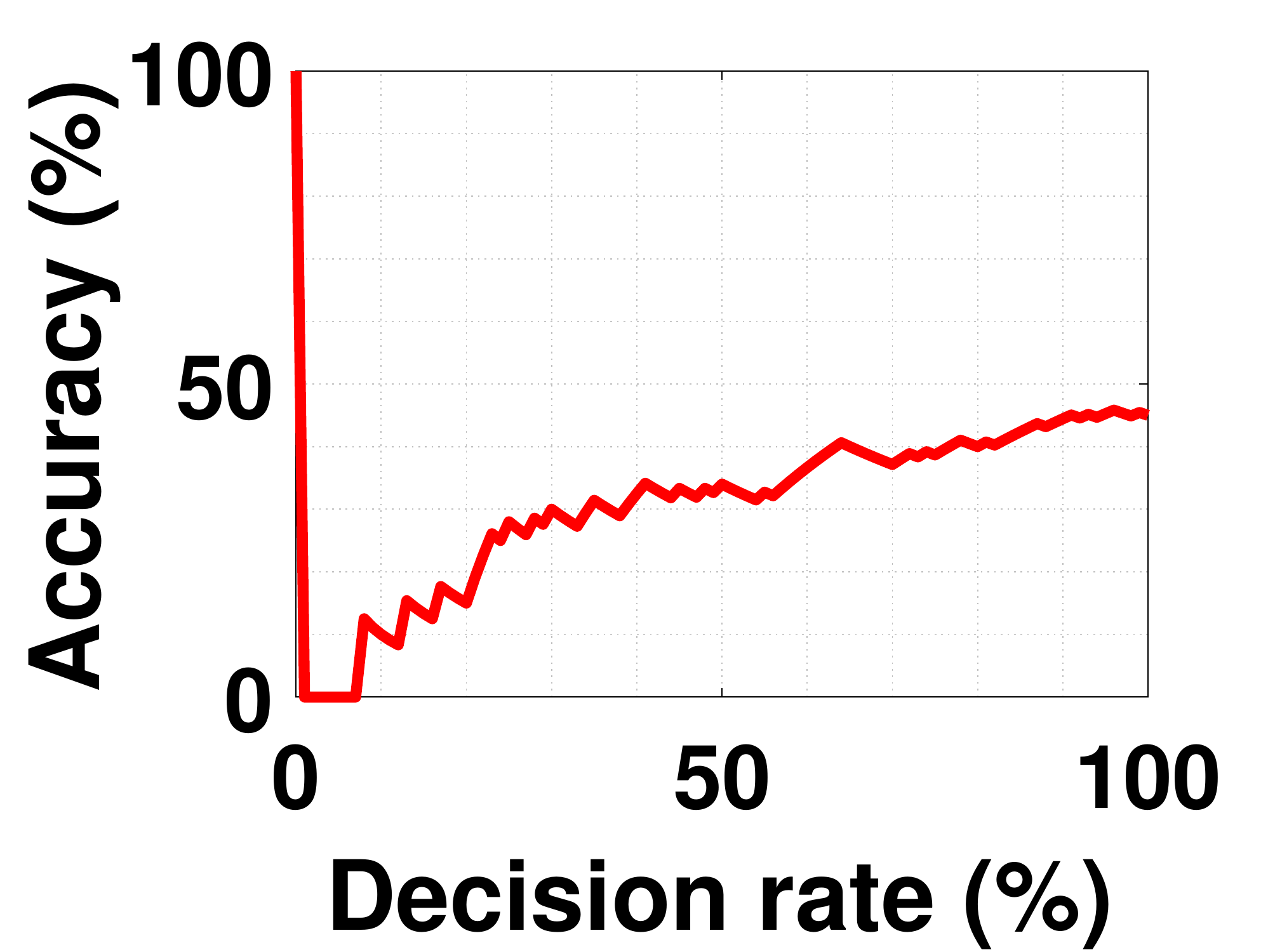}
		{\ttfamily SIM-ln}-SVR
		\includegraphics[width=1\linewidth]{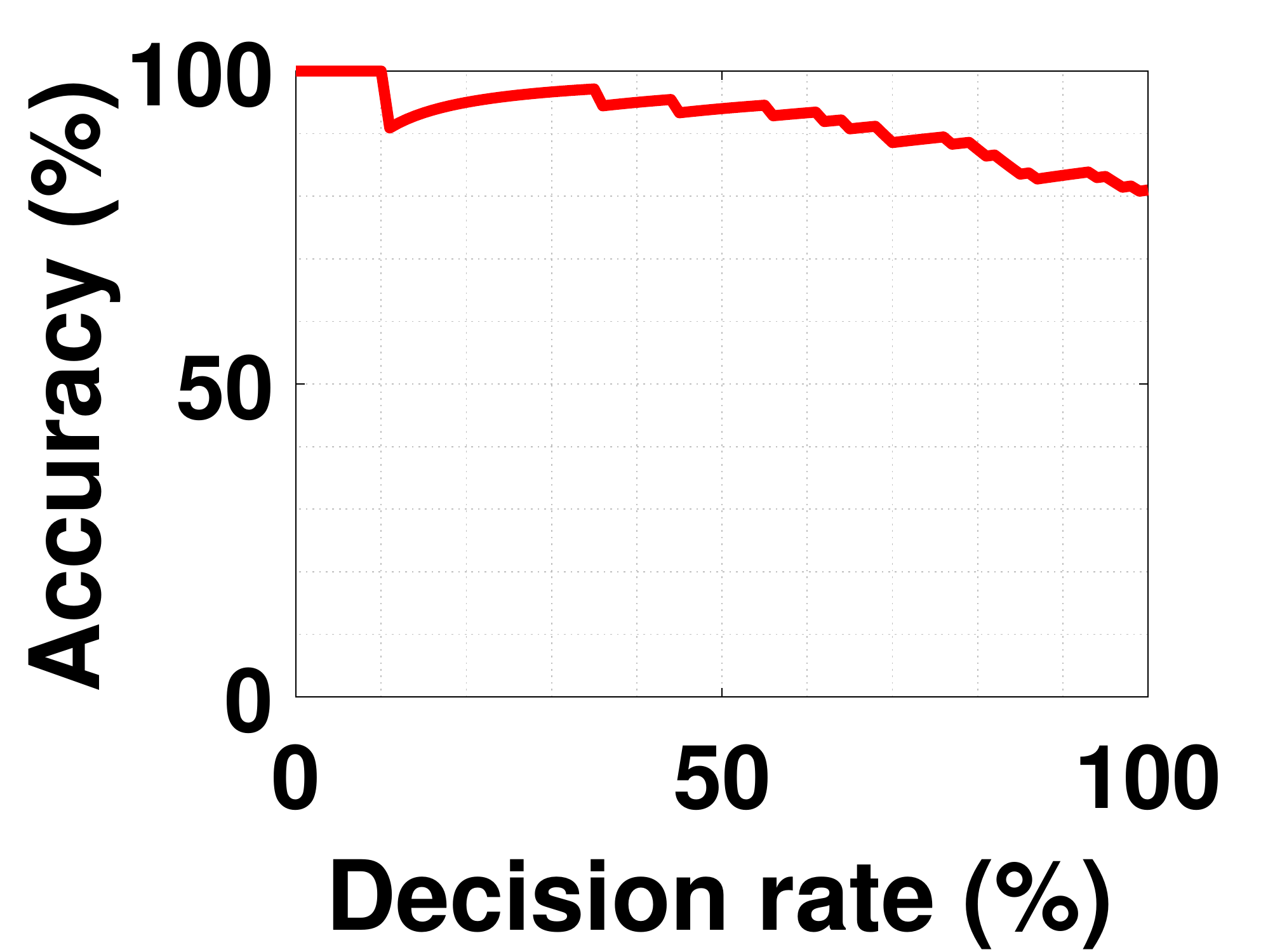}
		{\ttfamily SIM-ln}-NN
	\end{multicols}
	
	\begin{multicols}{5}
		\centering
		\includegraphics[width=1\linewidth]{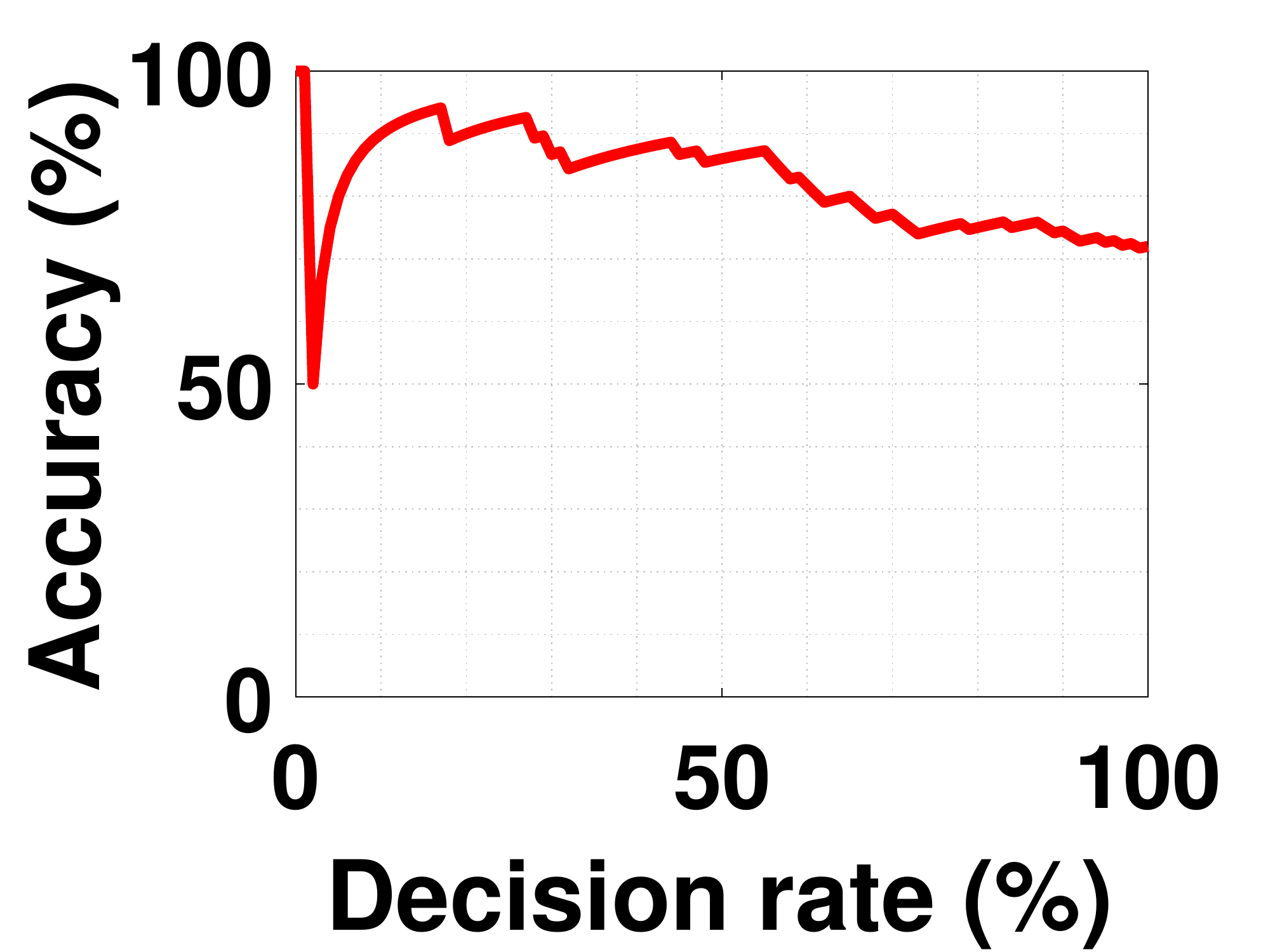}
		{\ttfamily SIM-G}-LOG
		\includegraphics[width=1\linewidth]{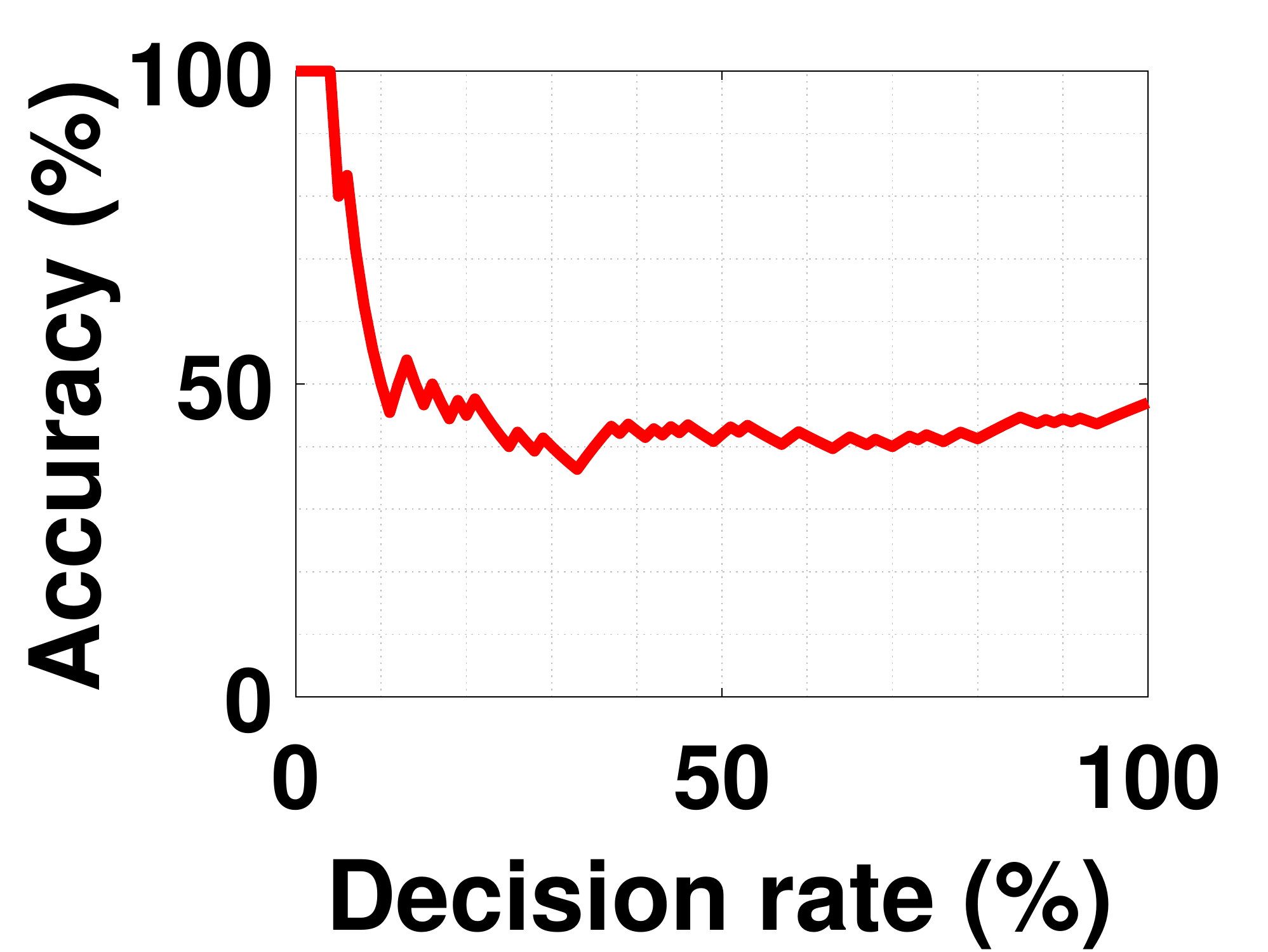}
		{\ttfamily SIM-G}-MON
		\includegraphics[width=1\linewidth]{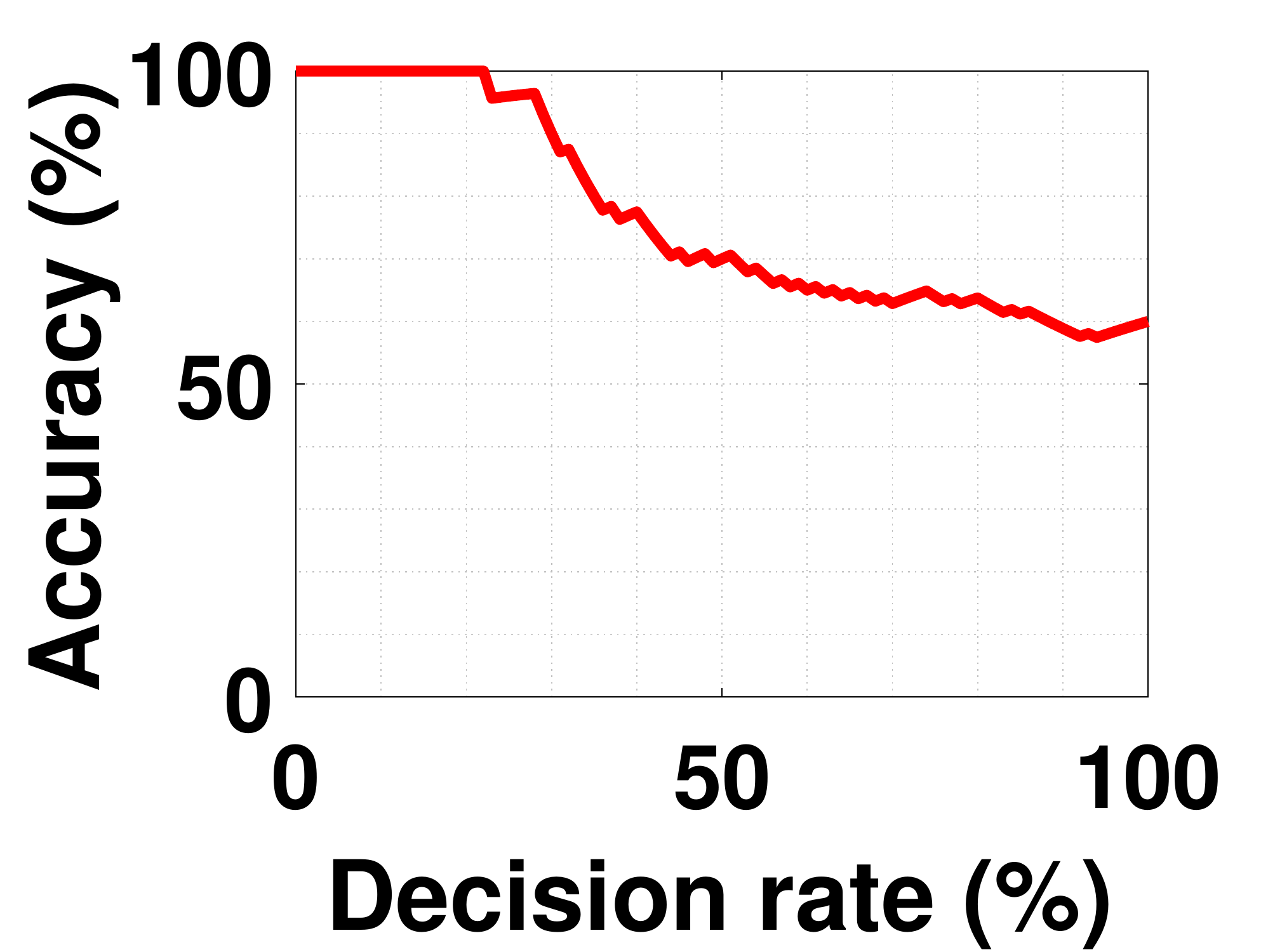}
		{\ttfamily SIM-G}-POLY
		\includegraphics[width=1\linewidth]{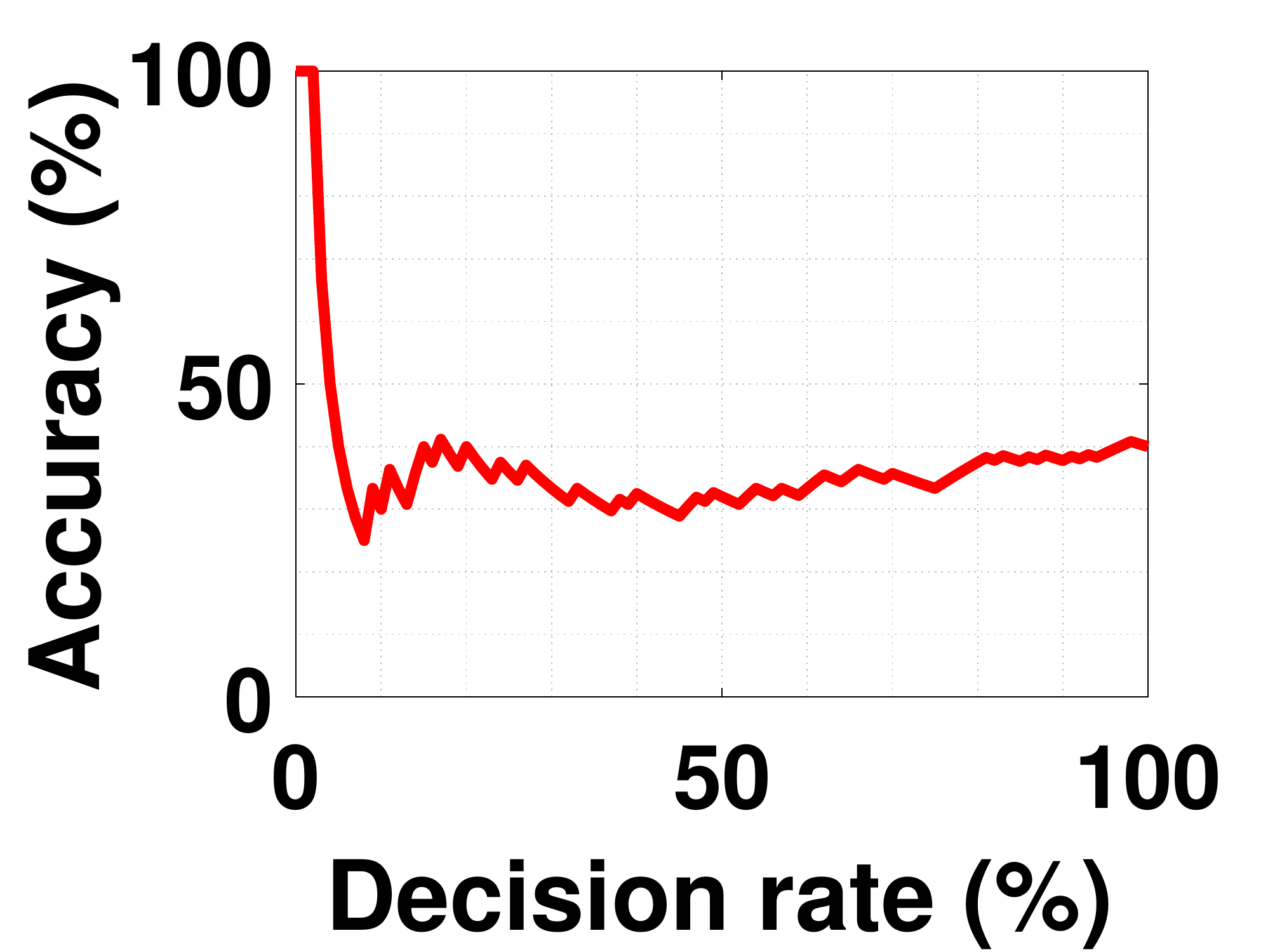}
		{\ttfamily SIM-G}-SVR
		\includegraphics[width=1\linewidth]{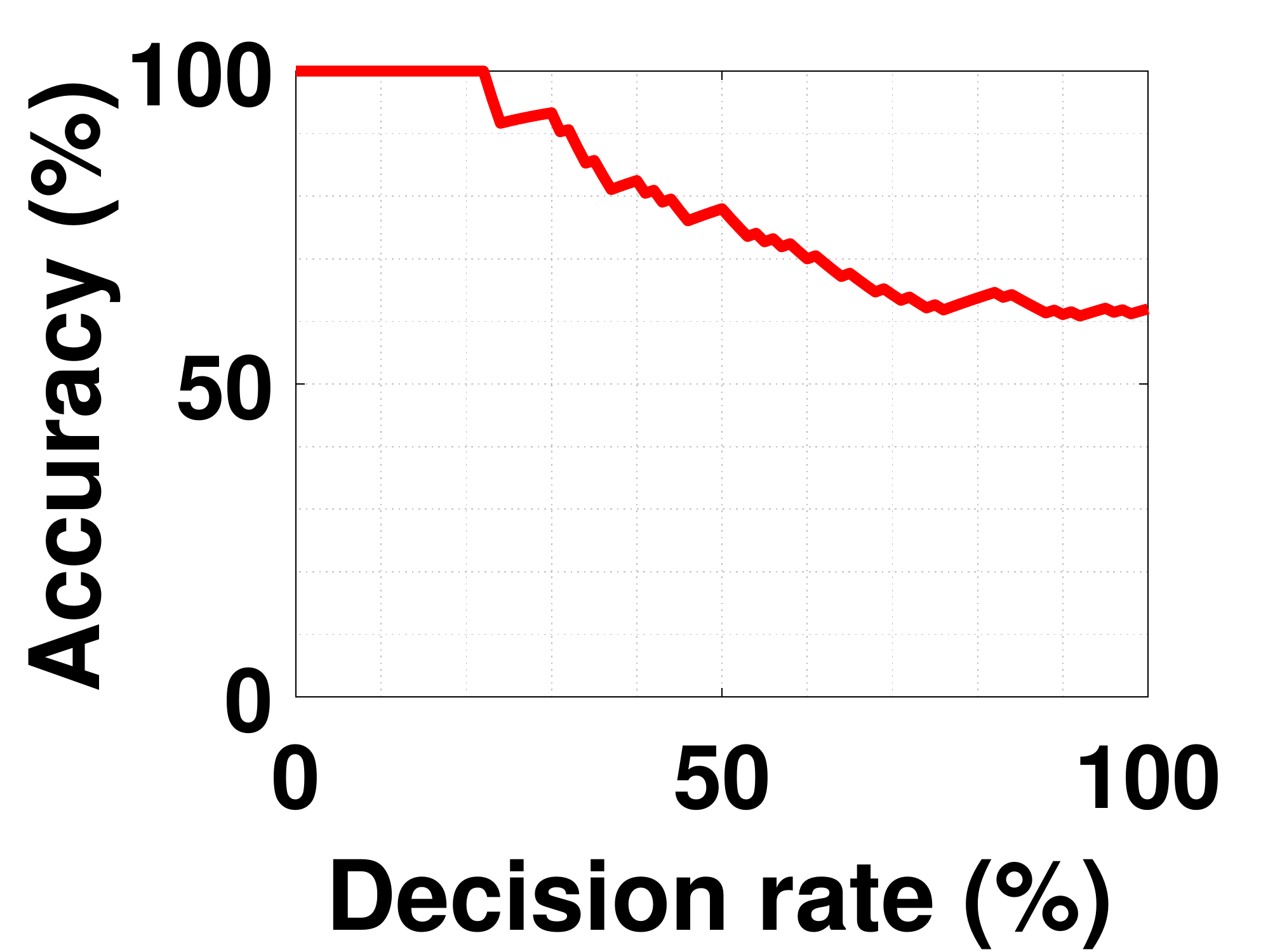}
		{\ttfamily SIM-G}-NN
	\end{multicols}
	\caption{The performance of RECI in the original data sets if a certain decisions rate is forced. Here, the decisions are ranked according to the confidence measure defined in (23).\label{fig:fig2}}
\end{figure}

\begin{figure}[H]
	\begin{multicols}{5}
		\centering
		\includegraphics[width=1\linewidth]{images/plotsRatio/Decision_CEP_OUTLIER_MEAN_LOG.pdf}
		{\ttfamily CEP}-LOG
		\includegraphics[width=1\linewidth]{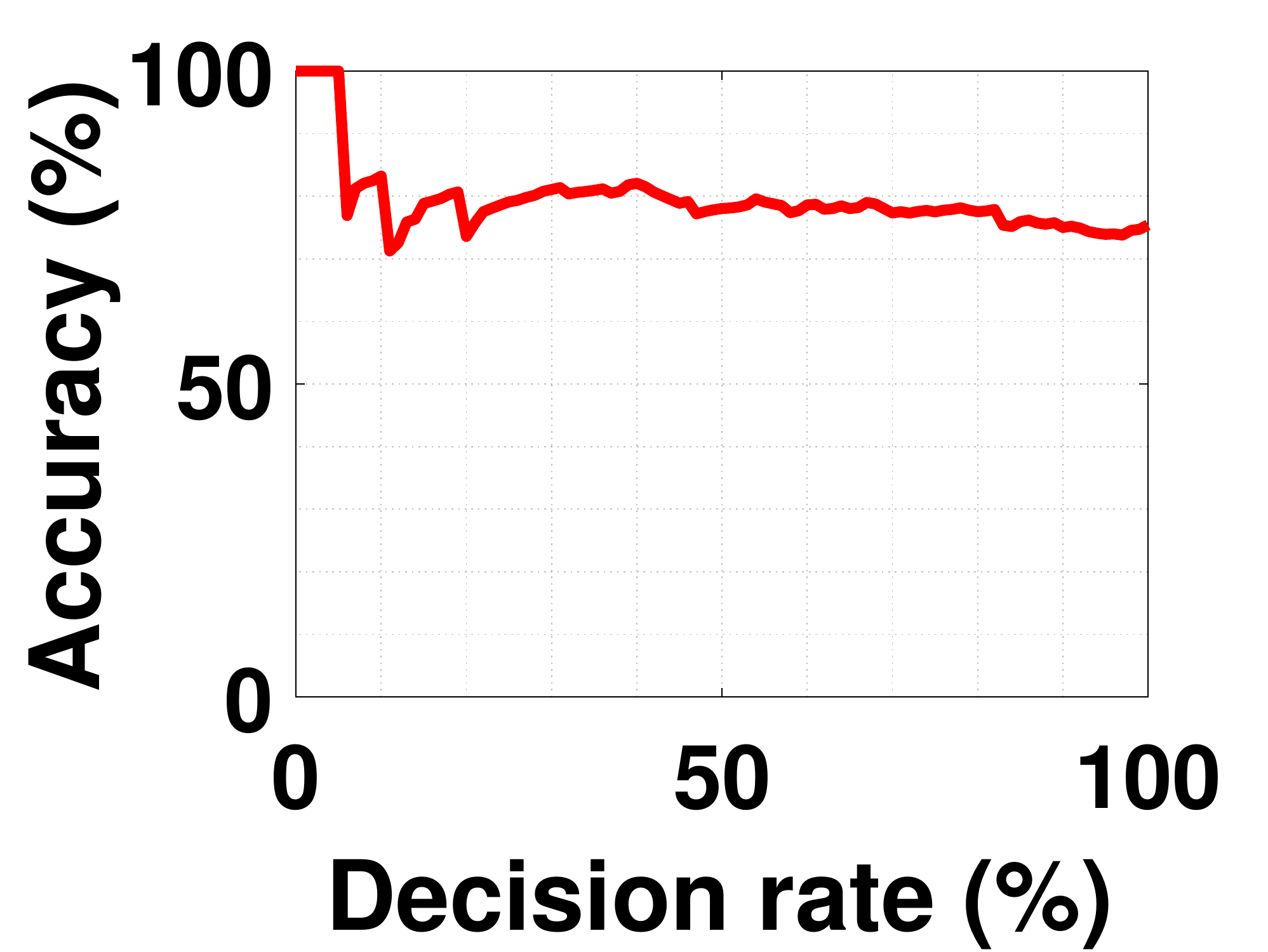}
		{\ttfamily CEP}-MON
		\includegraphics[width=1\linewidth]{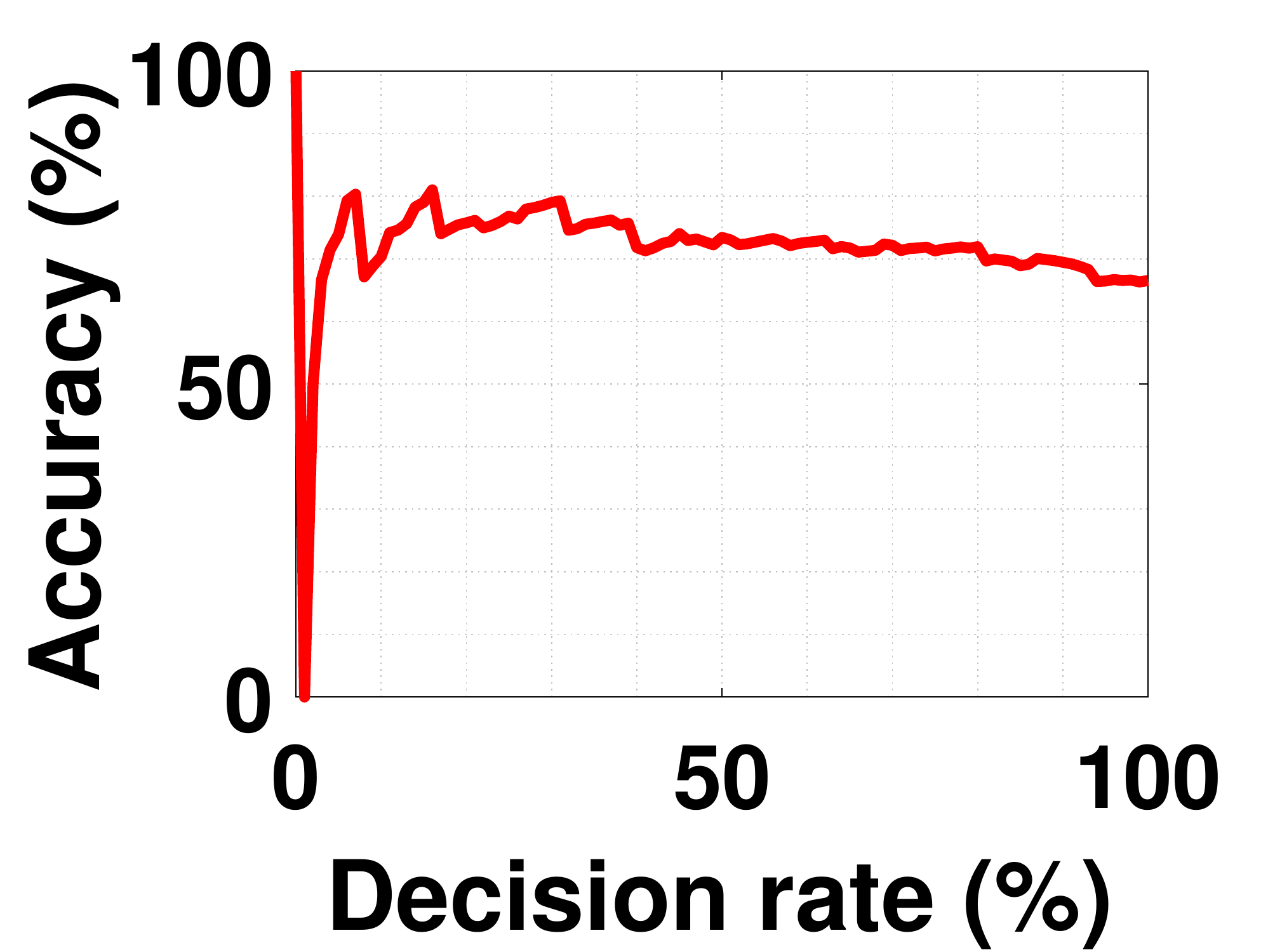}
		{\ttfamily CEP}-POLY
		\includegraphics[width=1\linewidth]{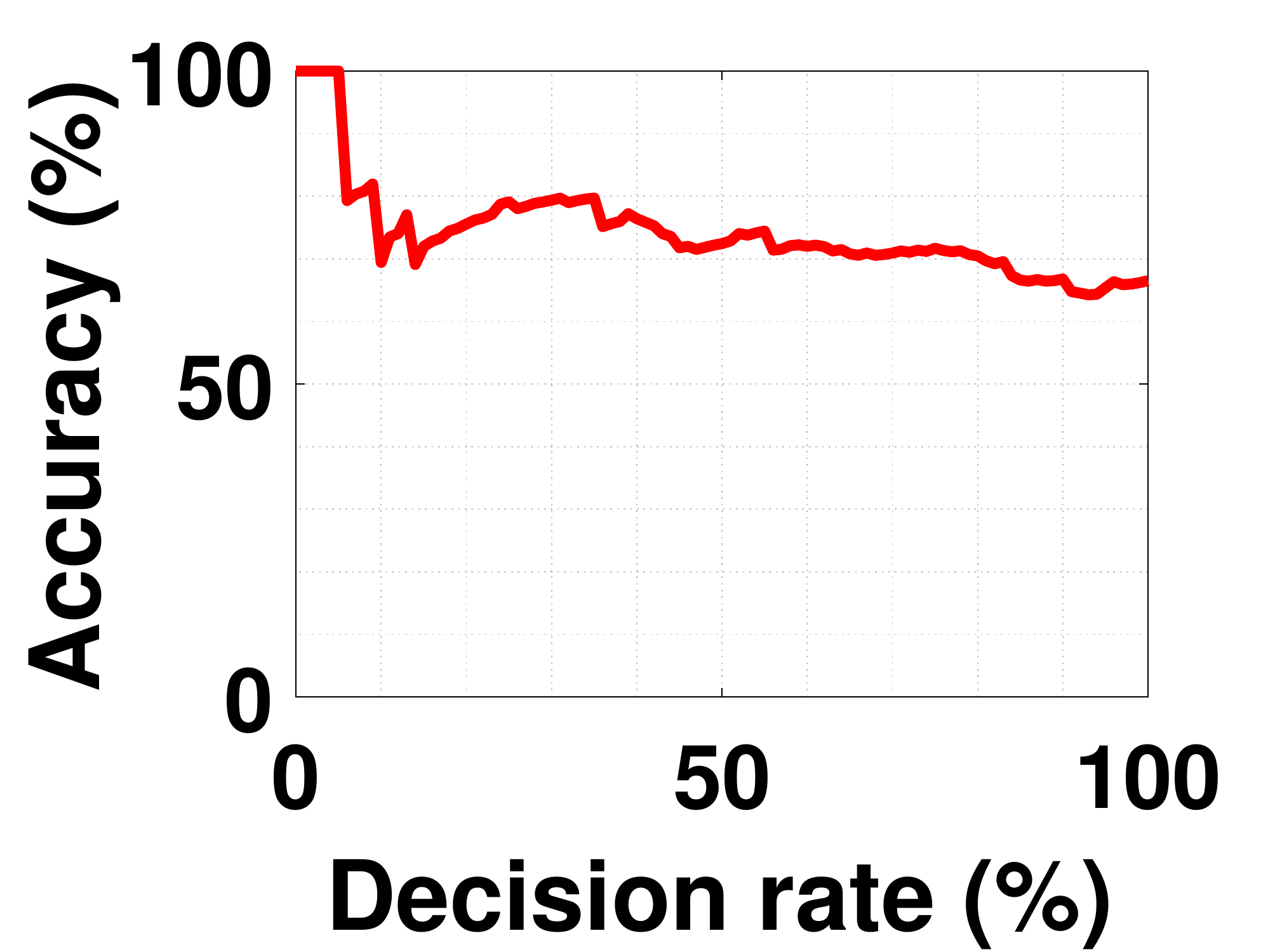}
		{\ttfamily CEP}-SVR
		\includegraphics[width=1\linewidth]{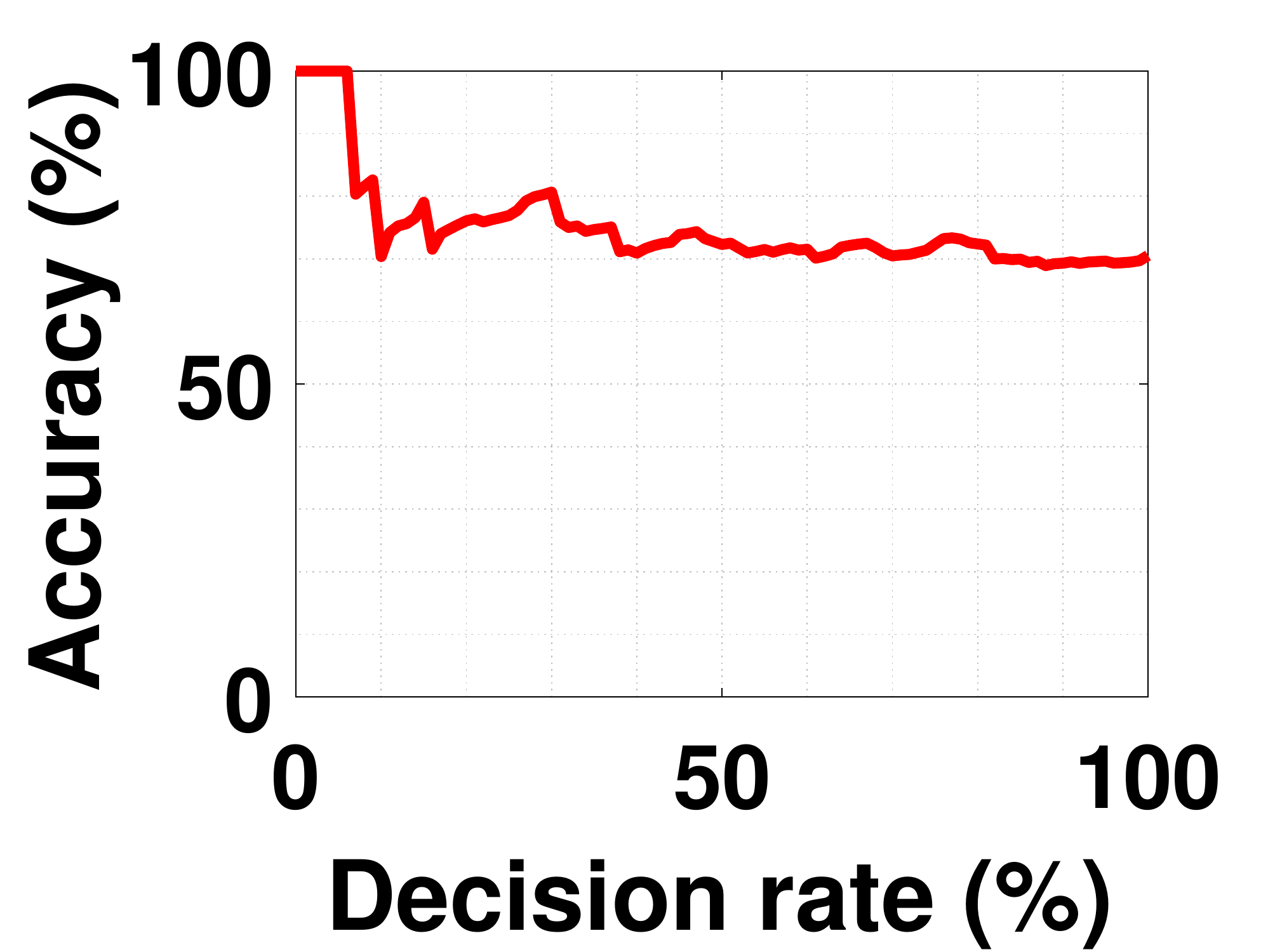}
		{\ttfamily CEP}-NN
	\end{multicols}
	
	\begin{multicols}{5}
		\centering
		\includegraphics[width=1\linewidth]{images/plotsRatio/Decision_SIM_OUTLIER_MEAN_LOG.pdf}
		{\ttfamily SIM}-LOG
		\includegraphics[width=1\linewidth]{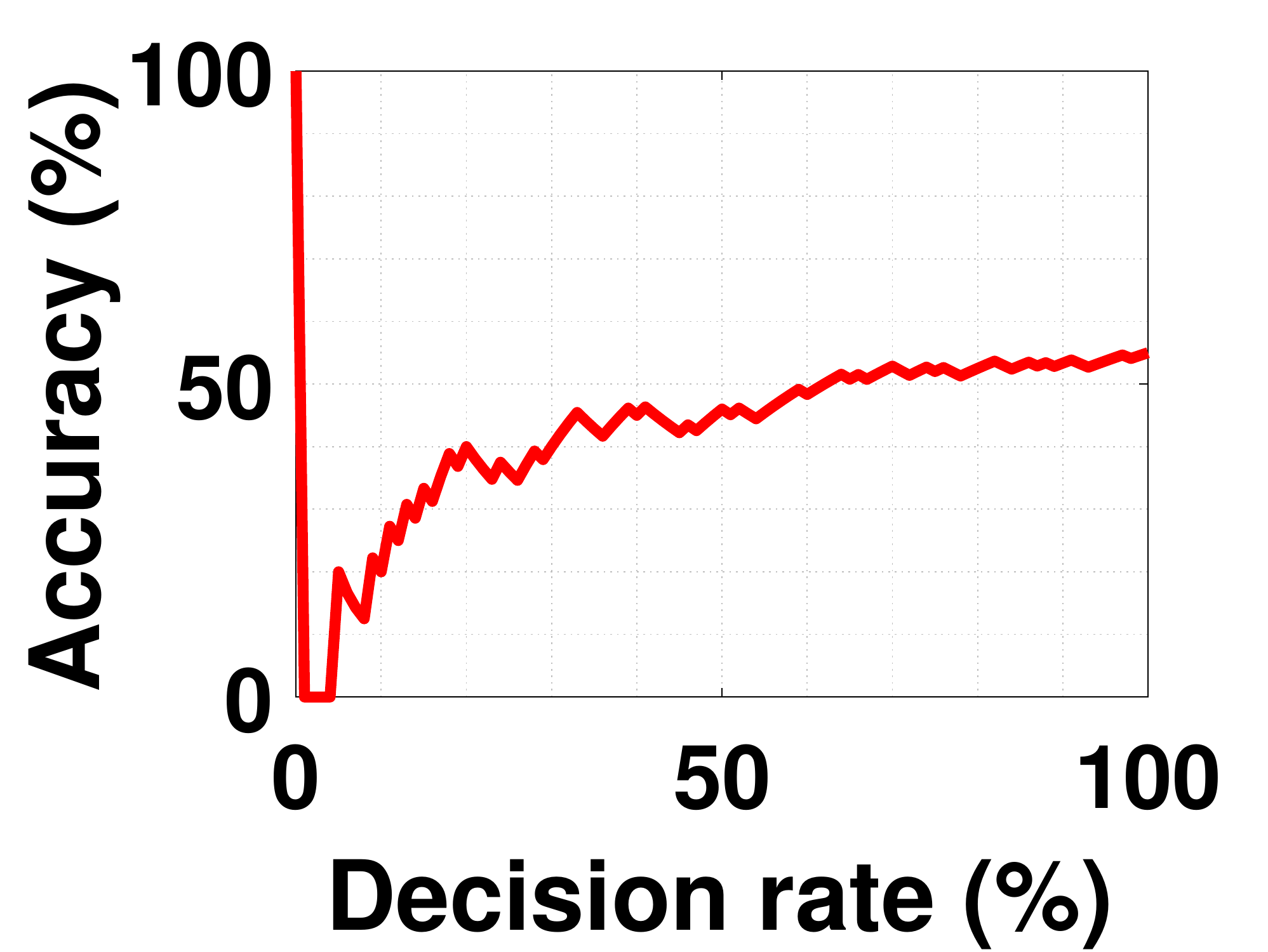}
		{\ttfamily SIM}-MON
		\includegraphics[width=1\linewidth]{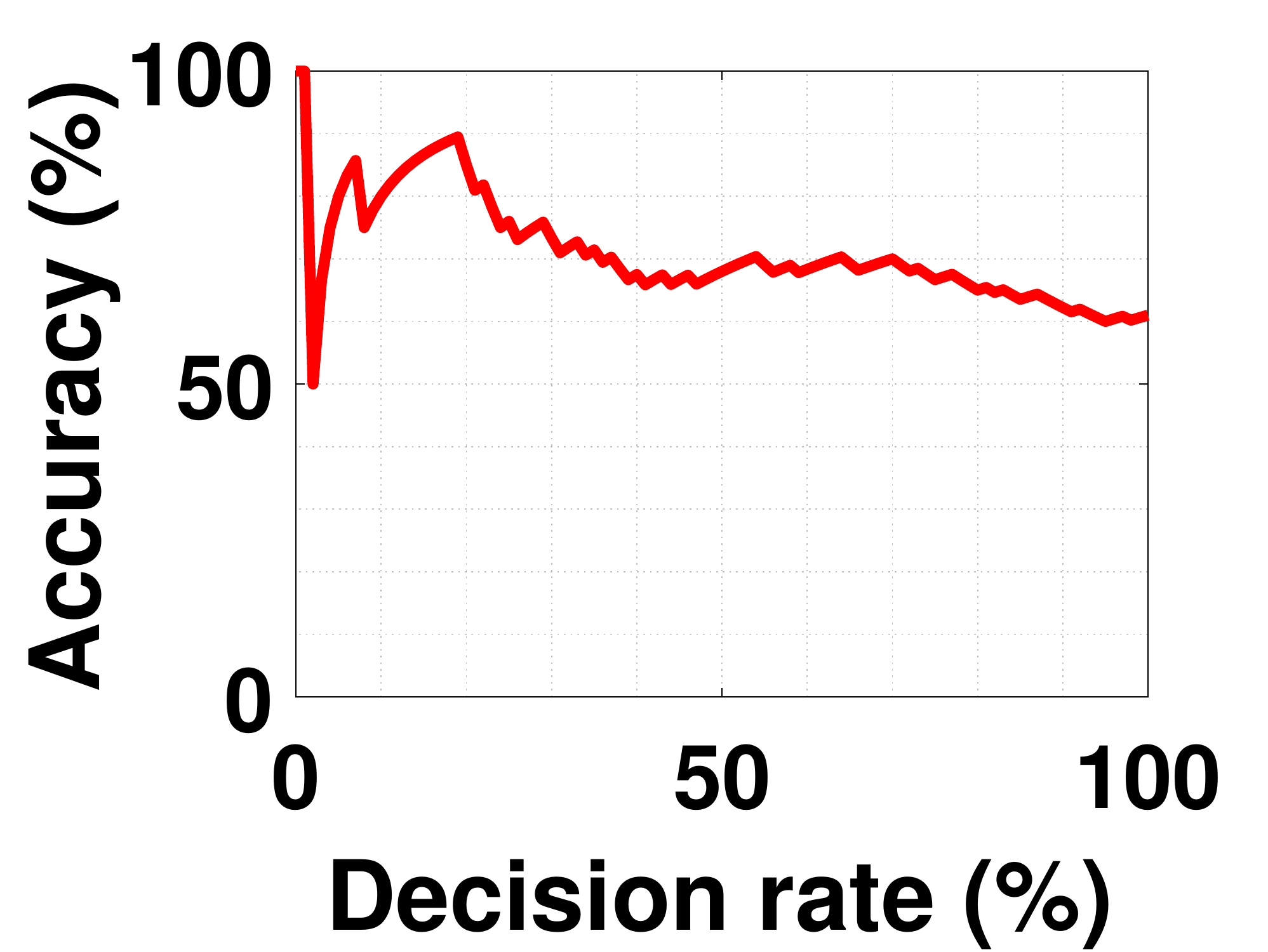}
		{\ttfamily SIM}-POLY
		\includegraphics[width=1\linewidth]{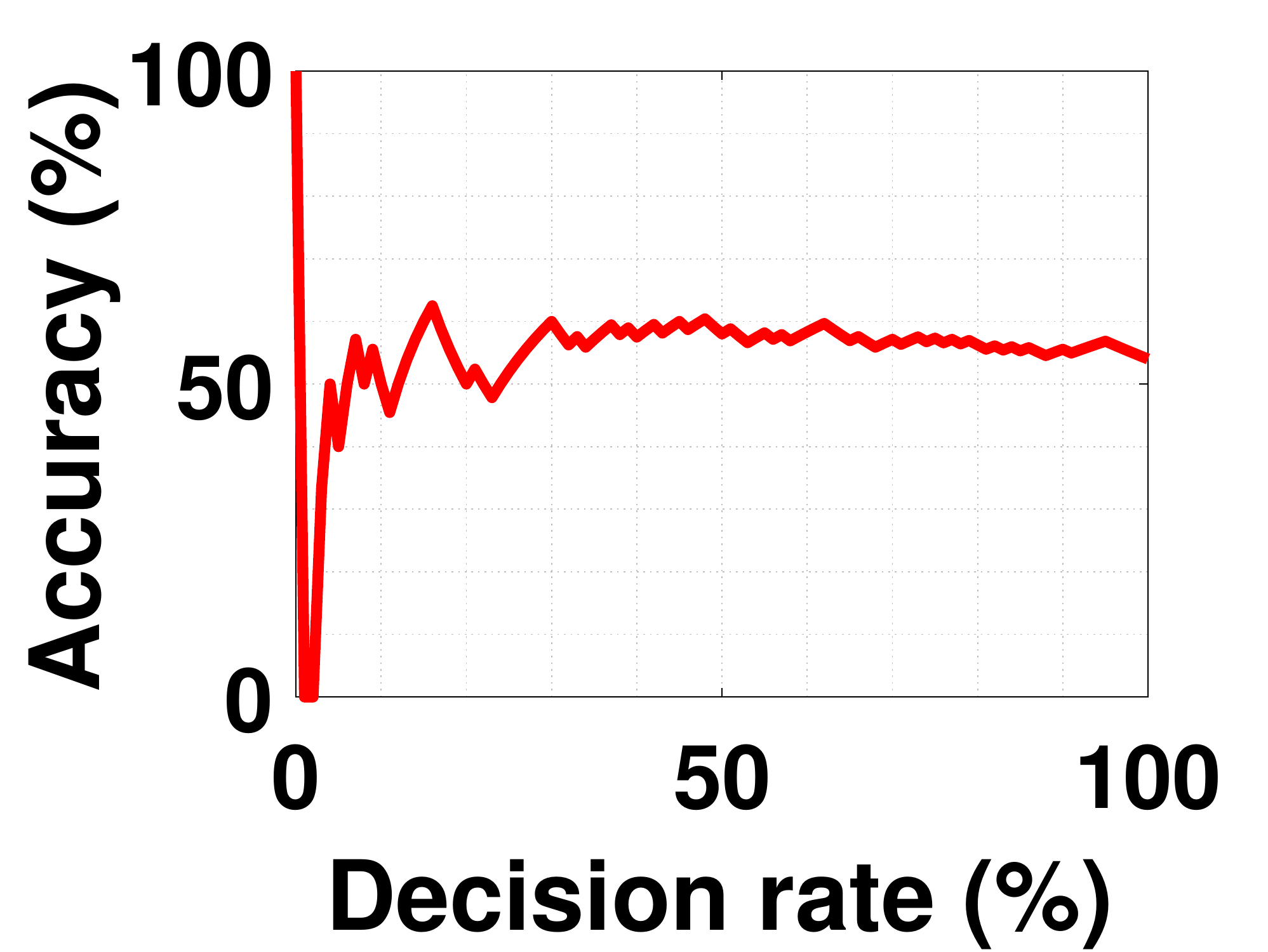}
		{\ttfamily SIM}-SVR
		\includegraphics[width=1\linewidth]{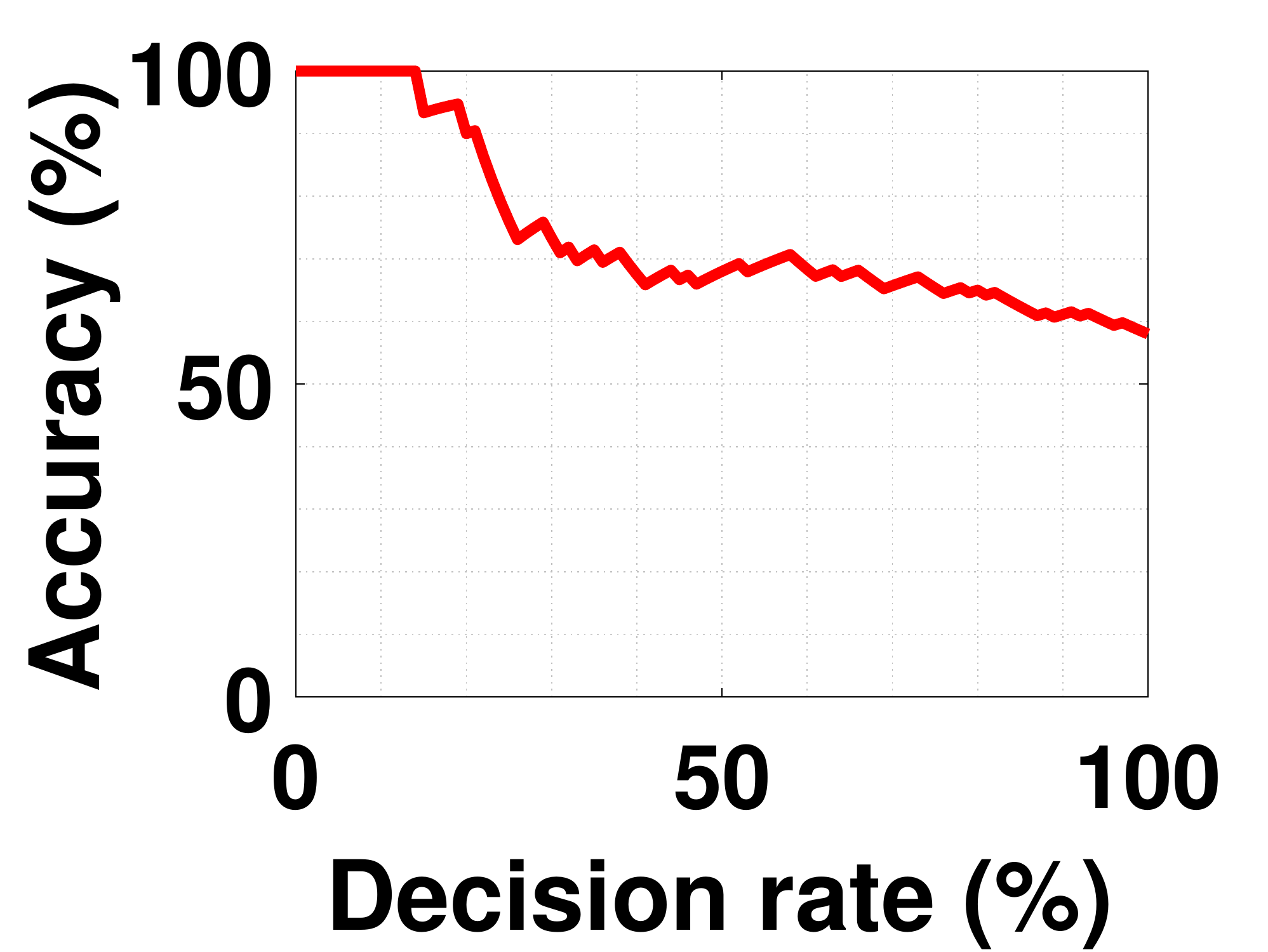}
		{\ttfamily SIM}-NN
	\end{multicols}
	
	\begin{multicols}{5}
		\centering
		\includegraphics[width=1\linewidth]{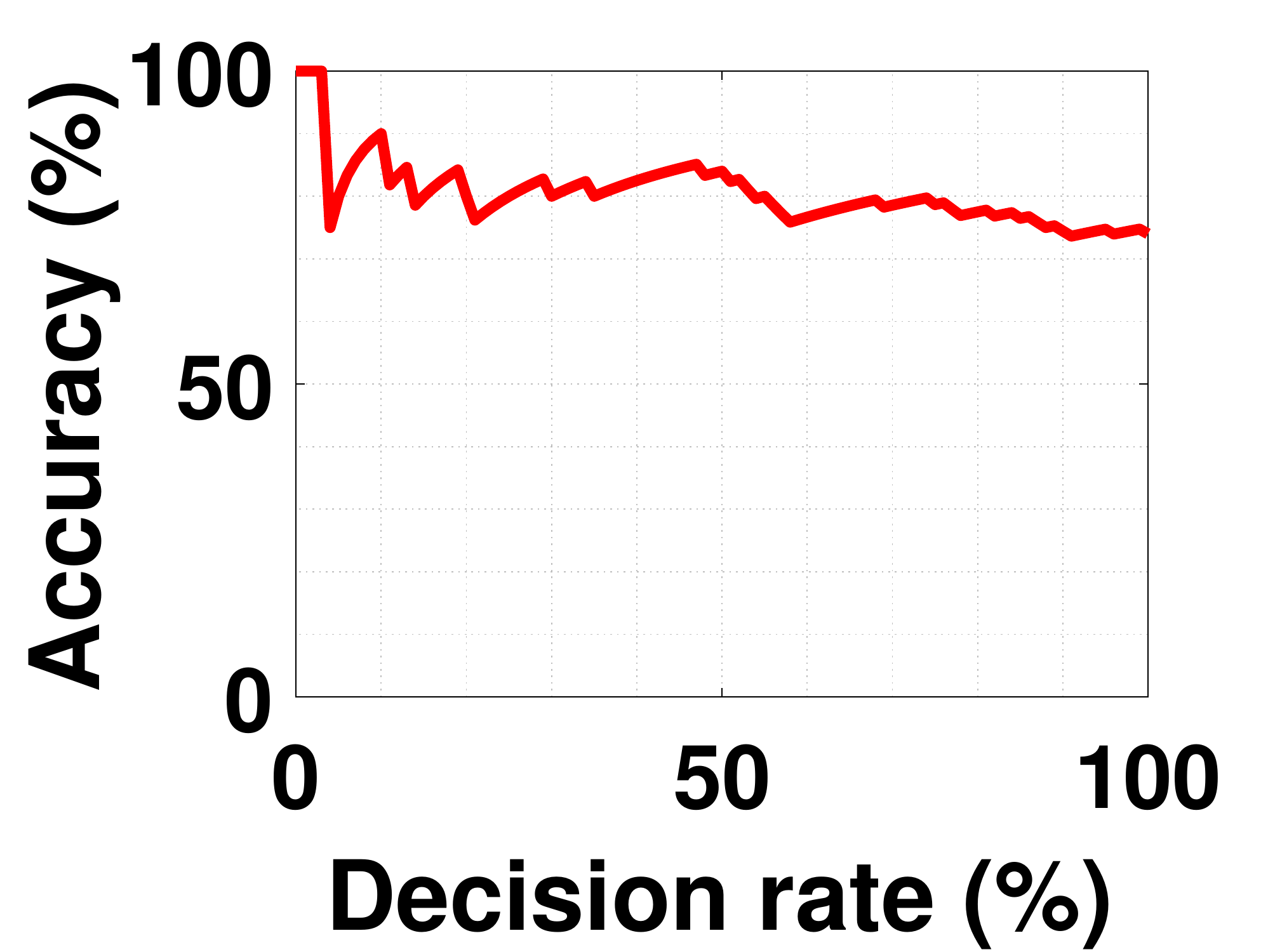}
		{\ttfamily SIM-c}-LOG
		\includegraphics[width=1\linewidth]{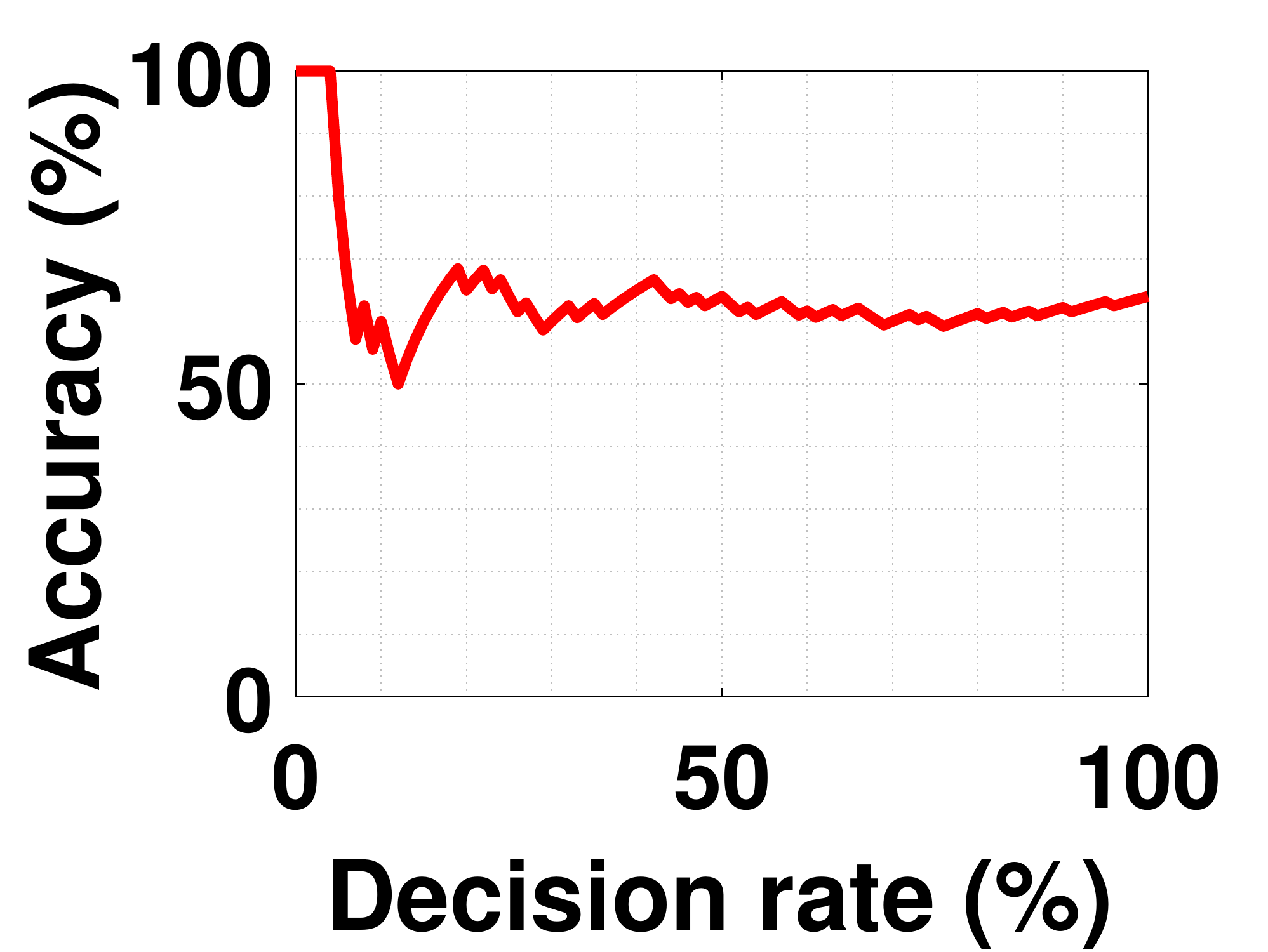}
		{\ttfamily SIM-c}-MON
		\includegraphics[width=1\linewidth]{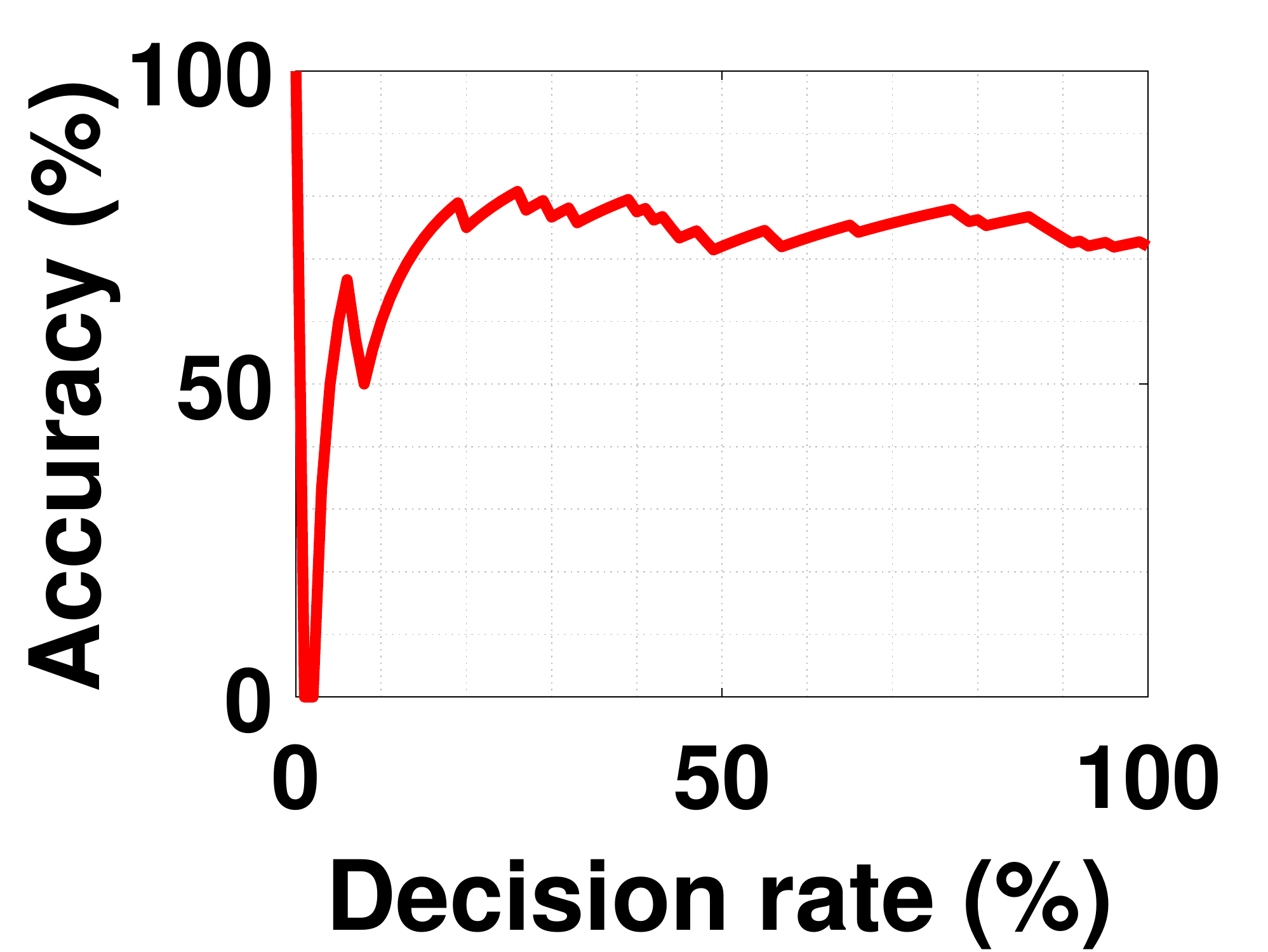}
		{\ttfamily SIM-c}-POLY
		\includegraphics[width=1\linewidth]{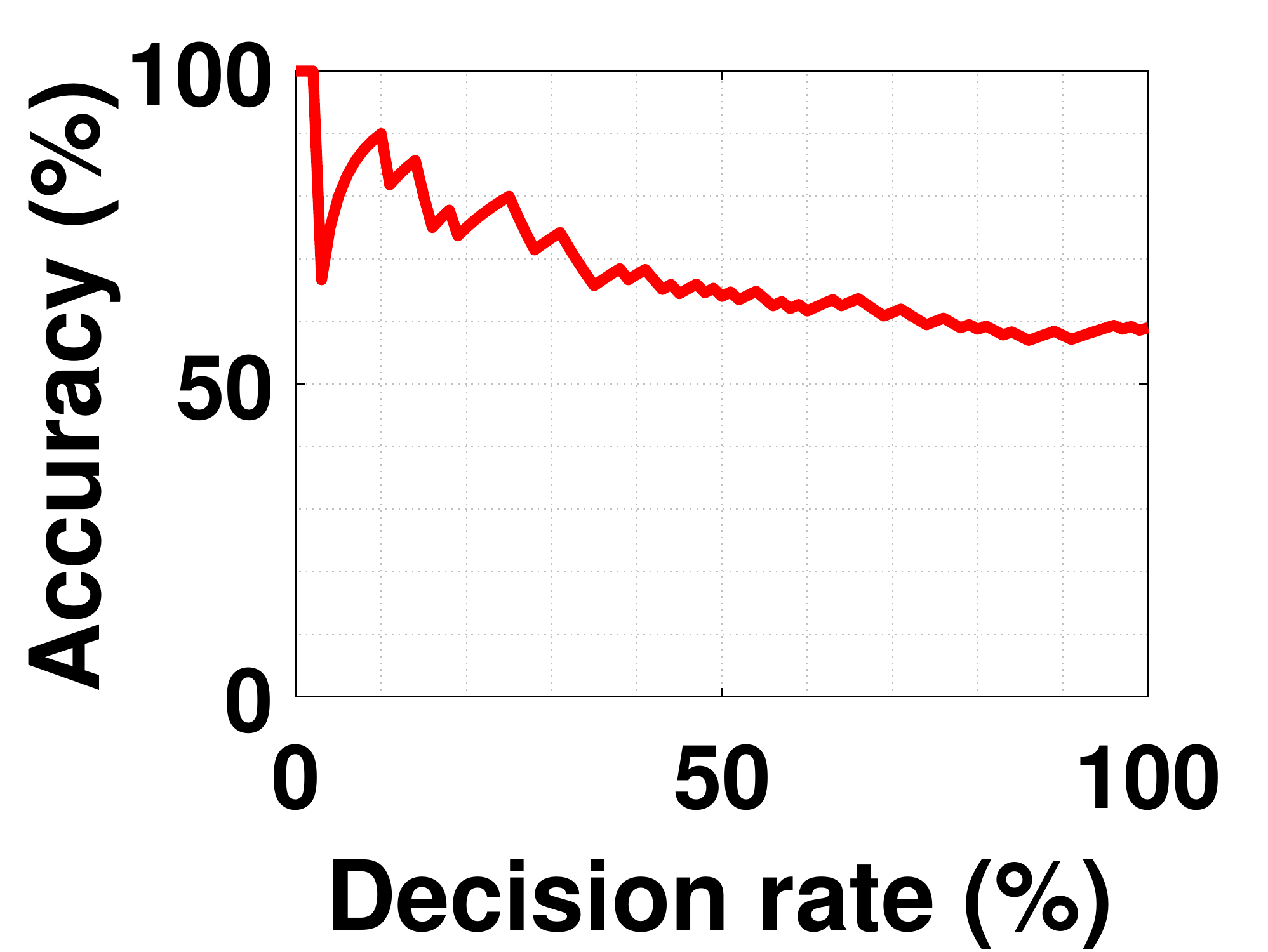}
		{\ttfamily SIM-c}-SVR
		\includegraphics[width=1\linewidth]{images/plotsRatio/Decision_SIMC_OUTLIER_MEAN_NN.pdf}
		{\ttfamily SIM-c}-NN
	\end{multicols}
	
	\begin{multicols}{5}
		\centering
		\includegraphics[width=1\linewidth]{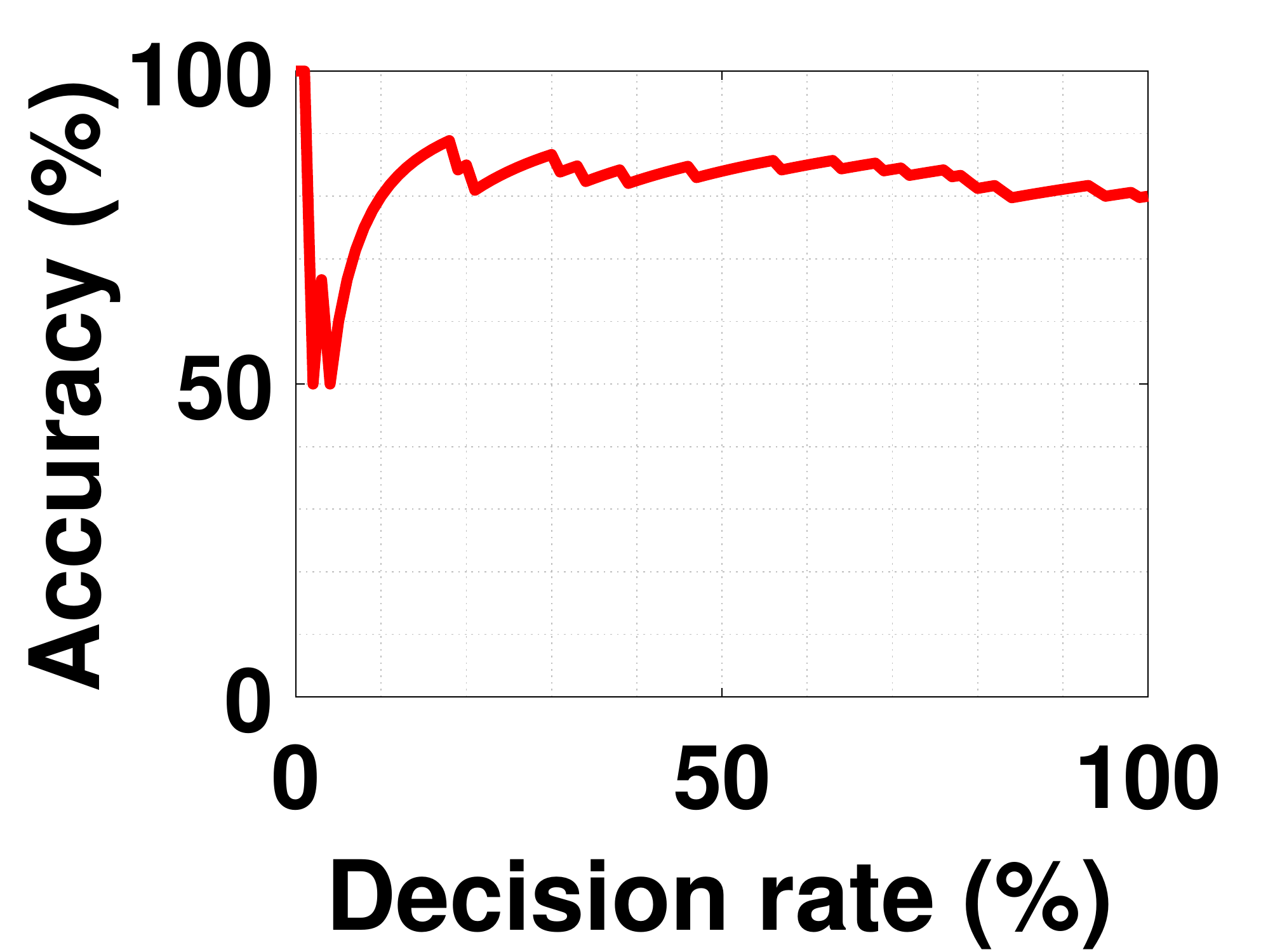}
		{\ttfamily SIM-ln}-LOG
		\includegraphics[width=1\linewidth]{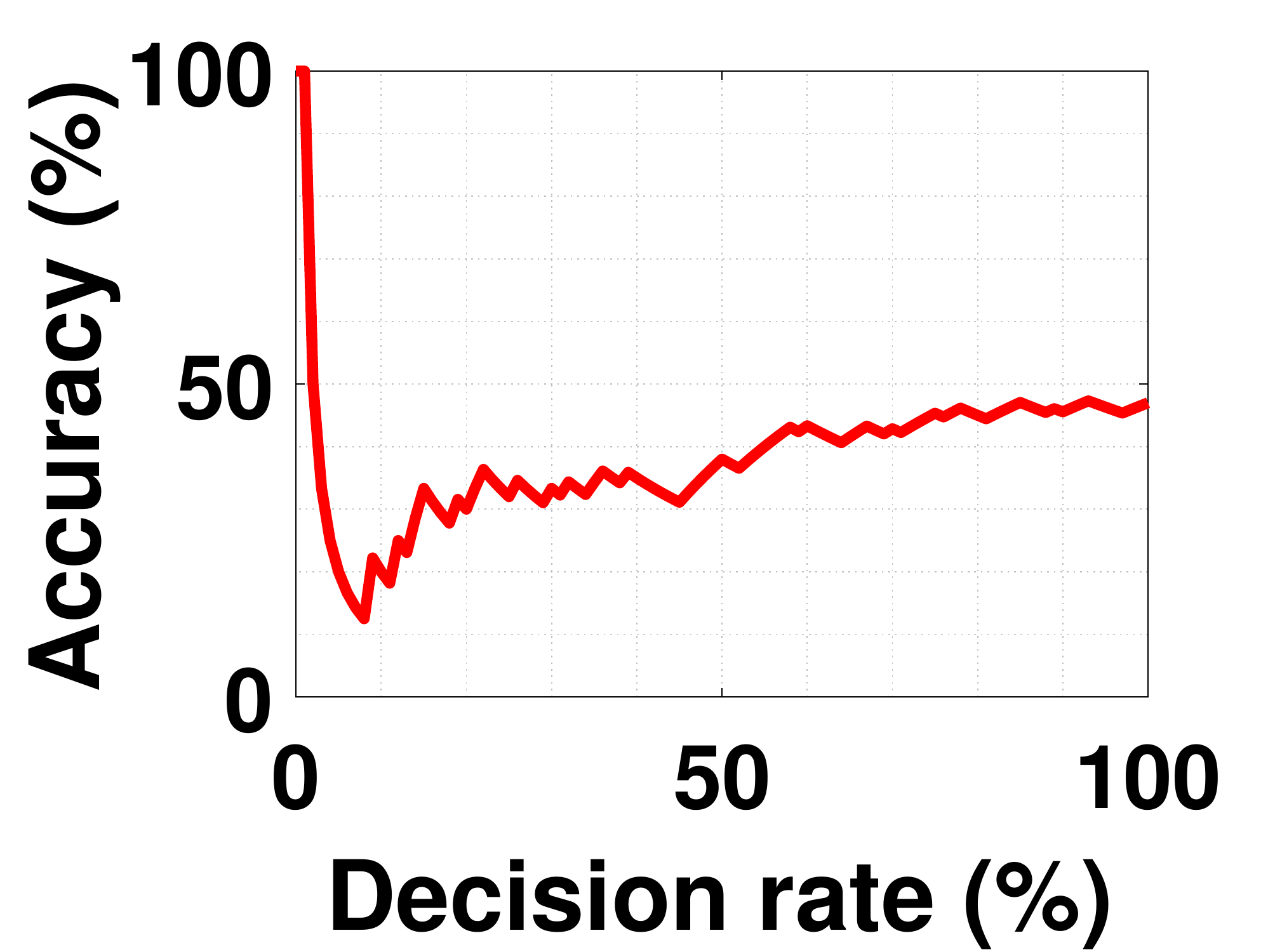}
		{\ttfamily SIM-ln}-MON
		\includegraphics[width=1\linewidth]{images/plotsRatio/Decision_SIMLN_OUTLIER_MEAN_POLY.pdf}
		{\ttfamily SIM-ln}-POLY
		\includegraphics[width=1\linewidth]{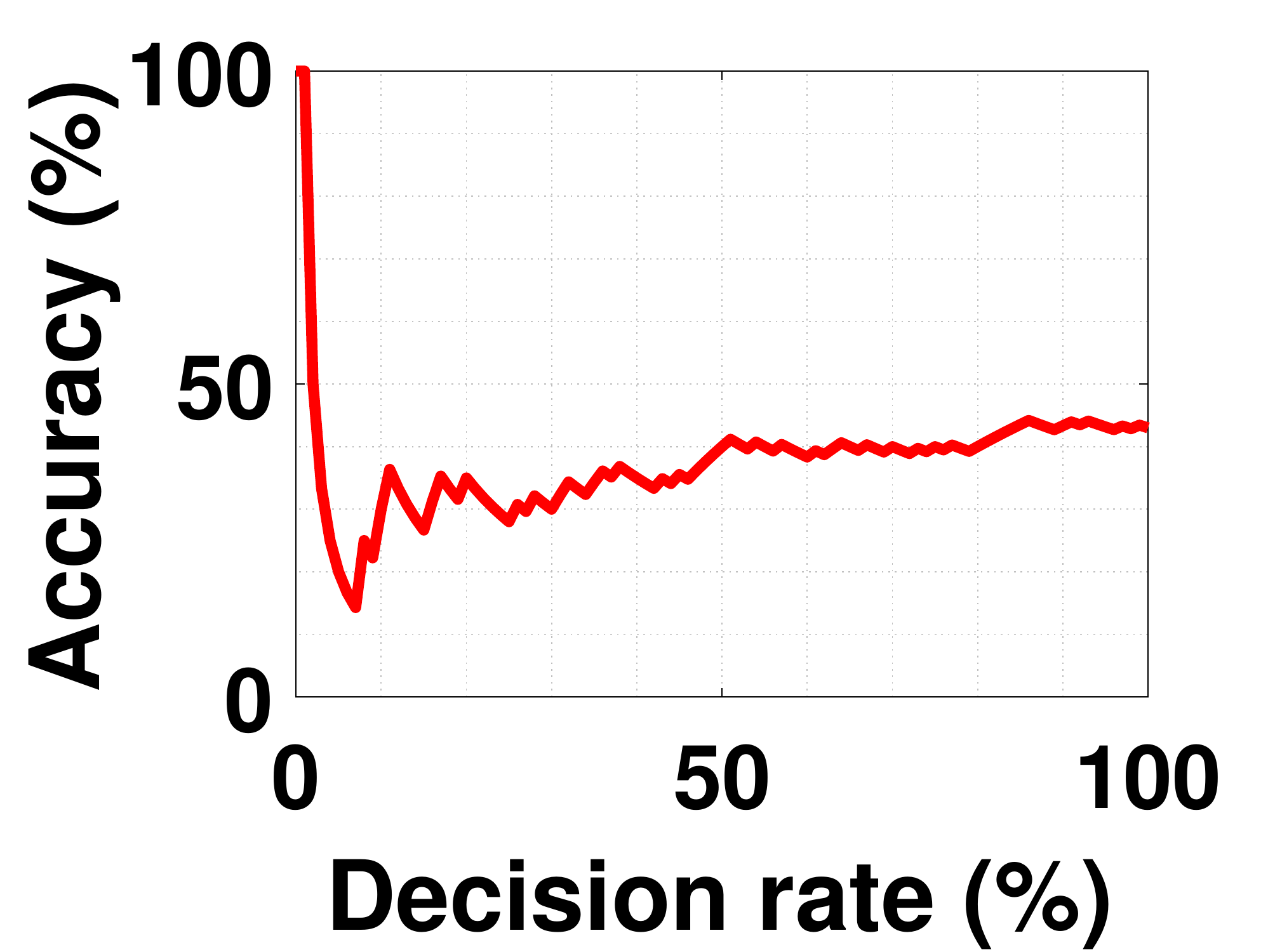}
		{\ttfamily SIM-ln}-SVR
		\includegraphics[width=1\linewidth]{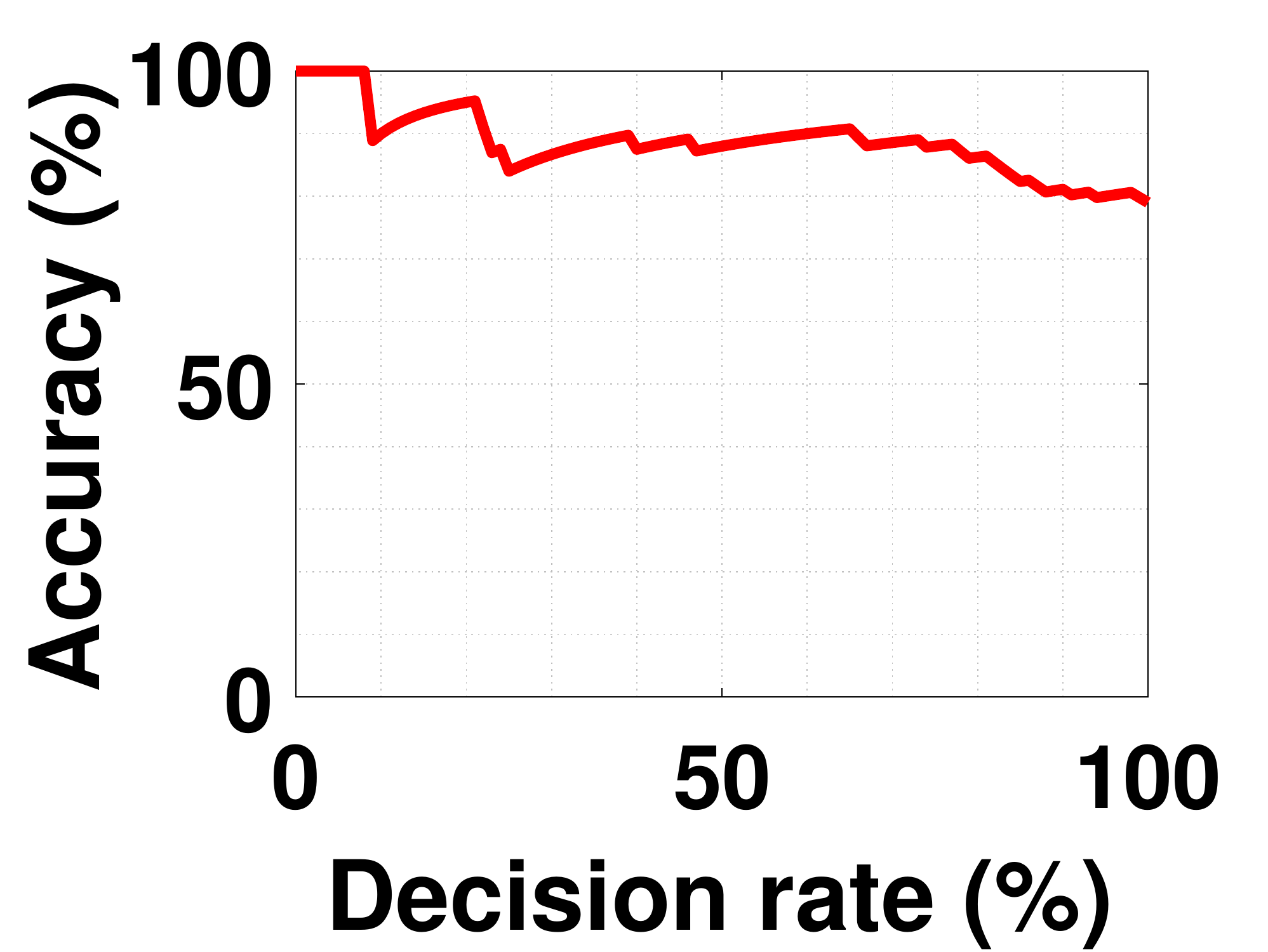}
		{\ttfamily SIM-ln}-NN
	\end{multicols}
	
	\begin{multicols}{5}
		\centering
		\includegraphics[width=1\linewidth]{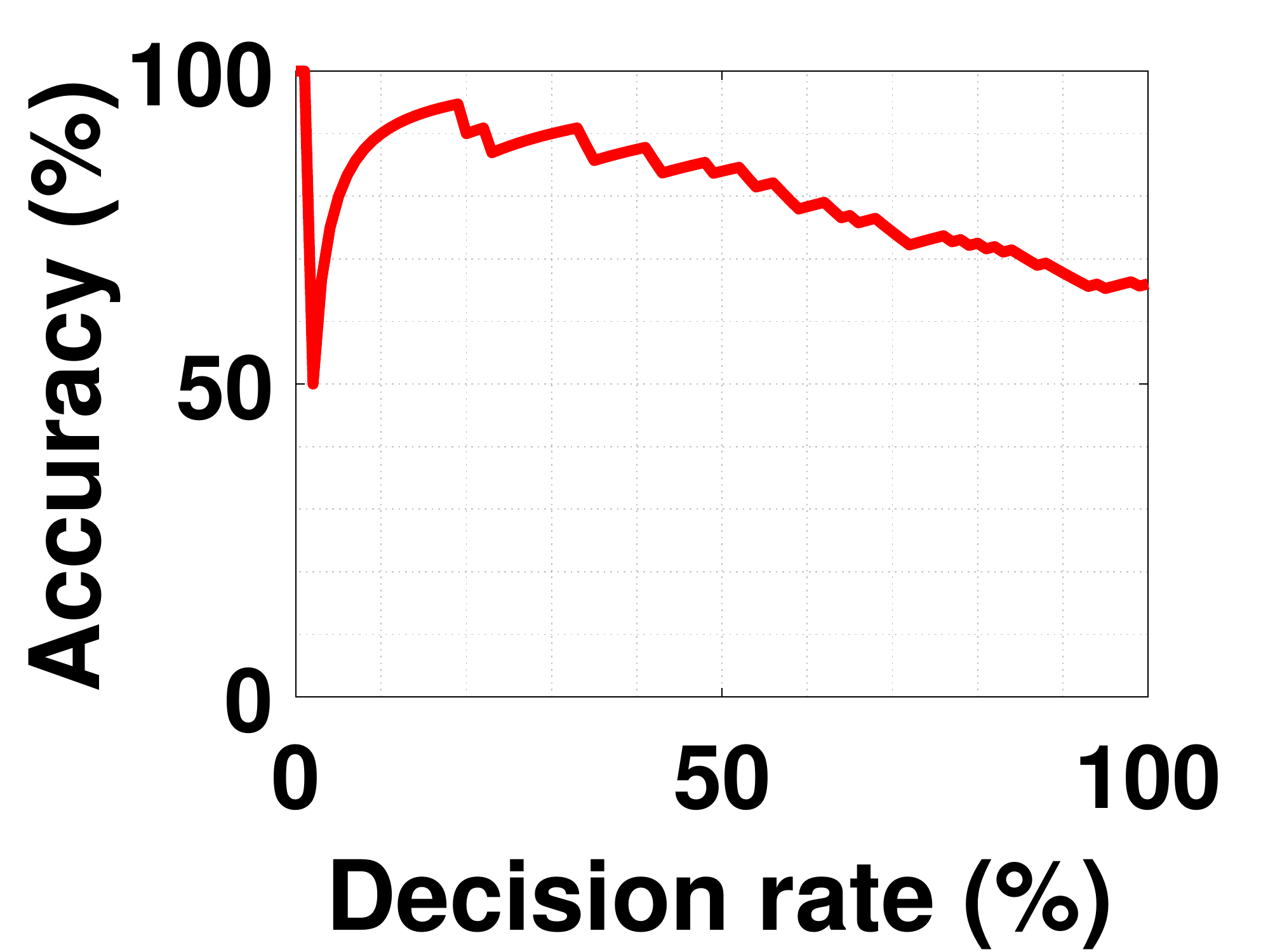}
		{\ttfamily SIM-G}-LOG
		\includegraphics[width=1\linewidth]{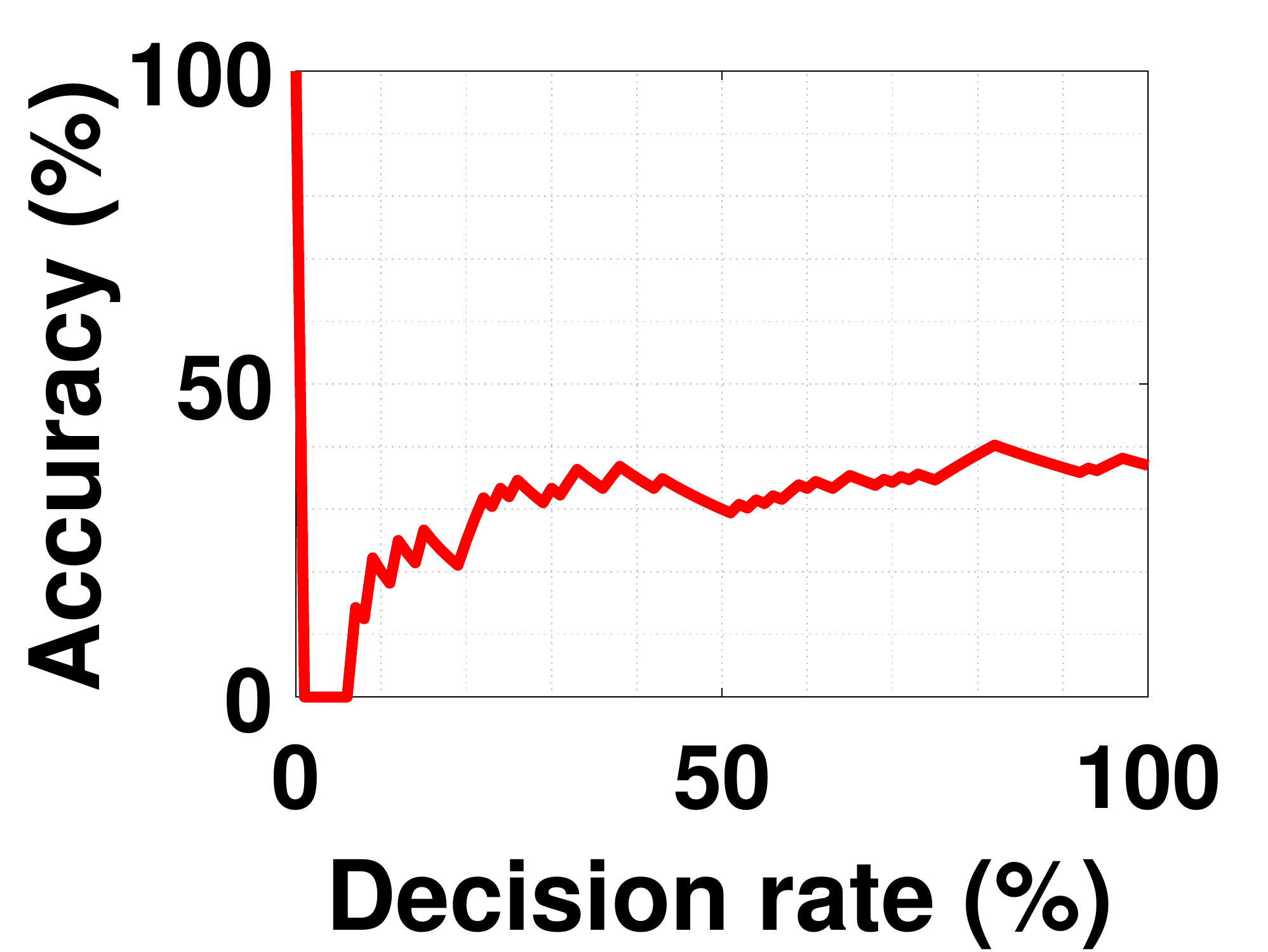}
		{\ttfamily SIM-G}-MON
		\includegraphics[width=1\linewidth]{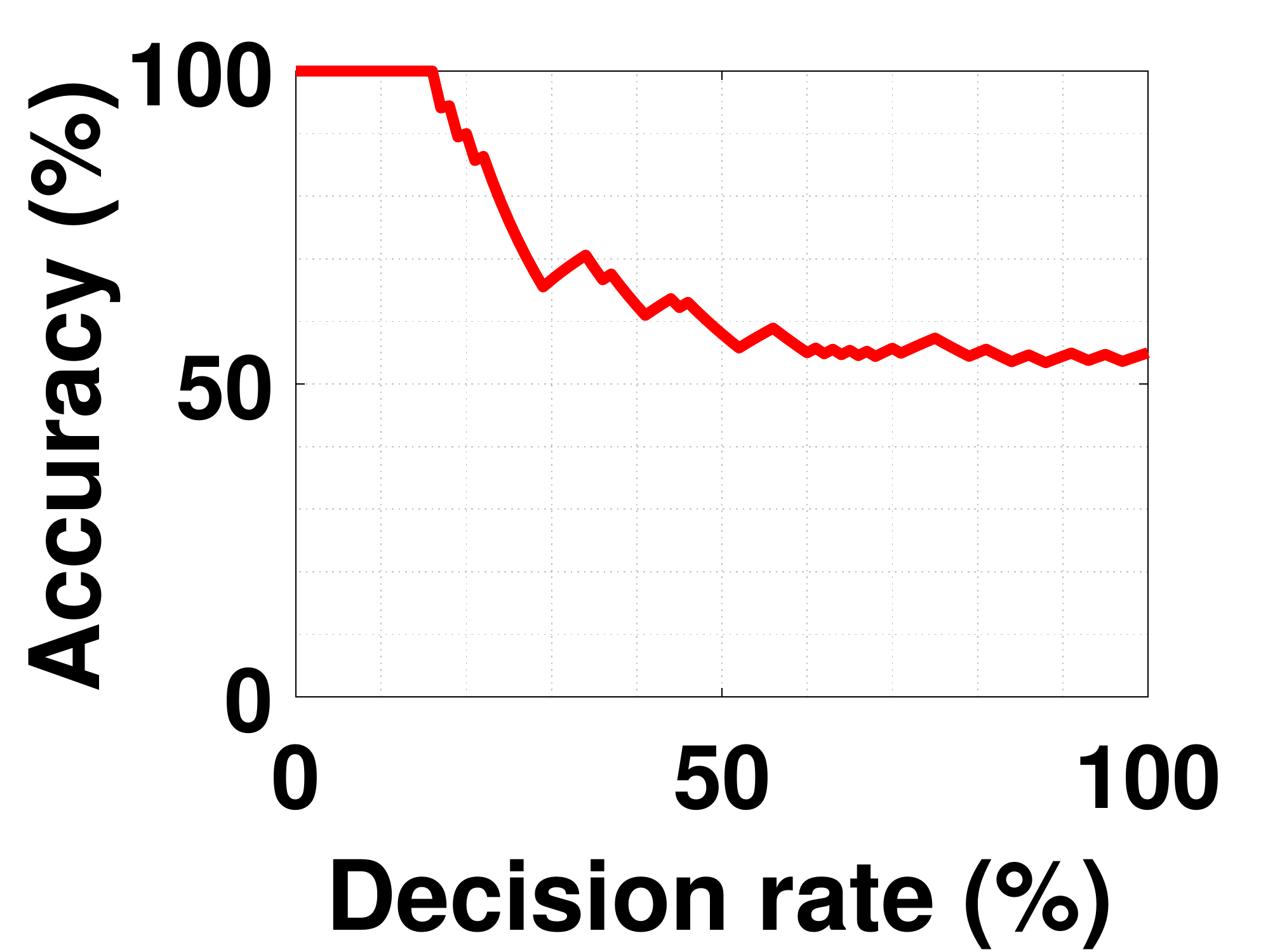}
		{\ttfamily SIM-G}-POLY
		\includegraphics[width=1\linewidth]{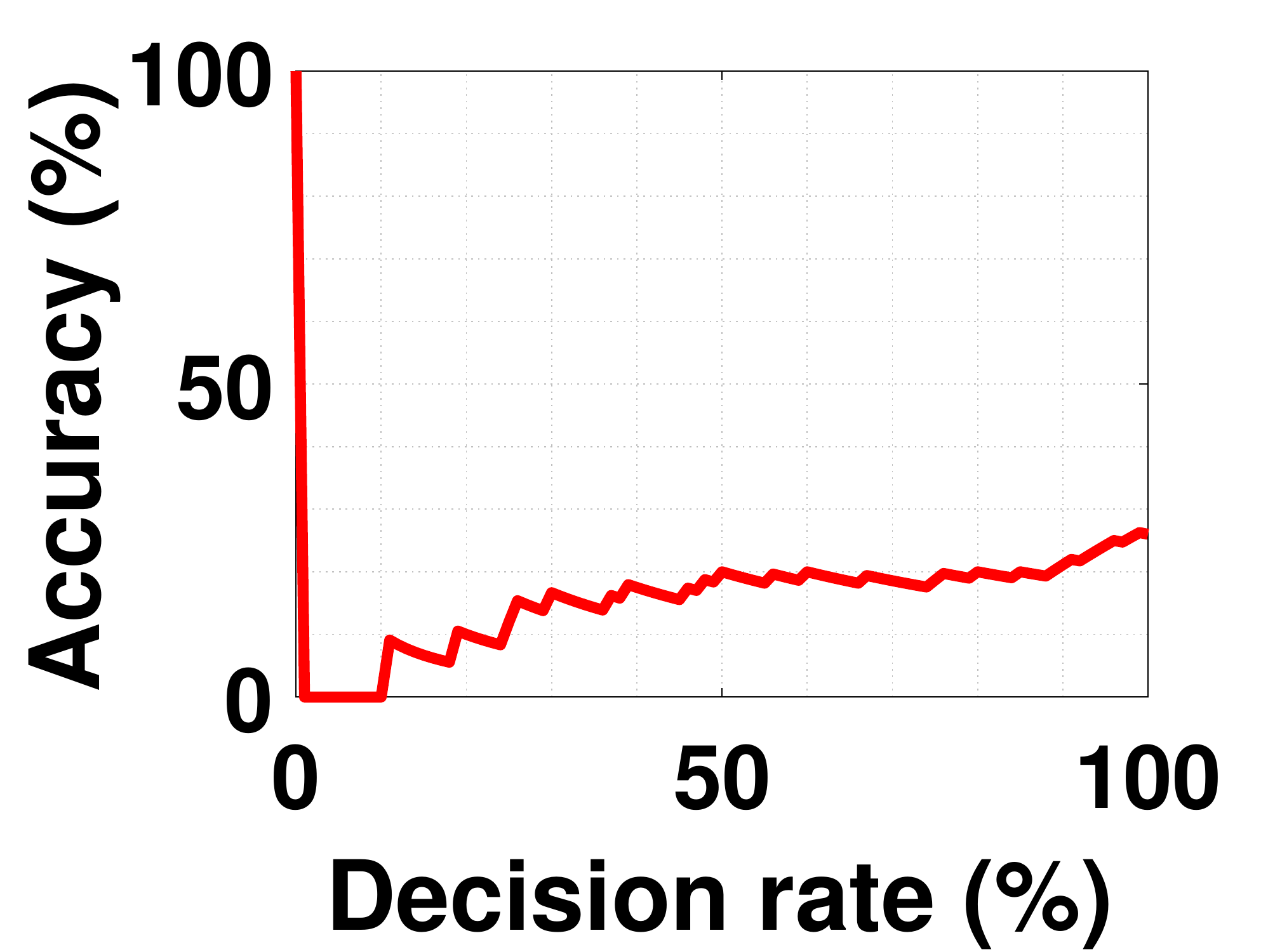}
		{\ttfamily SIM-G}-SVR
		\includegraphics[width=1\linewidth]{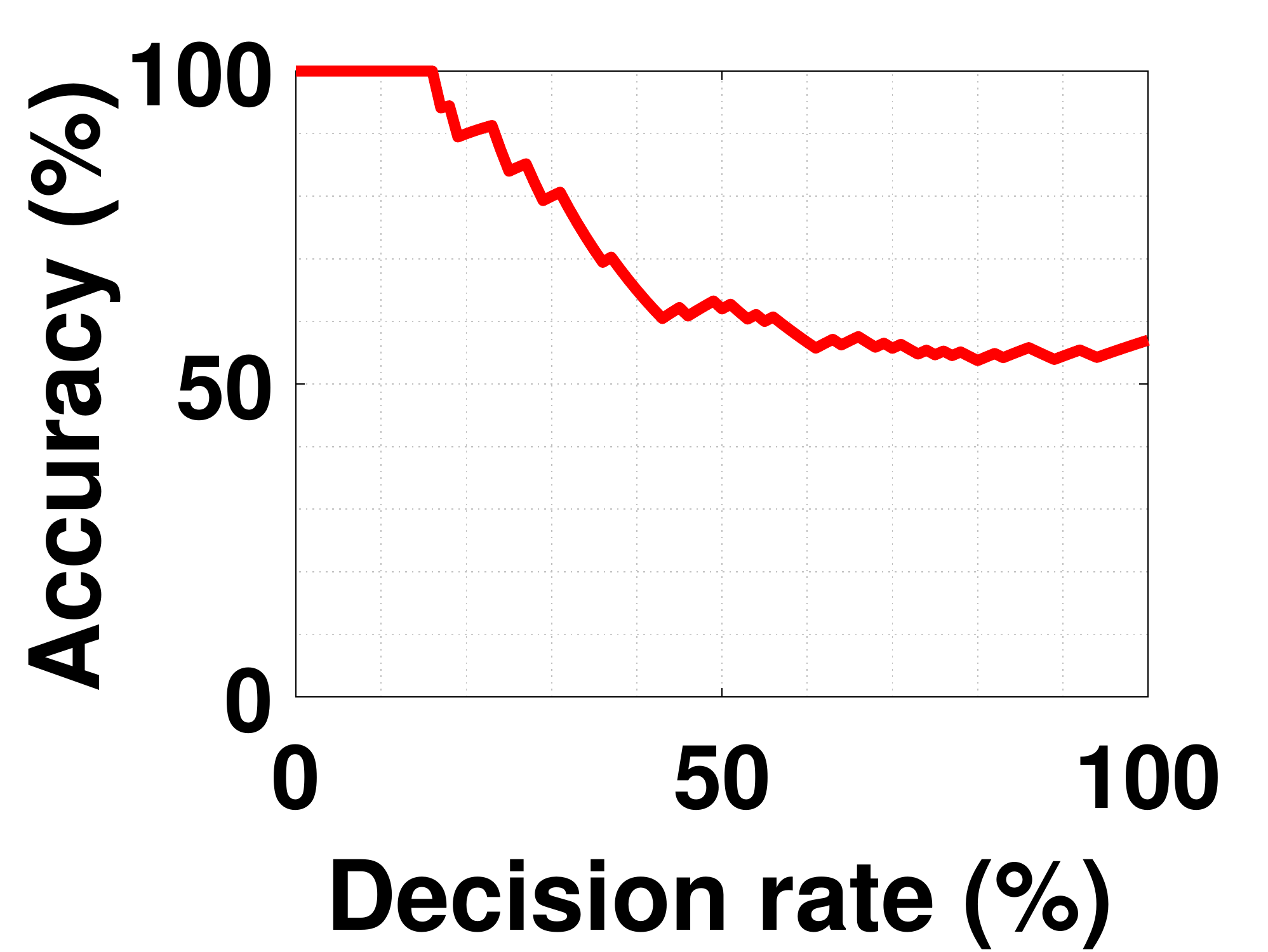}
		{\ttfamily SIM-G}-NN
	\end{multicols}
	\caption{The performance of RECI in the preprocessed data sets if a certain decisions rate is forced. Here, the decisions are ranked according to the confidence measure defined in (23).\label{fig:fig3}}
\end{figure}

\begin{figure}[H]
	\begin{multicols}{5}
		\centering
		\includegraphics[width=1\linewidth]{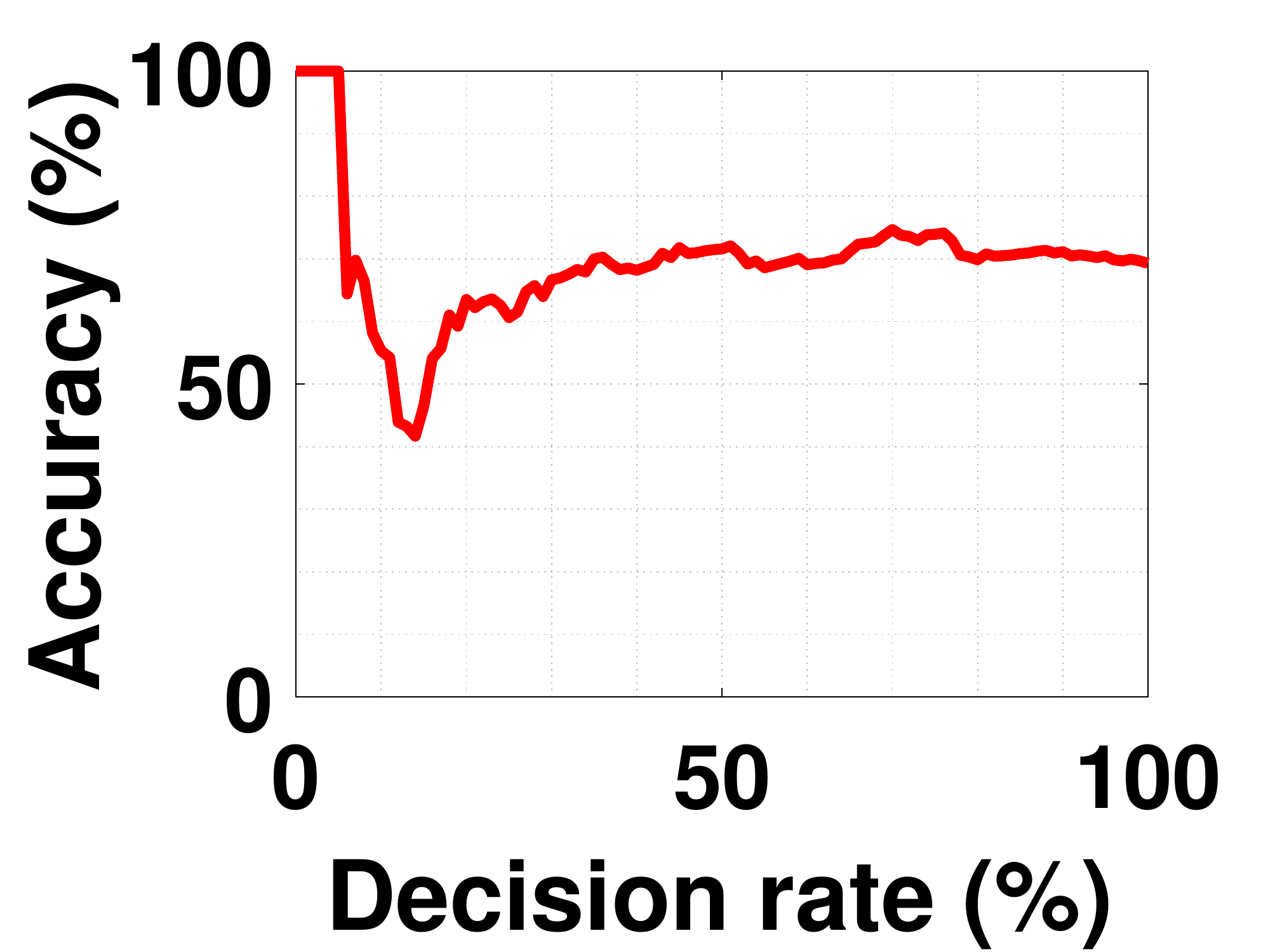}
		{\scriptsize Org. {\ttfamily CEP}-LOG}
		\includegraphics[width=1\linewidth]{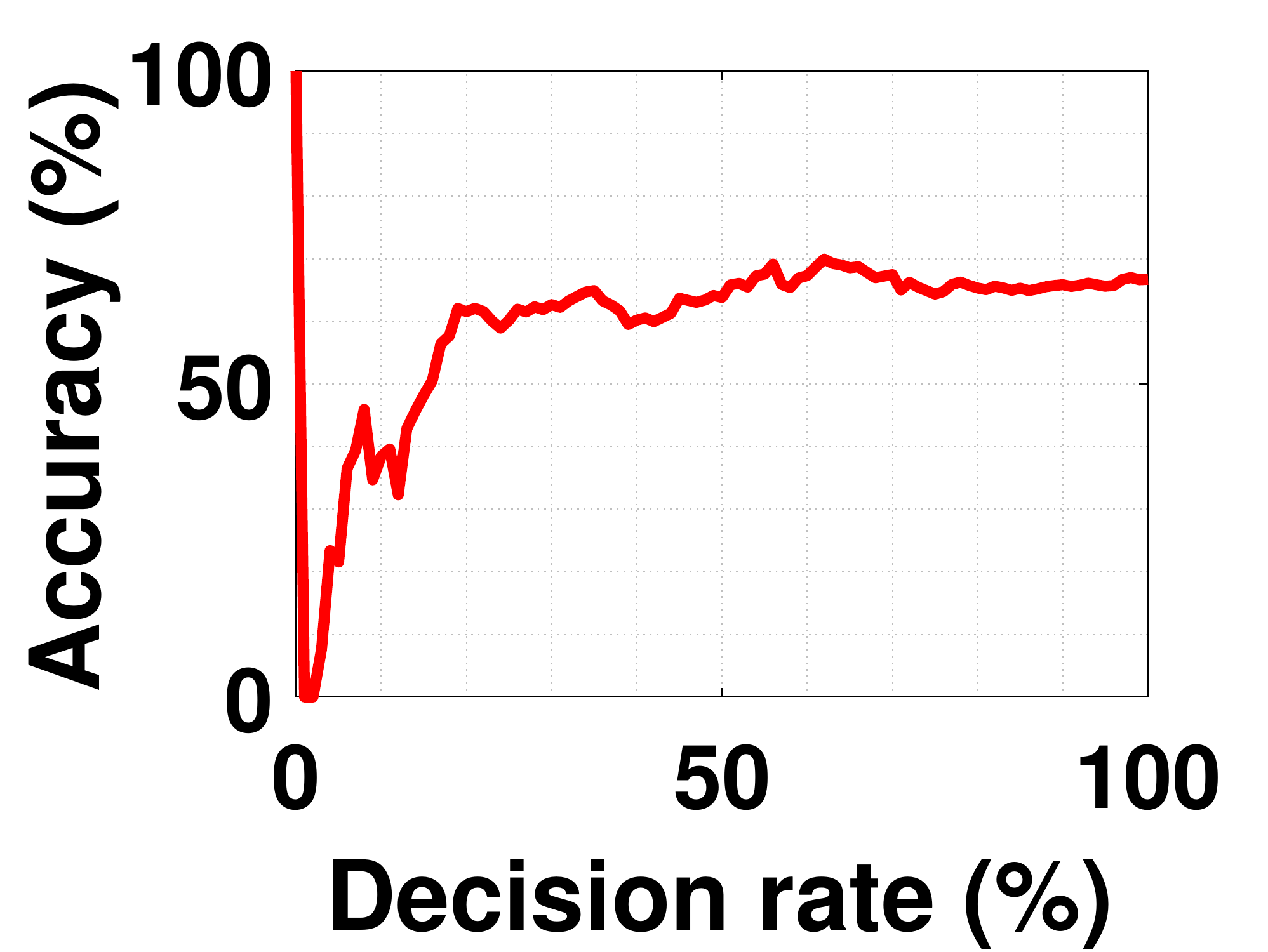}
		{\scriptsize Org. {\ttfamily CEP}-MON}
		\includegraphics[width=1\linewidth]{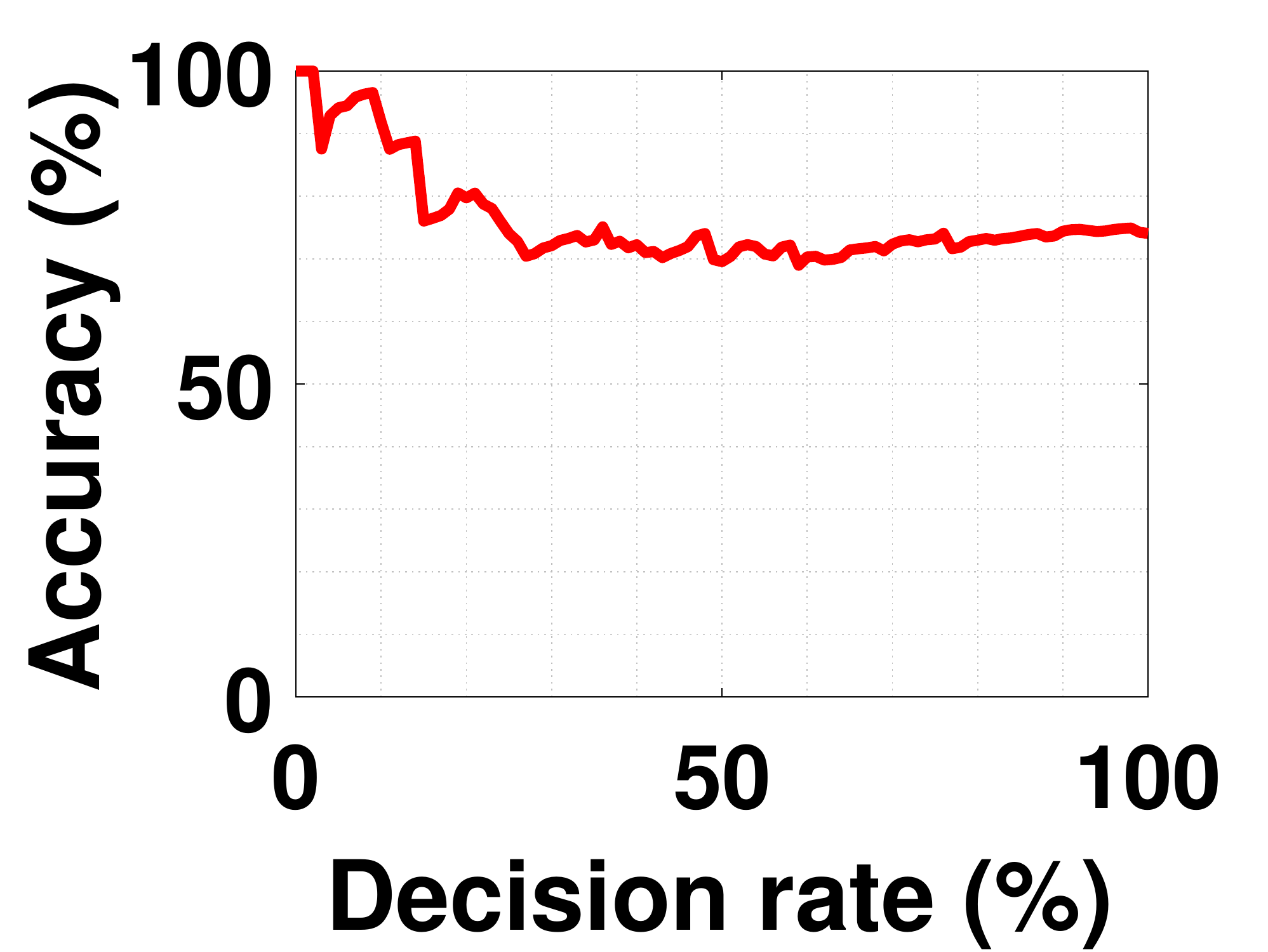}
		{\scriptsize Org. {\ttfamily CEP}-POLY}
		\includegraphics[width=1\linewidth]{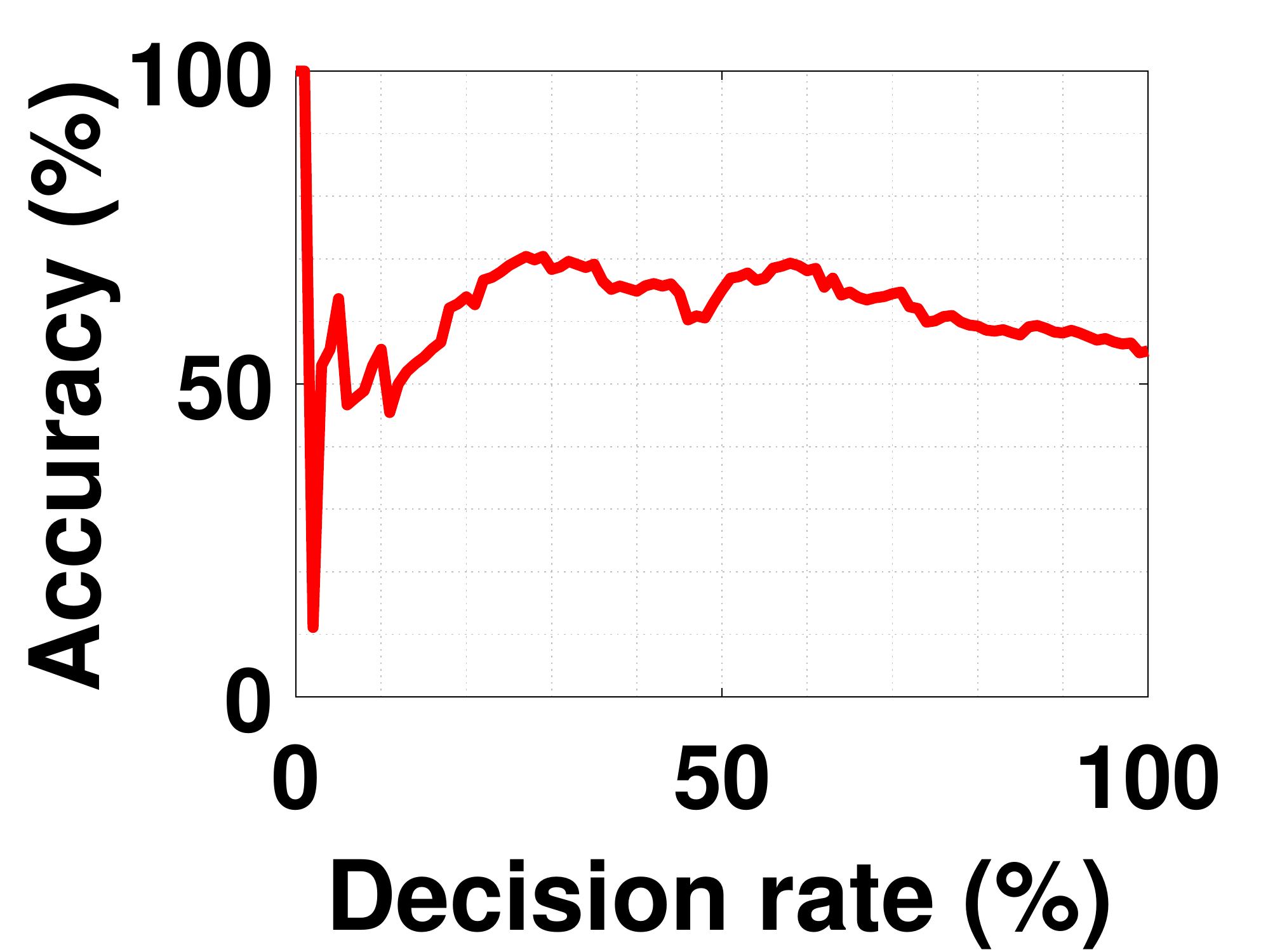}
		{\scriptsize Org. {\ttfamily CEP}-SVR}
		\includegraphics[width=1\linewidth]{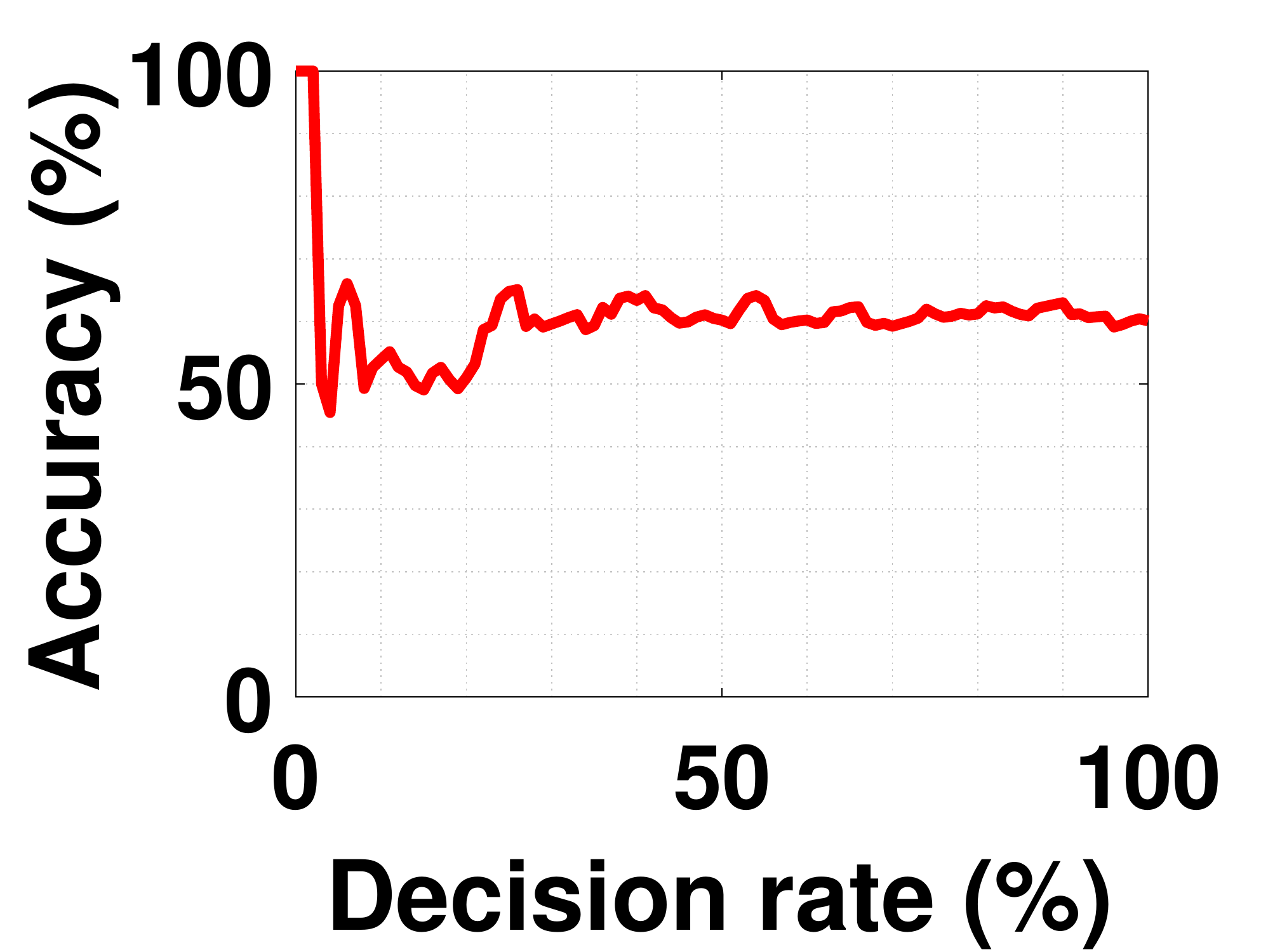}
		{\scriptsize Org. {\ttfamily CEP}-NN}
	\end{multicols}
	
	\begin{multicols}{5}
		\centering
		\includegraphics[width=1\linewidth]{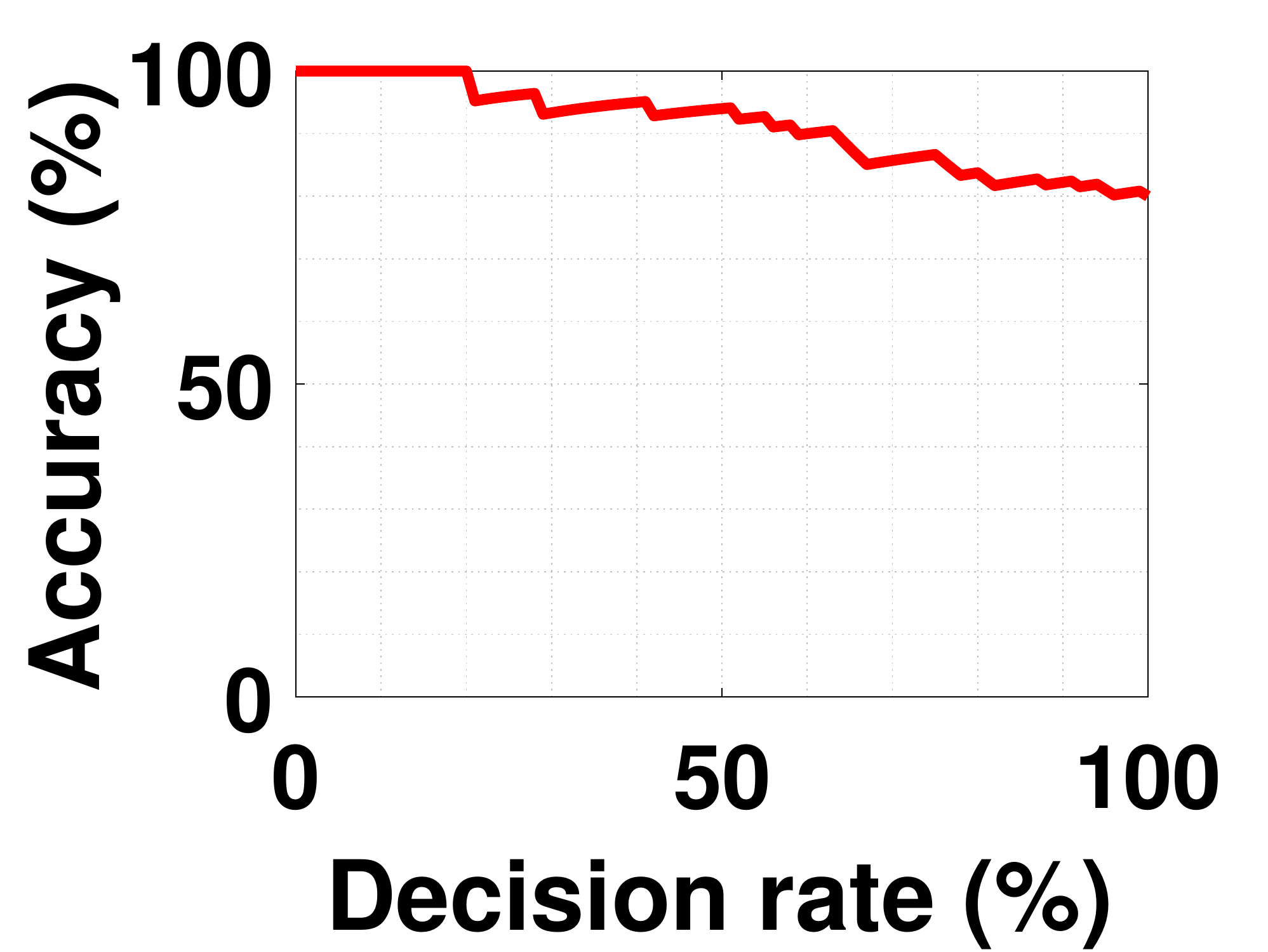}
		{\scriptsize Org. {\ttfamily SIM-G}-LOG}
		\includegraphics[width=1\linewidth]{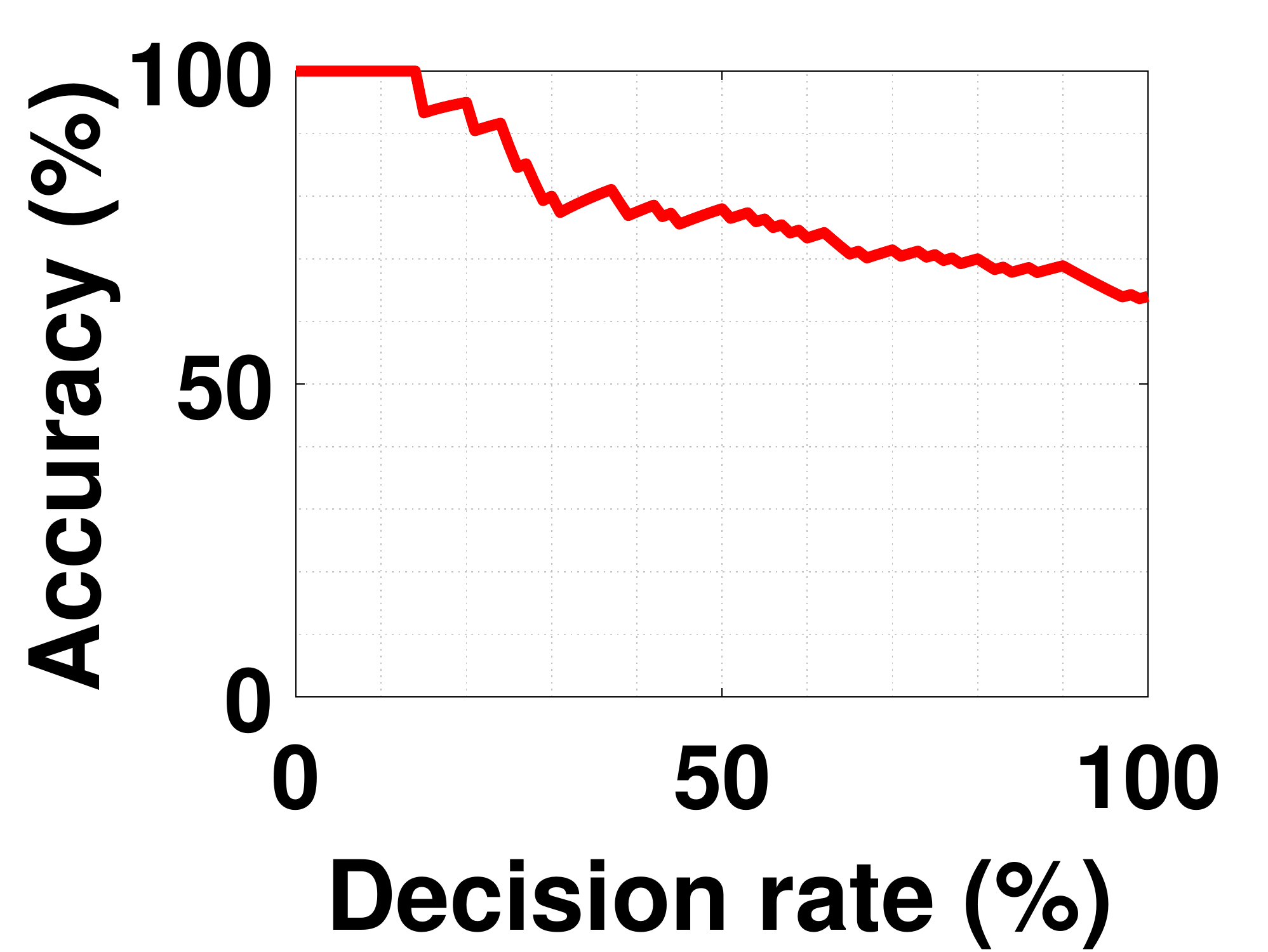}
		{\scriptsize Org. {\ttfamily SIM-G}-MON}
		\includegraphics[width=1\linewidth]{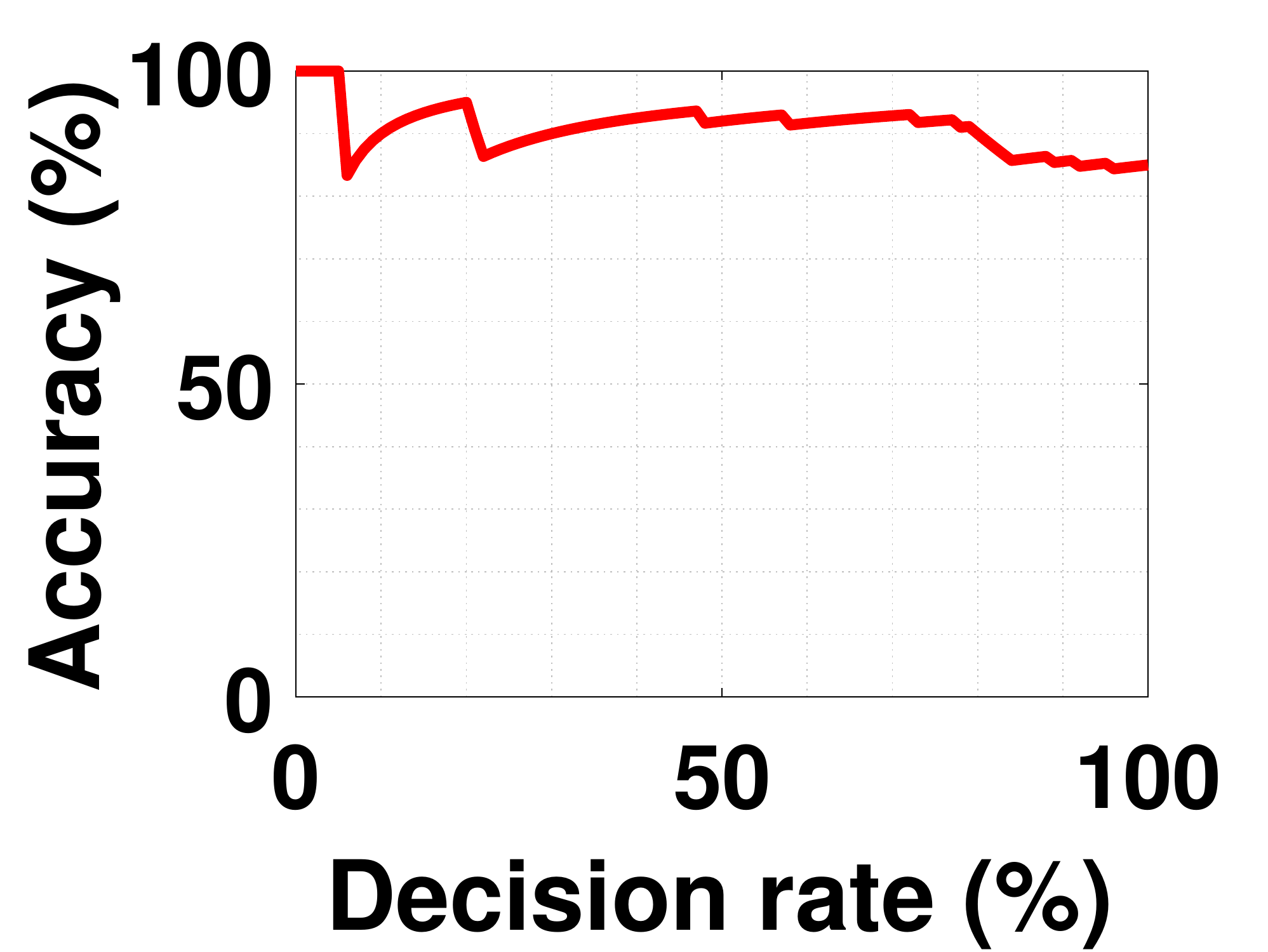}
		{\scriptsize Org. {\ttfamily SIM-G}-POLY}
		\includegraphics[width=1\linewidth]{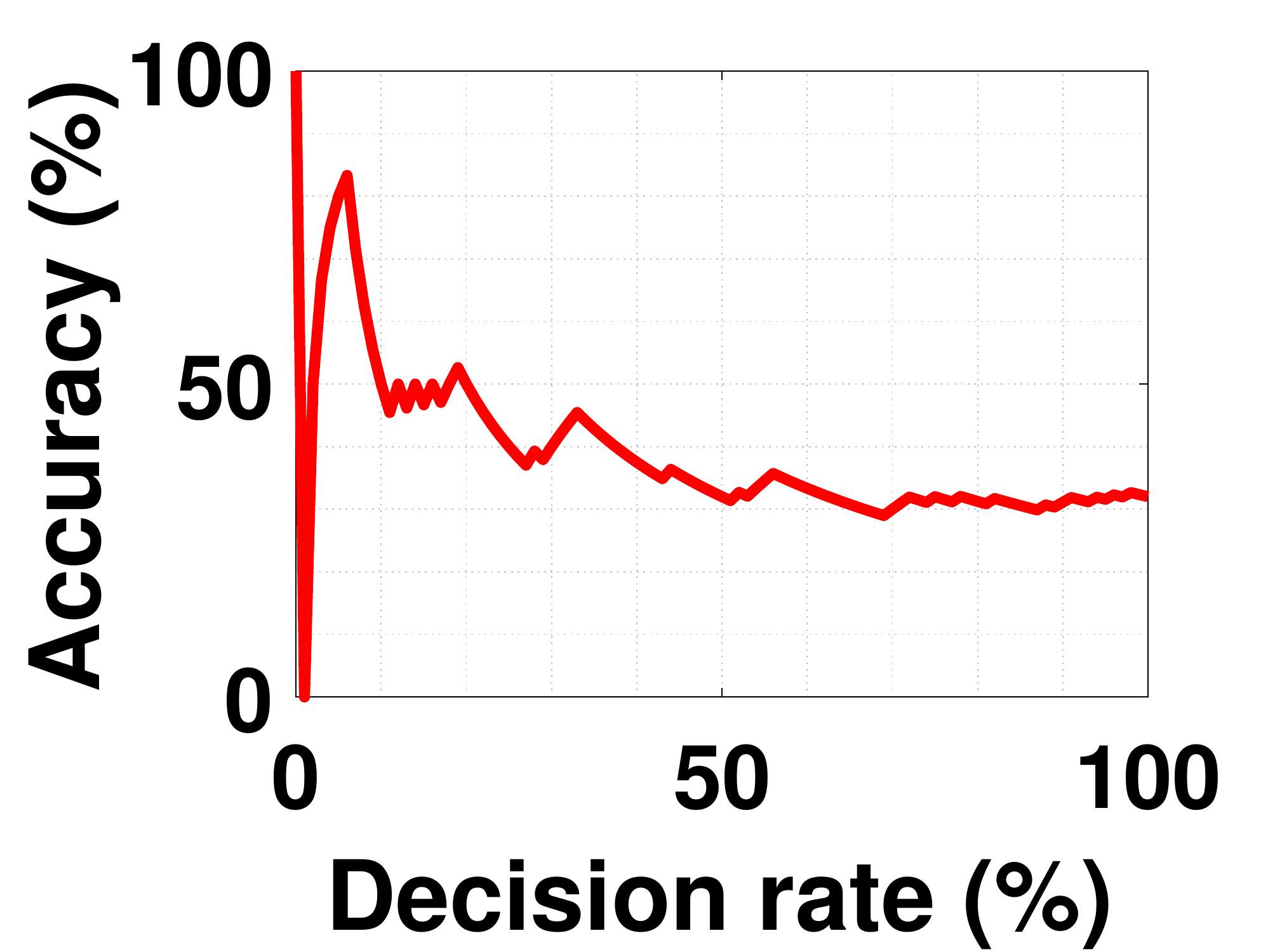}
		{\scriptsize Org. {\ttfamily SIM-G}-SVR}
		\includegraphics[width=1\linewidth]{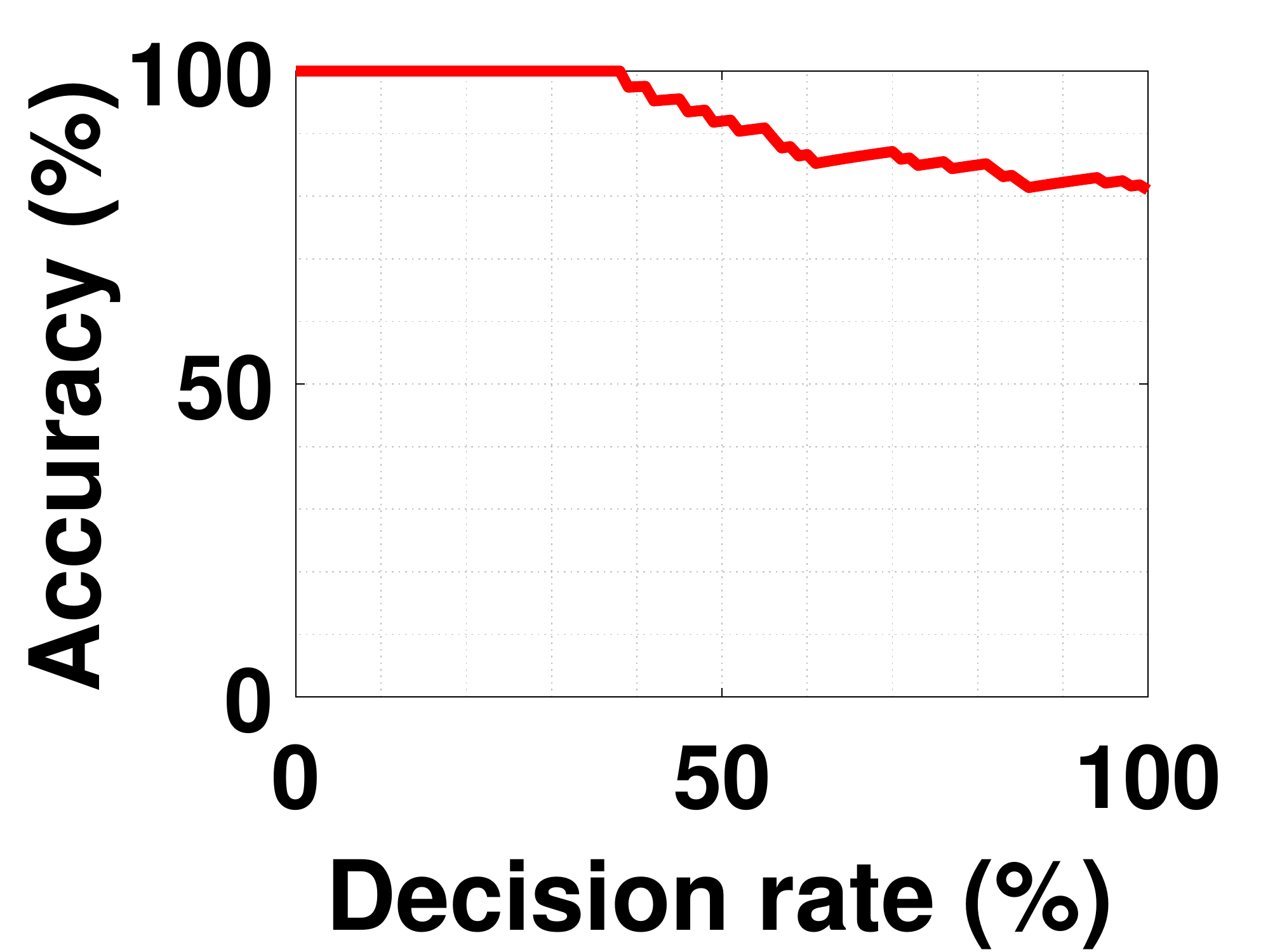}
		{\scriptsize Org. {\ttfamily SIM-G}-NN}
	\end{multicols}
	
	\begin{multicols}{5}
		\centering
		\includegraphics[width=1\linewidth]{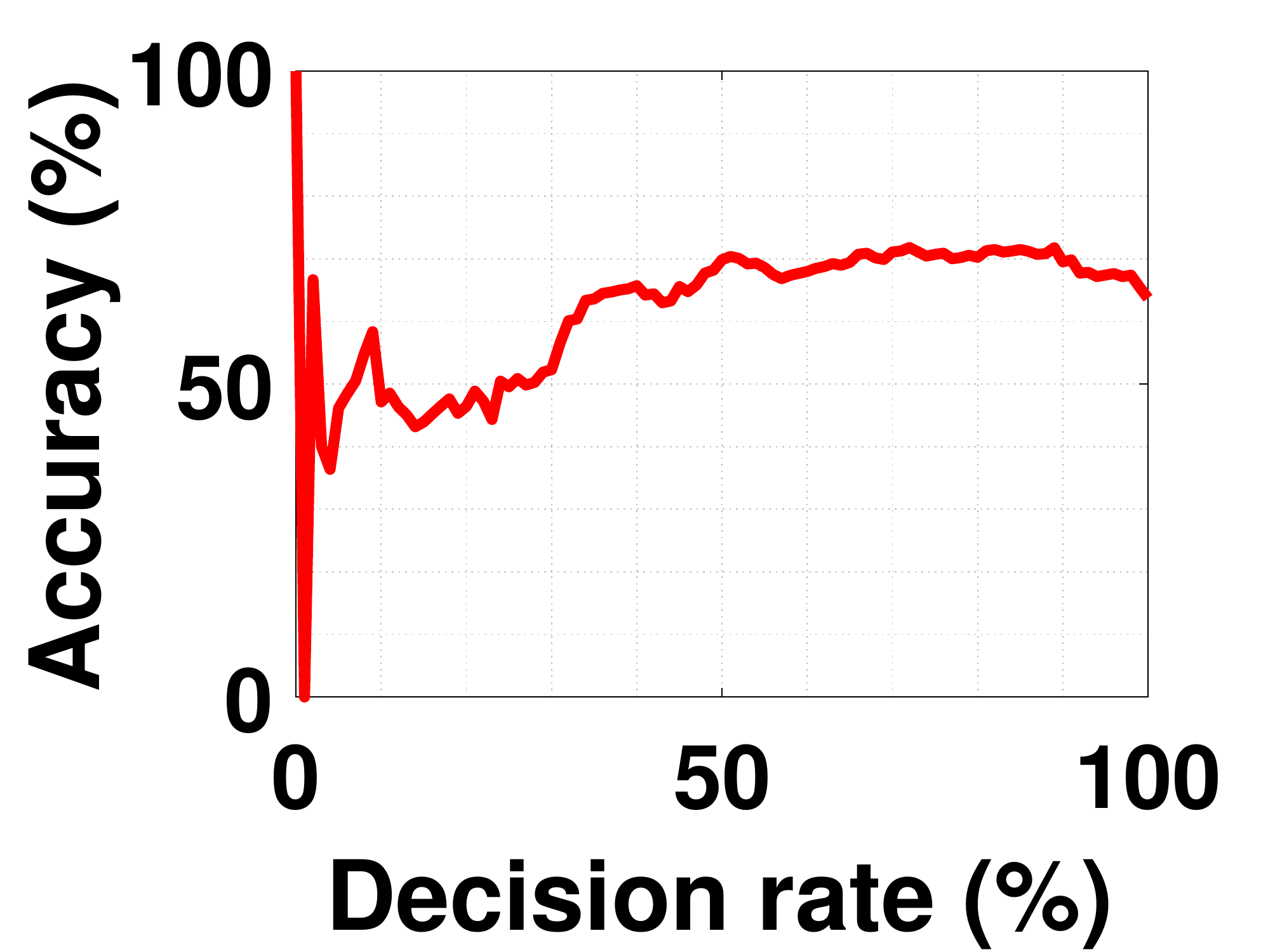}
		{\scriptsize Prep. {\ttfamily CEP}-LOG}
		\includegraphics[width=1\linewidth]{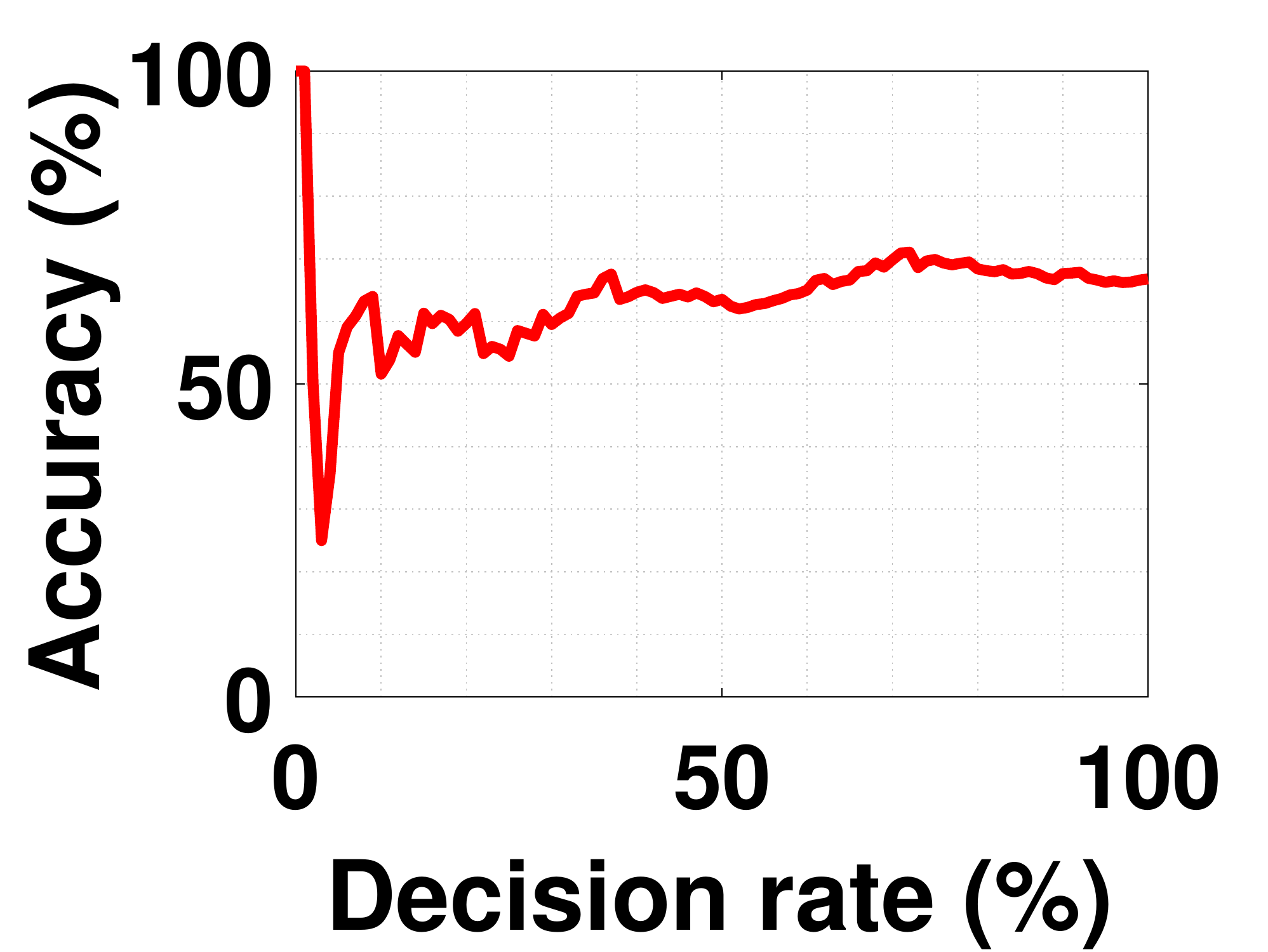}
		{\scriptsize Prep. {\ttfamily CEP}-MON}
		\includegraphics[width=1\linewidth]{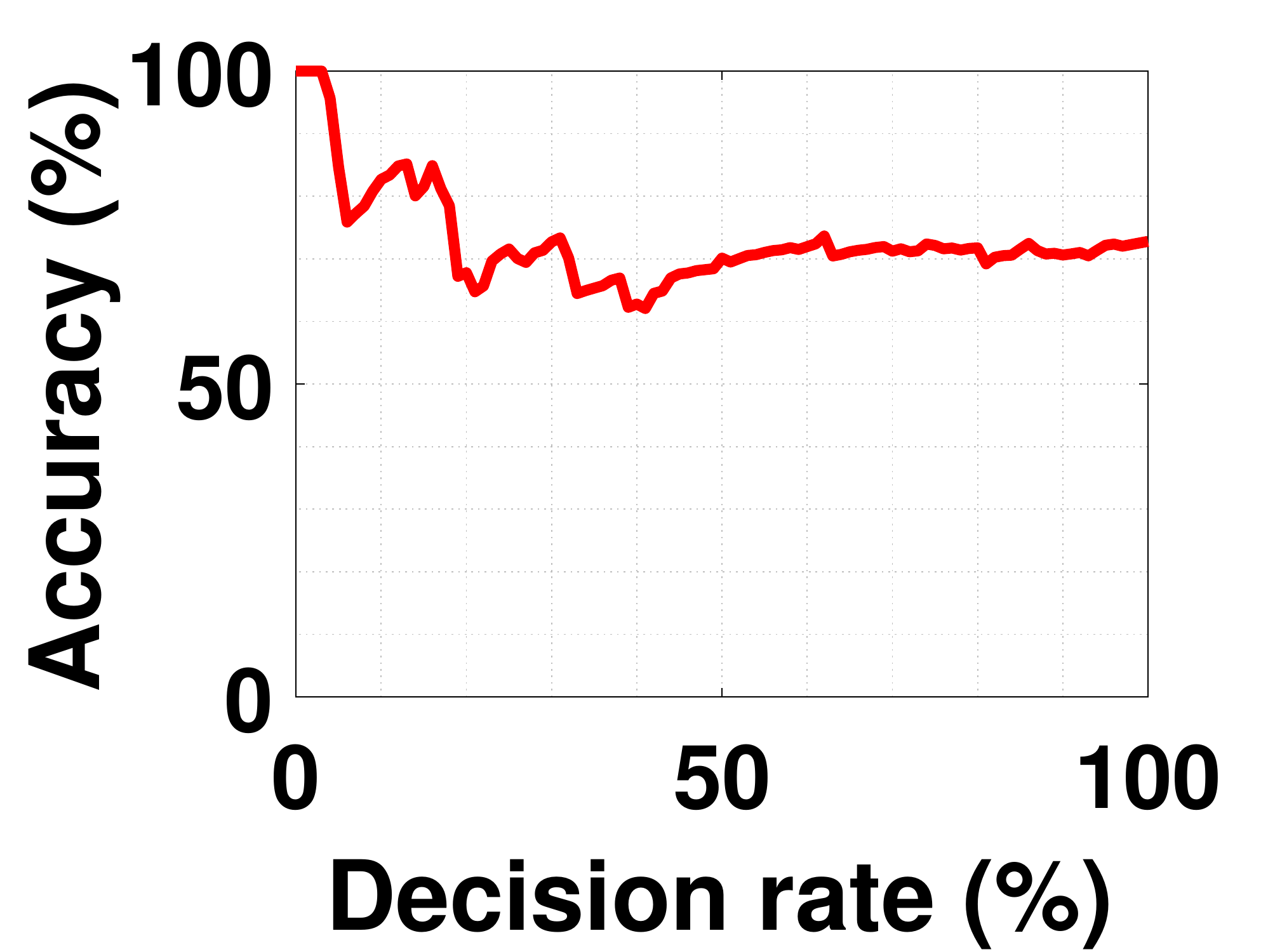}
		{\scriptsize Prep. {\ttfamily CEP}-POLY}
		\includegraphics[width=1\linewidth]{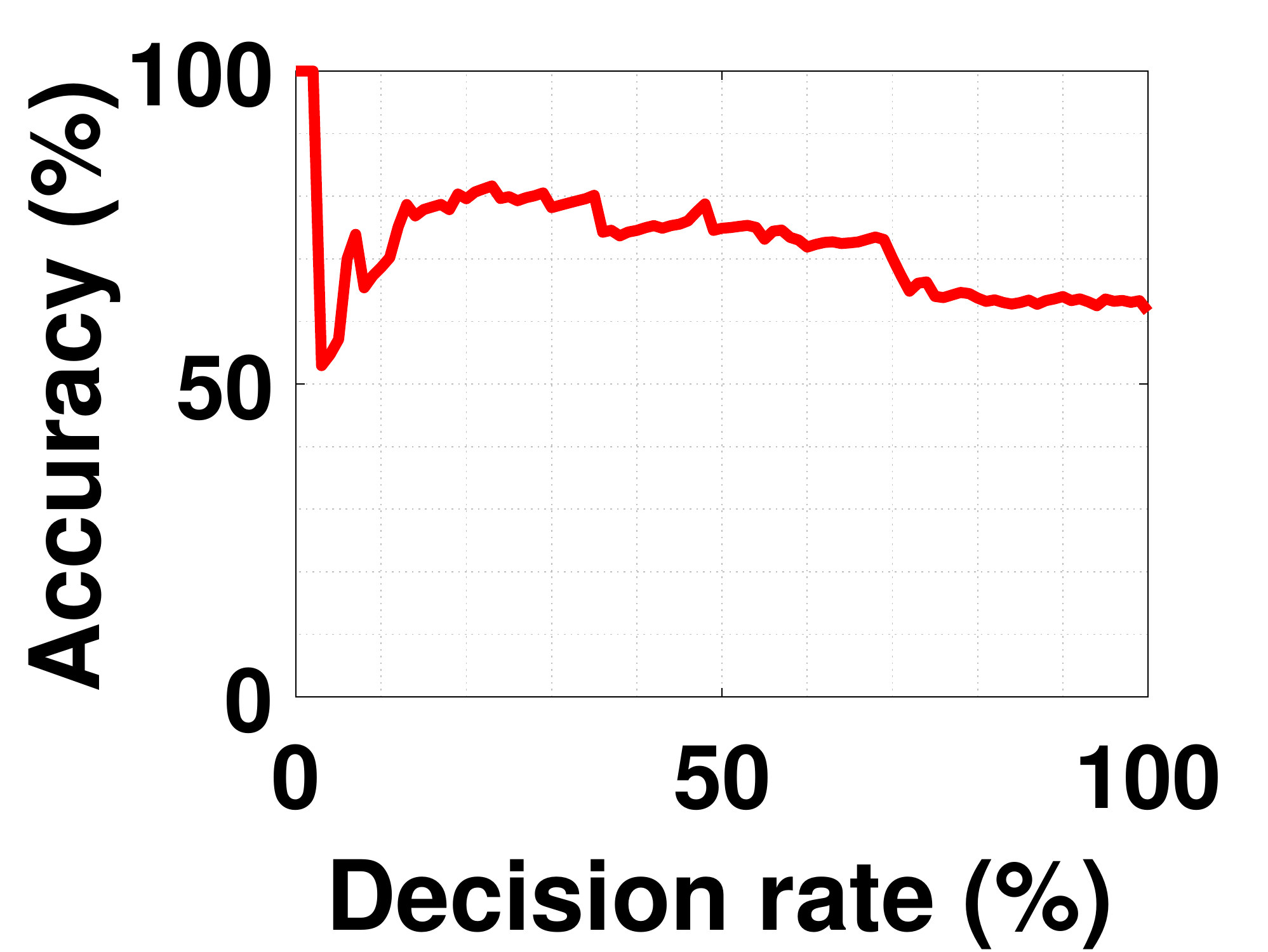}
		{\scriptsize Prep. {\ttfamily CEP}-SVR}
		\includegraphics[width=1\linewidth]{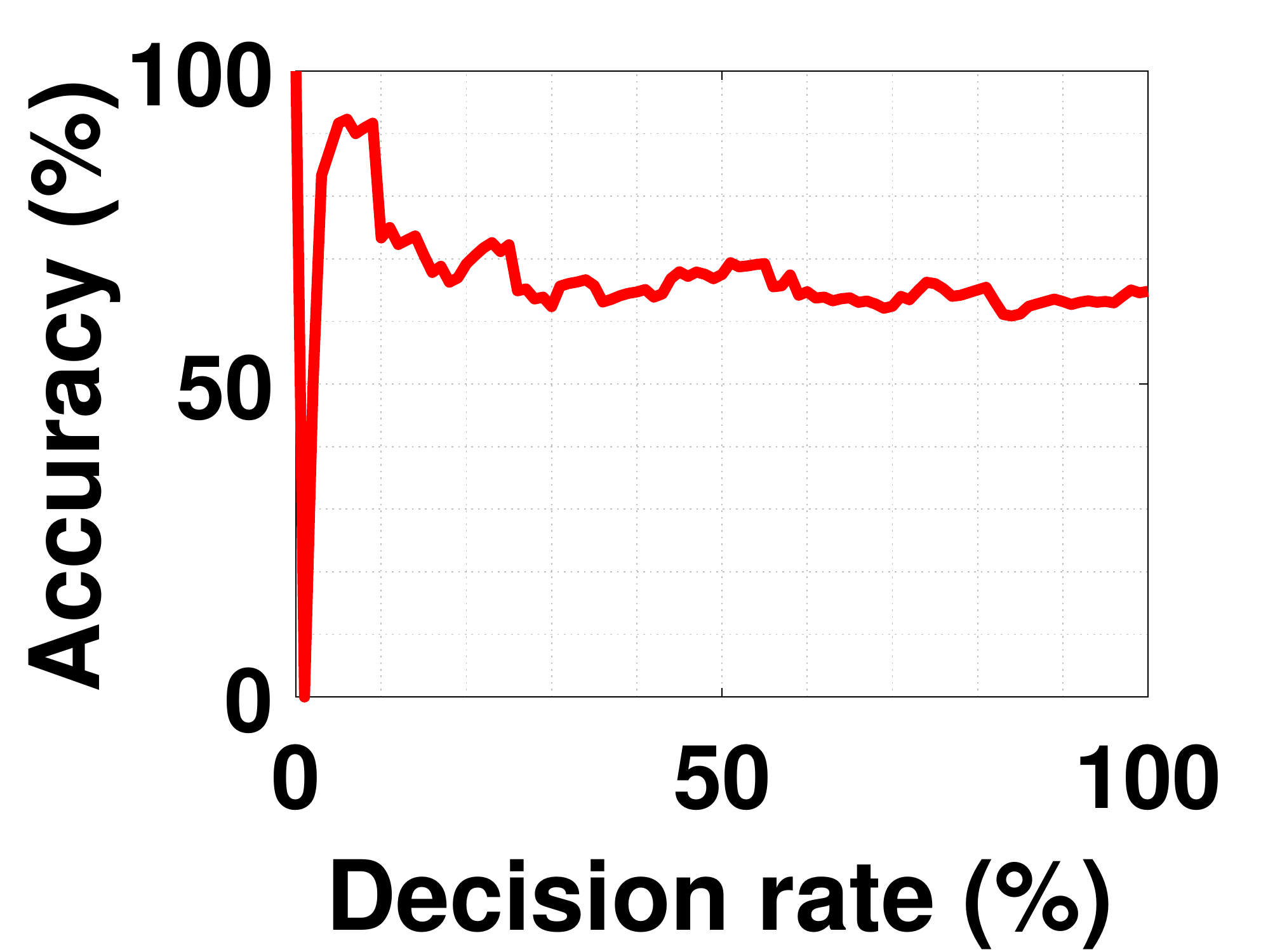}
		{\scriptsize Prep. {\ttfamily CEP}-NN}
	\end{multicols}
	
	\begin{multicols}{5}
		\centering
		\includegraphics[width=1\linewidth]{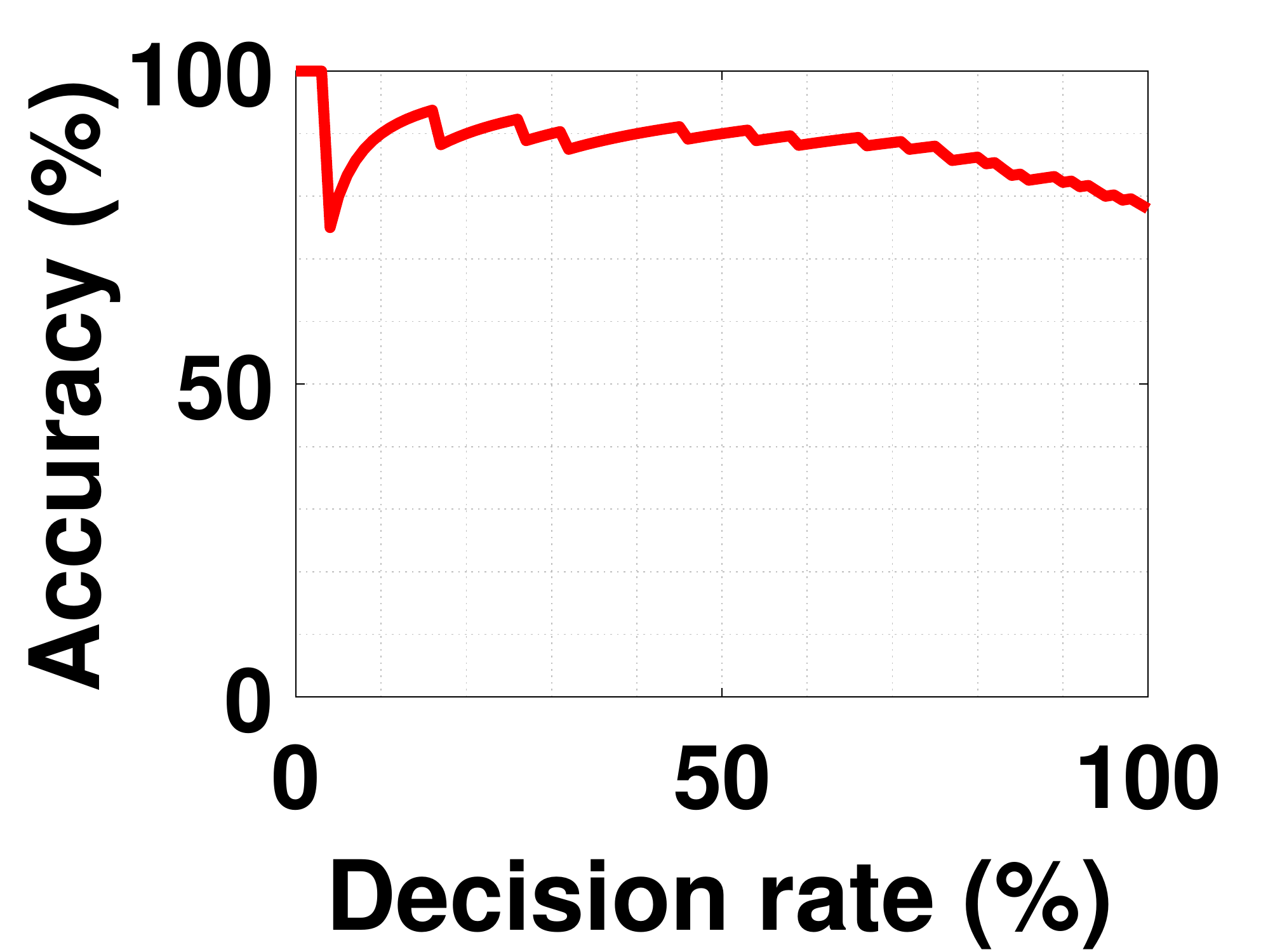}
		{\scriptsize Prep. {\ttfamily SIM-G}-LOG}
		\includegraphics[width=1\linewidth]{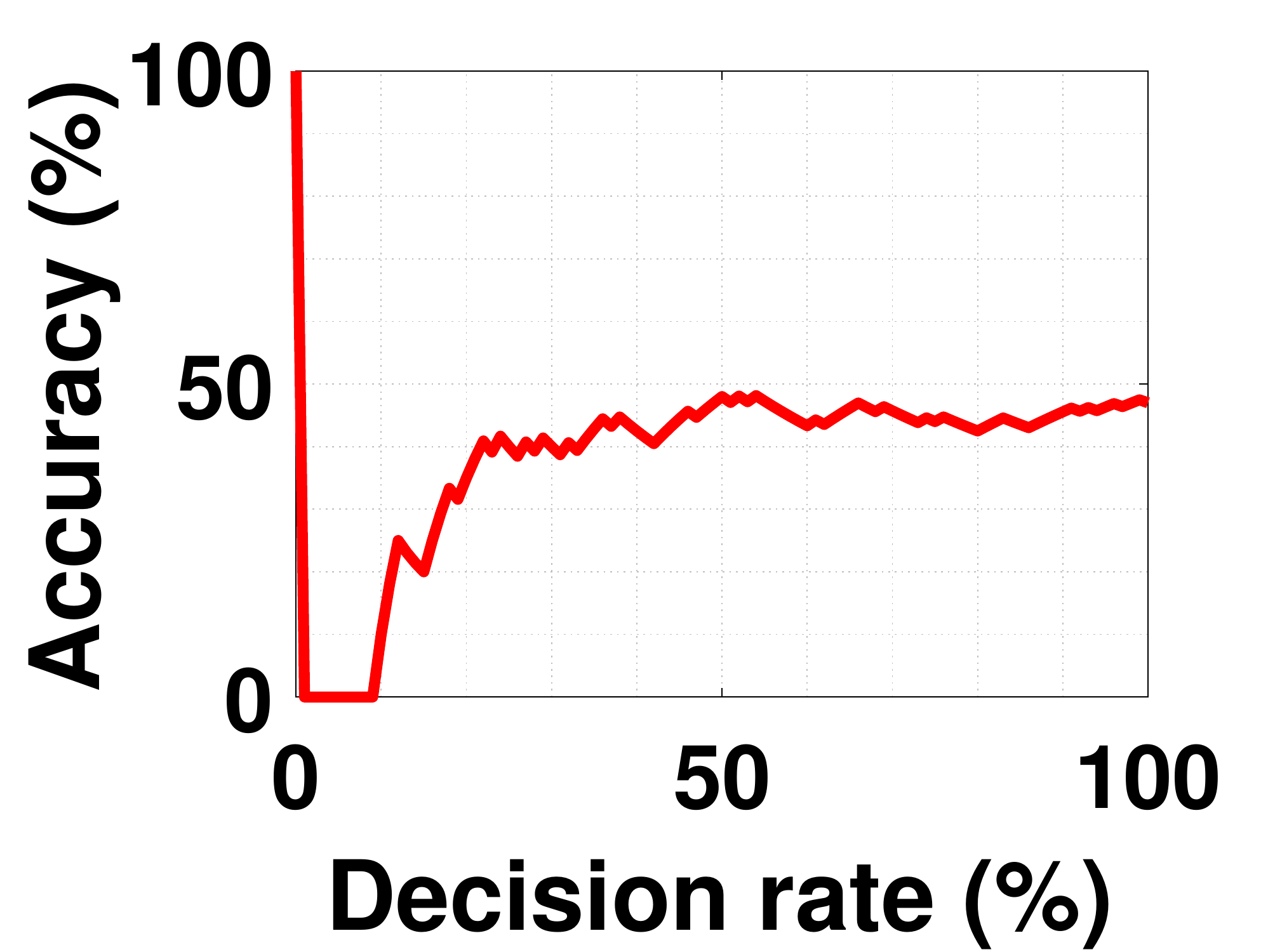}
		{\scriptsize Prep. {\ttfamily SIM-G}-MON}
		\includegraphics[width=1\linewidth]{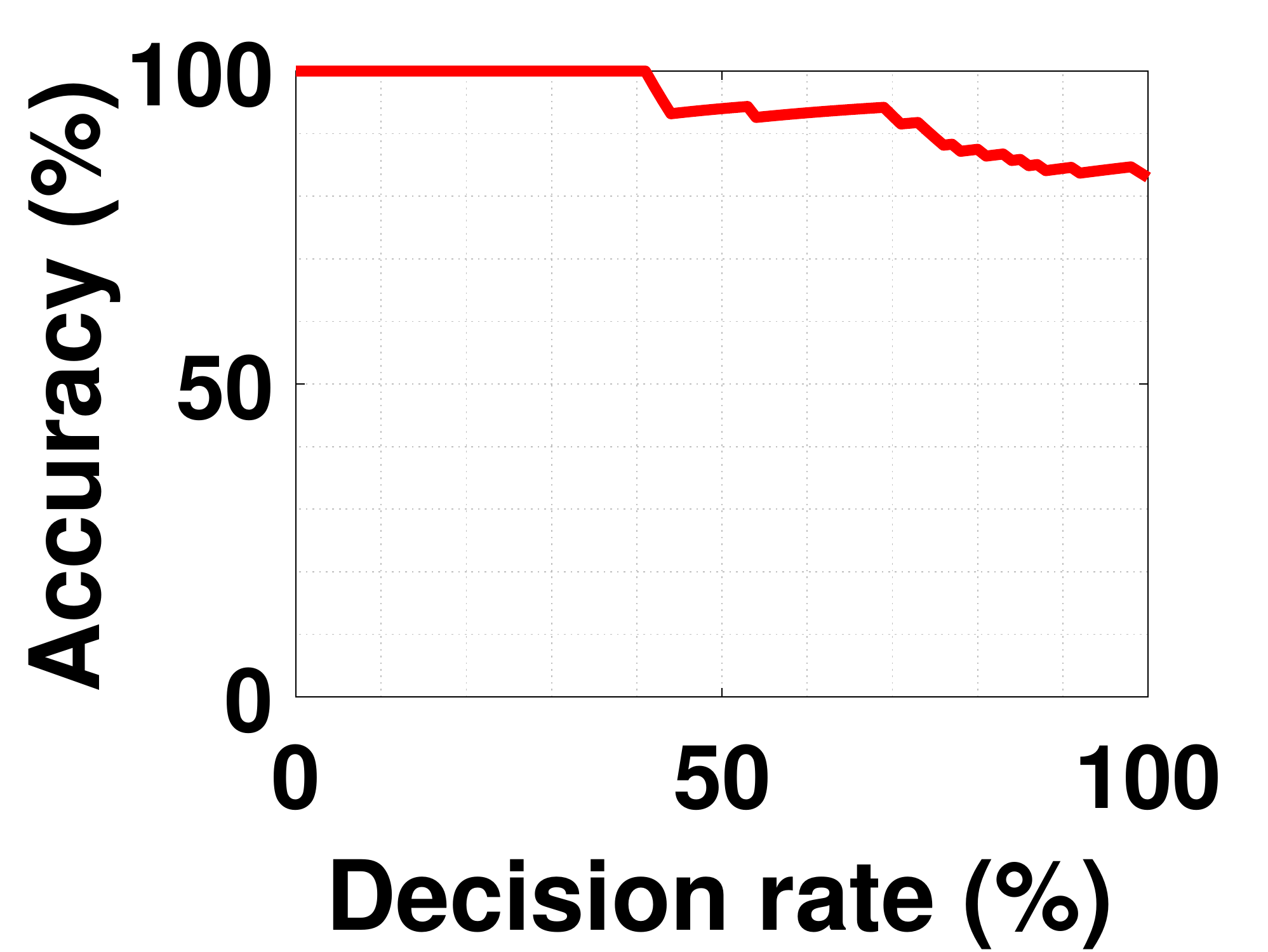}
		{\scriptsize Prep. {\ttfamily SIM-G}-POLY}
		\includegraphics[width=1\linewidth]{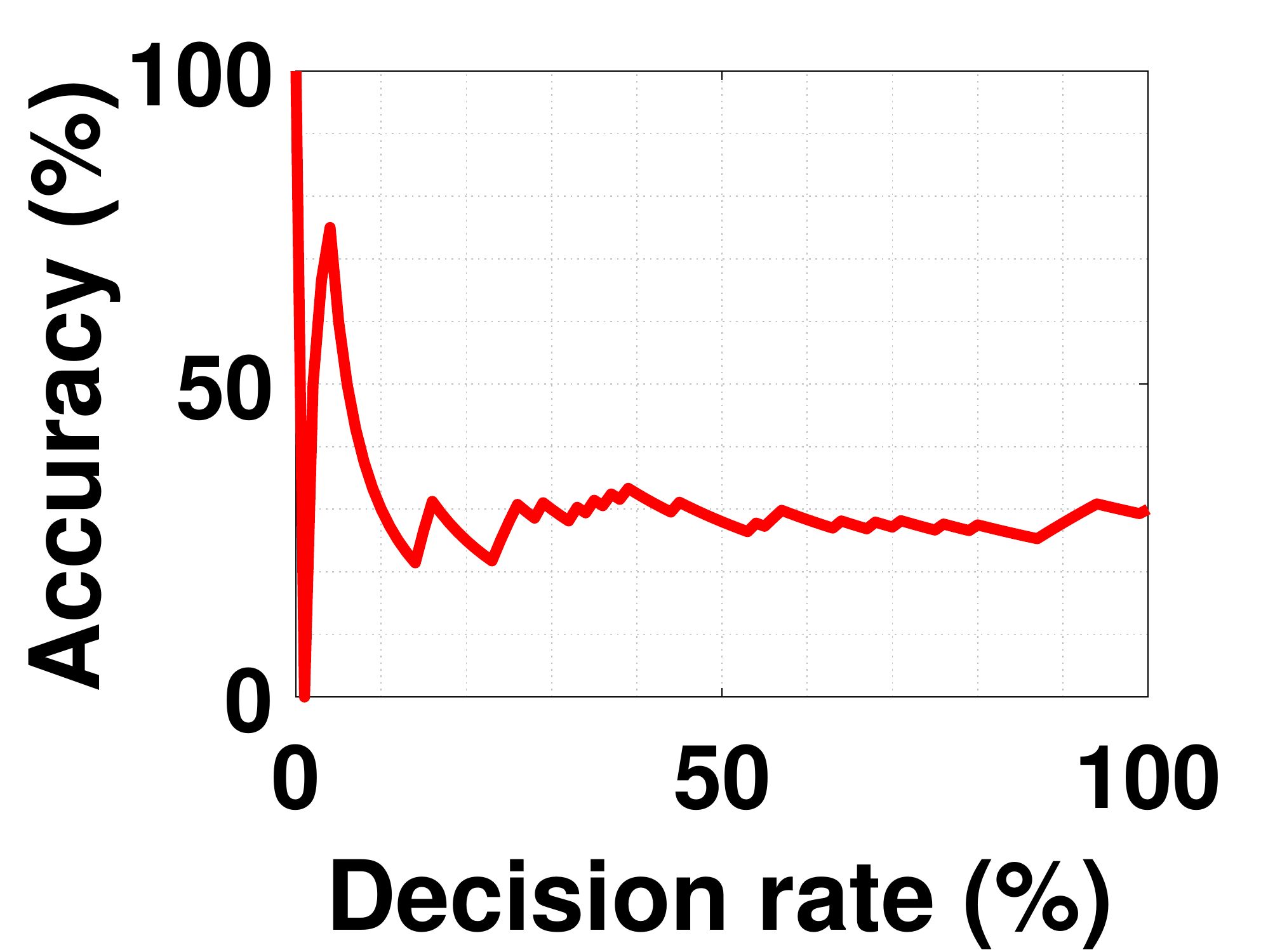}
		{\scriptsize Prep. {\ttfamily SIM-G}-SVR}
		\includegraphics[width=1\linewidth]{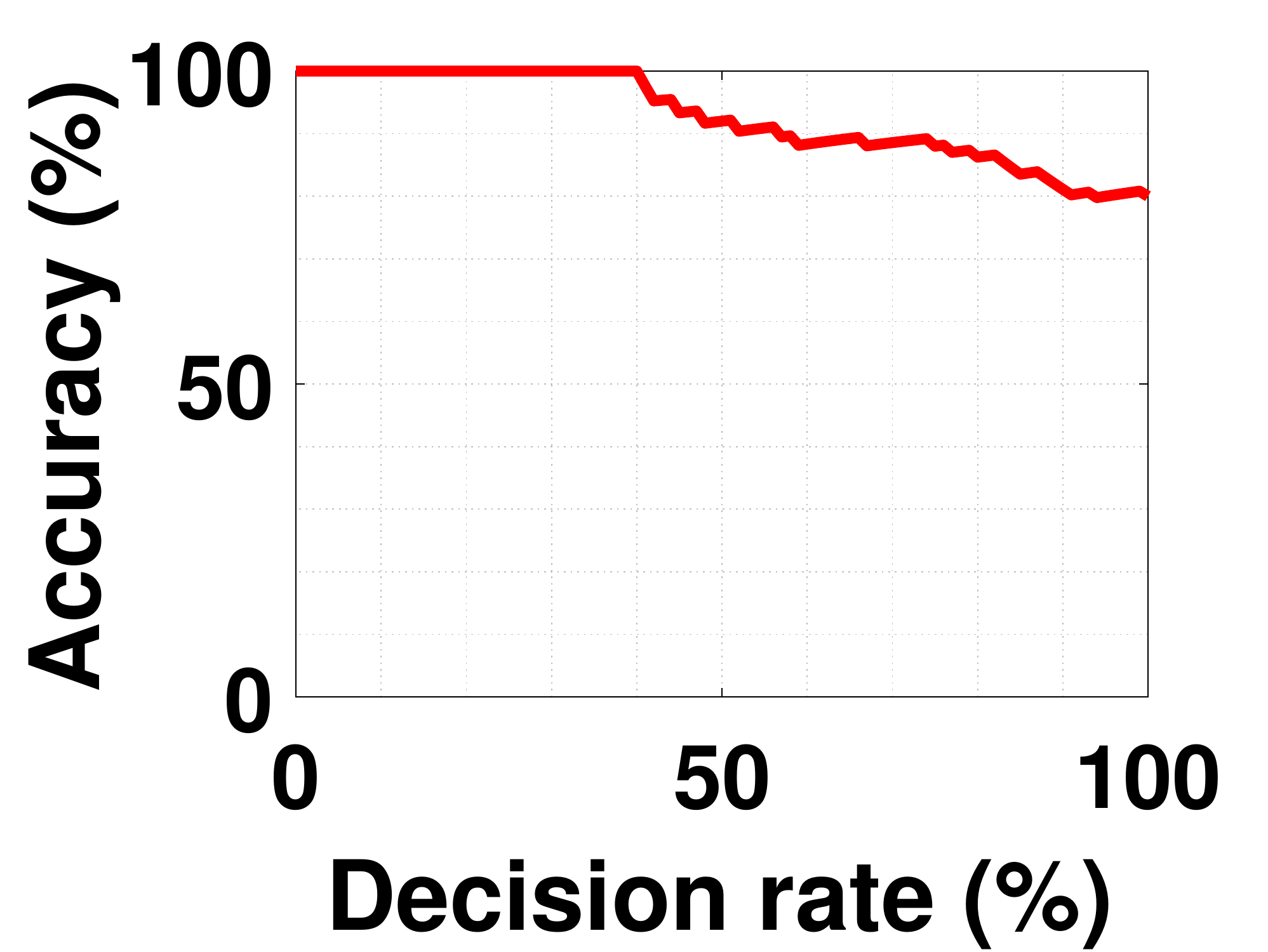}
		{\scriptsize Prep. {\ttfamily SIM-G}-NN}
	\end{multicols}
	\caption{The performance of RECI in the standardized data sets if a certain decisions rate is forced. Here, the decisions are ranked according to the confidence measure defined in (23).\label{fig:fig4}}
\end{figure}

\end{document}